# Implicit Abstraction Heuristics

**Michael Katz** DUGI@TX.TECHNION.AC.IL
**Carmel Domshlak** DCARMEL@IE.TECHNION.AC.IL
*Faculty of Industrial Engineering & Management,*
*Technion, Israel*

## Abstract

State-space search with explicit abstraction heuristics is at the state of the art of cost-optimal planning. These heuristics are inherently limited, nonetheless, because the size of the abstract space must be bounded by some, even if a very large, constant. Targeting this shortcoming, we introduce the notion of *(additive) implicit abstractions*, in which the planning task is abstracted by instances of tractable fragments of optimal planning. We then introduce a concrete setting of this framework, called *fork-decomposition*, that is based on two novel fragments of tractable cost-optimal planning. The induced admissible heuristics are then studied formally and empirically. This study testifies for the accuracy of the fork decomposition heuristics, yet our empirical evaluation also stresses the tradeoff between their accuracy and the runtime complexity of computing them. Indeed, some of the power of the explicit abstraction heuristics comes from precomputing the heuristic function offline and then determining $h(s)$ for each evaluated state $s$ by a very fast lookup in a "database." By contrast, while fork-decomposition heuristics can be calculated in polynomial time, computing them is far from being fast. To address this problem, we show that the time-per-node complexity bottleneck of the fork-decomposition heuristics can be successfully overcome. We demonstrate that an equivalent of the explicit abstraction notion of a "database" exists for the fork-decomposition abstractions as well, despite their exponential-size abstract spaces. We then verify empirically that heuristic search with the "databased" fork-decomposition heuristics favorably competes with the state of the art of cost-optimal planning.

## 1. Introduction

Heuristic search, either through progression in the space of world states or through regression in the space of subgoals, is a common and successful approach to classical planning. It is probably the most popular approach to *cost-optimal planning*, that is, finding a plan with a minimal total cost of its actions. The difference between various heuristic-search algorithms for optimal planning is mainly in the admissible heuristic functions they employ. In state-space search, such a heuristic estimates the cost of achieving the goal from a given state and guarantees not to overestimate that cost.

A useful heuristic function must be accurate as well as efficiently computable. Improving the accuracy of a heuristic function without substantially worsening the time complexity of computing it usually translates into faster search for optimal solutions. During the last decade, numerous computational ideas evolved into new admissible heuristics for classical planning; these include the delete-relaxing max heuristic $h_{\max}$ (Bonet & Geffner, 2001), critical path heuristics $h^m$ (Haslum & Geffner, 2000), landmark heuristics $h^{\mathrm{L}}$, $h^{\mathrm{LA}}$ (Karpas & Domshlak, 2009) and $h^{\text{LM-cut}}$ (Helmert & Domshlak, 2009), and abstraction heuristics such as pattern database heuristics (Edelkamp, 2001) and merge-and-shrink heuristics (Helmert, Haslum, & Hoffmann, 2007). Our focus in this work is on the *abstraction heuristics*.

Generally speaking, an abstraction of a planning task is given by a mapping $\alpha : S \rightarrow S^\alpha$ from the states of the planning task's transition system to the states of some "abstract transition system" such that, for all states $s, s' \in S$, the cost from $\alpha(s)$ to $\alpha(s')$ is upper-bounded by the cost from $s$ to $s'$. The abstraction heuristic value $h^\alpha(s)$ is then the cost from $\alpha(s)$ to the closest goal state of the abstract transition system. Perhaps the most well-known abstraction heuristics are pattern database (PDB) heuristics, which are based on projecting the planning task onto a subset of its state variables and then explicitly searching for optimal plans in the abstract space. Over the years, PDB heuristics have been shown to be very effective in several hard search problems, including cost-optimal planning (Culberson & Schaeffer, 1998; Edelkamp, 2001; Felner, Korf, & Hanan, 2004; Haslum, Botea, Helmert, Bonet, & Koenig, 2007). The conceptual limitation of these heuristics, however, is that the size of the abstract space and its dimensionality must be fixed.[1] The more recent merge-and-shrink abstractions generalize PDB heuristics to overcome the latter limitation (Helmert et al., 2007). Instead of perfectly reflecting just a few state variables, merge-and-shrink abstractions allow for imperfectly reflecting all variables. As demonstrated by the formal and empirical analysis of Helmert et al., this flexibility often makes the merge-and-shrink abstractions much more effective than PDBs. However, the merge-and-shrink abstract spaces are still searched explicitly, and thus they still have to be of fixed size. While quality heuristics estimates can still be obtained for many problems, this limitation is a critical obstacle for many others.

Our goal in this paper is to push the envelope of abstraction heuristics beyond explicit abstractions. We introduce a principled way to obtain abstraction heuristics that limit neither the dimensionality nor the size of the abstract spaces. The basic idea behind what we call *implicit abstractions* is simple and intuitive: instead of relying on abstract problems that are easy to solve because they are small, we can rely on abstract problems belonging to provably tractable fragments of optimal planning. The key point is that, at least theoretically, moving to implicit abstractions removes the requirement on the abstractions size to be small. Our contribution, however, is in showing that implicit abstractions are far from being of theoretical interest only. Specifically,

1. We specify *acyclic causal-graph decompositions*, a general framework for additive implicit abstractions that is based on decomposing the problem at hand along its causal graph. We then introduce a concrete family of such abstractions, called *fork decompositions*, that are based on two novel fragments of tractable cost-optimal planning. Following the type of analysis suggested by Helmert and Mattmüller (2008), we formally analyze the asymptotic performance ratio of the fork-decomposition heuristics and prove that their worst-case accuracy on selected domains is comparable with that of (even parametric) state-of-the-art admissible heuristics. We then empirically evaluate the accuracy of the fork-decomposition heuristics on a large set of domains from recent planning competitions and show that their accuracy is competitive with the state of the art.

1. This does not necessarily apply to *symbolic* PDBs which, on some tasks, may exponentially reduce the PDB's representation (Edelkamp, 2002).

2. The key attraction of explicit abstractions is that state-to-goal costs in the abstract space can be precomputed and stored in memory in a preprocessing phase so that heuristic evaluation during search can be done by a simple lookup. A necessary condition for this would seem to be the small size of the abstract space. However, we show that an equivalent of the PDB and merge-and-shrink's notion of "database" exists for the fork-decomposition abstractions as well, despite the exponential-size abstract spaces of the latter. These *databased implicit abstractions* are based on a proper partitioning of the heuristic computation into parts that can be shared between search states and parts that must be computed online per state. Our empirical evaluation shows that $A^*$ equipped with the "databased" fork-decomposition heuristics favorably competes with the state of the art of cost-optimal planning.

This work is a revision and extension of the formulation and results presented by Katz and Domshlak (2008, 2009), which in turn is based on ideas first sketched also by Katz and Domshlak (2007a).

## 2. Preliminaries

We consider classical planning tasks corresponding to state models with a single initial state and only deterministic actions. Specifically, we consider state models captured by the SAS$^+$ formalism (Bäckström & Nebel, 1995) with nonnegative action costs. Such a *planning task* is given by a quintuple $\Pi = \langle V, A, I, G, cost\rangle$, where:

- $V$ is a set of *state variables*, with each $v \in V$ being associated with a finite domain $\mathcal{D}(v)$. For a subset of variables $V' \subseteq V$, we denote the set of assignments to $V'$ by $\mathcal{D}(V') = \times_{v \in V'} \mathcal{D}(v)$. Each complete assignment to $V$ is called a *state*, and $S = \mathcal{D}(V)$ is the *state space* of $\Pi$. $I$ is an *initial state*. The *goal* $G$ is a partial assignment to $V$; a state $s$ is a *goal state* iff $G \subseteq s$.

- $A$ is a finite set of *actions*. Each action $a$ is a pair $\langle \mathsf{pre}(a), \mathsf{eff}(a)\rangle$ of partial assignments to $V$ called *preconditions* and *effects*, respectively. By $A_v \subseteq A$ we denote the actions affecting the value of $v$. $cost : A \to \mathbb{R}^{0+}$ is a real-valued, nonnegative *action cost* function.

For a variable $v$ and a value $\vartheta \in \mathcal{D}(v)$, instantiation of $v$ by $\vartheta$ is denoted by $v : \vartheta$. For a partial assignment $p$, $\mathcal{V}(p) \subseteq V$ denotes the subset of state variables instantiated by $p$. In turn, for any $V' \subseteq \mathcal{V}(p)$, by $p[V']$ we denote the value of $V'$ in $p$; if $V' = \{v\}$ is a singleton, we use $p[v]$ for $p[V']$. For any sequence of actions $\rho$ and variable $v \in V$, by $\rho_{\downarrow v}$ we denote the restriction of $\rho$ to actions changing the value of $v$; that is, $\rho_{\downarrow v}$ is the maximal subsequence of $\rho$ consisting only of actions in $A_v$.

An action $a$ is applicable in a state $s$ iff $s[v] = \mathsf{pre}(a)[v]$ for all $v \in \mathcal{V}(\mathsf{pre}(a))$. Applying $a$ changes the value of $v \in \mathcal{V}(\mathsf{eff}(a))$ to $\mathsf{eff}(a)[v]$. The resulting state is denoted by $s[\![a]\!]$; by $s[\![\langle a_1, \ldots, a_k\rangle]\!]$ we denote the state obtained from sequential application of the (respectively applicable) actions $a_1, \ldots, a_k$ starting at state $s$. Such an action sequence is an *s-plan* if $G \subseteq s[\![\langle a_1, \ldots, a_k\rangle]\!]$, and it is a *cost-optimal* (or, in what follows, *optimal*) $s$-plan if the sum of its action costs is minimal among all $s$-plans. The purpose of (optimal) planning is finding an (optimal) $I$-plan. For a pair of states $s_1, s_2 \in S$, by $cost(s_1, s_2)$ we refer to the cost of a cost-optimal plan from $s_1$ to $s_2$; $h^*(s) = \min_{s' \supseteq G} cost(s, s')$ is the custom notation for the cost of the optimal $s$-plan in $\Pi$. Finally, important roles in what follows are played by a pair of standard graphical structures induced by planning tasks.

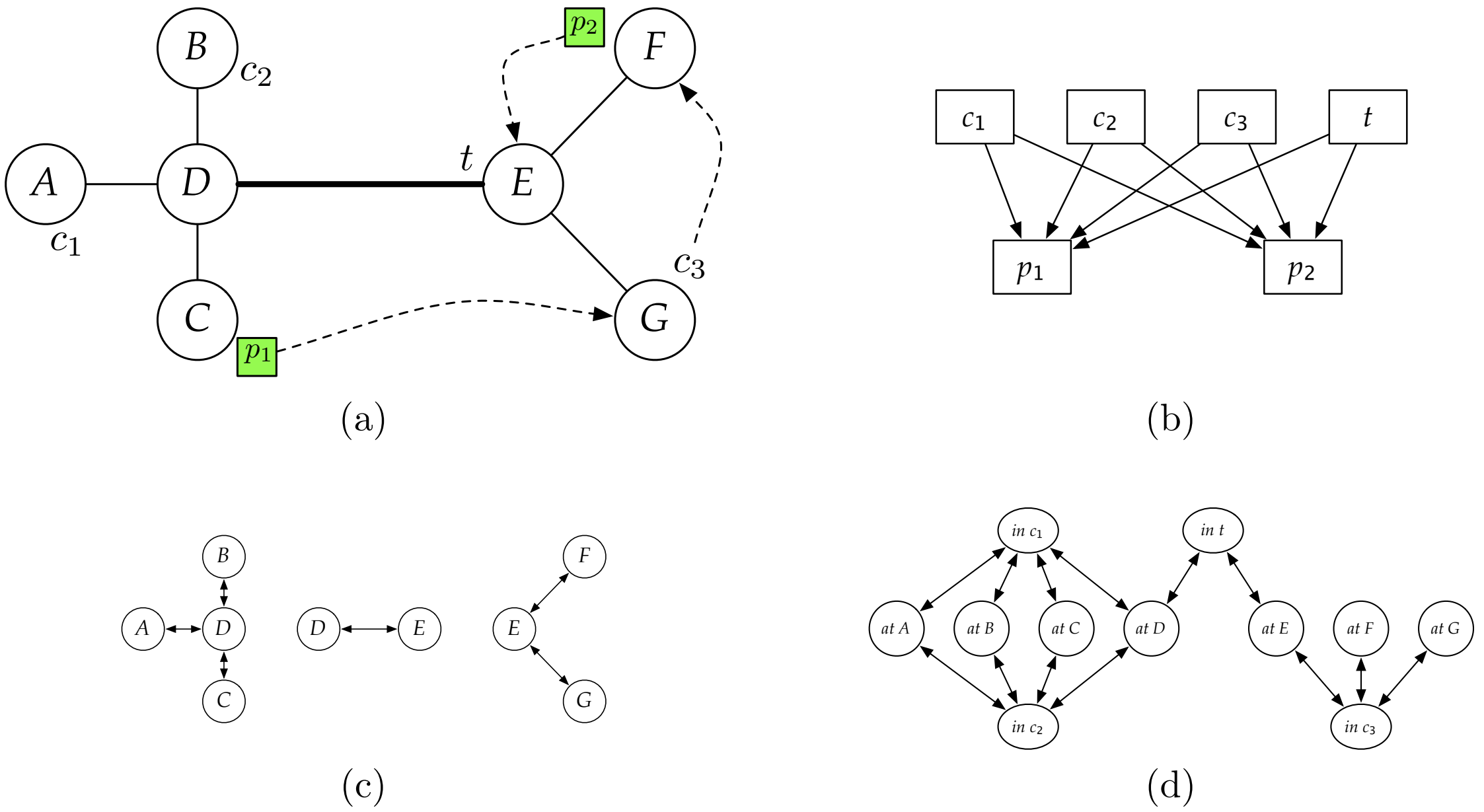

Figure 1: Logistics-style example adapted from Helmert (2006) and illustrated in (a). The goal is to deliver $p_1$ from $C$ to $G$ and $p_2$ from $F$ to $E$ using the cars $c_1, c_2, c_3$ and truck $t$, making sure that $c_3$ ends up at $F$. The cars may only use city roads (thin edges); the truck may only use the highway (thick edge). Figures (b), (c), and (d) depict, respectively, the causal graph of the problem, the domain transition graphs (labels omitted) of $c_1$ and $c_2$ (left), $t$ (center), and $c_3$ (right), and the identical domain transition graphs of of $p_1$ and $p_2$.

- The *causal graph* $CG(\Pi)$ of $\Pi$ is a digraph over nodes $V$. An arc $(v, v')$ is in $CG(\Pi)$ iff $v \neq v'$ and there exists an action $a \in A$ such that $(v, v') \in \mathcal{V}(\mathsf{eff}(a)) \cup \mathcal{V}(\mathsf{pre}(a)) \times \mathcal{V}(\mathsf{eff}(a))$. In this case, we say that $(v, v')$ is *induced* by $a$. By $\mathsf{succ}(v)$ and $\mathsf{pred}(v)$ we respectively denote the sets of immediate successors and predecessors of $v$ in $CG(\Pi)$.

- The *domain transition graph* $DTG(v, \Pi)$ of a variable $v \in V$ is an arc-labeled digraph over the nodes $\mathcal{D}(v)$ such that an arc $(\vartheta, \vartheta')$ labeled with $\mathsf{pre}(a)[V \setminus \{v\}]$ and $cost(a)$ exists in the graph iff both $\mathsf{eff}(a)[v] = \vartheta'$, and either $\mathsf{pre}(a)[v] = \vartheta$ or $v \notin \mathcal{V}(\mathsf{pre}(a))$.

To illustrate various constructs, we use a slight variation of a Logistics-style example from Helmert (2006). This example is depicted in Figure 1a, and in $\text{SAS}^+$ it has

$$
\begin{aligned}
V &= \{p_1, p_2, c_1, c_2, c_3, t\} \\
\mathcal{D}(p_1) &= \mathcal{D}(p_2) = \{A, B, C, D, E, F, G, c_1, c_2, c_3, t\} \\
\mathcal{D}(c_1) &= \mathcal{D}(c_2) = \{A, B, C, D\} \\
\mathcal{D}(c_3) &= \{E, F, G\} \\
\mathcal{D}(t) &= \{D, E\} \\
I &= \{p_1 : C, p_2 : F, t : E, c_1 : A, c_2 : B, c_3 : G\} \\
G &= \{p_1 : G, p_2 : E, c_3 : F\},
\end{aligned}
$$

and actions corresponding to all possible loads and unloads, as well as single-segment movements of the vehicles. For instance, if action $a$ captures loading $p_1$ into $c_1$ at $C$, then $\mathsf{pre}(a) = \{p_1 : C, c_1 : C\}$, and $\mathsf{eff}(a) = \{p_1 : c_1\}$. All actions in the example have unit cost. The causal graph of this example, as well as the domain transition graphs of the state variables, are depicted in Figures 1b-1d.

Heuristic functions are used by informed-search procedures to estimate the cost (of the cheapest path) from a search node to the nearest goal node. Our focus here is on state-dependent, admissible abstraction heuristics. A heuristic function $h$ is *state-dependent* if its estimate for a search node depends only on the problem state associated with that node, that is, $h : S \rightarrow \mathbb{R}^{0+} \cup \{\infty\}$. Most heuristics in use these days are state-dependent (though see, e.g., Richter, Helmert, & Westphal, 2008 and Karpas & Domshlak, 2009 for a different case). A heuristic $h$ is *admissible* if $h(s) \leq h^*(s)$ for all states $s$. If $h_1$ and $h_2$ are two admissible heuristics, and $h_2(s) \leq h_1(s)$ for all states $s$, we say that $h_1$ *dominates* $h_2$.

For any set of admissible heuristics $h_1, \dots, h_m$, their pointwise maximum is always an admissible heuristic, dominating each individual heuristic in the set. For some sets of admissible heuristics, their pointwise sum is also admissible and dominates their pointwise maximum. Many recent works on cost-optimal planning are based on *additive ensembles of admissible heuristics*, and this includes critical-path heuristics (Haslum, Bonet, & Geffner, 2005; Coles, Fox, Long, & Smith, 2008), pattern database heuristics (Edelkamp, 2001; Haslum et al., 2007), and landmark heuristics (Karpas & Domshlak, 2009; Helmert & Domshlak, 2009). In particular, Katz and Domshlak (2007a, 2008) and Yang et al. (2007, 2008) independently introduced a general criterion for admissible additive ensembles of heuristics, called in the former work *action cost partitioning*. This criterion can be formalized as follows. Let $\Pi = \langle V, A, I, G, cost \rangle$ be a planning task and $\{cost_i : A \rightarrow \mathbb{R}^{0+}\}_{i=1}^{m}$ a family of cost functions such that $\sum_{i=1}^{m} cost_i(a) \leq cost(a)$ for all actions $a \in A$. If $\{h_i\}_{i=1}^{m}$ is a set of arbitrary admissible heuristic functions for $\Pi_i = \langle V, A, I, G, cost_i \rangle$, respectively, then $\sum_{i=1}^{m} h_i$ is also an admissible heuristic for $\Pi$. The set of cost functions $\{cost_i\}_{i=1}^{m}$ can be seen as a partition of the action costs *cost*.

## 3. Abstractions and Abstraction Heuristics

The semantics of any planning task $\Pi$ is given by its induced state-transition model, often called the *transition graph* of $\Pi$.

**Definition 1** *A* **transition graph** *is a tuple* $\mathcal{T} = (S, L, Tr, s^0, S^\star, \varpi)$ *where* $S$ *is a finite set of* states, $L$ *is a finite set of* transition labels, $Tr \subseteq S \times L \times S$ *is a set of (labeled)* transitions, $s^0 \in S$ *is an* initial state, $S^\star \subseteq S$ *is a set of* goal states, *and* $\varpi : L \rightarrow \mathbb{R}^{0+}$ *is a* transition cost function.

- *For a state* $s \in S$ *and a subset of states* $S' \subseteq S$ *in* $\mathcal{T}$, $cost(s, S')$ *is the* **cost** *(of a cheapest with respect to* $\varpi$ *path) from* $s$ *to a state in* $S'$ *along the transitions of* $\mathcal{T}$*; if no state in* $S'$ *is reachable from* $s$*, then we have* $cost(s, S') = \infty$.
- *Any path from* $s^0$ *to* $S^\star$ *is a* **plan** *for* $\mathcal{T}$*, and cheapest such plans are called* **optimal**.

The states of the transition graph $\mathcal{T}(\Pi)$ induced by a planning task $\Pi = \langle V, A, I, G, cost\rangle$ are the states of $\Pi$. The transition labels of $\mathcal{T}(\Pi)$ are the actions $A$; there is a transition $(s, a, s[\![a]\!]) \in Tr$ iff $a$ is applicable in $s$; the initial state $s^0 = I$; the set of goal states $S^\star = \{s \in S \mid s \supseteq G\}$; and the transition cost function $\varpi = cost$. We now proceed with formally specifying the notion of abstraction. Our definition of abstraction resembles that of Prieditis (1993), and right from the beginning we specify a more general notion of *additive abstraction*. Informally, by additive abstraction we refer to a set of abstractions interconstrained by a requirement to *jointly* not overestimate the transition-path costs in the abstracted transition graph.

**Definition 2** *An* **additive abstraction** *of a transition graph* $\mathcal{T} = (S, L, Tr, s^0, S^\star, \varpi)$ *is a set of pairs* $\{\langle \mathcal{T}_i, \alpha_i\rangle\}_{i=1}^m$ *where, for* $1 \leq i \leq m$,

- $\mathcal{T}_i = (S_i, L_i, Tr_i, s_i^0, S_i^\star, \varpi_i)$ *is a transition graph,*
- $\alpha_i : S \rightarrow S_i$ *is a function, called* abstraction mapping*, such that*
  - $\alpha_i(s^0) = s_i^0$, $\alpha_i(s) \in S_i^\star$ *for all* $s \in S^\star$*, and,*
  - *for all pairs of states* $s, s' \in S$ *holds*

$$\sum_{i=1}^{m} cost(\alpha_i(s), \alpha_i(s')) \leq cost(s, s'). \tag{1}$$

A few words on why we use *this* particular notion of abstraction. The term "abstraction" is usually associated with simplifying the original system, reducing and factoring out details less crucial in the given context. Which details can be reduced and which should better be preserved depends, of course, on the context. For instance, in the context of formal verification, the abstract transition graphs are required not to decrease the reachability between the states; that is, if there is a path from $s$ to $s'$ in the original transition graph, then there should be a path from $\alpha(s)$ to $\alpha(s')$ in the abstract transition graph (Clarke, Grumberg, & Peled, 1999). In addition, the reachability should also be increased as little as possible. Beyond that, the precise relationship between the path costs in the original and abstract transition graphs is only of secondary importance. In contrast, when abstractions are designed to induce admissible heuristic functions for heuristic search, the relationship between the path costs as captured by Eq. 1 is what must be obeyed. However, requirements above and beyond the general requirement of Eq. 1 not to overestimate the distances between the states are unnecessary. Hence, in particular, Definition 2 generalizes the notion of abstraction by Helmert et al. (2007) by replacing the condition of preserving individual transitions and their labels, that is, $(\alpha(s), l, \alpha(s'))$ if $(s, l, s')$, with a weaker condition stated in Eq. 1. The reader, of course, may well ask whether the generality of the condition in Eq. 1 beyond the condition of Helmert et al. (2007) really delivers any practical gain, and later we show that the answer to this question is affirmative. For now, we proceed with adding further requirements essential to making abstraction usable as a basis for heuristic functions.

**Definition 3** *Let* $\Pi$ *be a planning task over states* $S$*, and let* $\{\langle \mathcal{T}_i, \alpha_i \rangle\}_{i=1}^{m}$ *be an additive abstraction of the transition graph* $\mathcal{T}(\Pi)$*. If* $m = O(poly(||\Pi||))$ *and, for all states* $s \in S$ *and all* $1 \leq i \leq m$*, the cost* $cost(\alpha_i(s), S_i^{\star})$ *in* $\mathcal{T}_i$ *is computable in time* $O(poly(||\Pi||))$*, then* $h^{\mathcal{A}}(s) = \sum_{i=1}^{m} cost(\alpha_i(s), S_i^{\star})$ *is an* **abstraction heuristic** *function for* $\Pi$.

Note that admissibility of $h^{\mathcal{A}}$ is implied by the cost conservation condition of Eq. 1. To further illustrate the connection between abstractions and admissible heuristics, consider three well-known mechanisms for devising admissible planning heuristics: delete relaxation (Bonet & Geffner, 2001), critical-path relaxation (Haslum & Geffner, 2000),[2] and pattern database heuristics (Edelkamp, 2001).

First, while typically not considered this way, the delete relaxation of a planning task $\Pi = \langle V, A, I, G, cost \rangle$ does correspond to an abstraction $\langle \mathcal{T}_+ = (S_+, L_+, Tr_+, s_+^0, S_+^{\star}, \varpi_+), \alpha_+ \rangle$ of the transition graph $\mathcal{T}(\Pi)$. Assuming unique naming of the variable values in $\Pi$ and denoting $\mathcal{D}_+ = \bigcup_{v \in V} \mathcal{D}(v)$, we have the abstract states $S_+$ being the power-set of $\mathcal{D}_+$, and the labels $L_+ = \{a, a_+ \mid a \in A\}$. The transitions come from two sources: for each abstract state $s_+ \in S_+$ and each original action $a \in A$ applicable in $s_+$, we have both $(s_+, a, s_+[\![a]\!]) \in Tr_+$ and $(s_+, a_+, s_+ \cup \mathsf{eff}(a)) \in Tr_+$. With a minor abuse of notation, the initial state and the goal states of the abstraction are $s_+^0 = I$ and $S_+^{\star} = \{s_+ \in S_+ \mid s_+ \supseteq G\}$, and the abstraction mapping $\alpha_+$ is simply the identity function. It is easy to show that, for any state $s$ of our planning task $\Pi$, we have $cost(\alpha_+(s), S_+^{\star}) = h^+(s)$, where $h^+(s)$ is the delete-relaxation estimate of the cost from $s$ to the goal. As an aside, we note that this "delete-relaxation abstraction" $\langle \mathcal{T}_+, \alpha_+ \rangle$ in particular exemplifies that nothing in Definition 2 requires the size of the abstract state space to be limited by the size of the original state space. In any event, however, the abstraction $\langle \mathcal{T}_+, \alpha_+ \rangle$ does *not* induce a heuristic in terms of Definition 3 because computing $h^+(s)$ is known to be NP-hard (Bylander, 1994).

The situation for critical-path relaxation is exactly the opposite. While computing the corresponding family of admissible estimates $h^m$ is polynomial-time for any fixed $m$, this computation is not based on computing the shortest paths in an abstraction of the planning task. The state graph over which $h^m$ is computed is an AND/OR-graph (and not an OR-graph such as transition graphs), and the actual computation of $h^m$ corresponds to computing a critical tree (and not a shortest path) to the goal. To the best of our knowledge, the precise relation between critical path and abstraction heuristics is currently an open question (Helmert & Domshlak, 2009).

Overall, the only abstraction heuristics in the toolbox of planning these days appear to be the *explicit homomorphism abstractions*, whose best-known representative is probably

2. We assume the reader is familiar with these two relaxations. If not, their discussion here can be safely skipped.

the *pattern database (PDB)* heuristics. Given a planning task $\Pi$ over state variables $V$, a PDB heuristic is based on projecting $\Pi$ onto a subset of its variables $V^\alpha \subseteq V$. Such a homomorphism abstraction $\alpha$ maps two states $s_1, s_2 \in S$ into the same abstract state iff $s_1[V^\alpha] = s_2[V^\alpha]$. Inspired by the (similarly named) domain-specific heuristics for search problems such as $(n^2-1)$-puzzles or Rubik's Cube (Culberson & Schaeffer, 1998; Hernadvölgyi & Holte, 1999; Felner et al., 2004), PDB heuristics have been successfully exploited in domain-independent planning as well (Edelkamp, 2001, 2002; Haslum et al., 2007). The key decision in constructing PDBs is what sets of variables the problem is projected to (Edelkamp, 2006; Haslum et al., 2007). However, apart from that need to automatically select good projections, the two limitations of PDB heuristics are the size of the abstract space $S^\alpha$ and its dimensionality. First, the number of abstract states should be small enough to allow reachability analysis in $S^\alpha$ by exhaustive search. Moreover, an $O(1)$ bound on $|S^\alpha|$ is typically set explicitly to fit the time and memory limitations of the system. Second, since PDB abstractions are projections, the explicit constraint on $|S^\alpha|$ implies a fixed-dimensionality constraint $|V^\alpha| = O(1)$. In planning tasks with, informally, many alternative resources, this limitation is a pitfall. For instance, suppose $\{\Pi_i\}_{i=1}^\infty$ is a sequence of Logistics problems of growing size with $|V_i| = i$. If each package in $\Pi_i$ can be transported by some $\Theta(i)$ vehicles, then starting from some $i$, $h^\alpha$ will not account *at all* for movements of vehicles essential for solving $\Pi_i$ (Helmert & Mattmüller, 2008).

Aiming at preserving the attractiveness of the PDB heuristic while eliminating the bottleneck of fixed dimensionality, Helmert et al. (2007) have generalized the methodology of Dräger, Finkbeiner, and Podelski (2006) and introduced the so called *merge-and-shrink (MS)* abstractions for planning. MS abstractions are homomorphisms that generalize PDB abstractions by allowing for more flexibility in selection of pairs of states to be contracted. The problem's state space is viewed as the synchronized product of its projections onto the single state variables. Starting with all such "atomic" abstractions, this product can be computed by iteratively composing two abstract spaces, replacing them with their product. While in a PDB the size of the abstract space $S^\alpha$ is controlled by limiting the number of product compositions, in MS abstractions it is controlled by interleaving the iterative composition of projections with abstraction of the partial composites. Helmert et al. (2007) have proposed a concrete strategy for this interleaved abstraction/refinement scheme and empirically demonstrated the power of the merge-and-shrink abstraction heuristics. Like PDBs, however, MS abstractions are explicit abstractions, and thus computing their heuristic values is also based on explicitly searching for optimal plans in the abstract spaces. Hence, while merge-and-shrink abstractions escape the fixed-dimensionality constraint of PDBs, the constraint on the abstract space to be of a fixed size still holds.

## 4. Implicit Abstractions

Focusing on the $O(1)$ bound posted by explicit abstractions on the size of the abstract space, our first observation is that explicit abstractions are not necessarily the only way to proceed with abstraction heuristics. Given a planning task $\Pi$ over states $S$, suppose we can transform it into a different planning task $\Pi^\alpha$ such that

1. the transformation induces an abstraction mapping $\alpha : S \to S^\alpha$ where $S^\alpha$ is the state space of $\Pi^\alpha$, and

2. both the transformation of $\Pi$ to $\Pi^\alpha$, as well as computing $\alpha$ for any state in $S$, can be done in time polynomial in $||\Pi||$.

Having such planning-task-to-planning-task transformations in mind, we define what we call (additive) *implicit abstractions*.

**Definition 4** *An* **additive implicit abstraction** *of a planning task* $\Pi$ *is a set of pairs* $\mathcal{A} = \{\langle \Pi_i, \alpha_i \rangle\}_{i=1}^m$ *such that* $\{\Pi_i\}_{i=1}^m$ *are some planning tasks and* $\{\langle \mathcal{T}(\Pi_i), \alpha_i \rangle\}_{i=1}^m$ *is an additive abstraction of* $\mathcal{T}(\Pi)$.

Let us now examine the notion of implicit abstractions more closely. First, implicit abstractions allow for a natural additive combination of admissible heuristics for the abstract tasks. This composition is formulated below by Theorem 1, extending the original criterion for admissibility of additive heuristics described in Section 2. Second, as formulated by Theorem 2, implicit abstractions can be composed via the functional composition of their abstraction mappings. These two easy-to-prove properties of implicit abstractions allow us then to take the desired step from implicit abstractions to implicit abstraction heuristics.

**Theorem 1 (Admissibility)** *Let* $\Pi$ *be a planning task and* $\mathcal{A} = \{\langle \Pi_i, \alpha_i \rangle\}_{i=1}^m$ *be an additive implicit abstraction of* $\Pi$. *If, for each* $1 \leq i \leq m$, $h_i$ *is an admissible heuristic for* $\Pi_i$, *then the function* $h(s) = \sum_{i=1}^m h_i(\alpha_i(s))$ *is an admissible heuristic for* $\Pi$.

**Proof:** The proof is straightforward. Let $\mathcal{T} = (S, L, Tr, s^0, S^\star, \varpi)$ be the transition graph of $\Pi$, and let $s$ be some state in $S$. For each $1 \leq i \leq m$, let $\mathcal{T}_i = (S_i, L_i, Tr_i, s_i^0, S_i^\star, \varpi_i)$ be the transition graph of $\Pi_i$.

First, if $h_i$ is an admissible heuristic for $\Pi_i$, then for all $s_i \in S_i^\star$,

$$h_i(\alpha_i(s)) \leq cost(\alpha_i(s), s_i).$$

Now, for each state $s' \in S^\star$, from Definition 2 we have $\alpha_i(s') \in S_i^\star$, and from Eq. 1 we have

$$\sum_{i=1}^m cost(\alpha_i(s), \alpha_i(s')) \leq cost(s, s'),$$

and thus

$$h(s) = \sum_{i=1}^m h_i(\alpha_i(s)) \leq \sum_{i=1}^m cost(\alpha_i(s), \alpha_i(s')) \leq cost(s, s'),$$

giving us an admissible estimate for $h^*(s)$. □

**Theorem 2 (Composition)** *Let* $\Pi$ *be a planning task and* $\mathcal{A} = \{\langle \Pi_i, \alpha_i \rangle\}_{i=1}^m$ *be an additive implicit abstraction of* $\Pi$. *If, for each* $1 \leq i \leq m$, $\mathcal{A}_i = \{\langle \Pi_{i,j}, \alpha_{i,j} \rangle\}_{j=1}^{m_i}$ *is an additive implicit abstraction of* $\Pi_i$, *then* $\mathcal{A}' = \bigcup_{i=1}^m \{\langle \Pi_{i,j}, \alpha_{i,j} \circ \alpha_i \rangle\}_{j=1}^{m_i}$ *is an additive implicit abstraction of* $\Pi$.

**Proof:** Let $\mathcal{T} = (S, L, Tr, s^0, S^\star, \varpi)$ be the transition graph of $\Pi$. For each $1 \leq i \leq m$, let $\mathcal{T}_i = (S_i, L_i, Tr_i, s_i^0, S_i^\star, \varpi_i)$ be the transition graph of $\Pi_i$, and for each $1 \leq j \leq m_i$, let $\mathcal{T}_{i,j} = (S_{i,j}, L_{i,j}, Tr_{i,j}, s_{i,j}^0, S_{i,j}^\star, \varpi_{i,j})$ be the transition graph of $\Pi_{i,j}$. We need to show that $\alpha_{i,j} \circ \alpha_i$ is an abstraction mapping as in Definition 2. From $\alpha_i$ and $\alpha_{i,j}$ being abstraction mappings, we have

- $s^0_{i,j} = \alpha_{i,j}(s^0_i) = \alpha_{i,j}(\alpha_i(s^0)) = \alpha_{i,j} \circ \alpha_i(s^0)$,
- for all $s \in S^\star$ we have $\alpha_i(s) \in S^\star_i$ and thus $\alpha_{i,j}(\alpha_i(s)) = \alpha_{i,j} \circ \alpha_i(s) \in S^\star_{i,j}$, and
- for all $s_i, s'_i \in S_i$, $cost(s_i, s'_i) \geq \sum_{j=1}^{m_i} cost(\alpha_{i,j}(s_i), \alpha_{i,j}(s'_i))$, and thus for all $s, s' \in S$,

$$\begin{aligned} cost(s,s') \geq \sum_{i=1}^{m} cost(\alpha_i(s), \alpha_i(s')) &\geq \sum_{i=1}^{m}\sum_{j=1}^{m_i} cost(\alpha_{i,j}(\alpha_i(s)), \alpha_{i,j}(\alpha_i(s'))) \\ &= \sum_{i=1}^{m}\sum_{j=1}^{m_i} cost(\alpha_{i,j} \circ \alpha_i(s), \alpha_{i,j} \circ \alpha_i(s')). \end{aligned}$$

□

Together, Theorems 1 and 2 suggest the following scheme for deriving abstraction heuristics. Given an additive implicit abstraction $\mathcal{A} = \{\langle \Pi_i, \alpha_i \rangle\}_{i=1}^m$, if all its individual abstract tasks belong to some *tractable fragments of optimal planning*, then we can use in practice the (sum of the) true costs in all $\Pi_i$ as the admissible estimates for the costs in $\Pi$. Otherwise, if optimal planning for some abstract tasks $\Pi_i$ in $\mathcal{A}$ cannot be proven polynomial-time solvable, then we can further abstract these tasks, obtaining admissible estimates for the true costs in $\Pi_i$.

**Definition 5** *Let $\Pi$ be a planning task over states $S$, and let $\mathcal{A} = \{\langle \Pi_i, \alpha_i \rangle\}_{i=1}^m$ be an additive implicit abstraction of $\Pi$. If $m = O(poly(||\Pi||))$, and, for all states $s \in S$ and all $1 \leq i \leq m$, $h^*(\alpha_i(s))$ is polynomial-time computable, then $h^{\mathcal{A}}(s) = \sum_{i=1}^m h^*(\alpha_i(s))$ is an* **implicit abstraction heuristic** *function for $\Pi$.*

Compared to explicit abstraction heuristics such as PDB heuristics and merge-and-shrink heuristics, the direction of implicit abstraction heuristics is, at least in principle, appealing because neither the dimensionality nor even the size of the state spaces induced by implicit abstractions are required to be bounded by something restrictive, if at all. The pitfall, however, is that implicit abstraction heuristics correspond to tractable fragments of optimal planning, and the palette of such known fragments is extremely limited (Bäckström & Nebel, 1995; Bylander, 1994; Jonsson & Bäckström, 1998; Jonsson, 2007; Katz & Domshlak, 2007b). In fact, none so far has appeared to us very convenient for automatically devising useful problem transformations as above. Fortunately, we show next that the boundaries of tractability can be expanded in the right way, allowing us to successfully materialize the idea of implicit abstraction heuristics.

In the following, a key role is played by the causal graphs induced by the planning tasks. Informally, the basic idea behind what we call *causal-graph decompositions* is to abstract the given planning task $\Pi$ along a subgraph of $\Pi$'s causal graph, with the goal of obtaining abstract problems of *specific structure*. Naturally, there are numerous possibilities for obtaining such structure-oriented abstractions. We now present one such decomposition that is tailored to abstractions around *acyclic subgraphs*. Informally, this decomposition can be seen as a sequential application of two kinds of task transformations: dropping preconditions (Pearl, 1984) and (certain form of) breaking actions with conjunctive effects into single-effect actions.

**Definition 6** *Let* $\Pi = \langle V, A, I, G, cost\rangle$ *be a planning task, and let* $\mathcal{G} = (V_{\mathcal{G}}, E_{\mathcal{G}})$ *be an acyclic subgraph of the causal graph* $CG(\Pi)$*. A planning task* $\Pi_{\mathcal{G}} = \langle V_{\mathcal{G}}, A_{\mathcal{G}}, I_{\mathcal{G}}, G_{\mathcal{G}}, cost_{\mathcal{G}}\rangle$ *is an* **acyclic causal-graph decomposition** *of* $\Pi$ *with respect to* $\mathcal{G}$ *if*

1. $I_{\mathcal{G}} = I[V_{\mathcal{G}}]$, $G_{\mathcal{G}} = G[V_{\mathcal{G}}]$,
2. $A_{\mathcal{G}} = \bigcup_{a\in A} A_{\mathcal{G}}(a)$ *where each* $A_{\mathcal{G}}(a) = \{a^1, \dots, a^{l(a)}\}$ *is a set of actions over* $V_{\mathcal{G}}$ *such that, for a topological with respect to* $\mathcal{G}$ *ordering of the variables* $\{v_1, \dots, v_{l(a)}\} = \mathcal{V}(\mathsf{eff}(a)) \cap V_{\mathcal{G}}$*, and* $1 \le i \le l(a)$,

$$\mathsf{eff}(a^i)[v] = \begin{cases} \mathsf{eff}(a)[v], & v = v_i \\ \text{unspecified}, & \text{otherwise} \end{cases}$$
$$\mathsf{pre}(a^i)[v] = \begin{cases} \mathsf{pre}(a)[v], & (v, v_i) \in E_{\mathcal{G}} \wedge v \notin \mathcal{V}(\mathsf{eff}(a)) \text{ or } v = v_i \\ \mathsf{eff}(a)[v], & (v, v_i) \in E_{\mathcal{G}} \wedge v \in \mathcal{V}(\mathsf{eff}(a)) \\ \text{unspecified}, & \text{otherwise} \end{cases} \tag{2}$$

3. *For each action* $a \in A$,

$$\sum_{a' \in A_{\mathcal{G}}(a)} cost_{\mathcal{G}}(a') \le cost(a). \tag{3}$$

It is not hard to verify from Definition 6 that for any planning task $\Pi$ and any acyclic causal-graph decomposition $\Pi_{\mathcal{G}}$ of $\Pi$, the causal graph $CG(\Pi_{\mathcal{G}})$ is exactly the subgraph $\mathcal{G}$ underlying the decomposition. To illustrate the notion of acyclic causal-graph decomposition, we consider a planning task $\Pi = \langle V, A, I, G, cost\rangle$ over five state variables $V = \{u, v, x, y, z\}$, two unit-cost actions $A = \{a_1, a_2\}$ as in Figure 2a, initial state $I = \{u{:}0, v{:}0, x{:}0, y{:}0, z{:}0\}$, and goal $G = \{u{:}1, v{:}1, x{:}0, y{:}1, z{:}1\}$. The causal graph $CG(\Pi)$ is depicted in Figure 2a. Figures 2b-c show two subgraphs $\mathcal{G}_1$ and $\mathcal{G}_2$ of $CG(\Pi)$, respectively, as well as the action sets $A_{\mathcal{G}_1}(a_1) = \{a_1^1, a_1^2, a_1^3\}$ and $A_{\mathcal{G}_1}(a_2) = \{a_2^1, a_2^2, a_2^3\}$ in Figure 2(b), and the action sets $A_{\mathcal{G}_2}(a_1) = \{a_1^1, a_1^2, a_1^3\}$ and $A_{\mathcal{G}_2}(a_2) = \{a_2^1, a_2^2, a_2^3\}$ in Figure 2(c). For $i \in \{1, 2\}$, let $\Pi_i = \langle V, A_i, I, G, cost_i\rangle$ be the planning task with $A_i = A_{\mathcal{G}_i}(a_1) \cup A_{\mathcal{G}_i}(a_2)$ and $cost_i(a) = 1/3$ for all $a \in A_i$. These two planning tasks $\Pi_i$ (individually) satisfy the conditions of Definition 6 with respect to $\Pi$ and $\mathcal{G}_i$, and thus they are acyclic causal-graph decompositions of $\Pi$ with respect to $\mathcal{G}_i$.

We now proceed with specifying implicit abstractions defined via acyclic causal-graph decompositions.

**Definition 7** *Let* $\Pi = \langle V, A, I, G, cost\rangle$ *be a planning task over states* $S$*, and let* $\mathbf{G} = \{\mathcal{G}_i = (V_{\mathcal{G}_i}, E_{\mathcal{G}_i})\}_{i=1}^m$ *be a set of acyclic subgraphs of the causal graph* $CG(\Pi)$*.* $\mathcal{A} = \{\langle \Pi_{\mathcal{G}_i}, \alpha_i\rangle\}_{i=1}^m$ *is an* **acyclic causal-graph abstraction** *of* $\Pi$ *over* $\mathbf{G}$ *if, for some set of cost functions* $\{cost_i : A \to \mathbb{R}^{0+}\}_{i=1}^m$ *satisfying*

$$\forall a \in A: \quad \sum_{i=1}^m cost_i(a) \le cost(a), \tag{4}$$

*we have, for* $1 \le i \le m$,

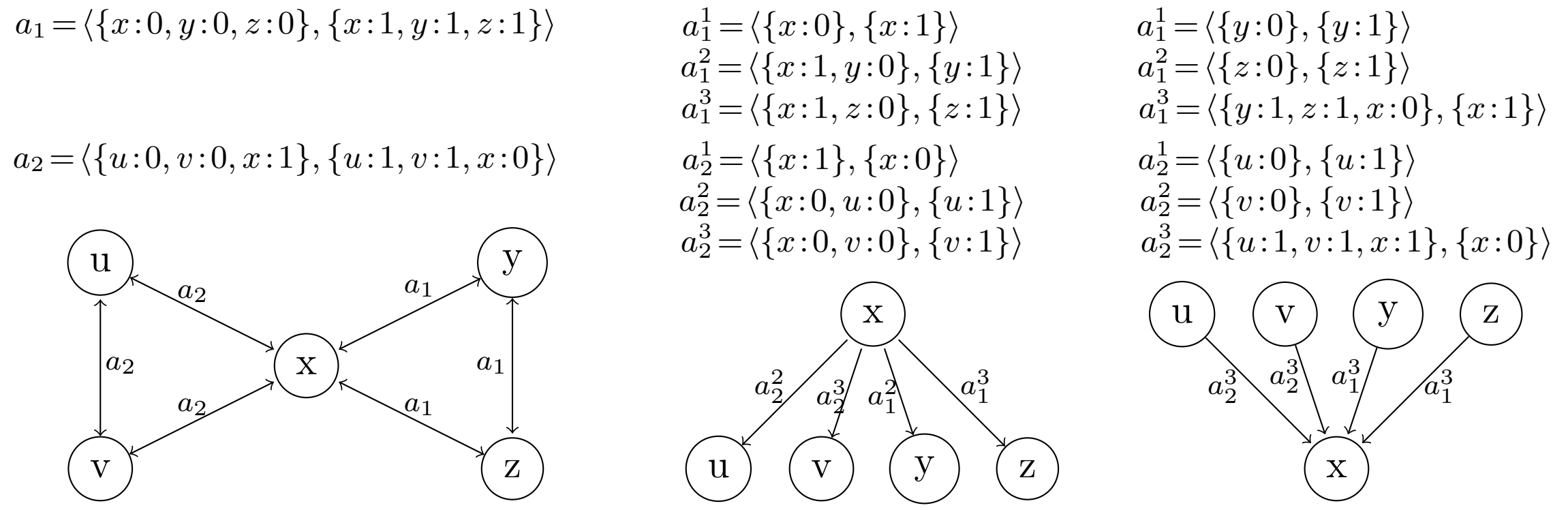

(a) (b) (c)

Figure 2: (a) The actions and causal graph $CG(\Pi)$ of the planning graph in the example illustrating Definition 2. (b) Subgraph $\mathcal{G}_1$ of $CG(\Pi)$ and the induced action sets $A_{\mathcal{G}_1}(a_1)$ and $A_{\mathcal{G}_1}(a_2)$. (c) Subgraph $\mathcal{G}_2$ of $CG(\Pi)$ and the induced action sets $A_{\mathcal{G}_2}(a_1)$ and $A_{\mathcal{G}_2}(a_2)$. The arcs of both $CG(\Pi)$ and its subgraphs $\mathcal{G}_1$ and $\mathcal{G}_2$ are labeled with the actions inducing the arcs.

- $\Pi_{\mathcal{G}_i} = \langle V_{\mathcal{G}_i}, A_{\mathcal{G}_i}, I_{\mathcal{G}_i}, G_{\mathcal{G}_i}, cost_{\mathcal{G}_i}\rangle$ *is an acyclic causal-graph decomposition of* $\Pi_i = \langle V, A, I, G, cost_i\rangle$ *with respect to* $\mathcal{G}_i$, *and*
- *the abstraction mapping* $\alpha_i : S \to S_i$ *is the projection mapping* $\alpha_i(s) = s[V_{\mathcal{G}_i}]$.

**Theorem 3** *Acyclic causal-graph abstractions of the planning tasks are additive implicit abstractions of these tasks.*

**Proof:** Let $\Pi = \langle V, A, I, G, cost\rangle$ be a planning task, and let $\mathcal{A} = \{\langle\Pi_{\mathcal{G}_i}, \alpha_i\rangle\}_{i=1}^m$ be an acyclic causal-graph abstraction of $\Pi$ over a set of subgraphs $\mathbf{G} = \{\mathcal{G}_i = (V_{\mathcal{G}_i}, E_{\mathcal{G}_i})\}_{i=1}^m$. Let $\mathcal{T} = (S, L, Tr, s^0, S^\star, \varpi)$ be the transition graph of $\Pi$, and, for $1 \leq i \leq m$, $\mathcal{T}_i = (S_i, L_i, Tr_i, s_i^0, S_i^\star, \varpi_i)$ be the transition graph of $\Pi_{\mathcal{G}_i}$. We need to show that $\alpha_i$ is an abstraction mapping as in Definition 2.

First, from Definitions 6 and 7, we have

- $s_i^0 = I_{\mathcal{G}_i} = I[V_{\mathcal{G}_i}] = s^0[V_{\mathcal{G}_i}] = \alpha_i(s^0)$, and
- for all $s \in S^\star$ we have $s \supseteq G$ and thus $\alpha_i(s) = s[V_{\mathcal{G}_i}] \supseteq G[V_{\mathcal{G}_i}] = G_{\mathcal{G}_i}$, providing us with $\alpha_i(s) \in S_i^\star$.

Now, if $s$ is a state of $\Pi$ and $a \in A$ is an action with $\mathsf{pre}(a) \subseteq s$, then $\alpha_i(s)$ is a state of $\Pi_{\mathcal{G}_i}$ and $\mathsf{pre}(a)[V_{\mathcal{G}_i}] \subseteq \alpha_i(s)$. Let the action sequence $\rho = \langle a^1, a^2, \ldots, a^{l(a)}\rangle$ be constructed from $a$ as in Eq. 2. We inductively prove that $\rho$ is applicable in $\alpha_i(s)$. First, for each $v \in V_{\mathcal{G}_i}$, either $\mathsf{pre}(a^1)[v] = \mathsf{pre}(a)[v]$, or $\mathsf{pre}(a^1)[v]$ is unspecified, and thus $\rho_1 = \langle a^1\rangle$ is applicable in $\alpha_i(s)$. The inductive hypothesis is now that $\rho_j = \langle a^1, a^2, \ldots, a^j\rangle$ is applicable in $\alpha_i(s)$, and let $s' = \alpha_i(s)[\![\rho_j]\!]$. From Eq. 2, for each $1 \leq j' \leq j$, $a^{j'}$ changes the value of $v_{j'}$ to $\mathsf{eff}(a)[v_{j'}]$, and that is the only change of $v_{j'}$ along $\rho_j$. Likewise, since all the actions constructed as in Eq. 2 are unary-effect, $\{v_1, \ldots, v_j\}$ are the only variables in $V_{\mathcal{G}_i}$ affected along $\rho_j$. Hence, for all $v \in V_{\mathcal{G}_i}$, if $v = v_{j'}$, $1 \le j' \le j$, then $s'[v] = \mathsf{eff}(a)[v] = \mathsf{pre}(a^{j+1})[v]$, and otherwise, $s'[v] = \alpha_i(s)[v]$, and if $\mathsf{pre}(a^{j+1})[v]$ is specified, then $\mathsf{pre}(a^{j+1})[v] = \mathsf{pre}(a)[v] = \alpha_i(s)[v]$. This implies that $a^{j+1}$ is applicable in $s'$ and, as a result, $\rho_{j+1} = \langle a^1, a^2, \ldots, a^{j+1}\rangle$ is applicable in $\alpha_i(s)$, finalizing the inductive proof. Likewise, exactly the same arguments on the affect of $\{a^j\}_{j=1}^{l(a)}$ on $\alpha_i(s)$ immediately imply that, if $\rho = \langle a^1, a^2, \ldots, a^{l(a)}\rangle$, then $\alpha_i(s[\![a]\!]) = \alpha_i(s)[\![\rho]\!]$.

Next, for each $a \in A$, from Eqs. 3 and 4 we have

$$\sum_{i=1}^{m} \sum_{a' \in A_{\mathcal{G}_i}(a)} cost_{\mathcal{G}_i}(a') \le \sum_{i=1}^{m} cost_i(a) \le cost(a). \tag{5}$$

Now, let $s, s' \in S$ be a pair of original states such that $cost(s, s') < \infty$, and let $\varrho = \langle a_1, \ldots, a_k\rangle$ be the sequence of labels along a cheapest path from $s$ to $s'$ in $\mathcal{T}$. From that, $cost(s, s') = cost(\varrho) = \sum_{j=1}^{k} cost(a_j)$. The decomposition of such a path to the sequences of actions as in Eq. 2 is a (not neccesarily cheapest) path from $\alpha_i(s)$ to $\alpha_i(s')$ in $\mathcal{T}_i$, and thus $cost(\alpha_i(s), \alpha_i(s')) \le \sum_{j=1}^{k} \sum_{a' \in A_{\mathcal{G}_i}(a_j)} cost_{\mathcal{G}_i}(a')$, providing us with

$$\begin{aligned}\sum_{i=1}^{m} cost(\alpha_i(s), \alpha_i(s')) &\le \sum_{i=1}^{m} \sum_{j=1}^{k} \sum_{a' \in A_{\mathcal{G}_i}(a_j)} cost_{\mathcal{G}_i}(a') = \sum_{j=1}^{k} \sum_{i=1}^{m} \sum_{a' \in A_{\mathcal{G}_i}(a_j)} cost_{\mathcal{G}_i}(a') \\ &\overset{(5)}{\le} \sum_{j=1}^{k} cost(a_j) = cost(s, s').\end{aligned}$$

□

Thus, if we can decompose the given task $\Pi$ into a set of tractable acyclic causal-graph decompositions $\mathbf{\Pi} = \{\Pi_{\mathcal{G}_1}, \ldots, \Pi_{\mathcal{G}_m}\}$, then we can solve all these tasks in polynomial time, and derive an additive admissible heuristic for $\Pi$. Before we proceed with considering concrete acyclic causal-graph decomposition, note that Definition 2 leaves the decision about the actual partition of the action costs rather open. In what follows we adopt the most straightforward, *uniform* action cost partition in which the cost of each action $a$ is *equally split* among all the non-redundant representatives of $a$ in $\bigcup_{i=1}^{m} A_{\mathcal{G}_i}(a)$. However, a better choice of action cost partition can sometimes be made. In fact, sometimes it can even be optimized (Katz & Domshlak, 2010)

## 5. Fork Decompositions

We now proceed with introducing two concrete acyclic causal-graph decompositions that, when combined with certain variable domain abstractions, provide us with implicit abstraction heuristics. These so called *fork-decomposition heuristics* are based on two novel fragments of tractable cost-optimal planning for tasks with fork and inverted-fork structured causal graphs.

**Definition 8** *For a planning task $\Pi$ over variables $V$, and a variable $v \in V$,*

*(1)* $v$-**fork** *of* $\Pi$ *is the subgraph* $\mathcal{G}^{\mathsf{f}}_v$ *of* $CG(\Pi)$ *over nodes* $V_{\mathcal{G}^{\mathsf{f}}_v} = \{v\} \cup \mathsf{succ}(v)$ *and edges* $E_{\mathcal{G}^{\mathsf{f}}_v} = \{(v, u) \mid u \in \mathsf{succ}(v)\}$*, and*

*(2)* $v$-**ifork** *(short for inverted fork) of* $\Pi$ *is a subgraph* $\mathcal{G}^{\mathsf{i}}_v$ *of* $CG(\Pi)$ *over nodes* $V_{\mathcal{G}^{\mathsf{i}}_v} = \{v\} \cup \mathsf{pred}(v)$ *and edges* $E_{\mathcal{G}^{\mathsf{i}}_v} = \{(u, v) \mid u \in \mathsf{pred}(v)\}$.

*The sets of all* $v$*-forks and all* $v$*-iforks of* $\Pi$ *are denoted by* $\mathbf{G}_{\mathcal{F}} = \{\mathcal{G}^{\mathsf{f}}_v\}_{v \in V}$ *and* $\mathbf{G}_{\mathcal{I}} = \{\mathcal{G}^{\mathsf{i}}_v\}_{v \in V}$*, respectively.*

For any planning task and each of its state variables $v$, both $v$-fork and $v$-ifork are acyclic digraphs, allowing us to define our three implicit abstractions as follows.

**Definition 9** *For any planning task* $\Pi = \langle V, A, I, G, cost\rangle$,

*(1) any acyclic causal-graph abstraction* $\mathcal{A}_{\mathcal{F}} = \{\langle \Pi^{\mathsf{f}}_v, \alpha^{\mathsf{f}}_v\rangle\}_{v \in V}$ *of* $\Pi$ *over* $\mathbf{G}_{\mathcal{F}}$ *is called* $\mathcal{F}$**-abstraction**, *and the set of abstract planning tasks* $\mathbf{\Pi}_{\mathcal{F}} = \{\Pi^{\mathsf{f}}_v\}_{v \in V}$ *is called* $\mathcal{F}$**-decomposition** *of* $\Pi$*;*

*(2) any acyclic causal-graph abstraction* $\mathcal{A}_{\mathcal{I}} = \{\langle \Pi^{\mathsf{i}}_v, \alpha^{\mathsf{i}}_v\rangle\}_{v \in V}$ *of* $\Pi$ *over* $\mathbf{G}_{\mathcal{I}}$ *is called* $\mathcal{I}$**-abstraction**, *and the set of abstract planning tasks* $\mathbf{\Pi}_{\mathcal{I}} = \{\Pi^{\mathsf{i}}_v\}_{v \in V}$ *is called* $\mathcal{I}$**-decomposition** *of* $\Pi$*;*

*(3) any acyclic causal-graph abstraction* $\mathcal{A}_{\mathcal{FI}} = \{\langle \Pi^{\mathsf{f}}_v, \alpha^{\mathsf{f}}_v\rangle, \langle \Pi^{\mathsf{i}}_v, \alpha^{\mathsf{i}}_v\rangle\}_{v \in V}$ *of* $\Pi$ *over* $\mathbf{G}_{\mathcal{FI}} = \mathbf{G}_{\mathcal{F}} \cup \mathbf{G}_{\mathcal{I}}$ *is called* $\mathcal{FI}$**-abstraction**, *and the set of abstract planning tasks* $\mathbf{\Pi}_{\mathcal{FI}} = \{\Pi^{\mathsf{f}}_v, \Pi^{\mathsf{i}}_v\}_{v \in V}$ *is called* $\mathcal{FI}$**-decomposition** *of* $\Pi$.

Definition 9 can be better understood by considering the $\mathcal{FI}$-abstraction of the problem $\Pi$ from our Logistics example; Figure 3 schematically illustrates the process. To simplify the example, here we as if eliminate from $\mathbf{G}_{\mathcal{FI}}$ all the single-node subgraphs, obtaining

$$\mathcal{A}_{\mathcal{FI}} = \{\langle \Pi^{\mathsf{f}}_{c_1}, \alpha^{\mathsf{f}}_{c_1}\rangle, \{\langle \Pi^{\mathsf{f}}_{c_2}, \alpha^{\mathsf{f}}_{c_2}\rangle, \{\langle \Pi^{\mathsf{f}}_{c_3}, \alpha^{\mathsf{f}}_{c_3}\rangle, \{\langle \Pi^{\mathsf{f}}_{t}, \alpha^{\mathsf{f}}_{t}\rangle, \{\langle \Pi^{\mathsf{i}}_{p_1}, \alpha^{\mathsf{i}}_{p_1}\rangle, \{\langle \Pi^{\mathsf{i}}_{p_2}, \alpha^{\mathsf{i}}_{p_2}\rangle\}.$$

Considering the action sets of the problems in $\mathbf{\Pi}_{\mathcal{FI}} = \{\Pi^{\mathsf{f}}_{c_1}, \Pi^{\mathsf{f}}_{c_2}, \Pi^{\mathsf{f}}_{c_3}, \Pi^{\mathsf{f}}_{t}, \Pi^{\mathsf{i}}_{p_1}, \Pi^{\mathsf{i}}_{p_2}\}$, we see that each original driving action has one nonredundant (that is, "changing some variable") representative in three of the abstract planning tasks, while each load/unload action has one nonredundant representative in five of these tasks. For instance, the action *drive-*$c_1$*-from-A-to-D* has one nonredundant representative in each of the tasks $\{\Pi^{\mathsf{f}}_{c_1}, \Pi^{\mathsf{i}}_{p_1}, \Pi^{\mathsf{i}}_{p_2}\}$, and the action *load-*$p_1$*-into-*$c_1$*-at-A* has one nonredundant representative in each of the tasks $\{\Pi^{\mathsf{f}}_{c_1}, \Pi^{\mathsf{f}}_{c_2}, \Pi^{\mathsf{f}}_{c_3}, \Pi^{\mathsf{f}}_{t}, \Pi^{\mathsf{i}}_{p_1}\}$. Since we assume a uniform partition of the action costs, the cost of each driving and load/unload action in each relevant abstract planning task is thus set to 1/3 and 1/5, respectively. From Theorem 3 we have $\mathcal{A}_{\mathcal{FI}}$ being an additive implicit abstraction of $\Pi$, and from Theorem 1 we then have

$$h^{\mathcal{FI}} = \sum_{v \in V} \left( h^*_{\Pi^{\mathsf{f}}_v} + h^*_{\Pi^{\mathsf{i}}_v} \right), \tag{6}$$

being an admissible estimate of $h^*$ in $\Pi$. The question now is how good this estimate is. The optimal cost of solving our running example is 19. Taking as a reference the well-known admissible heuristics $h_{\max}$ (Bonet & Geffner, 2001) and $h^2$ (Haslum & Geffner, 2000), we have $h_{\max}(I) = 8$ and $h^2(I) = 13$. Considering our $\mathcal{FI}$-abstraction, the optimal plans for the tasks in $\mathbf{\Pi}_{\mathcal{FI}}$ are as follows.

$\Pi^{\mathsf{f}}_{c_1}$ : *load-$p_1$-into-$c_2$-at-C, unload-$p_1$-from-$c_2$-at-D, load-$p_1$-into-t-at-D, unload-$p_1$-from-t-at-E, load-$p_1$-into-$c_3$-at-E, unload-$p_1$-from-$c_3$-at-G, load-$p_2$-into-$c_3$-at-F, unload-$p_2$-from-$c_3$-at-E.*

$\Pi^{\mathsf{f}}_{c_2}$ : *load-$p_1$-into-$c_1$-at-C, unload-$p_1$-from-$c_1$-at-D, load-$p_1$-into-t-at-D, unload-$p_1$-from-t-at-E, load-$p_1$-into-$c_3$-at-E, unload-$p_1$-from-$c_3$-at-G, load-$p_2$-into-$c_3$-at-F, unload-$p_2$-from-$c_3$-at-E.*

$\Pi^{\mathsf{f}}_{c_3}$ : *load-$p_1$-into-$c_1$-at-C, unload-$p_1$-from-$c_1$-at-D, load-$p_1$-into-t-at-D, unload-$p_1$-from-t-at-E, drive-$c_3$-from-G-to-E, load-$p_1$-into-$c_3$-at-E, drive-$c_3$-from-E-to-G, unload-$p_1$-from-$c_3$-at-G, drive-$c_3$-from-G-to-E, drive-$c_3$-from-E-to-F, load-$p_2$-into-$c_3$-at-F, drive-$c_3$-from-F-to-E, unload-$p_2$-from-$c_3$-at-E, drive-$c_3$-from-E-to-F.*

$\Pi^{\mathsf{f}}_{t}$ : *load-$p_1$-into-$c_1$-at-C, unload-$p_1$-from-$c_1$-at-D, drive-t-from-E-to-D, load-$p_1$-into-t-at-D, drive-t-from-D-to-E, unload-$p_1$-from-t-at-E, load-$p_1$-into-$c_3$-at-E, unload-$p_1$-from-$c_3$-at-G, load-$p_2$-into-$c_3$-at-F, unload-$p_2$-from-$c_3$-at-E.*

$\Pi^{\mathsf{i}}_{p_1}$ : *drive-$c_1$-from-A-to-D, drive-$c_1$-from-D-to-C, load-$p_1$-into-$c_1$-at-C, drive-$c_1$-from-C-to-D, unload-$p_1$-from-$c_1$-at-D, drive-t-from-E-to-D, load-$p_1$-into-t-at-D, drive-t-from-D-to-E, unload-$p_1$-from-t-at-E, drive-$c_3$-from-G-to-E, load-$p_1$-into-$c_3$-at-E, drive-$c_3$-from-E-to-G, unload-$p_1$-from-$c_3$-at-G, drive-$c_3$-from-G-to-E, drive-$c_3$-from-E-to-F.*

$\Pi^{\mathsf{i}}_{p_2}$ : *drive-$c_3$-from-G-to-E, drive-$c_3$-from-E-to-F, load-$p_2$-into-$c_3$-at-F, drive-$c_3$-from-F-to-E, unload-$p_2$-from-$c_3$-at-E, drive-$c_3$-from-E-to-F.*

Hence, we have

$$\begin{aligned} h^{\mathfrak{FI}} &= h^*_{\Pi^{\mathsf{f}}_{c_1}} + h^*_{\Pi^{\mathsf{f}}_{c_2}} + h^*_{\Pi^{\mathsf{f}}_{c_3}} + h^*_{\Pi^{\mathsf{f}}_{t}} + h^*_{\Pi^{\mathsf{i}}_{p_1}} + h^*_{\Pi^{\mathsf{f}}_{p_2}} \\ &= \tfrac{8}{5} + \tfrac{8}{5} + \tfrac{8}{5} + \tfrac{6}{3} + \tfrac{8}{5} + \tfrac{2}{3} + \tfrac{6}{5} + \tfrac{9}{3} + \tfrac{2}{5} + \tfrac{4}{3} = 15, \end{aligned} \tag{7}$$

and so $h^{\mathfrak{FI}}$ appears at least promising.

Unfortunately, despite the seeming simplicity of the planning tasks in $\mathbf{\Pi}_{\mathfrak{FI}}$, it turns out that implicit fork-decomposition abstractions as in Definitions 9 do not fit the requirements of implicit abstraction heuristics as in Definition 5. The causal graphs of the planning tasks in $\mathbf{\Pi}_{\mathfrak{F}}$ and $\mathbf{\Pi}_{\mathfrak{I}}$ form directed forks and directed inverted forks, respectively, and, in general, the number of variables in each such planning task can be as large as $\Theta(|V|)$. The problem is that even satisficing planning for SAS$^+$ fragments with fork and inverted fork causal graphs is NP-complete (Domshlak & Dinitz, 2001). In fact, recent results by Chen and Gimenez (2008) show that planning for any SAS$^+$ fragment characterized by any nontrivial form of causal graph is NP-hard. Moreover, even if the domain transition graphs of all the state variables are strongly connected (as in our example), optimal planning for fork and inverted fork structured problems remain NP-hard (see Helmert 2003, and 2004 for the respective results). Next, however, we show that this is not the end of the story for fork decompositions.

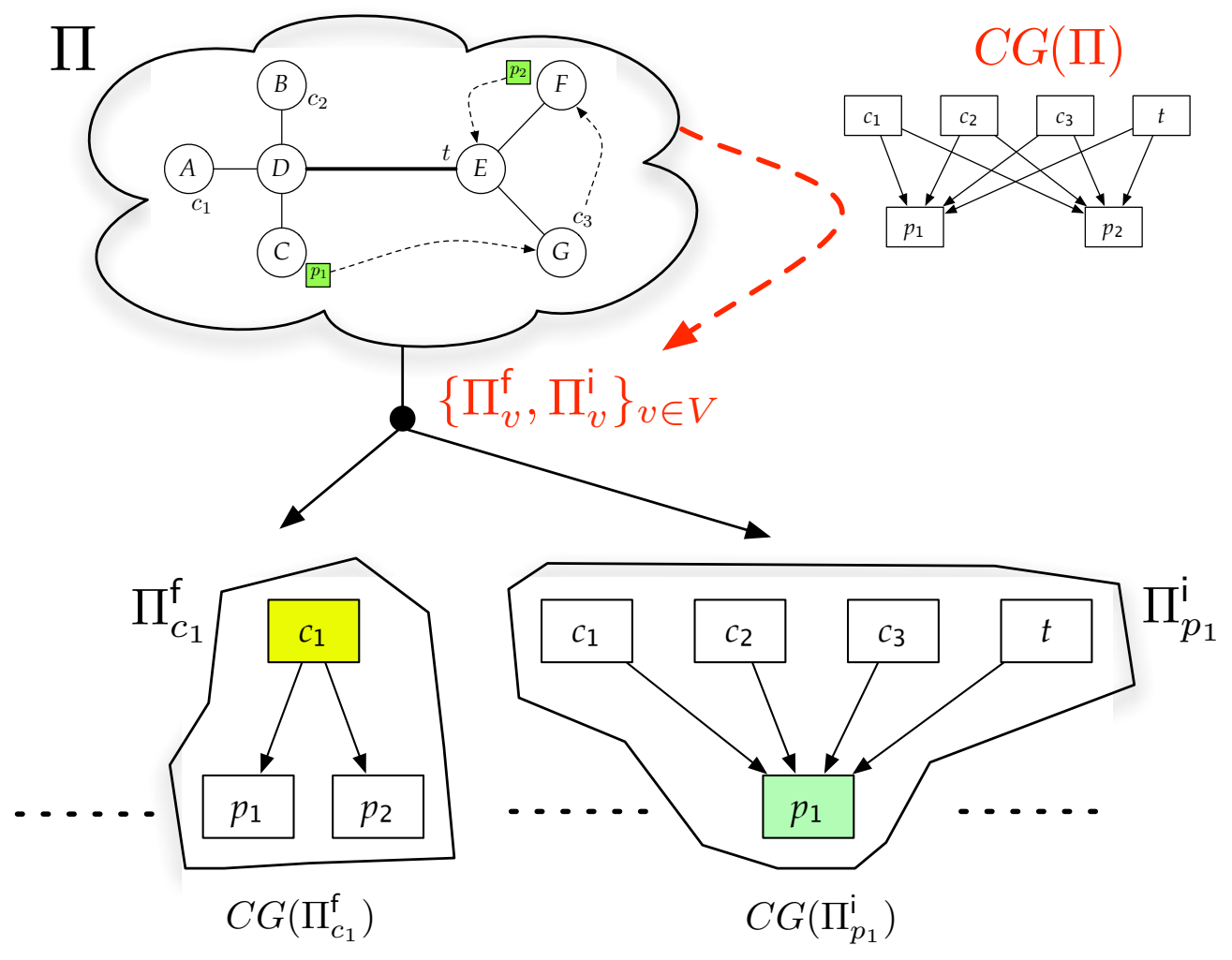

Figure 3: Schematic illustration of $\mathfrak{FI}$-decomposition for our running Logistics example

While the hardness of optimal planning for problems with fork and inverted fork causal graphs casts a shadow on the relevance of fork decompositions, a closer look at the proofs of the corresponding hardness results of Domshlak and Dinitz (2001) and Helmert (2003, 2004) reveals that they in particular rely on root variables having large domains. Exploiting this observation, we now show that this reliance is not incidental and characterize two substantial islands of tractability within the structural fragments of SAS$^+$.

**Theorem 4 (Tractable Forks)** *Given a planning task* $\Pi = \langle V, A, I, G, cost \rangle$ *with a fork causal graph rooted at* $r \in V$*, if* $|\mathcal{D}(r)| = 2$*, the time complexity of the cost-optimal planning for* $\Pi$ *is polynomial in* $||\Pi||$*.*

**Proof:** Observe that, for any planning task $\Pi$ as in the theorem, the fork structure of the causal graph $CG(\Pi)$ implies that all the actions in $\Pi$ are unary-effect, and each leaf variable $v \in \mathsf{succ}(r)$ preconditions only the actions affecting $v$ itself. The algorithm below is based on the following three properties satisfied by the optimal plans $\rho$ for $\Pi$.

(i) For any leaf variable $v \in \mathsf{succ}(r)$, the path $\rho_{\downarrow_v}$ from $I[v]$ to $G[v]$ induced by $\rho$ in $DTG(v, \Pi)$ is either cycle-free or contains only zero-cost cycles. This is the case because otherwise all the nonzero-cost cycles can be eliminated from $\rho_{\downarrow_v}$ while preserving its validity, violating the assumed optimality of $\rho$. Without loss of generality, in what follows we assume that this path $\rho_{\downarrow_v}$ in $DTG(v, \Pi)$ is cycle-free; in the case of fork causal graphs, we can always select an optimal $\rho$ that satisfies this requirement for all $v \in \mathsf{succ}(r)$. Thus, we have $|\rho_{\downarrow_v}| \leq |\mathcal{D}(v)| - 1$.

(ii) *Having fixed* a sequence of value changes of $r$, the fork's leaves become mutually independent; that is, our ability to change the value of one of them does not affect our ability to change the value of all the others.

(iii) *Because $r$ is binary-valued*, if $v \in V \setminus \{r\}$ is the "most demanding" leaf variable in terms of the number of value changes required from $r$ by the action preconditions along $\rho_{\downarrow v}$, then these are the only value changes of $r$ along $\rho$, except for, possibly, a final value change to $G[r]$. Thus, in particular, we have $|\rho_{\downarrow r}| \leq \max_{v \in \mathsf{succ}(r)} |\mathcal{D}(v)|$.

We begin with introducing some auxiliary notations. With $|\mathcal{D}(r)| = 2$, let $\mathcal{D}(r) = \{0, 1\}$ with $I[r] = 0$. Let $\sigma(r)$ be an alternating 0/1 sequence starting with 0, and having 0 in all odd and 1 in all even positions. This sequence $\sigma(r)$ is such that $|\sigma(r)| = 1$ if no action in $A$ can change $r$'s value to 1, $|\sigma(r)| = 2$ if some action can change $r$'s value to 1 but no action can then restore it to value 0, and otherwise, $|\sigma(r)| = 1 + \max_{v \in \mathsf{succ}(r)} |\mathcal{D}(v)|$. Let $\trianglerighteq[\sigma(r)]$ be the set of all nonempty prefixes of $\sigma(r)$ if $G[r]$ is unspecified; otherwise, let it be the set of all nonempty prefixes of $\sigma(r)$ ending with $G[r]$. Note that, if $\trianglerighteq[\sigma(r)] = \emptyset$, then the problem is trivially unsolvable; in what follows we assume this is not the case. For each $v \in \mathsf{succ}(r)$, let $DTG_v^0$ and $DTG_v^1$ be the subgraphs of the domain transition graphs $DTG(v, \Pi)$, obtained by removing from $DTG(v, \Pi)$ all the arcs labeled with $r:1$ and $r:0$, respectively.

The algorithm below incrementally constructs a set $\mathcal{R}$ of valid plans for $\Pi$, starting with $\mathcal{R} = \emptyset$.

(1) For each $v \in \mathsf{succ}(r)$, and each pair of $v$'s values $x, y \in \mathcal{D}(v)$, compute the cheapest (that is, cost-minimal) paths $\pi_v^0(x, y)$ and $\pi_v^1(x, y)$ from $x$ to $y$ in $DTG_v^0$ and $DTG_v^1$, respectively. For some pairs of values $x, y$, one or even both these paths may, of course, not exist.

(2) For each sequence $\sigma \in \trianglerighteq[\sigma(r)]$, and each $v \in \mathsf{succ}(r)$, construct a layered digraph $\mathcal{L}_v(\sigma)$ with $|\sigma| + 1$ node layers $L_0, \ldots, L_{|\sigma|}$, where $L_0$ consists of only $I[v]$, and for $1 \leq i \leq |\sigma|$, $L_i$ consists of all nodes $y \in \mathcal{D}(v)$ for which a path $\pi_v^{\sigma[i]}(x, y)$ from some node $x \in L_{i-1}$ has been constructed in step (1). For each $x \in L_{i-1}, y \in L_i$, $\mathcal{L}_v(\sigma)$ contains an arc $(x, y)$ weighted with $cost(\pi_v^{\sigma[i]}(x, y))$.

(3) For each $\sigma \in \trianglerighteq[\sigma(r)]$, let $k = |\sigma|$. A candidate plan $\rho_\sigma$ for $\Pi$ is constructed as follows.

   (a) For each $v \in \mathsf{succ}(r)$, find a cost-minimal path from $I[v]$ to $G[v]$ in $\mathcal{L}_v(\sigma)$. If no such path exists, then proceed with the next prefix in $\trianglerighteq[\sigma(r)]$. Otherwise, note that the $i$-th edge on this path (taking us from some $x \in L_{i-1}$ to some $y \in L_i$) corresponds to the cost-minimal path $\pi_v^{\sigma[i]}(x, y)$ from $x$ to $y$. Let us denote this path from $x$ to $y$ by $S_v^i$.

   (b) Set $\mathcal{R} = \mathcal{R} \cup \{\rho_\sigma\}$, where $\rho_\sigma = S^1 \cdot a_{\sigma[2]} \cdot S^2 \cdot \ldots \cdot a_{\sigma[k]} \cdot S^k$, each sequence $S^i$ is obtained by an *arbitrary* merge of the sequences $\{S_v^i\}_{v \in \mathsf{succ}(r)}$, and $a_\vartheta$ is the cheapest action changing the value of $r$ to value $\vartheta$.

(4) If $\mathcal{R} = \emptyset$, then fail, otherwise return $\rho = \mathrm{argmin}_{\rho_\sigma \in \mathcal{R}}\, cost(\rho_\sigma)$.

It is straightforward to verify that the complexity of the above procedure is polynomial in the description size of $\Pi$. To prove correctness, we show that the procedure returns a plan for any solvable task $\Pi$, and that the returned plan $\rho'$ satisfies $cost(\rho') \leq cost(\rho)$ for any optimal plan $\rho$ for $\Pi$.

Given a solvable task $\Pi$, let $\rho$ be an optimal plan for $\Pi$ with all $\rho_{\downarrow v}$ for the leaf variables $v$ being cycle-free. Let $\rho_{\downarrow r} = \langle a_2 \ldots, a_k \rangle$; the numbering of actions along $\rho_{\downarrow r}$ starts with $a_2$ to simplify indexing later on. For each $v \in \mathsf{succ}(r)$, the actions of $\rho_{\downarrow r}$ divide $\rho_{\downarrow v}$ into subsequences of $v$-changing actions $\rho_{\downarrow v} = \rho_v^1 \cdot \ldots \cdot \rho_v^k$, separated by the value changes required from $r$. That is, for each $1 \leq i \leq k$, all actions in $\rho_v^i$ are preconditioned by the same value of $r$, if any, and if two actions $a \in \rho_v^i$ and $a' \in \rho_v^{i+1}$ are preconditioned by $r$, then $\mathsf{pre}(a)[r] \neq \mathsf{pre}(a')[r]$. Let $\sigma \in \trianglerighteq[\sigma(r)]$ be a value sequence such that $|\sigma| = k = |\rho_{\downarrow r}| + 1$. For each $v \in \mathsf{succ}(r)$, $\rho_{\downarrow v}$ is a path from $I[v]$ to $G[v]$ in $\mathcal{L}_v(\sigma)$, and therefore some $\rho_\sigma$ is added into $\mathcal{R}$ by the algorithm, meaning that the algorithm finds a solution. Now, if $\rho_\sigma \in \mathcal{R}$, then, for each $v \in \mathsf{succ}(r)$, let $S_v^1 \cdot S_v^2 \cdot \ldots \cdot S_v^k$ be a cost-minimal path from $I[v]$ to $G[v]$ in $\mathcal{L}_v(\sigma)$ such that $S_v^i$ is the sequence of actions changing the value of $v$ and preconditioned either by $r\!:\!0$ or nothing for odd $i$, and by $r\!:\!1$ or nothing for even $i$. Thus,

$$cost(S_v^1 \cdot S_v^2 \cdot \ldots \cdot S_v^k) = \sum_{i=1}^{k} cost(S_v^i) \leq cost(\rho_{\downarrow v}).$$

Because sequence $S^i$ is obtained by an arbitrary merge of the sequences $\{S_v^i\}_{v \in \mathsf{succ}(r)}$, and $a_\vartheta$ is the cheapest action changing the value of $r$ to $\vartheta$, then $\rho_\sigma = S^1 \cdot a_{\sigma[2]} \cdot S^2 \cdot \ldots \cdot a_{\sigma[k]} \cdot S^k$ is an applicable sequence of actions that achieves the goal values for each $v \in \mathsf{succ}(r)$ as well as for $r$, and

$$\begin{aligned} cost(\rho_\sigma) = cost(S^1 \cdot a_{\sigma[2]} \cdot S^2 \cdot \ldots \cdot a_{\sigma[k]} \cdot S^k) = &\sum_{i=2}^{k} cost(a_{\sigma[i]}) + \sum_{i=1}^{k} cost(S^i) \leq \\ \leq &cost(\rho_{\downarrow r}) + \sum_{v \in \mathsf{succ}(r)} cost(\rho_{\downarrow v}) = cost(\rho). \end{aligned}$$

Hence, if $\Pi$ is solvable, then the algorithm returns a plan for $\Pi$, and this plan must be optimal. Finally, if $\Pi$ is not solvable, then $\mathcal{R}$ necessarily remains empty, and thus the algorithm fails. □

While Theorem 4 concerns the tractability tasks with fork-structured causal graphs and roots with binary domains, in our earlier work we also reported an additional tractability result for fork-structured causal graphs with the domains of all variables being of a fixed size, though not necessarily binary-valued (Katz & Domshlak, 2008). We do not discuss this result here in detail because, at least so far, we have not found it very helpful in the context of devising effective abstraction heuristics.

**Theorem 5 (Tractable Inverted Forks)** *Given a planning task $\Pi = \langle V, A, I, G, cost \rangle$ with an inverted fork causal graph with sink $r \in V$, if $|\mathcal{D}(r)| = O(1)$, the time complexity of the cost-optimal planning for $\Pi$ is polynomial in $||\Pi||$.*

**Proof:** Let $|\mathcal{D}(r)| = d$. Observe that the inverted-fork structure of the causal graph $CG(\Pi)$ implies all the actions in $\Pi$ are unary-effect, and that the sink $r$ preconditions only the actions affecting $r$ itself. Hence, in what follows we assume that $G[r]$ is specified; otherwise

Given a path $\langle a_1, \ldots, a_m \rangle$ from $I[r]$ to $G[r]$ in $DTG(r, \Pi)$:
$\rho := \langle\rangle$
$a_{m+1} := \langle G[\mathsf{pred}(r)], \emptyset \rangle$
**foreach** $v \in \mathsf{pred}(r)$ **do** $x_v := I[v]$
**for** $i := 1$ to $m + 1$ **do**
    **foreach** $v \in \mathsf{pred}(r)$ **do**
        **if** $\mathsf{pre}(a_i)[v]$ is specified **and** $\mathsf{pre}(a_i)[v] \neq x_v$ **then**
            **if** $\mathsf{pre}(a_i)[v]$ is not reachable from $x_v$ in $DTG(v, \Pi)$ **then fail**
            **append** to $\rho$ the actions induced by some cost-minimal path
                from $\mathsf{pre}(a_i)[v]$ to $x_v$ in $DTG(v, \Pi)$
            $x_v := \mathsf{pre}(a_i)[v]$
    **if** $i < m + 1$ **then append** to $\rho$ the action $a_i$
**return** $\rho$

Figure 4: Detailed outline of step (3) of the planning algorithm for inverted-fork structured task.

$\Pi$ breaks down to a set of trivial planning problems over a single variable each. Likewise, from the above properties of $\Pi$ it follows that, if $\rho$ is an optimal plan for $\Pi$, then the path $\rho_{\downarrow r}$ from $I[r]$ to $G[r]$ induced by $\rho$ in $DTG(r, \Pi)$ is either cycle-free or contains only zero-cost cycles. The latter can be safely eliminated from $\rho$, and thus we can assume that $\rho_{\downarrow r}$ is cycle-free. Given that, a simple algorithm that finds a cost-optimal plan for $\Pi$ in time $\Theta(||\Pi||^d + ||\Pi||^3)$ is as follows.

(1) Create all $\Theta(|A_r|^{d-1})$ cycle-free paths from $I[r]$ to $G[r]$ in $DTG(r, \Pi)$.

(2) For each variable $v \in \mathsf{pred}(r)$, and each pair of $v$'s values $x, y \in \mathcal{D}(v)$, compute the cost-minimal path from $x$ to $y$ in $DTG(v, \Pi)$. The whole set of such cost-minimal paths can be computed using $\Theta(d|V|)$ applications of the Floyd-Warshall algorithm on the domain transition graphs of the sink's parents $\mathsf{pred}(r)$.

(3) For each $I[r]$-to-$G[r]$ path in $DTG(r, \Pi)$ generated in step (1), construct a plan for $\Pi$ based on that path for $r$, and the cheapest paths computed in (2). This simple construction, depicted in Figure 4, is possible because the values of each parent variable can be changed independently of the values of all other variables in the inverted fork.

(4) Take the cheapest plan among those constructed in (3). If no plan was constructed in step (3), then $\Pi$ is unsolvable.

We have already observed that, for each cost-optimal plan $\rho$, $\rho_{\downarrow r}$ is one of the $I[r]$-to-$G[r]$ paths generated in step (1). For each $v \in \mathsf{pred}(r)$, let $S_v$ denote the sequence of values from $\mathcal{D}(v)$ that is required by the preconditions of the actions along $\rho_{\downarrow r}$. For each $v \in \mathsf{pred}(r)$, we have $\rho_{\downarrow v}$ corresponding to a (possibly cyclic) path from $I[v]$ to $G[v]$ in $DTG(v, \Pi)$, traversing the values (= nodes) from $S_v$ in the order required by $S_v$. In turn, the plan for $\Pi$ generated in (3) consists of cost-minimal such paths for all $v \in \mathsf{pred}(r)$. Therefore, at least one of the plans generated in (3) must be cost-optimal for Π, and the minimization step (4) will select one of them. □

Theorems 4 and 5 clarify the gap between fork decompositions and implicit abstraction heuristics, and now we can bridge this gap by further abstracting each task in the given fork decomposition of Π. We do that by abstracting domains of the fork roots and inverted-fork sinks to meet the requirements of the tractable fragments. We note that, in itself, the idea of domain decomposition is not very new in general (Hernadvölgyi & Holte, 1999) and in domain-independent planning in particular (Domshlak, Hoffmann, & Sabharwal, 2009). In fact, the shrinking step of the algorithm for building the merge-and-shrink abstractions is precisely a variable domain abstraction for meta-variables constructed in the merging steps (Helmert et al., 2007).

**Definition 10** *Let $\Pi = \langle V, A, I, G, cost\rangle$ be a planning task over states $S$, $v \in V$ be a state variable, and $\Phi = \{\phi_1, \dots, \phi_m\}$ be a set of mappings from $\mathcal{D}(v)$ to some sets $\Gamma_1, \dots, \Gamma_m$, respectively. $\mathcal{A} = \{\langle \Pi_{\phi_i}, \alpha_i\rangle\}_{i=1}^m$ is a* **domain abstraction** *of $\Pi$ over $\Phi$ if, for some set of cost functions $\{cost_i : A \to \mathbb{R}^{0+}\}_{i=1}^m$ satisfying*

$$\forall a \in A : \quad \sum_{i=1}^{m} cost_i(a) \le cost(a), \tag{8}$$

*we have, for $1 \le i \le m$,*

- *the abstraction mapping $\alpha_i$ of states $S$ is*

$$\forall u \in V : \quad \alpha_i(s)[u] = \begin{cases} \phi_i(s[u]), & u = v \\ s[u], & u \ne v \end{cases},$$

  *and, extending $\alpha_i$ to partial assignments on $V' \subseteq V$ as $\alpha_i(s[V']) = \alpha_i(s)[V']$,*

- *$\Pi_{\phi_i} = \langle V, A_{\phi_i}, I_{\phi_i}, G_{\phi_i}, cost_{\phi_i}\rangle$ is a planning task with*

  1. *$I_{\phi_i} = \alpha_i(I)$, $G_{\phi_i} = \alpha_i(G)$,*
  2. *$A_{\phi_i} = \{a_{\phi_i} = \langle \alpha_i(\mathsf{pre}(a)), \alpha_i(\mathsf{eff}(a))\rangle \mid a \in A\}$, and*
  3. *for each action $a \in A$,*

$$cost_{\phi_i}(a_{\phi_i}) = cost_i(a). \tag{9}$$

*We say that $\Pi_{\phi_i}$ is a* **domain decomposition** *of $\Pi_i = \langle V, A, I, G, cost_i\rangle$ with respect to $\phi_i$.*

**Theorem 6** *Domain abstractions of the planning tasks are additive implicit abstractions of these tasks.*

**Proof:** Let $\Pi = \langle V, A, I, G, cost\rangle$ be a planning task and $\mathcal{A} = \{\langle \Pi_{\phi_i}, \alpha_i\rangle\}_{i=1}^m$ be a domain abstraction of $\Pi$ over $\Phi = \{\phi_1, \dots, \phi_m\}$. Let $\mathcal{T} = (S, L, Tr, s^0, S^\star, \varpi)$ be the transition graph of $\Pi$. For each $1 \le i \le m$, let $\mathcal{T}_i = (S_i, L_i, Tr_i, s_i^0, S_i^\star, \varpi_i)$ be the transition graph of $\Pi_{\phi_i}$. We need to show that $\alpha_i$ is an abstraction mapping as in Definition 2.

First, from Definition 10 we have

- $s_i^0 = I_{\phi_i} = \alpha_i(I) = \alpha_i(s^0)$, and
- for all $s \in S^\star$ we have $s \supseteq G$ and thus $\alpha_i(s) \supseteq \alpha_i(G) = G_{\phi_i}$, providing us with $\alpha_i(s) \in S_i^\star$.

Now, if $s$ is a state of $\Pi$ and $a \in A$ is an action with $\mathsf{pre}(a) \subseteq s$, then $\alpha_i(s)$ is a state of $\Pi_{\phi_i}$ and $\mathsf{pre}(a_{\phi_i}) = \alpha_i(\mathsf{pre}(a)) \subseteq \alpha_i(s)$. Thus, $a_{\phi_i}$ is applicable in $\alpha_i(s)$, and now we show that applying $a_{\phi_i}$ in $\alpha_i(s)$ results in $\alpha_i(s)[\![a_{\phi_i}]\!] = \alpha_i(s[\![a]\!])$.

1. For the effect variables $v \in \mathcal{V}(\mathsf{eff}(a)) = \mathcal{V}(\mathsf{eff}(a_{\phi_i}))$, we have $\mathsf{eff}(a_{\phi_i}) \subseteq \alpha_i(s)[\![a_{\phi_i}]\!]$ and $\mathsf{eff}(a_{\phi_i}) = \alpha_i(\mathsf{eff}(a)) \subseteq \alpha_i(s[\![a]\!])$.
2. For all other variables $v \notin \mathcal{V}(\mathsf{eff}(a))$, we have $s[\![a]\!][v] = s[v]$ and $\alpha_i(s)[\![a_{\phi_i}]\!][v] = \alpha_i(s)[v]$, and thus

$$\alpha_i(s)[\![a_{\phi_i}]\!][v] = \alpha_i(s)[v] = \alpha_i(s[v]) = \alpha_i(s[\![a]\!][v]) = \alpha_i(s[\![a]\!])[v].$$

Next, for each $a \in A$, from Eqs. 8 and 9 we have

$$\sum_{i=1}^{m} cost_{\phi_i}(a_{\phi_i}) = \sum_{i=1}^{m} cost_i(a) \leq cost(a). \tag{10}$$

Now, let $s, s' \in S$ be a pair of states such that $cost(s, s') \leq \infty$, and let $\varrho = \langle a^1, \dots, a^l \rangle$ be the sequence of labels along a cheapest path from $s$ to $s'$ in $\mathcal{T}$. From that, $cost(s, s') = cost(\varrho) = \sum_{j=1}^{l} cost(a^j)$. The decomposition of such a path to the actions as in Definition 10 is a (not neccesarily cheapest) path from $\alpha_i(s)$ to $\alpha_i(s')$ in $\mathcal{T}_i$, and thus $cost(\alpha_i(s), \alpha_i(s')) \leq \sum_{j=1}^{l} cost_i(a^j)$, providing us with

$$\sum_{i=1}^{m} cost(\alpha_i(s), \alpha_i(s')) \leq \sum_{i=1}^{m}\sum_{j=1}^{l} cost_{\phi_i}(a^j_{\phi_i}) = \sum_{j=1}^{l}\sum_{i=1}^{m} cost_{\phi_i}(a^j_{\phi_i}) \overset{(10)}{\leq} \sum_{j=1}^{l} cost(a^j) = cost(s, s').$$

□

Having put the notion of domain abstraction in the framework of implicit abstractions, we are now ready to connect fork decompositions and implicit abstraction heuristics. Given a $\mathcal{FI}$-abstraction $\mathcal{A}_{\mathcal{FI}} = \{\langle \Pi_v^{\mathsf{f}}, \alpha_v^{\mathsf{f}} \rangle, \langle \Pi_v^{\mathsf{i}}, \alpha_v^{\mathsf{i}} \rangle\}_{v \in V}$ of a planning task $\Pi = \langle V, A, I, G, cost \rangle$,

- for each $\Pi_v^{\mathsf{f}} \in \mathbf{\Pi}_{\mathcal{FI}}$, we associate the root $v$ of $CG(\Pi_v^{\mathsf{f}})$ with mappings $\Phi_v^{\mathsf{f}} = \{\phi_{v,1}^{\mathsf{f}}, \dots, \phi_{v,k_v}^{\mathsf{f}}\}$ such that $k_v = O(poly(||\Pi||))$ and all $\phi_{v,i}^{\mathsf{f}} : \mathcal{D}(v) \to \{0, 1\}$, and then abstract $\Pi_v^{\mathsf{f}}$ with $\mathcal{A}_v^{\mathsf{f}} = \{\langle \Pi_{v,i}^{\mathsf{f}}, \alpha_{v,i}^{\mathsf{f}} \rangle\}_{i=1}^{k_v}$, and
- for each $\Pi_v^{\mathsf{i}} \in \mathbf{\Pi}_{\mathcal{FI}}$, we associate the sink $v$ of $CG(\Pi_v^{\mathsf{i}})$ with mappings $\Phi_v^{\mathsf{i}} = \{\phi_{v,1}^{\mathsf{i}}, \dots, \phi_{v,k'_v}^{\mathsf{i}}\}$ such that $k'_v = O(poly(||\Pi||))$ and all $\phi_{v,i}^{\mathsf{i}} : \mathcal{D}(v) \to \{0, 1, \dots, b_{v,i}\}$, $b_{v,i} = O(1)$, and then abstract $\Pi_v^{\mathsf{i}}$ with $\mathcal{A}_v^{\mathsf{i}} = \{\langle \Pi_{v,i}^{\mathsf{i}}, \alpha_{v,i}^{\mathsf{i}} \rangle\}_{i=1}^{k'_v}$.

From Theorem 3, Theorem 6, and the composition Theorem 2, we then immediately have

$$\mathcal{A}_{\mathfrak{FI}} = \bigcup_{v \in V} \left[ \bigcup_{i=1}^{k_v} \{\langle \Pi^{\mathsf{f}}_{v,i}, \alpha^{\mathsf{f}}_{v,i} \circ \alpha^{\mathsf{f}}_{v} \rangle\} \cup \bigcup_{i=1}^{k'_v} \{\langle \Pi^{\mathsf{i}}_{v,i}, \alpha^{\mathsf{i}}_{v,i} \circ \alpha^{\mathsf{i}}_{v} \rangle\} \right] \tag{11}$$

being an additive implicit abstraction of $\Pi$. Hence, from Theorem 1,

$$h^{\mathfrak{FI}} = \sum_{v \in V} \left[ \sum_{i=1}^{k_v} h^*_{\Pi^{\mathsf{f}}_{v,i}} + \sum_{i=1}^{k'_v} h^*_{\Pi^{\mathsf{i}}_{v,i}} \right] \tag{12}$$

is an admissible estimate of $h^*$ for $\Pi$, and, from Theorems 4 and 5, $h^{\mathfrak{FI}}$ is also computable in time $O(poly(||\Pi||))$.

This finalizes our construction of a concrete family of implicit abstraction heuristics. To illustrate the mixture of acyclic causal-graph and domain abstractions as above, we again use our running Logistics example. One bothersome question is to what extent further abstracting fork decompositions using domain abstractions affects the informativeness of the heuristic estimate. Though generally a degradation here is unavoidable, below we show that the answer to this question can sometimes be somewhat surprising.

To begin with an extreme setting, let all the domain abstractions for roots of forks and sinks of inverted forks be to *binary-valued* domains. Among multiple options for choosing the mapping sets $\{\Phi^{\mathsf{f}}_v\}$ and $\{\Phi^{\mathsf{i}}_v\}$, here we use a simple choice of distinguishing between different values of each variable $v$ on the basis of their cost from $I[v]$ in $DTG(v, \Pi)$. Specifically, for each $v \in V$, we set $\Phi^{\mathsf{f}}_v = \Phi^{\mathsf{i}}_v$, and, for each value $\vartheta \in \mathcal{D}(v)$ and each $1 \le i \le \max_{\vartheta' \in \mathcal{D}(v)} d(I[v], \vartheta')$,

$$\phi^{\mathsf{f}}_{v,i}(\vartheta) = \phi^{\mathsf{i}}_{v,i}(\vartheta) = \begin{cases} 0, & d(I[v], \vartheta) < i \\ 1, & \text{otherwise} \end{cases} \tag{13}$$

For example, the problem $\Pi^{\mathsf{f}}_{c_1}$ is decomposed (see the domain transition graph of $c_1$ on the left in Figure 1c) into two problems, $\Pi^{\mathsf{f}}_{c_1,1}$ and $\Pi^{\mathsf{f}}_{c_1,2}$, with the binary abstract domains of $c_1$ corresponding to the partitions $\{\{A\}, \{B, C, D\}\}$ and $\{\{A, D\}, \{B, C\}\}$ of $\mathcal{D}(c_1)$, respectively. As yet another example, the problem $\Pi^{\mathsf{i}}_{p_1}$ is decomposed (see the domain transition graph of $p_1$ in Figure 1d) into six problems $\Pi^{\mathsf{i}}_{p_1,1}, \dots, \Pi^{\mathsf{i}}_{p_1,6}$ along the abstractions of $\mathcal{D}(p_1)$ depicted in Figure 5a. Now, given the $\mathfrak{FI}$-decomposition of $\Pi$ and mappings $\{\Phi^{\mathsf{f}}_v, \Phi^{\mathsf{i}}_v\}_{v \in V}$ as above, consider the problem $\Pi^{\mathsf{i}}_{p_1,1}$, obtained from abstracting $\Pi$ along the inverted fork of $p_1$ and then abstracting $\mathcal{D}(p_1)$ using

$$\phi^{\mathsf{i}}_{p_1,1}(\vartheta) = \begin{cases} 0, & \vartheta \in \{C\} \\ 1, & \vartheta \in \{A, B, D, E, F, G, c_1, c_2, c_3, t\} \end{cases}.$$

It is not hard to verify that, from the original actions affecting $p_1$, we are left in $\Pi^{\mathsf{i}}_{p_1,1}$ with only actions conditioned by $c_1$ and $c_2$. If so, then no information is lost[3] if we remove from $\Pi^{\mathsf{i}}_{p_1,1}$ both variables $c_3$ and $t$, as well as the actions changing (only) these variables,

3. No information is lost here because we still keep either fork or inverted fork for each variable of $\Pi$.

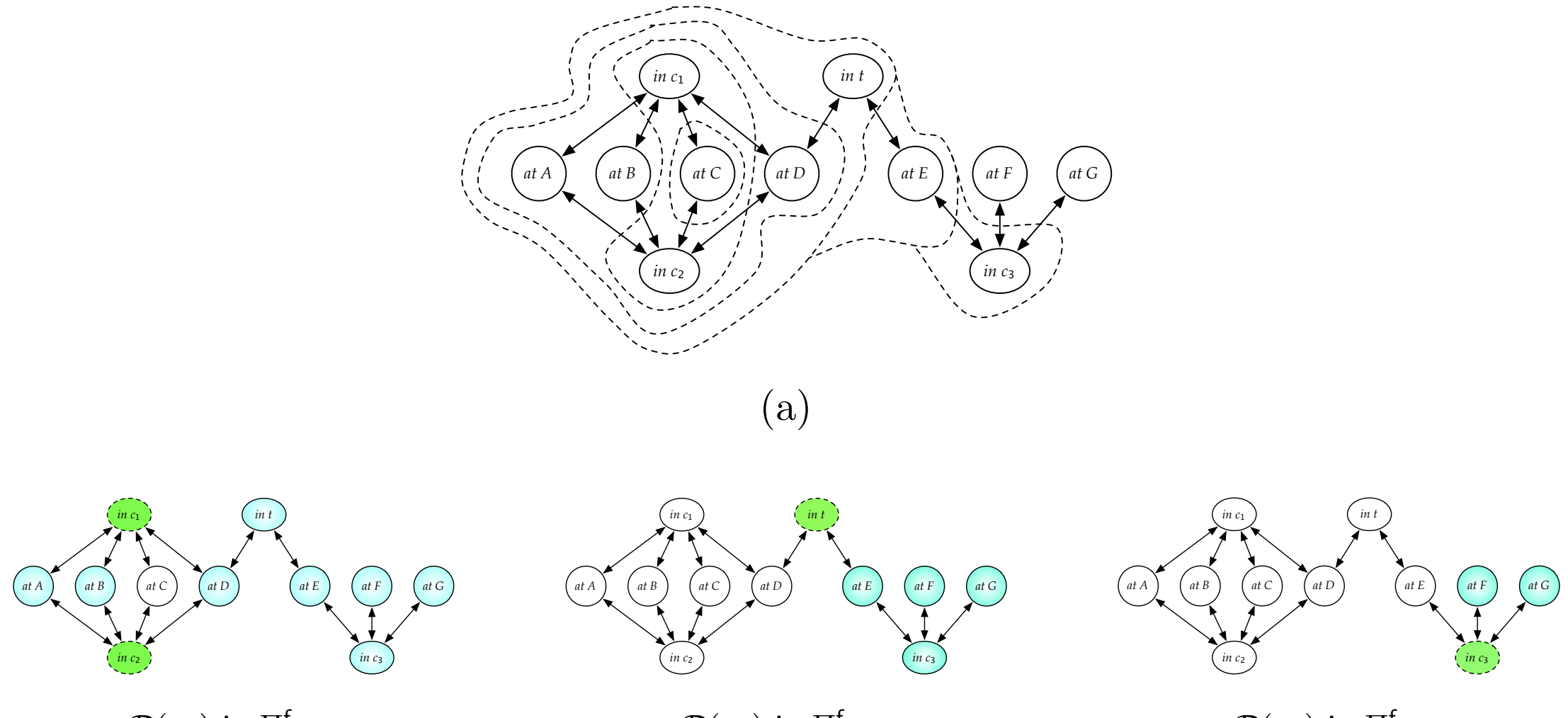

(a)

$\mathcal{D}(p_1)$ in $\Pi^{\mathsf{f}}_{p_1,1}$ $\mathcal{D}(p_1)$ in $\Pi^{\mathsf{f}}_{p_1,2}$ $\mathcal{D}(p_1)$ in $\Pi^{\mathsf{f}}_{p_1,3}$

(b)

Figure 5: Domain abstractions for $\mathcal{D}(p_1)$. (a) Binary-valued domain abstractions: the values inside and outside each dashed contour are mapped to 0 and 1, respectively. (b) Ternary-valued domain abstractions: values that are mapped to the same abstract value are shown as nodes with the same color and borderline.

and redistribute the cost of the removed actions between all other representatives of their originals in $\Pi$. The latter revision of the action cost partition can be obtained directly by replacing the cost-partitioning steps corresponding to Eqs. 3-4 and 8-9 by a single, joint action cost partitioning applied over the final additive implicit abstraction $\mathcal{A}_{\mathfrak{FI}}$ as in Eq. 11 and satisfying

$$cost(a) \geq \sum_{v \in V} \left[ \sum_{i=1}^{k_v} \sum_{a' \in A_{\mathcal{G}^{\mathsf{f}}_v}(a)} cost^{\mathsf{f}}_{v,i}(\phi^{\mathsf{f}}_{v,i}(a')) \quad + \sum_{i=1}^{k'_v} \sum_{a' \in A_{\mathcal{G}^{\mathsf{i}}_v}(a)} cost^{\mathsf{i}}_{v,i}(\phi^{\mathsf{i}}_{v,i}(a')) \right]. \tag{14}$$

In what follows, by uniform action cost partition we refer to a partition in which the cost of each action is equally split among all its nonredundant representatives in the *final* additive implicit abstraction.

Overall, computing $h^{\mathfrak{FI}}$ as in Eq. 12 under our "all binary-valued domain abstractions" and such a uniform action cost partition provides us with $h^{\mathfrak{FI}}(I) = 12\frac{7}{15}$, and knowing that the original costs are all integers we can safely adjust it to $h^{\mathfrak{FI}}(I) = 13$. Hence, even under the most severe domain abstractions as above, the estimate of $h^{\mathfrak{FI}}$ in our example task is not lower than that of $h^2$.

Let us now slightly refine our domain abstractions for the sinks of the inverted forks to be to a ternary range $\{0, 1, 2\}$. While mappings $\{\Phi^{\mathsf{f}}_v\}$ remain unchanged, $\{\Phi^{\mathsf{i}}_v\}$ are set to

$$\forall \vartheta \in \mathcal{D}(v): \quad \phi^{\mathsf{i}}_{v,i}(\vartheta) = \begin{cases} 0, d(I[v], \vartheta) < 2i - 1 \\ 1, d(I[v], \vartheta) = 2i - 1 \\ 2, d(I[v], \vartheta) > 2i - 1 \end{cases} . \quad (15)$$

For example, the problem $\Pi^{\mathsf{i}}_{p_1}$ is now decomposed into $\Pi^{\mathsf{i}}_{p_1,1}, \ldots, \Pi^{\mathsf{i}}_{p_1,3}$ along the abstractions of $\mathcal{D}(p_1)$ depicted in Figure 5b. Applying now the same computation of $h^{\mathfrak{FI}}$ as in Eq. 12 over the new set of domain abstractions gives $h^{\mathfrak{FI}}(I) = 15\frac{1}{2}$, which, again, can be safely adjusted to $h^{\mathfrak{FI}}(I) = 16$. Note that this value is *higher* than $h^{\mathfrak{FI}} = 15$ obtained using the (generally intractable) "pure" fork-decomposition abstractions as in Eq. 6. At first view, this outcome may seem counterintuitive as the domain abstractions are applied *over* the fork decomposition, and one would expect a coarser abstraction to provide less precise estimates. This, however, is not necessarily the case when the employed action cost partition is ad hoc. For instance, domain abstraction for the sink of an inverted fork may create independence between the sink and its parent variables, and exploiting such domain-abstraction specific independence relations leads to more targeted action cost partition via Eq. 14.

To see why this surprising "estimate improvement" has been obtained, note that before the domain abstraction in Eq. 15 is applied on our example, the truck-moving actions *drive-t-from-D-to-E* and *drive-t-from-E-to-D* appear in *three* abstractions $\Pi^{\mathsf{f}}_{t}$, $\Pi^{\mathsf{i}}_{p_1}$ and $\Pi^{\mathsf{i}}_{p_2}$, while after domain abstraction they appear in *five* abstractions $\Pi^{\mathsf{f}}_{t,1}$, $\Pi^{\mathsf{i}}_{p_1,1}$, $\Pi^{\mathsf{i}}_{p_1,2}$, $\Pi^{\mathsf{i}}_{p_1,3}$ and $\Pi^{\mathsf{i}}_{p_2,1}$. However, a closer look at the action sets of these five abstractions reveals that the dependencies of $p_1$ in $CG(\Pi^{\mathsf{i}}_{p_1,1})$ and $CG(\Pi^{\mathsf{i}}_{p_1,3})$, and of $p_2$ in $CG(\Pi^{\mathsf{i}}_{p_2,1})$ on $t$ are redundant, and thus keeping representatives of *move-D-E* and *move-E-D* in the corresponding abstract tasks is entirely unnecessary. Hence, after all, the two truck-moving actions appear only in *two* post-domain-abstraction tasks. Moreover, in both these abstractions the truck-moving actions are fully counted, in contrast to the predomain-abstraction tasks where the portion of the cost of these actions allocated to $\Pi^{\mathsf{i}}_{p_2}$ simply gets lost.

## 6. Experimental Evaluation: Take I

To evaluate the practical attractiveness of the fork-decomposition heuristics, we have conducted an empirical study on a wide sample of planning domains from the International Planning Competitions (IPC) 1998-2006, plus a non-IPC Schedule-STRIPS domain.[4] The domains were selected to allow a comparative evaluation with other, both baseline and state-of-the-art, approaches/planners, not all of which supported all the PDDL features at the time of our evaluation.

Later we formally prove that, under ad hoc action cost partitions such as our uniform partition, none of the three fork decompositions as in Definition 9 is dominated by the other two. Hence, we have implemented three additive fork-decomposition heuristics, $h^{\mathfrak{F}}$, $h^{\mathfrak{I}}$, and $h^{\mathfrak{FI}}$, within the standard heuristic forward search framework of the Fast Downward planner (Helmert, 2006) using the $A^*$ algorithm with full duplicate elimination. The $h^{\mathfrak{F}}$ heuristic corresponds to the ensemble of all (not clearly redundant) fork subgraphs of the

4. Schedule-STRIPS appears in the domains' distribution of IPC-2000. Later we became aware of the fact that this domain was excluded from the competition because its encoding generated problems for various planners.

| domain | $S$ | $h^{\mathfrak{F}}$ | | $h^{\mathfrak{I}}$ | | $h^{\mathfrak{FI}}$ | | $MS$-$10^4$ | | $MS$-$10^5$ | | $\text{HSP}^*_{\text{F}}$ | | Gamer | | blind | | $h_{\max}$ | |
|---|---|---|---|---|---|---|---|---|---|---|---|---|---|---|---|---|---|---|---|
| | | $s$ | $\%S$ | $s$ | $\%S$ | $s$ | $\%S$ | $s$ | $\%S$ | $s$ | $\%S$ | $s$ | $\%S$ | $s$ | $\%S$ | $s$ | $\%S$ | $s$ | $\%S$ |
| airport-ipc4 | 21 | 11 | 52 | 14 | 67 | 11 | 52 | 19 | 90 | 17 | 81 | 15 | 71 | 11 | 52 | 18 | 86 | 20 | 95 |
| blocks-ipc2 | 30 | 17 | 57 | 15 | 50 | 15 | 50 | 18 | 60 | 20 | 67 | 30 | 100 | 30 | 100 | 18 | 60 | 18 | 60 |
| depots-ipc3 | 7 | 2 | 29 | 2 | 29 | 2 | 29 | 7 | 100 | 4 | 57 | 4 | 57 | 4 | 57 | 4 | 57 | 4 | 57 |
| driverlog-ipc3 | 12 | 9 | 75 | 10 | 83 | 9 | 75 | 12 | 100 | 12 | 100 | 9 | 75 | 11 | 92 | 7 | 58 | 8 | 67 |
| freecell-ipc3 | 5 | 3 | 60 | 2 | 40 | 2 | 40 | 5 | 100 | 1 | 20 | 5 | 100 | 2 | 40 | 4 | 80 | 5 | 100 |
| grid-ipc1 | 2 | 1 | 50 | 1 | 50 | 1 | 50 | 2 | 100 | 2 | 100 | 0 | 0 | 2 | 100 | 1 | 50 | 2 | 100 |
| gripper-ipc1 | 20 | 5 | 25 | 5 | 25 | 5 | 25 | 7 | 35 | 7 | 35 | 6 | 30 | 20 | 100 | 7 | 35 | 7 | 35 |
| logistics-ipc1 | 6 | 3 | 50 | 2 | 33 | 2 | 33 | 4 | 67 | 5 | 83 | 3 | 50 | 6 | 100 | 2 | 33 | 2 | 33 |
| logistics-ipc2 | 22 | 21 | 95 | 15 | 68 | 14 | 64 | 16 | 73 | 21 | 95 | 16 | 73 | 20 | 91 | 10 | 45 | 10 | 45 |
| miconic-strips-ipc2 | 85 | 45 | 53 | 42 | 49 | 40 | 47 | 54 | 64 | 55 | 65 | 45 | 53 | 85 | 100 | 50 | 59 | 50 | 59 |
| mprime-ipc1 | 24 | 17 | 71 | 17 | 71 | 17 | 71 | 21 | 88 | 12 | 50 | 8 | 33 | 9 | 38 | 19 | 79 | 24 | 100 |
| mystery-ipc1 | 21 | 16 | 76 | 15 | 71 | 16 | 76 | 17 | 81 | 13 | 62 | 11 | 52 | 8 | 38 | 18 | 86 | 18 | 86 |
| openstacks-ipc5 | 7 | 7 | 100 | 7 | 100 | 7 | 100 | 7 | 100 | 7 | 100 | 7 | 100 | 7 | 100 | 7 | 100 | 7 | 100 |
| pathways-ipc5 | 4 | 4 | 100 | 4 | 100 | 4 | 100 | 3 | 75 | 4 | 100 | 4 | 100 | 4 | 100 | 4 | 100 | 4 | 100 |
| pipes-notank-ipc4 | 21 | 9 | 43 | 11 | 52 | 8 | 38 | 20 | 95 | 12 | 57 | 13 | 62 | 11 | 52 | 14 | 67 | 17 | 81 |
| pipes-tank-ipc4 | 14 | 6 | 43 | 6 | 43 | 6 | 43 | 13 | 93 | 7 | 50 | 7 | 50 | 6 | 43 | 10 | 71 | 10 | 71 |
| psr-small-ipc4 | 50 | 47 | 94 | 48 | 96 | 47 | 94 | 50 | 100 | 50 | 100 | 50 | 100 | 47 | 94 | 48 | 96 | 49 | 98 |
| rovers-ipc5 | 7 | 5 | 71 | 6 | 86 | 6 | 86 | 6 | 86 | 7 | 100 | 6 | 86 | 5 | 71 | 5 | 71 | 6 | 86 |
| satellite-ipc4 | 6 | 6 | 100 | 6 | 100 | 5 | 83 | 6 | 100 | 6 | 100 | 5 | 83 | 6 | 100 | 4 | 67 | 5 | 83 |
| schedule-strips | 43 | 42 | 98 | 35 | 81 | 39 | 91 | 22 | 51 | 1 | 2 | 11 | 26 | 3 | 7 | 29 | 67 | 31 | 72 |
| tpp-ipc5 | 6 | 5 | 83 | 5 | 83 | 5 | 83 | 6 | 100 | 6 | 100 | 5 | 83 | 5 | 83 | 5 | 83 | 6 | 100 |
| trucks-ipc5 | 9 | 5 | 56 | 5 | 56 | 5 | 56 | 6 | 67 | 5 | 56 | 9 | 100 | 3 | 33 | 5 | 56 | 7 | 78 |
| zenotravel-ipc3 | 11 | 8 | 73 | 9 | 82 | 8 | 73 | 11 | 100 | 11 | 100 | 8 | 73 | 10 | 91 | 7 | 64 | 8 | 73 |
| total | 433 | 294 | | 282 | | 274 | | 332 | | 285 | | 277 | | 315 | | 296 | | 318 | |

Table 1: A summary of the experimental results. Per domain, $S$ denotes the number of tasks solved by *any* planner. Per planner/domain, the number of tasks solved by that planner is given both by the absolute number ($s$) and by the percentage from "solved by some planners" ($\%S$). The last row summarize the number of solved instances.

causal graph, with the domains of the roots being abstracted using the "leave-one-value-out" binary-valued domain decompositions as follows:

$$\forall \vartheta_i \in \mathcal{D}(v): \quad \phi^{\mathsf{f}}_{v,i}(\vartheta) = \begin{cases} 0, & \vartheta = \vartheta_i \\ 1, & \text{otherwise} \end{cases}. \tag{16}$$

The $h^{\mathfrak{I}}$ heuristic is the same but for the inverted fork subgraphs, with the domains of the sinks being abstracted using the "distance-to-goal-value" ternary-valued domain decompositions[5] as in Eq. 17.

$$\forall \vartheta \in \mathcal{D}(v): \quad \phi^{\mathsf{i}}_{v,i}(\vartheta) = \begin{cases} 0, & d(\vartheta, G[v]) < 2i-1 \\ 1, & d(\vartheta, G[v]) = 2i-1 \\ 2, & d(\vartheta, G[v]) > 2i-1 \end{cases}. \tag{17}$$

The ensemble of the $h^{\mathfrak{FI}}$ heuristic is the union of these for $h^{\mathfrak{F}}$ and $h^{\mathfrak{I}}$. The action cost partition in all three heuristics was what we call "uniform."

We make a comparison with two baseline approaches, namely "blind $A^*$" with heuristic value 0 for goal states and 1 otherwise, and $A^*$ with the $h_{\max}$ heuristic (Bonet & Geffner, 2001), as well as with state-of-the-art abstraction heuristics, represented by the merge-and-shrink abstractions of Helmert et al. (2007). The latter were constructed under the

5. While "distance-from-initial-value" is reasonable for the evaluation of just the initial state, "leave-one-value-out" for fork roots and "distance-to-goal-value" for inverted-fork sinks should typically be much more attractive for the evaluation of all the states examined by $A^*$.

linear, $f$-preserving abstraction strategy proposed by these authors, and this under two fixed bounds on the size of the abstract state spaces, notably $|S^{\alpha}| < 10^4$ and $|S^{\alpha}| < 10^5$. These four (baseline and merge-and-shrink) heuristics were implemented by Helmert et al. (2007) within the same planning system as our fork-decomposition heuristics, allowing for a fairly unbiased comparison. We also compare to the Gamer (Edelkamp & Kissmann, 2009) and $\mathrm{HSP}^*_{\mathrm{F}}$ (Haslum, 2008) planners, the winner and the runner-up at the sequential optimization track of IPC-2008. On the algorithmic side, Gamer is based on a bidirectional blind search using sophisticated symbolic-search techniques, and $\mathrm{HSP}^*_{\mathrm{F}}$ uses $A^*$ with an additive critical-path heuristic. The experiments were conducted on a 3GHz Intel E8400 CPU with 2 GB memory, using 1.5 GB memory limit and 30 minute timeout. The only exception was Gamer, for which we used similar machines but with 4 GB memory and 2 GB memory limit; this was done to provide Gamer with the environment for which it was configured.

Table 1 summarizes our experimental results in terms of the number of tasks solved by each planner. Our impression of fork-decomposition heuristics from Table 1 is somewhat mixed. On the one hand, the performance of all three fork-decomposition based planners was comparable to one of the settings of the merge-and-shrink heuristic, and this clearly testifies for that the framework of implicit abstractions is not of theoretical interest only. On the other hand, all the planners, except for $A^*$ with the merge-and-shrink heuristic with $|S^{\alpha}| < 10^4$, failed to outperform $A^*$ with the baseline $h_{\max}$ heuristic. More important for us is that, unfortunately, all three fork-decomposition based planners failed to outperform even the basic blind search.

This, however, is not the end of the story for the fork-decomposition heuristics. Some hope can be found in the detailed results in Tables 9-14 in the appendix. As it appears from Table 10, on, e.g., the Logistics-ipc2 domain, $h^{\mathfrak{F}}$ almost consistently leads to expanding fewer search nodes than the (better between the two merge-and-shrink heuristics on this domain) $MS$-$10^5$, with the difference hitting four orders of magnitude. However, the time complexity of $h^{\mathfrak{F}}$ per search node is substantially higher than that of $MS$-$10^5$, with the two expanding at a rate of approximately 40 and 100000 nodes per second, respectively. The outcome is simple: while with no time limits (and only memory limit of 1.5 GB) $h^{\mathfrak{F}}$ solves more tasks in Logistics-ipc2 than $MS$-$10^5$ (task 12-1 is solved with $h^{\mathfrak{F}}$ in 2519.01 seconds), this is not so with a standard time limit of half an hour used for Table 10. In what follows we examine the possibility of exploiting the informativeness of fork-decomposition heuristics while not falling into the trap of costly per-node heuristic evaluation.

## 7. Back to Theory: $h$-Partitions and Databased Implicit Abstraction

Accuracy and low time complexity are both desired yet competing properties of heuristic functions. For many powerful heuristics, and abstraction heuristics in particular, computing $h(s)$ for each state $s$ in isolation is impractical: while computing $h(s)$ is polynomial in the description size of $\Pi$, it is often not efficient enough to be performed at each search node. However, for some costly heuristics this obstacle can be largely overcome by sharing most of the computation between the evaluations of $h$ on different states. If that is possible, the shared parts of computing $h$ for *all* problem states are precomputed and memorized before the search, and then reused during the search by the evaluations of $h$ on different states. Such a mixed offline/online heuristic computation is henceforth called **$h$-partition**, and we define the time complexity of an $h$-partition as the complexity of computing $h$ for a *set* of states. Given a subset of $k$ problem states $S' \subseteq S$, the $h$-partition's time complexity of computing $\{h(s) \mid s \in S'\}$ is expressed as $O(X + kY)$, where $O(X)$ and $O(Y)$ are, respectively, the complexity of the (offline) pre-search and (online) per-node parts of computing $h(s)$.

These days $h$-partitions are being adopted by various optimal planners using critical-path heuristics $h^m$ for $m > 1$ (Haslum et al., 2005), landmark heuristics $h^L$ and $h^{LA}$ (Karpas & Domshlak, 2009), and PDB and merge-and-shrink abstraction heuristics (Edelkamp, 2001; Helmert et al., 2007). Without effective $h$-partitions, optimal search with these heuristics would not scale up well, while with such $h$-partitions it constitutes the state of the art of cost-optimal planning. For instance, a very attractive property of PDB abstractions is the complexity of their natural $h^\alpha$-partition. Instead of computing $h^\alpha(s) = h^*(\alpha(s))$ from scratch for each evaluated state $s$ (impractical for all but tiny projections), the practice is to precompute and store $h^*(s')$ for *all* abstract states $s' \in S_\alpha$, after which the per-node computation of $h^\alpha(s)$ boils down to a hash-table lookup for $h^*(\alpha(s))$ with a perfect hash function. In our terms, the time and space complexity of that PDB $h^\alpha$-partition for a set of $k$ states is $O(|S^\alpha|(\log(|S^\alpha|) + |A|) + k)$ and $O(|S^\alpha|)$, respectively. This is precisely what makes PDB heuristics so attractive in practice. In that respect, the picture with merge-and-shrink abstractions is very much similar. While the order in which composites are formed and the choice of abstract states to contract are crucial to the complexity of their natural $h^\alpha$-partitions, the time and space complexity for the concrete linear abstraction strategy of Helmert *et al.* are respectively $O(|V||S^\alpha|(\log(|S^\alpha|) + |A|) + k \cdot |V|)$ and $O(|S^\alpha|)$. Similarly to PDB abstractions, the per-node computation of $h^\alpha(s)$ with a merge-and-shrink abstraction $\alpha$ is just a lookup in a data structure storing $h^*(\alpha(s))$ for all abstract states $\alpha(s) \in S_\alpha$. Hence, while the pre-search computation with MS abstractions can be more costly than with PDBs, the online part of computing heuristic values is still extremely efficient. This per-node efficiency provides the merge-and-shrink heuristics with impressive practical effectiveness on numerous IPC domains (Helmert et al., 2007).

To sum up, we can say that the fixed size of abstract spaces induced by explicit abstractions such as PDBs and merge-and-shrink is not only a limitation but also a key to obtaining effective $h$-partitions. In contrast, escaping that limitation with implicit abstractions might trap us into having to pay a high price for each search-node evaluation. We now show, however, that the time-per-node complexity bottleneck of fork-decomposition heuristics can be successfully overcome. Specifically, we show that an equivalent of PDB's and merge-and-shrink notion of "database" exists for fork-decomposition abstractions as well, despite their exponential-size abstract spaces. Of course, unlike with PDB and merge-and-shrink abstractions, the *databased fork-decomposition heuristics* do not (and cannot) provide us with a purely lookup online computation of $h^\alpha(s)$. The online part of the $h^\alpha$-partition has to be nontrivial in the sense that its complexity cannot be $O(1)$. In what comes next we prove the existence of such effective $h$-partitions for fork and inverted fork abstractions. In Section 8 we then empirically show that these $h$-partitions lead to fast pre-search and per-node computations, allowing the informativeness of the fork-decomposition heuristics to be successfully exploited in practice.

**Theorem 7** *Let* $\Pi = \langle V, A, I, G, cost \rangle$ *be a planning task with a fork causal graph rooted at a binary-valued variable* $r$. *There exists an* $h^*$*-partition for* $\Pi$ *such that, for any set of* $k$ *states, the time and space complexity of that* $h^*$*-partition is, respectively,* $O(d^3|V| + |A_r| + kd|V|)$ *and* $O(d^2|V|)$, *where* $d = \max_v \mathcal{D}(v)$.

**Proof:** The proof is by a modification of the polynomial-time algorithm for computing $h^*(s)$ for a state $s$ of such a task $\Pi$ used in the proof of Theorem 4 (Tractable Forks). *Given a state* $s$, let $\mathcal{D}(r) = \{0, 1\}$, where $s[r] = 0$. In what follows, for each of the two root's values $\vartheta \in \mathcal{D}(r)$, $\neg\vartheta$ denotes the opposite value $1 - \vartheta$; $\sigma(r)$, $\trianglerighteq[\sigma(r)]$, $DTG_v^0$ and $DTG_v^1$ are defined exactly as in the proof of Theorem 4.

(1) For each of the two values $\vartheta_r \in \mathcal{D}(r)$ of the root variable, each leaf variable $v \in V \setminus \{r\}$, and each pair of values $\vartheta, \vartheta' \in \mathcal{D}(v)$, let $p_{\vartheta,\vartheta';\vartheta_r}$ be the cost of the cheapest sequence of actions changing $v$ from $\vartheta$ to $\vartheta'$ *provided* $r : \vartheta_r$. The whole set $\{p_{\vartheta,\vartheta';\vartheta_r}\}$ for all the leaves $v \in V \setminus \{r\}$ can be computed by a straightforward variant of the all-pairs-shortest-paths, Floyd-Warshall algorithm on $DTG_v^{\vartheta_r}$ in time $O(d^3|V|)$.

(2) For each leaf variable $v \in V \setminus \{r\}$, $1 \le i \le d + 1$, and $\vartheta \in \mathcal{D}(v)$, let $g_{\vartheta;i}$ be the cost of the cheapest sequence of actions changing $s[v]$ to $\vartheta$ *provided* a sequence $\sigma \in \trianglerighteq[\sigma(r)]$, $|\sigma| = i$, of value changes of $r$. Having the values $\{p_{\vartheta,\vartheta';\vartheta_r}\}$ from step (1), the set $\{g_{\vartheta;i}\}$ is given by the solution of the recursive equation

$$g_{\vartheta;i} = \begin{cases} p_{s[v],\vartheta;s[r]}, & i = 1 \\ \min_{\vartheta'} \left[ g_{\vartheta';i-1} + p_{\vartheta',\vartheta;s[r]} \right], & 1 < i \le \delta_\vartheta,\ i \text{ is odd} \\ \min_{\vartheta'} \left[ g_{\vartheta';i-1} + p_{\vartheta',\vartheta;\neg s[r]} \right], & 1 < i \le \delta_\vartheta,\ i \text{ is even} \\ g_{\vartheta;i-1}, & \delta_\vartheta < i \le d + 1 \end{cases},$$

where $\delta_\vartheta = |\mathcal{D}(v)| + 1$. Given that, we have

$$h^*(s) = \min_{\sigma \in \trianglerighteq[\sigma(r)]} \left[ cost(\sigma) + \sum_{v \in V \setminus \{r\}} g_{G[v];|\sigma|} \right],$$

with $cost(\sigma) = \sum_{i=2}^{|\sigma|} cost(a_{\sigma[i]})$, where $a_{\sigma[i]} \in A$ is the cheapest action changing the value of $r$ from $\sigma[i-1]$ to $\sigma[i]$.

Note that step (1) is already state-independent, but the heavy step (2) is not. However, the state dependence of step (2) can mostly be overcome as follows. For each $v \in V \setminus \{r\}$, $\vartheta \in \mathcal{D}(v)$, $1 \le i \le d + 1$, and $\vartheta_r \in \mathcal{D}(r)$, let $\tilde{g}_{\vartheta;i}(\vartheta_r)$ be the cost of the cheapest sequence of actions changing $\vartheta$ to $G[v]$ *provided* the value changes of $r$ induce a 0/1 sequence of length $i$ starting with $\vartheta_r$. The set $\{\tilde{g}_{\vartheta;i}(\vartheta_r)\}$ is given by the solution of the recursive equation

$$\tilde{g}_{\vartheta;i}(\vartheta_r) = \begin{cases} p_{\vartheta,G[v];\vartheta_r}, & i = 1 \\ \min_{\vartheta'} \left[ \tilde{g}_{\vartheta';i-1}(\neg\vartheta_r) + p_{\vartheta,\vartheta';\vartheta_r} \right], & 1 < i \le \delta_\vartheta \\ \tilde{g}_{\vartheta;i-1}(\vartheta_r), & \delta_\vartheta < i \le d + 1 \end{cases}, \quad (18)$$

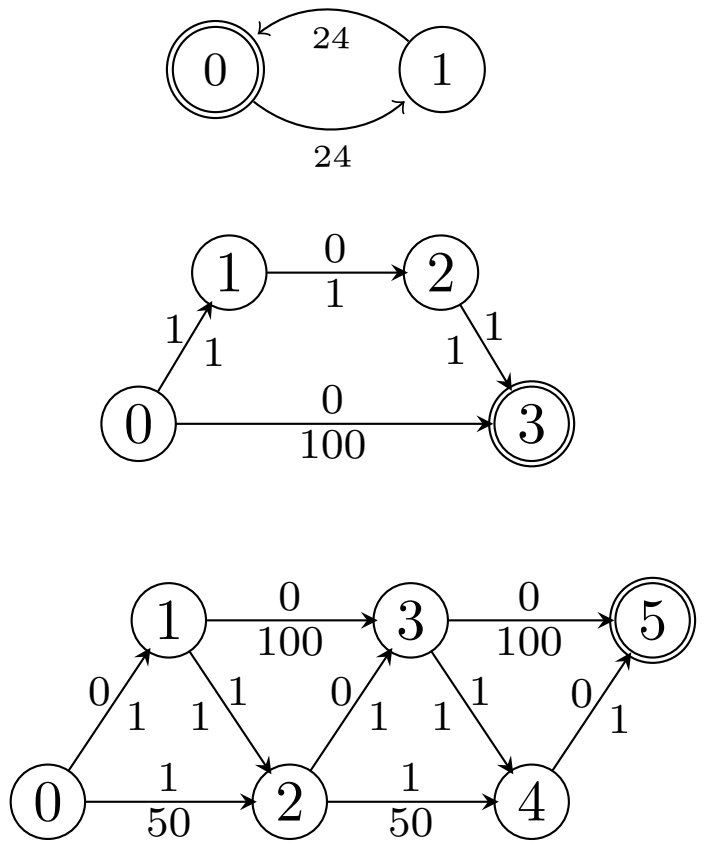

(a)

| $r$ | $\|\sigma\|$ | $cost(\sigma)$ | $v:0$ | $v:1$ | $v:2$ | $v:3$ | $u:0$ | $u:1$ | $u:2$ | $u:3$ | $u:4$ | $u:5$ |
|---|---|---|---|---|---|---|---|---|---|---|---|---|
| 0 | 1 | 0 | 100 | $\infty$ | $\infty$ | 0 | 201 | 200 | 101 | 100 | 1 | 0 |
| | 2 | 24 | 100 | 2 | 1 | 0 | 201 | 200 | 101 | 100 | 1 | 0 |
| | 3 | 48 | 100 | 2 | 1 | 0 | 53 | 102 | 3 | 2 | 1 | 0 |
| | 4 | 72 | 100 | 2 | 1 | 0 | 53 | 102 | 3 | 2 | 1 | 0 |
| | 5 | 96 | 3 | 2 | 1 | 0 | 5 | 4 | 3 | 2 | 1 | 0 |
| | 6 | 120 | 3 | 2 | 1 | 0 | 5 | 4 | 3 | 2 | 1 | 0 |
| | 7 | 144 | 3 | 2 | 1 | 0 | 5 | 4 | 3 | 2 | 1 | 0 |
| 1 | 1 | 0 | $\infty$ | $\infty$ | 1 | 0 | $\infty$ | $\infty$ | $\infty$ | $\infty$ | $\infty$ | 0 |
| | 2 | 24 | 100 | $\infty$ | 1 | 0 | 101 | 52 | 51 | 2 | 1 | 0 |
| | 3 | 48 | 3 | 2 | 1 | 0 | 101 | 52 | 51 | 2 | 1 | 0 |
| | 4 | 72 | 3 | 2 | 1 | 0 | 53 | 4 | 3 | 2 | 1 | 0 |
| | 5 | 96 | 3 | 2 | 1 | 0 | 53 | 4 | 3 | 2 | 1 | 0 |
| | 6 | 120 | 3 | 2 | 1 | 0 | 5 | 4 | 3 | 2 | 1 | 0 |
| | 7 | 144 | 3 | 2 | 1 | 0 | 5 | 4 | 3 | 2 | 1 | 0 |

(b)

Figure 6: The database for a fork-structured problem with a binary-valued root variable $r$ and two children $v$ and $u$, and $G[r] = 0$, $G[v] = 3$, and $G[u] = 5$. (a) depicts the domain transition graphs of $r$ (top), $v$ (middle), and $u$ (bottom); the numbers above and below each edge are the precondition on $r$ and the cost of the respective action. (b) depicts the database created by the algorithm. For instance, the entry in row $r:0 \wedge |\sigma| = 5$ and column $v:0$ captures the value of $\tilde{g}_{v0;5}(r:0)$ computed as in Eq. 18. The shaded entries are those examined during the online computation of $h^*(r:0, v:0, u:0)$.

which can be solved in time $O(d^3|V|)$. Note that this equation is now independent of the evaluated state $s$, and yet $\{\tilde{g}_{\vartheta;i}(\vartheta_r)\}$ allow for computing $h^*(s)$ for a given state $s$ via

$$h^*(s) = \min_{\sigma \in \trianglerighteq[\sigma(r|s[r])]} \left[ cost(\sigma) + \sum_{v \in V \setminus \{r\}} \tilde{g}_{s[v];|\sigma|}(s[r]) \right] \tag{19}$$

where $\sigma(r|\vartheta_r)$ is defined similarly to $\sigma(r)$ but with respect to the initial value $\vartheta_r$ of $r$.

With the new formulation, the only computation that has to be performed online, per search node, is the final minimization over $\trianglerighteq[\sigma(r|s[r])]$ in Eq. 19, and this is the lightest part of the whole algorithm anyway. The major computations, notably those of $\{p_{\vartheta,\vartheta';\vartheta_r}\}$ and $\{\tilde{g}_{\vartheta;i}(\vartheta_r)\}$, can now be performed offline and shared between the evaluated states. The space required to store this information is $O(d^2|V|)$ as it contains only a fixed amount of information per pair of values of each variable. The time complexity of the offline computation is $O(d^3|V| + |A_r|)$; the $|A_r|$ component stems from precomputing the costs $cost(\sigma)$. The time complexity of the online computation per state is $O(d|V|)$; $|V|$ comes from the internal summation and $d$ comes from the size of $\trianglerighteq[\sigma(r|s[r])]$.

□

Figure 6b shows the database created for a fork-structured problem with a binary-valued root $r$, two children $v$ and $u$, and $G[r] = 0$, $G[v] = 3$, and $G[u] = 5$; the domain transition graphs of $v$ and $u$ are depicted in Figure 6(a). Online computation of $h^*(s)$ as in Eq. 19 for $s = (r:0, v:0, u:0)$ sums over the shaded entries of each of the four rows having such entries, and minimizes over the resulting four sums, with the minimum being obtained in the row $r:0 \wedge |\sigma| = 5$.

**Theorem 8** *Let $\Pi = \langle V, A, I, G, cost\rangle$ be a planning task with an inverted fork causal graph with sink $r$ and $|\mathcal{D}(r)| = b = O(1)$. There exists an $h^*$-partition for $\Pi$ such that, for any set of $k$ states, the time and space complexity of that $h^*$-partition is $O(b|V||A_r|^{b-1} + d^3|V| + k|V||A_r|^{b-1})$ and $O(|V||A_r|^{b-1} + d^2|V|)$, respectively, where $d = \max_v \mathcal{D}(v)$.*

**Proof:** Like the proof of Theorem 7, the proof of Theorem 8 is based on a modification of the polynomial-time algorithm for computing $h^*(s)$ used for the proof of Theorem 5 (Tractable Inverted Forks).

(1) For each parent variable $v \in V \setminus \{r\}$, and each pair of its values $\vartheta, \vartheta' \in \mathcal{D}(v)$, let $p_{\vartheta,\vartheta'}$ be the cost of the cheapest sequence of actions changing $\vartheta$ to $\vartheta'$. The whole set $\{p_{\vartheta,\vartheta'}\}$ can be computed using the Floyd-Warshall algorithm on the domain transition graph of $v$ in time $O(d^3|V|)$.

(2) Given a state $s$, for each cycle-free path $\pi = \langle a_1, \ldots, a_m\rangle$ from $s[r]$ to $G[r]$ in $DTG(v, \Pi)$, let $g_\pi$ be the cost of the cheapest plan from $s$ in $\Pi$ based on $\pi$, and the cheapest paths $\{p_{\vartheta,\vartheta'}\}$ computed in step (1). Each $g_\pi$ can be computed as

$$g_\pi = \sum_{i=1}^{m} cost(a_i) + \sum_{i=0}^{m} \sum_{v \in V \setminus \{r\}} p_{\mathsf{pre}_i[v], \mathsf{pre}_{i+1}[v]},$$

where $\mathsf{pre}_0, \ldots, \mathsf{pre}_{m+1}$ are the values required from the parents of $r$ along the path $\pi$. That is, for each $v \in V \setminus \{r\}$, and $0 \le i \le m+1$,

$$\mathsf{pre}_i[v] = \begin{cases} s[v], & i = 0 \\ G[v], & i = m+1, \text{ and } G[v] \text{ is specified} \\ \mathsf{pre}(a_i)[v], & 1 \le i \le m, \text{ and } \mathsf{pre}(a_i)[v] \text{ is specified} \\ \mathsf{pre}_{i-1}[v] & \text{otherwise} \end{cases}.$$

From that, we have $h^*(s) = \min_\pi g_\pi$.

Note that step (1) is state-independent, but step (2) is not. However, the dependence of step (2) on the evaluated state can be substantially relaxed. As there are only $O(1)$ different values of $r$, it is possible to consider cycle-free paths to $G[r]$ from *all* values of $r$. For each such path $\pi$, and each parent variable $v \in V \setminus \{r\}$, we know what the first value of $v$ required by $\pi$ would be. Given that, we can precompute the cost-optimal plans induced by each $\pi$ *assuming the parents start at their first required values.* The remainder of the computation of $h^*(s)$ is delegated to online, and the modified step (2) is as follows.

For each $\vartheta_r \in \mathcal{D}(r)$ and each cycle-free path $\pi = \langle a_1, \ldots, a_m\rangle$ from $\vartheta_r$ to $G[r]$ in $DTG(r, \Pi)$, let a "proxy" state $s_\pi$ be

$$s_\pi[v] = \begin{cases} \vartheta_r, & v = r \\ G[v], & \forall 1 \le i \le m : \mathsf{pre}(a_i)[v] \text{ is unspecified} \\ \mathsf{pre}(a_i)[v], & i = \text{argmin}_j \{\mathsf{pre}(a_j)[v] \text{ is specified}\} \end{cases},$$

that is, the nontrivial part of $s_\pi$ captures the first values of $V \setminus \{r\}$ required along $\pi$.[6] Given that, let $g_\pi$ be the cost of the cheapest plan from $s_\pi$ in $\Pi$ based on $\pi$, and the cheapest paths $\{p_{\vartheta,\vartheta'}\}$ computed in (1). Each $g_\pi$ can be computed as

$$g_\pi = \sum_{i=1}^{m} \left[ cost(a_i) + \sum_{v \in V\setminus\{r\}} p_{\mathsf{pre}_i[v],\mathsf{pre}_{i+1}[v]} \right],$$

where, for each $v \in V \setminus \{r\}$, and $1 \leq i \leq m+1$,

$$\mathsf{pre}_i[v] = \begin{cases} s_\pi[v], & i = 1 \\ G[v], & i = m+1, \text{ and } G[v] \text{ is specified} \\ \mathsf{pre}(a_i)[v], & 2 \leq i \leq m, \text{ and } \mathsf{pre}(a_i)[v] \text{ is specified} \\ \mathsf{pre}_{i-1}[v], & \text{otherwise} \end{cases}.$$

Storing the pairs $(g_\pi, s_\pi)$ accomplishes the offline part of the computation. Now, given a search state $s$, we can compute

$$h^*(s) = \min_{\substack{\pi \text{ s.t.} \\ s_\pi[r]=s[r]}} \left[ g_\pi + \sum_{v \in V\setminus\{r\}} p_{s[v],s_\pi[v]} \right]. \tag{20}$$

The number of cycle-free paths to $G[r]$ in $DTG(r, \Pi)$ is $\Theta(|A_r|^{b-1})$, and $g_\pi$ for each such path $\pi$ can be computed in time $O(b|V|)$. Hence, the overall offline time complexity is $O(b|V||A_r|^{b-1} + d^3|V|)$, and the space complexity (including the storage of the proxy states $s_\pi$) is $O(|V||A_r|^{b-1} + d^2|V|)$. The time complexity of the online computation per state via Eq. 20 is $O(|V||A_r|^{b-1})$; $|V|$ comes from the internal summation and $|A_r|^{b-1}$ from the upper bound on the number of cycle-free paths from $s[r]$ to $G[r]$. □

Figure 7(b) shows the database created for an inverted fork structured problem with a ternary-valued sink variable $r$, two parents $u$ and $v$, and $G[r] = 2$, $G[u] = 0$, and $G[v] = 2$. The domain transition graphs of $u$ and $v$ are depicted at the top of Figure 7(a); the actual identities of actions affecting these two parents are not important here. The actions affecting the sink $r$ are

$$\begin{aligned} a_1 &= \langle \{u:1, r:0\}, \{r:1\} \rangle \\ a_2 &= \langle \{v:1, r:0\}, \{r:1\} \rangle \\ a_3 &= \langle \{u:2, r:1\}, \{r:2\} \rangle \\ a_4 &= \langle \{v:1, r:1\}, \{r:2\} \rangle. \end{aligned}$$

The domain transition graph of $r$ is depicted at the bottom of Figure 7(a). Online computation of $h^*(s)$ as in Eq. 20 for $s = (r:0, v:0, u:0)$ sums over the shaded entries of each of the four rows having such entries, and minimizes over the resulting four sums, with the minimum being obtained in the lowest such row.

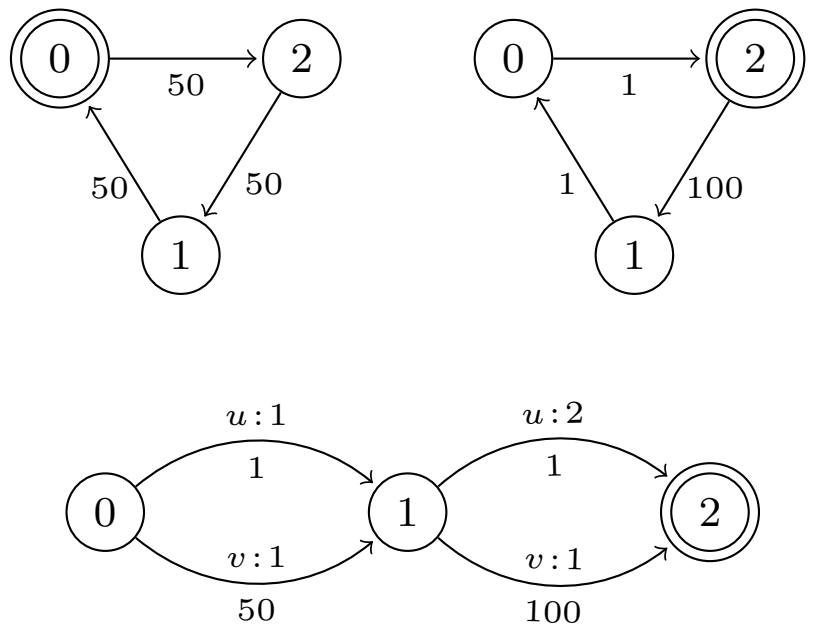

(a)

| $r$ | $\pi$ | $s_\pi$ | $g_\pi$ | $u:0$ | $u:1$ | $u:2$ | $v:0$ | $v:1$ | $v:2$ |
|---|---|---|---|---|---|---|---|---|---|
| 0 | $\langle a_1, a_3 \rangle$ | $u:1, v:2$ | 202 | 100 | 0 | 50 | 1 | 2 | 0 |
| | $\langle a_1, a_4 \rangle$ | $u:1, v:1$ | 153 | 100 | 0 | 50 | 101 | 0 | 100 |
| | $\langle a_2, a_3 \rangle$ | $u:2, v:1$ | 153 | 50 | 100 | 0 | 101 | 0 | 100 |
| | $\langle a_2 \cdot a_4 \rangle$ | $u:0, v:1$ | 152 | 0 | 50 | 100 | 101 | 0 | 100 |
| 1 | $\langle a_3 \rangle$ | $u:2, v:2$ | 101 | 50 | 100 | 0 | 1 | 2 | 0 |
| | $\langle a_4 \rangle$ | $u:0, v:1$ | 102 | 0 | 50 | 100 | 101 | 0 | 100 |

(b)

Figure 7: The database for an inverted fork-structured problem with a $O(1)$ bounded sink variable $r$ and two parents $u$ and $v$, and $G[r] = 2$, $G[u] = 0$, and $G[v] = 2$. (a) depicts the domain transition graphs of $u$ (top left), $v$ (top right), and $r$ (bottom); the numbers above and below each edge are the preconditions and the cost of the respective action, respectively. (b) depicts the database created by the algorithm. The shaded entries are those examined during the online computation of $h^*(r:0, u:0, v:0)$.

## 8. Experimental Evaluation: Take II

To evaluate the practical attractiveness of the databased fork-decomposition heuristics, we have repeated our empirical evaluation as in Section 6, but now for the databased versions of the heuristics. The detailed results of this evaluation are relegated to Tables 15-20 in the appendix, but they are summarized here in Table 2. For each domain, the $S$ column captures the number of tasks in that domain that were solved by at least one planner in the suite. Per planner/domain, the number of tasks solved by that planner is given both by the absolute number ($s$) and by the percentage from "solved by some planners" (%$S$). Boldfaced results indicate the best performance within the corresponding domain. The last three rows summarize the performance of the planners via three measures. The first is the number of tasks solved in all the 23 domains; this is basically the performance evaluation measure used in the optimization track at IPC-2008. As domains are not equally challenging and do not equally discriminate between the planners' performance, the second is a "domain-normalized" performance measure

$$\hat{s}(p) = \sum_{\text{domain } \mathcal{D}} \frac{\#\text{tasks in } \mathcal{D} \text{ solved by planner } p}{\#\text{tasks in } \mathcal{D} \text{ solved by some planners}}.$$

Finally, the third measure corresponds to the number of domains $w$ in which the planner in question solved at least as many tasks as any other planner.

Overall, Table 2 clearly suggests that heuristic search with "databased" fork-decomposition heuristics favorably competes with the state of the art of optimal planning. In particular,

6. For ease of presentation, we omit here the case where $v$ is required neither along $\pi$, nor by the goal; such variables should be simply ignored when accounting for the cost of $\pi$.

| domain | $S$ | $h^{\mathfrak{F}}$ | | $h^{\mathfrak{I}}$ | | $h^{\mathfrak{FI}}$ | | $MS$-$10^4$ | | $MS$-$10^5$ | | $\text{HSP}^*_\text{F}$ | | Gamer | | blind | | $h_{\max}$ | |
|---|---|---|---|---|---|---|---|---|---|---|---|---|---|---|---|---|---|---|---|
| | | $s$ | $\%S$ | $s$ | $\%S$ | $s$ | $\%S$ | $s$ | $\%S$ | $s$ | $\%S$ | $s$ | $\%S$ | $s$ | $\%S$ | $s$ | $\%S$ | $s$ | $\%S$ |
| airport-ipc4 | 22 | **22** | 100 | 20 | 91 | 21 | 95 | 19 | 86 | 17 | 77 | 15 | 68 | 11 | 50 | 18 | 82 | 20 | 91 |
| blocks-ipc2 | 30 | 21 | 70 | 18 | 60 | 18 | 60 | 18 | 60 | 20 | 67 | **30** | 100 | **30** | 100 | 18 | 60 | 18 | 60 |
| depots-ipc3 | 7 | **7** | 100 | 4 | 57 | **7** | 100 | **7** | 100 | 4 | 57 | 4 | 57 | 4 | 57 | 4 | 57 | 4 | 57 |
| driverlog-ipc3 | 12 | **12** | 100 | **12** | 100 | **12** | 100 | **12** | 100 | **12** | 100 | 9 | 75 | 11 | 92 | 7 | 58 | 8 | 67 |
| freecell-ipc3 | 5 | **5** | 100 | 4 | 80 | 4 | 80 | **5** | 100 | 1 | 20 | **5** | 100 | 2 | 40 | 4 | 80 | **5** | 100 |
| grid-ipc1 | 2 | **2** | 100 | 1 | 50 | 1 | 50 | **2** | 100 | **2** | 100 | 0 | 0 | **2** | 100 | 1 | 50 | **2** | 100 |
| gripper-ipc1 | 20 | 7 | 35 | 7 | 35 | 7 | 35 | 7 | 35 | 7 | 35 | 6 | 30 | **20** | 100 | 7 | 35 | 7 | 35 |
| logistics-ipc2 | 22 | **22** | 100 | 16 | 73 | 16 | 73 | 16 | 73 | 21 | 95 | 16 | 73 | 20 | 91 | 10 | 45 | 10 | 45 |
| logistics-ipc1 | 7 | **6** | 86 | 4 | 57 | 5 | 71 | 4 | 57 | 5 | 71 | 3 | 43 | **6** | 86 | 2 | 29 | 2 | 29 |
| miconic-strips-ipc2 | 85 | 51 | 60 | 50 | 59 | 50 | 59 | 54 | 64 | 55 | 65 | 45 | 53 | **85** | 100 | 50 | 59 | 50 | 59 |
| mprime-ipc1 | 24 | 23 | 96 | 22 | 92 | 21 | 88 | 21 | 88 | 12 | 50 | 8 | 33 | 9 | 38 | 19 | 79 | **24** | 100 |
| mystery-ipc1 | 21 | **21** | 100 | 18 | 86 | **21** | 100 | 17 | 81 | 13 | 62 | 11 | 52 | 8 | 38 | 18 | 86 | 18 | 86 |
| openstacks-ipc5 | 7 | **7** | 100 | **7** | 100 | **7** | 100 | **7** | 100 | **7** | 100 | **7** | 100 | **7** | 100 | **7** | 100 | **7** | 100 |
| pathways-ipc5 | 4 | **4** | 100 | **4** | 100 | **4** | 100 | 3 | 75 | **4** | 100 | **4** | 100 | **4** | 100 | **4** | 100 | **4** | 100 |
| pipes-notank-ipc4 | 21 | 17 | 81 | 15 | 71 | 16 | 76 | **20** | 95 | 12 | 57 | 13 | 62 | 11 | 52 | 14 | 67 | 17 | 81 |
| pipes-tank-ipc4 | 14 | 11 | 79 | 9 | 64 | 9 | 64 | **13** | 93 | 7 | 50 | 7 | 50 | 6 | 43 | 10 | 71 | 10 | 71 |
| psr-small-ipc4 | 50 | 49 | 98 | 49 | 98 | 49 | 98 | **50** | 100 | **50** | 100 | **50** | 100 | 47 | 94 | 48 | 96 | 49 | 98 |
| rovers-ipc5 | 7 | 6 | 86 | **7** | 100 | 6 | 86 | 6 | 86 | **7** | 100 | 6 | 86 | 5 | 71 | 5 | 71 | 6 | 86 |
| satellite-ipc4 | 6 | **6** | 100 | **6** | 100 | **6** | 100 | **6** | 100 | **6** | 100 | 5 | 83 | **6** | 100 | 4 | 67 | 5 | 83 |
| schedule-strips | 46 | **46** | 100 | 40 | 87 | **46** | 100 | 22 | 48 | 1 | 2 | 11 | 24 | 3 | 7 | 29 | 63 | 31 | 67 |
| tpp-ipc5 | 6 | **6** | 100 | **6** | 100 | **6** | 100 | **6** | 100 | **6** | 100 | 5 | 83 | 5 | 83 | 5 | 83 | **6** | 100 |
| trucks-ipc5 | 9 | 6 | 67 | 7 | 78 | 7 | 78 | 6 | 67 | 5 | 56 | **9** | 100 | 3 | 33 | 5 | 56 | 7 | 78 |
| zenotravel-ipc3 | 11 | **11** | 100 | **11** | 100 | **11** | 100 | **11** | 100 | **11** | 100 | 8 | 73 | 10 | 91 | 7 | 64 | 8 | 73 |
| total | 438 | **368** | | 337 | | 350 | | 332 | | 285 | | 277 | | 315 | | 296 | | 318 | |
| $\hat{s}$ | | **20.56** | | 18.38 | | 19.13 | | 19.07 | | 16.64 | | 15.45 | | 16.66 | | 15.58 | | 17.66 | |
| $w$ | | **14** | | 7 | | 9 | | 11 | | 9 | | 6 | | 8 | | 2 | | 6 | |

Table 2: A summary of the experimental results with databased versions of the fork-decomposition heuristics. Per domain, $S$ denotes the number of tasks solved by *any* planner. Per planner/domain, the number of tasks solved by that planner is given both by the absolute number ($s$) and by the percentage from "solved by some planners" ($\%S$). Boldfaced results indicate the best performance within the corresponding domain. The last three rows summarize the number of solved instances, the domain-normalized measure of solved instances ($\hat{s}$), and the number of domains in which the planners achieved superior performance ($w$).

$A^*$ with the "only forks" heuristic $h^{\mathfrak{F}}$ exhibited the best overall performance according to all three measures. In terms of the absolute number of solved instances, $A^*$ with all three fork-decomposition heuristics outperformed all other planners in the suite. The contribution of databasing to the success of the fork-decomposition heuristics was dramatic. Looking back at the results with "fully online" heuristic computation depicted in Table 1, note that the total number of solved instances for the fork-decomposition heuristics $h^{\mathfrak{F}}$, $h^{\mathfrak{I}}$, and $h^{\mathfrak{FI}}$ increased by 74, 55, and 76, respectively, and this made the whole difference.

We have also performed a comparative evaluation on the planning domains from the recent IPC-2008. The IPC-2008 domains differ from the previous domains in that actions had various costs, and, more importantly, many actions had zero cost. The latter is an issue for heuristic-search planners because heuristic functions cannot differentiate between subplans that have the same cost of zero, but differ in length. In any case, the comparative side of our evaluation on the IPC-2008 domains differ on several points from the previous one. First, neither for merge-and-shrink nor for $h_{\max}$ heuristics, we had implementation supporting arbitrary action costs. Hence, our comparison here is only with Gamer, $\text{HSP}^*_\text{F}$, and blind search. Second, to ensure admissibility of the blind search, the latter has been modified to return on non-goal states the cost of the cheapest applicable action. Finally, all the planners were run on a 3GHz Intel E8400 CPU with 4 GB memory, using 2 GB memory limit and 30 minute timeout. The results of this evaluation are summarized in Table 3; for the detailed results we refer the reader to Tables 21-22 in the appendix. Overall, these results show that $A^*$ with the fork-decomposition heuristics are very much competitive on the IPC-2008 domains as well.

| domain | $S$ | $h^{\mathfrak{F}}$ | | $h^{\mathfrak{I}}$ | | $h^{\mathfrak{FI}}$ | | $\text{HSP}^*_{\text{F}}$ | | Gamer | | blind | |
|---|---|---|---|---|---|---|---|---|---|---|---|---|---|
| | | $s$ | %$S$ | $s$ | %$S$ | $s$ | %$S$ | $s$ | %$S$ | $s$ | %$S$ | $s$ | %$S$ |
| elevators-strips-ipc6 | 22 | 18 | 82 | 14 | 64 | 15 | 68 | 7 | 32 | **22** | 100 | 11 | 50 |
| openstacks-strips-ipc6 | 21 | 19 | 90 | 19 | 90 | 19 | 90 | **21** | 100 | 19 | 90 | 19 | 90 |
| parcprinter-strips-ipc6 | 16 | 14 | 88 | 13 | 81 | 13 | 81 | **16** | 100 | 9 | 56 | 10 | 63 |
| pegsol-strips-ipc6 | 27 | **27** | 100 | **27** | 100 | **27** | 100 | **27** | 100 | 24 | 89 | **27** | 100 |
| scanalyzer-strips-ipc6 | 12 | **12** | 100 | 6 | 50 | 6 | 50 | 6 | 50 | 11 | 92 | **12** | 100 |
| sokoban-strips-ipc6 | 28 | 25 | 89 | 26 | 93 | **27** | 96 | 13 | 46 | 20 | 71 | 20 | 71 |
| transport-strips-ipc6 | 11 | **11** | 100 | **11** | 100 | **11** | 100 | 9 | 82 | **11** | 100 | **11** | 100 |
| woodworking-strips-ipc6 | 14 | 8 | 57 | 8 | 57 | 8 | 57 | 9 | 64 | **14** | 100 | 7 | 50 |
| total | 152 | **134** | | 124 | | 126 | | 108 | | 130 | | 117 | |
| $\hat{s}$ | | **7.06** | | 6.35 | | 6.43 | | 5.74 | | 6.99 | | 6.24 | |
| $w$ | | **3** | | 2 | | **3** | | **3** | | **3** | | **3** | |

Table 3: A summary of the experimental results. Per domain, $S$ denotes the number of tasks solved by *any* planner. Per planner/domain, the number of tasks solved by that planner is given both by the absolute number ($s$) and by the percentage from "solved by some planners" (%$S$). Boldfaced results indicate the best performance within the corresponding domain. The last three rows summarize the number of solved instances, the domain-normalized measure of solved instances ($\hat{s}$), and the number of domains in which the planners achieved superior performance ($w$).

## 9. Formal Analysis: Asymptotic Performance Ratios

Empirical evaluation on a concrete set of benchmark tasks is a standard and important methodology for assessing the effectiveness of heuristic estimates: it allows us to study the tradeoff between the accuracy of the heuristics and the complexity of computing them. However, as rightfully noted by Helmert and Mattmüller (2008), such evaluations almost never lead to absolute statements of the type "Heuristic $h$ is well-suited for solving problems from benchmark suite $X$," but only to relative statements of the type "Heuristic $h$ expands fewer nodes than heuristic $h'$ on benchmark suite $X$." Moreover, one would probably like to obtain formal evidence of the effectiveness of a heuristic before proceeding with its implementation, especially for very complicated heuristic procedures such as those underlying the proofs of Theorems 7 and 8. Our formal analysis of the effectiveness of the fork-decomposition heuristics using the methodology suggested and exploited by Helmert and Mattmüller was motivated primarily by this desire for formal evidence.

Given a planning domain $\mathcal{D}$ and heuristic $h$, Helmert and Mattmüller (2008) consider the *asymptotic performance ratio* of $h$ in $\mathcal{D}$. The goal is to find a value $\alpha(h, \mathcal{D}) \in [0, 1]$ such that

(1) for all states $s$ in all problems $\Pi \in \mathcal{D}$, $h(s) \geq \alpha(h, \mathcal{D}) \cdot h^*(s) + o(h^*(s))$, and

(2) there is a family of problems $\{\Pi_n\}_{n \in \mathbb{N}} \subseteq \mathcal{D}$ and solvable, non-goal states $\{s_n\}_{n \in \mathbb{N}}$ such that $s_n \in \Pi_n$, $\lim_{n \to \infty} h^*(s_n) = \infty$, and $h(s_n) \leq \alpha(h, \mathcal{D}) \cdot h^*(s_n) + o(h^*(s_n))$.

| Domain | $h^+$ | $h^k$ | $h^{\mathrm{PDB}}$ | $h^{\mathrm{PDB}}_{\mathrm{add}}$ | $h^{\mathfrak{F}}$ | $h^{\mathfrak{I}}$ | $h^{\mathfrak{FI}}$ |
|---|---|---|---|---|---|---|---|
| Gripper | 2/3 | 0 | 0 | 2/3 | 2/3 | 0 | 4/9 |
| Logistics | 3/4 | 0 | 0 | 1/2 | 1/2 | 1/2 | 1/2 |
| Blocksworld | 1/4 | 0 | 0 | 0 | 0 | 0 | 0 |
| Miconic-Strips | 6/7 | 0 | 0 | 1/2 | 5/6 | 1/2 | 1/2 |
| Satellite | 1/2 | 0 | 0 | 1/6 | 1/6 | 1/6 | 1/6 |

Table 4: Performance ratios of multiple heuristics in selected planning domains; ratios for $h^+$, $h^k$, $h^{\mathrm{PDB}}$, $h^{\mathrm{PDB}}_{\mathrm{add}}$ are by Helmert and Mattmüller (2008).

In other words, $h$ is *never* worse than $\alpha(,domain,\cdot)h^*$ (plus a sublinear term), and it can become as bad as $\alpha(h,\mathcal{D}) \cdot h^*$ (plus a sublinear term) for arbitrarily large inputs; note that both the existence and uniqueness of $\alpha(h,\mathcal{D})$ are guaranteed for any $h$ and $\mathcal{D}$.

Helmert and Mattmüller (2008) study the asymptotic performance ratio of some standard admissible heuristics on a set of well-known benchmark domains from the first four IPCs. Their results for Gripper, Logistics, Blocksworld, Miconic, and Satellite are shown in the first four columns of Table 4.

- The $h^+$ estimate corresponds to the optimal cost of solving the well-known "delete relaxation" of the original planning task, which is generally NP-hard to compute (Bylander, 1994).
- The $h^k$, $k \in \mathbb{N}^+$, family of heuristics is based on a relaxation where the cost of achieving a partial assignment is approximated by the highest cost of achieving its sub-assignment of size $k$ (Haslum & Geffner, 2000); computing $h^k$ is exponential only in $k$.
- The $h^{\mathrm{PDB}}$ and $h^{\mathrm{PDB}}_{\mathrm{add}}$ heuristics are regular (maximized over) and additive pattern database heuristics where the size of each pattern is assumed to be $O(\log(n))$ where $n = |V|$, and, importantly, the choice of the patterns is assumed to be optimal.

These results provide us with a baseline for evaluating our fork-decomposition heuristics $h^{\mathfrak{F}}$, $h^{\mathfrak{I}}$, and $h^{\mathfrak{FI}}$. First, however, Theorem 9 shows that these three heuristics are worth analyzing because each alone can be strictly more informative than the other two, depending on the planning task and/or the state being evaluated.[7]

**Theorem 9 (Undominance)** *Under uniform action cost partition, none of the heuristic functions $h^{\mathfrak{F}}$, $h^{\mathfrak{I}}$, and $h^{\mathfrak{FI}}$ dominates another.*

**Proof:** The proof is by example of two tasks, $\Pi^1$ and $\Pi^2$, which illustrate the following two cases: $h^{\mathfrak{F}}(I) > h^{\mathfrak{FI}}(I) > h^{\mathfrak{I}}(I)$ and $h^{\mathfrak{F}}(I) < h^{\mathfrak{FI}}(I) < h^{\mathfrak{I}}(I)$. These two tasks are defined over the same set of binary-valued variables $V = \{v_1, v_2, v_3, u_1, u_2, u_3\}$, have the same initial state $I = \{v_1 : 0, v_2 : 0, v_3 : 0, u_1 : 0, u_2 : 0, u_3 : 0\}$, and have the same goal

7. Theorem 9 is formulated and proven under the uniform action cost partition that we use throughout the paper, including the experiments. For per-step optimal action cost partitions (Katz & Domshlak, 2010), it is trivial to show that $h^{\mathfrak{FI}}$ dominates both $h^{\mathfrak{F}}$ and $h^{\mathfrak{I}}$ for all planning tasks.

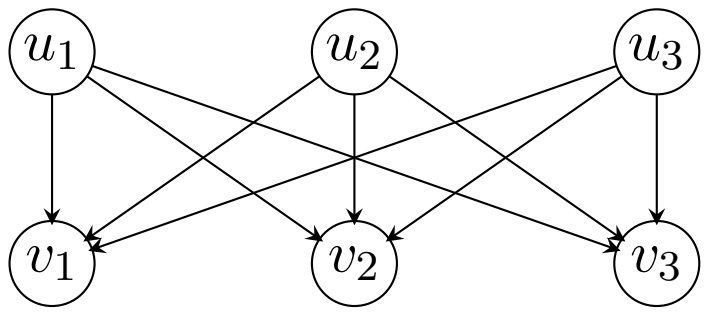

(a)

| | $A_1$ | $\mathbf{\Pi}^1_{\mathcal{F}}$ | $\mathbf{\Pi}^1_{\mathcal{I}}$ | $\mathbf{\Pi}^1_{\mathcal{FI}}$ |
|---|---|---|---|---|
| $a_1$ | $\langle\{v_1:0, u_1:0, u_2:0, u_3:0\}, \{v_1:1\}\rangle$ | 1/3 | 1 | 1/4 |
| $a_2$ | $\langle\{v_2:0, u_1:1, u_2:0, u_3:1\}, \{v_2:1\}\rangle$ | 1/3 | 1 | 1/4 |
| $a_3$ | $\langle\{v_3:0, u_1:1, u_2:1, u_3:0\}, \{v_3 1\}\rangle$ | 1/3 | 1 | 1/4 |
| $a_4$ | $\langle\{u_1:0\}, \{u_1:1\}\rangle$ | 1 | 1/3 | 1/4 |
| $a_5$ | $\langle\{u_1:1\}, \{u_1:0\}\rangle$ | 1 | 1/3 | 1/4 |
| $a_6$ | $\langle\{u_2:0\}, \{u_2:1\}\rangle$ | 1 | 1/3 | 1/4 |
| $a_7$ | $\langle\{u_2:1\}, \{u_2:0\}\rangle$ | 1 | 1/3 | 1/4 |
| $a_8$ | $\langle\{u_3:0\}, \{u_3:1\}\rangle$ | 1 | 1/3 | 1/4 |
| $a_9$ | $\langle\{u_3:1\}, \{u_3:0\}\rangle$ | 1 | 1/3 | 1/4 |

(c)

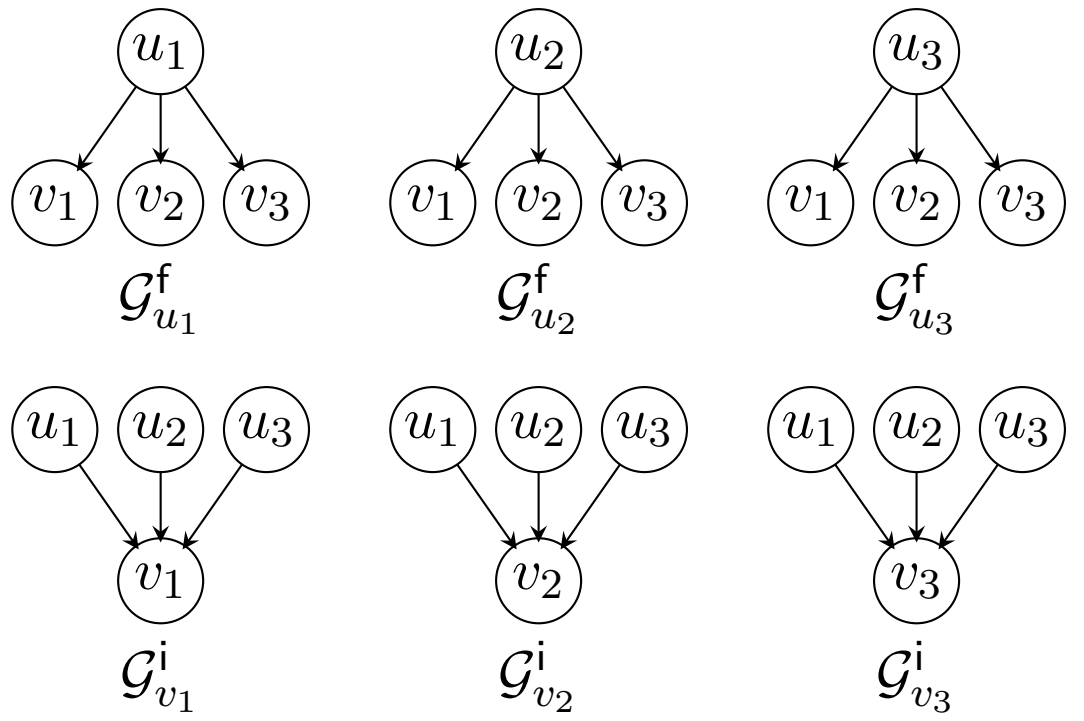

(b)

| | $A_2$ | $\mathbf{\Pi}^2_{\mathcal{F}}$ | $\mathbf{\Pi}^2_{\mathcal{I}}$ | $\mathbf{\Pi}^2_{\mathcal{FI}}$ |
|---|---|---|---|---|
| $a_1$ | $\langle\{v_1:0, u_1:1\}, \{v_1:1\}\rangle$ | 1/3 | 1 | 1/4 |
| $a_2$ | $\langle\{v_1:0, u_2:1\}, \{v_1:1\}\rangle$ | 1/3 | 1 | 1/4 |
| $a_3$ | $\langle\{v_1:0, u_3:1\}, \{v_1:1\}\rangle$ | 1/3 | 1 | 1/4 |
| $a_4$ | $\langle\{v_2:0, u_1:1\}, \{v_2:1\}\rangle$ | 1/3 | 1 | 1/4 |
| $a_5$ | $\langle\{v_2:0, u_2:1\}, \{v_2:1\}\rangle$ | 1/3 | 1 | 1/4 |
| $a_6$ | $\langle\{v_2:0, u_3:1\}, \{v_2:1\}\rangle$ | 1/3 | 1 | 1/4 |
| $a_7$ | $\langle\{v_3:0, u_1:1\}, \{v_3:1\}\rangle$ | 1/3 | 1 | 1/4 |
| $a_8$ | $\langle\{v_3:0, u_2:1\}, \{v_3:1\}\rangle$ | 1/3 | 1 | 1/4 |
| $a_9$ | $\langle\{v_3:0, u_3:1\}, \{v_3:1\}\rangle$ | 1/3 | 1 | 1/4 |
| $a_{10}$ | $\langle\{u_1:0\}, \{u_1:1\}\rangle$ | 1 | 1/3 | 1/4 |
| $a_{11}$ | $\langle\{u_2:0\}, \{u_2:1\}\rangle$ | 1 | 1/3 | 1/4 |
| $a_{12}$ | $\langle\{u_3:0\}, \{u_3:1\}\rangle$ | 1 | 1/3 | 1/4 |

(d)

Figure 8: Illustrations for the proof of Theorem 9: (a) causal graphs of $\Pi^1$ and $\Pi^2$, (b) fork and inverted fork subgraphs of the (same) causal graph of $\Pi^1$ and $\Pi^2$, and the action sets of (c) $\Pi^1$ and (d) $\Pi^2$, as well as the costs of the action representatives in each abstract problem along these subgraphs. Considering for example the first row of table (c), the action $a_1$ in $\Pi^1$ has a single representative in each of the three fork abstractions, as well as a representative in the inverted-fork abstraction $\Pi^1_{\mathcal{G}^{\mathsf{i}}_{v_1}}$. Hence, the cost of each of its representatives in $\mathcal{F}$-decomposition is 1/3, while the cost of its sole representative in $\mathcal{I}$-decomposition is 1.

$G = \{v_1:1, v_2:1, v_3:1\}$. The difference between $\Pi^1$ and $\Pi^2$ is in the action sets, listed in Figure 8c-d, with all the actions being unit-cost actions. The two tasks induce identical causal graphs, depicted in Figure 8a. Hence, the collections of $v$-forks and $v$-iforks of both tasks are also identical; these are depicted in Figure 8b. The fractional costs of the tasks' action representatives in the corresponding abstract problems are given in Figure 8c-d.

Figure 9 shows the optimal plans for all the abstract problems in $\mathcal{F}$-decompositions $\mathbf{\Pi}^1_{\mathcal{F}} = \{\Pi^1_{\mathcal{G}^{\mathsf{f}}_{u_1}}, \Pi^1_{\mathcal{G}^{\mathsf{f}}_{u_2}}, \Pi^1_{\mathcal{G}^{\mathsf{f}}_{u_3}}\}$ and $\mathbf{\Pi}^2_{\mathcal{F}} = \{\Pi^2_{\mathcal{G}^{\mathsf{f}}_{u_1}}, \Pi^2_{\mathcal{G}^{\mathsf{f}}_{u_2}}, \Pi^2_{\mathcal{G}^{\mathsf{f}}_{u_3}}\}$, $\mathcal{I}$-decompositions $\mathbf{\Pi}^1_{\mathcal{I}} = \{\Pi^1_{\mathcal{G}^{\mathsf{i}}_{v_1}}, \Pi^1_{\mathcal{G}^{\mathsf{i}}_{v_2}}, \Pi^1_{\mathcal{G}^{\mathsf{i}}_{v_3}}\}$ and $\mathbf{\Pi}^2_{\mathcal{I}} = \{\Pi^2_{\mathcal{G}^{\mathsf{i}}_{v_1}}, \Pi^2_{\mathcal{G}^{\mathsf{i}}_{v_2}}, \Pi^2_{\mathcal{G}^{\mathsf{i}}_{v_3}}\}$, and $\mathcal{FI}$-decompositions $\mathbf{\Pi}^1_{\mathcal{FI}} = \mathbf{\Pi}^1_{\mathcal{F}} \cup \mathbf{\Pi}^1_{\mathcal{I}}$ and $\mathbf{\Pi}^2_{\mathcal{FI}} = \mathbf{\Pi}^2_{\mathcal{F}} \cup \mathbf{\Pi}^2_{\mathcal{I}}$. The last column in both tables captures the estimates of the three heuristics for the initial states of $\Pi^1$ and $\Pi^2$, respectively. Together, these two cases show that none of the fork-decomposition heuristic functions $h^{\mathcal{F}}$, $h^{\mathcal{I}}$, and $h^{\mathcal{FI}}$ dominates any other, and, since all the

| $h$ | task | optimal plan | cost | $h(I)$ |
|---|---|---|---|---|
| $h^{\mathcal{F}}$ | $\Pi^1_{\mathcal{G}^{\mathsf{f}}_{u_1}}$ | $\langle a_1, a_4, a_2, a_3 \rangle$ | 2 | 6 |
| | $\Pi^1_{\mathcal{G}^{\mathsf{f}}_{u_2}}$ | $\langle a_1, a_2, a_6, a_3 \rangle$ | 2 | |
| | $\Pi^1_{\mathcal{G}^{\mathsf{f}}_{u_3}}$ | $\langle a_1, a_3, a_8, a_2 \rangle$ | 2 | |
| $h^{\mathcal{I}}$ | $\Pi^1_{\mathcal{G}^{\mathsf{i}}_{v_1}}$ | $\langle a_1 \rangle$ | 1 | $4\frac{1}{3}$ |
| | $\Pi^1_{\mathcal{G}^{\mathsf{i}}_{v_2}}$ | $\langle a_4, a_8, a_2 \rangle$ | 5/3 | |
| | $\Pi^1_{\mathcal{G}^{\mathsf{i}}_{v_3}}$ | $\langle a_4, a_6, a_3 \rangle$ | 5/3 | |
| $h^{\mathcal{FI}}$ | $\Pi^1_{\mathcal{G}^{\mathsf{f}}_{u_1}}$ | $\langle a_1, a_4, a_2, a_3 \rangle$ | 1 | $4\frac{3}{4}$ |
| | $\Pi^1_{\mathcal{G}^{\mathsf{f}}_{u_2}}$ | $\langle a_1, a_2, a_6, a_3 \rangle$ | 1 | |
| | $\Pi^1_{\mathcal{G}^{\mathsf{f}}_{u_3}}$ | $\langle a_1, a_3, a_8, a_2 \rangle$ | 1 | |
| | $\Pi^1_{\mathcal{G}^{\mathsf{i}}_{v_1}}$ | $\langle a_1 \rangle$ | 1/4 | |
| | $\Pi^1_{\mathcal{G}^{\mathsf{i}}_{v_2}}$ | $\langle a_4, a_8, a_2 \rangle$ | 3/4 | |
| | $\Pi^1_{\mathcal{G}^{\mathsf{i}}_{v_3}}$ | $\langle a_4, a_6, a_3 \rangle$ | 3/4 | |

(a)

| $h$ | task | optimal plan | cost | $h(I)$ |
|---|---|---|---|---|
| $h^{\mathcal{F}}$ | $\Pi^2_{\mathcal{G}^{\mathsf{f}}_{u_1}}$ | $\langle a_2, a_5, a_8 \rangle$ | 1 | 3 |
| | $\Pi^2_{\mathcal{G}^{\mathsf{f}}_{u_2}}$ | $\langle a_1, a_4, a_7 \rangle$ | 1 | |
| | $\Pi^2_{\mathcal{G}^{\mathsf{f}}_{u_3}}$ | $\langle a_1, a_4, a_7 \rangle$ | 1 | |
| $h^{\mathcal{I}}$ | $\Pi^2_{\mathcal{G}^{\mathsf{i}}_{v_1}}$ | $\langle a_{10}, a_1 \rangle$ | 4/3 | 4 |
| | $\Pi^2_{\mathcal{G}^{\mathsf{i}}_{v_2}}$ | $\langle a_{10}, a_4 \rangle$ | 4/3 | |
| | $\Pi^2_{\mathcal{G}^{\mathsf{i}}_{v_3}}$ | $\langle a_{10}, a_7 \rangle$ | 4/3 | |
| $h^{\mathcal{FI}}$ | $\Pi^2_{\mathcal{G}^{\mathsf{f}}_{u_1}}$ | $\langle a_2, a_5, a_8 \rangle$ | 3/4 | 15/4 |
| | $\Pi^2_{\mathcal{G}^{\mathsf{f}}_{u_2}}$ | $\langle a_1, a_4, a_7 \rangle$ | 3/4 | |
| | $\Pi^2_{\mathcal{G}^{\mathsf{f}}_{u_3}}$ | $\langle a_1, a_4, a_7 \rangle$ | 3/4 | |
| | $\Pi^2_{\mathcal{G}^{\mathsf{i}}_{v_1}}$ | $\langle a_{10}, a_1 \rangle$ | 1/2 | |
| | $\Pi^2_{\mathcal{G}^{\mathsf{i}}_{v_2}}$ | $\langle a_{10}, a_4 \rangle$ | 1/2 | |
| | $\Pi^2_{\mathcal{G}^{\mathsf{i}}_{v_3}}$ | $\langle a_{10}, a_7 \rangle$ | 1/2 | |

(b)

Figure 9: Illustrations for the proof of Theorem 9: Optimal plans for all the abstract problems of (a) $\Pi^1$, where we have $h^{\mathcal{F}}(I) > h^{\mathcal{FI}}(I) > h^{\mathcal{I}}(I)$, and (b) $\Pi^2$, where we have $h^{\mathcal{F}}(I) < h^{\mathcal{FI}}(I) < h^{\mathcal{I}}(I)$.

variables above are binary-valued, the claim holds in conjunction with *arbitrary* variable domain abstractions. □

One conclusion from Theorem 9 is that it is worth studying the asymptotic performance ratios for all three heuristics. The last three columns of Table 4 present our results for $h^{\mathcal{F}}$, $h^{\mathcal{I}}$, and $h^{\mathcal{FI}}$ for the Gripper, Logistics, Blocksworld, Miconic, and Satellite domains. We also studied the performance ratios of $\max\{h^{\mathcal{F}}, h^{\mathcal{I}}, h^{\mathcal{FI}}\}$, and in these five domains they appear to be identical to those of $h^{\mathcal{F}}$. (Note that "ratio of max" should not necessarily be identical to "max of ratios," and thus this analysis is worthwhile.) Taking a conservative position, the performance ratios for the fork-decomposition heuristics in Table 4 are "worst-case" in the sense that

(i) here we neither optimize the action cost partition (setting it to uniform as in the rest of the paper) nor eliminate clearly redundant abstractions, and

(ii) we use domain abstractions to (up to) ternary-valued abstract domains only.

The domains of the fork roots are all abstracted using the "leave-one-out" binary-valued domain decompositions as in Eq. 16 while the domains of the inverted-fork sinks are all abstracted using the "distance-from-initial-value" ternary-valued domain decompositions as in Eq. 15.

Overall, the results for fork-decomposition heuristics in Table 4 are gratifying. First, note that the performance ratios for $h^k$ and $h^{\mathrm{PDB}}$ are all 0. This is because every subgoal set of size $k$ (for $h^k$) and size $\log(n)$ (for $h^{\mathrm{PDB}}$) can be reached in the number of steps that only depends on $k$ (respectively, $\log(n)$), and not $n$, while $h^*(s_n)$ grows linearly in $n$ in all the five domains. This leaves us with $h^{\mathrm{PDB}}_{\mathrm{add}}$ being the only state-of-the-art (tractable and) admissible heuristic to compare with. Table 4 shows that the asymptotic performance ratio of $h^{\mathfrak{F}}$ heuristic is at least as good as that of $h^{\text{PDB}}_{\text{add}}$ in *all* five domains, while $h^{\mathfrak{F}}$ is superior to $h^{\text{PDB}}_{\text{add}}$ in Miconic, getting here quite close to $h^{+}$. When comparing $h^{\text{PDB}}_{\text{add}}$ and fork-decomposition heuristics, it is crucial to recall that the ratios devised by Helmert and Mattmüller for $h^{\text{PDB}}_{\text{add}}$ are with respect to optimal, manually-selected set of patterns. By contrast, the selection of variable subsets for fork-decomposition heuristics is completely nonparametric, and thus requires no tuning of the abstraction-selection process.

In the rest of the section we prove these asymptotic performance ratios of $h^{\mathfrak{F}}$, $h^{\mathfrak{I}}$, and $h^{\mathfrak{FI}}$ in Table 4 for the five domains. We begin with a very brief outline of how the results are obtained. *Some* familiarity with the domains is assumed. Next, each domain is addressed in detail: we provide an informal domain description as well as its SAS$^{+}$ representation, and then prove lower and upper bounds on the ratios for all three heuristics.

**Gripper** Assuming $n > 0$ balls should be moved from one room to another, all three heuristics $h^{\mathfrak{F}}, h^{\mathfrak{I}}, h^{\mathfrak{FI}}$ account for all the required pickup and drop actions, and only for $O(1)$-portion of move actions. However, the former actions are responsible for 2/3 of the optimal-plan length (= cost). Now, with the basic uniform action-cost partition, $h^{\mathfrak{F}}$, $h^{\mathfrak{I}}$, and $h^{\mathfrak{FI}}$ account for the whole, $O(1/n)$, and 2/3 of the total pickup/drop actions cost, respectively, providing the ratios in Table 4.[8]

**Logistics** An optimal plan contains at least as many load/unload actions as move actions, and all three heuristics $h^{\mathfrak{F}}, h^{\mathfrak{I}}, h^{\mathfrak{FI}}$ fully account for the former, providing a lower bound of 1/2. An instance on which all three heuristics achieve exactly 1/2 consists of two trucks $t_1, t_2$, no airplanes, one city, and $n$ packages such that the initial and goal locations of all the packages and trucks are all pair-wise different.

**Blocksworld** Arguments similar to those of Helmert and Mattmüller (2008) for $h^{\text{PDB}}_{\text{add}}$.

**Miconic** All three heuristics fully account for all the loads/unload actions. In addition, $h^{\mathfrak{F}}$ accounts for the full cost of all the move actions to the passengers' initial locations, and for half of the cost of all the other move actions. This provides us with lower bounds of 1/2 and 5/6, respectively. Tightness of 1/2 for $h^{\mathfrak{I}}$ and $h^{\mathfrak{FI}}$ is shown on a task consisting of $n$ passengers, $2n + 1$ floors, and all the initial and goal locations being pair-wise different. Tightness of 5/6 for $h^{\mathfrak{F}}$ is shown on a task consisting of $n$ passengers, $n+1$ floors, the elevator and all the passengers are initially at floor $n+1$, and each passenger $i$ wishes to get to floor $i$.

**Satellite** The length of an optimal plan for a problem with $n$ images to be taken and $k$ satellites to be moved to some end-positions is $\leq 6n + k$. All three heuristics fully account for all the image-taking actions and one satellite-moving action per satellite as above, providing a lower bound of $\frac{1}{6}$. Tightness of 1/6 for all three heuristics is shown on a task as follows: Two satellites with instruments $\{i\}_{i=1}^{l}$ and $\{i\}_{i=l+1}^{2l}$, respectively, where $l = n - \sqrt{n}$. Each pair of instruments $\{i, l+i\}$ can take images in modes $\{m_0, m_i\}$. There is a set of directions $\{d_j\}_{j=0}^{n}$ and a set of image objectives

8. We note that a very slight modification of the uniform action-cost partition results in a ratio of 2/3 for all three heuristics. Such optimizations, however, are outside of our scope here.

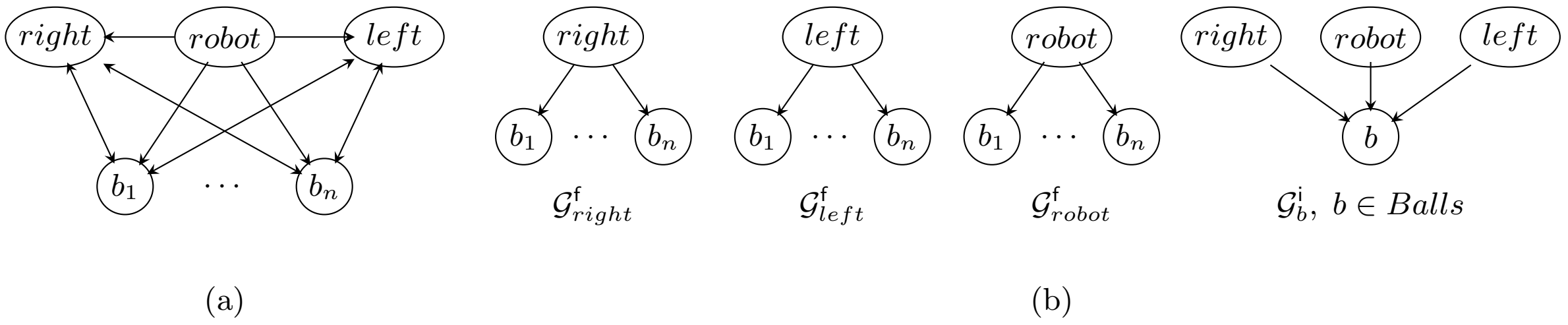

Figure 10: Gripper's (a) causal graph and (b) the corresponding collection of $v$-forks and $v$-iforks

$\{o_i\}_{i=1}^{n}$ such that, for $1 \leq i \leq l$, $o_i = (d_0, m_i)$ and, for $l < i \leq n$, $o_i = (d_i, m_0)$. Finally, the calibration direction for each pair of instruments $\{i, l+i\}$ is $d_i$.

### 9.1 Gripper

The domain consists of one robot *robot* with two arms $Arms = \{right, left\}$, two rooms $Rooms = \{r1, r2\}$, and a set *Balls* of $n$ balls. The robot can pick up a ball with an arm $arm \in Arms$ if *arm* is empty, release a ball $b \in Balls$ from the arm *arm* if *arm* currently holds $b$, and move from one room to another. All balls and the robot are initially in room $r1$, both arms are empty, and the goal is to move all the balls to room $r2$. A natural description of this planning task in SAS$^+$ is as follows.

- Variables $V = \{robot\} \bigcup Arms \bigcup Balls$ with domains

$$\mathcal{D}(robot) = Rooms$$
$$\mathcal{D}(left) = \mathcal{D}(right) = Balls \cup \{\mathsf{empty}\}$$
$$\forall b \in Balls: \ \mathcal{D}(b) = Rooms \cup \{\mathsf{robot}\}.$$

- Initial state $I = \{b{:}\mathsf{r1} \mid b \in Balls\} \cup \{robot{:}\mathsf{r1}, right{:}\mathsf{empty}, left{:}\mathsf{empty}\}$.

- Goal $G = \{b{:}\mathsf{r2} \mid b \in Balls\}$.

- Actions

$$A = \{Move(r, r') \mid \{r, r'\} \subseteq Rooms\} \bigcup \{Pickup(b, arm, r), Drop(b, arm, r) \mid b \in Balls, arm \in Arms, r \in Rooms\},$$

  where

  - move robot: $Move(r, r') = \langle\{robot{:}\mathsf{r}\}, \{robot{:}\mathsf{r}'\}\rangle$,
  - pickup ball: $Pickup(b, arm, r) = \langle\{b{:}\mathsf{r}, arm{:}\mathsf{empty}, robot{:}\mathsf{r}\}, \{b{:}\mathsf{robot}, arm{:}\mathsf{b}\}\rangle$, and
  - drop ball: $Drop(b, arm, r) = \langle\{b{:}\mathsf{robot}, arm{:}\mathsf{b}, robot{:}\mathsf{r}\}, \{b{:}\mathsf{r}, arm{:}\mathsf{empty}\}\rangle$.

The (parametric in $n$) causal graph of this task is depicted in Figure 10a.

| Action | $\Pi^{\mathsf{f}}_{robot}$ | $\Pi^{\mathsf{f}}_{arm,\mathsf{empty}}$ | $\Pi^{\mathsf{f}}_{arm,b}$ | $\Pi^{\mathsf{f}}_{arm,b'}$ | $\Pi^{\mathsf{f}}_{arm',\vartheta}$ | $\Pi^{\mathsf{i}}_{b}$ | $\Pi^{\mathsf{i}}_{b'}$ | $\mathbf{\Pi}_{\mathcal{F}}$ | $\mathbf{\Pi}_{\mathcal{I}}$ | $\mathbf{\Pi}_{\mathcal{FI}}$ |
|---|---|---|---|---|---|---|---|---|---|---|
| $Move(r, r')$ | 1 | 0 | 0 | 0 | 0 | 1 | 1 | 1 | $\frac{1}{n}$ | $\frac{1}{n+1}$ |
| $Pickup(b, arm, r)$ | 1 | 2 | 2 | 1 | 1 | 2 | 1 | $\frac{1}{2n+5}$ | $\frac{1}{n+1}$ | $\frac{1}{3n+6}$ |
| $Drop(b, arm, r)$ | 1 | 2 | 2 | 1 | 1 | 2 | 1 | $\frac{1}{2n+5}$ | $\frac{1}{n+1}$ | $\frac{1}{3n+6}$ |

Table 5: Number of representatives for each original Gripper action in each abstract task, as well as the partition of the action costs between these representatives

| | |
|---|---|
| $\Pi^{\mathsf{f}}_{robot}$ | $Pickup(b, right, \mathsf{r1}) = \langle\{robot\!:\!\mathsf{r1}, b\!:\!\mathsf{r1}\}, \{b\!:\!\mathsf{robot}\}\rangle$ |
| $\Pi^{\mathsf{f}}_{right,\mathsf{empty}}$ | $Pickup(b, right, \mathsf{r1})^1 = \langle\{right\!:\!\mathsf{empty}\}, \{right\!:\!\mathsf{b}\}\rangle$,<br>$Pickup(b, right, \mathsf{r1})^2 = \langle\{right\!:\!\mathsf{b}, b\!:\!\mathsf{r1}\}, \{b\!:\!\mathsf{robot}\}\rangle$ |
| $\Pi^{\mathsf{f}}_{right,b}$ | $Pickup(b, right, \mathsf{r1})^1 = \langle\{right\!:\!\mathsf{empty}\}, \{right\!:\!\mathsf{b}\}\rangle$,<br>$Pickup(b, right, \mathsf{r1})^2 = \langle\{right\!:\!\mathsf{b}, b\!:\!\mathsf{r1}\}, \{b\!:\!\mathsf{robot}\}\rangle$ |
| $\Pi^{\mathsf{f}}_{right,b'}$ | $Pickup(b, right, \mathsf{r1}) = \langle\{right\!:\!\mathsf{b}, b\!:\!\mathsf{r1}\}, \{b\!:\!\mathsf{robot}\}\rangle$ |
| $\Pi^{\mathsf{f}}_{left,\vartheta}$ | $Pickup(b, right, \mathsf{r1}) = \langle\{right\!:\!\mathsf{b}, b\!:\!\mathsf{r1}\}, \{b\!:\!\mathsf{robot}\}\rangle$ |
| $\Pi^{\mathsf{i}}_{b}$ | $Pickup(b, right, \mathsf{r1})^1 = \langle\{right\!:\!\mathsf{empty}\}, \{right\!:\!b\}\rangle$,<br>$Pickup(b, right, \mathsf{r1})^2 = \langle\{right\!:\!\mathsf{b}, robot\!:\!\mathsf{r1}, b\!:\!\mathsf{r1}\}, \{b\!:\!\mathsf{robot}\}\rangle$ |
| $\Pi^{\mathsf{i}}_{b'}$ | $Pickup(b, right, \mathsf{r1}) = \langle\{right\!:\!\mathsf{empty}\}, \{right\!:\!\mathsf{b}\}\rangle$ |

Table 6: The sets of representatives of the original action $Pickup(b, right, \mathsf{r1})$ in the abstract tasks

### 9.1.1 Fork Decomposition

Since the variables $robot$, $right$, and $left$ have no goal value, the collection of $v$-forks and $v$-iforks is as in Figure 10b. The domains of inverted fork sinks are ternary valued. The domains of fork roots are abstracted as in Eq. 16 ("leave one out"), and thus

$$
\begin{aligned}
\mathbf{\Pi}_{\mathcal{F}} &= \{\Pi^{\mathsf{f}}_{robot}\} \cup \{\Pi^{\mathsf{f}}_{right,\vartheta}, \Pi^{\mathsf{f}}_{left,\vartheta} \mid \vartheta \in \{\mathsf{empty}\} \cup Balls\},\\
\mathbf{\Pi}_{\mathcal{I}} &= \{\Pi^{\mathsf{i}}_{b} \mid b \in Balls\},\\
\mathbf{\Pi}_{\mathcal{FI}} &= \{\Pi^{\mathsf{f}}_{robot}\} \cup \{\Pi^{\mathsf{f}}_{right,\vartheta}, \Pi^{\mathsf{f}}_{left,\vartheta} \mid \vartheta \in \{\mathsf{empty}\} \cup Balls\} \cup \{\Pi^{\mathsf{i}}_{b} \mid b \in Balls\}.
\end{aligned}
$$

For each original action, the number of its representatives in each abstract task, as well as the cost assigned to each such representative, are listed in Table 5. Table 6 illustrates derivation of these numbers via decomposition of an example action $Pickup(b, right, \mathsf{r1})$ in each of the fork decomposition abstractions. That action has one nonredundant representative in $\Pi^{\mathsf{f}}_{robot}$, two such representatives in each of $\Pi^{\mathsf{f}}_{right,\mathsf{empty}}$ and $\Pi^{\mathsf{f}}_{right,b}$, one representative in each $\Pi^{\mathsf{f}}_{right,b'}$ for $b' \in Balls \setminus \{b\}$, one representative in each $\Pi^{\mathsf{f}}_{left,\vartheta}$ for $\vartheta \in Balls \cup \{\mathsf{empty}\}$, two representatives in $\Pi^{\mathsf{i}}_{b}$, and one representative in each $\Pi^{\mathsf{i}}_{b'}$ for $b' \in Balls \setminus \{b\}$. This results in cost $\frac{1}{2n+5}$ for each representative in $\mathbf{\Pi}_{\mathcal{F}}$, $\frac{1}{n+1}$ for each representative in $\mathbf{\Pi}_{\mathcal{I}}$, and $\frac{1}{3n+6}$ for each representative in $\mathbf{\Pi}_{\mathcal{FI}}$.

Given that, the optimal plans for the abstract tasks are as follows.

| $h$ | task | optimal plan | cost | # | $h(I)$ |
|---|---|---|---|---|---|
| $h^{\mathfrak{F}}$ | $\Pi^{\mathsf{f}}_{robot}$ | $\langle Pickup(b_1, right, \mathsf{r1}), \ldots, Pickup(b_n, right, \mathsf{r1}), Move(\mathsf{r1}, \mathsf{r2}), Drop(b_1, right, \mathsf{r2}), \ldots, Drop(b_n, right, \mathsf{r2})\rangle$ | $\frac{4n+5}{2n+5}$ | 1 | $2n - \frac{2n-5}{2n+5}$ |
| | $\Pi^{\mathsf{f}}_{right,\vartheta}$ | $\langle Pickup(b_1, left, \mathsf{r1}), \ldots, Pickup(b_n, left, \mathsf{r1}), Drop(b_1, left, \mathsf{r2}), \ldots, Drop(b_n, left, \mathsf{r2})\rangle$ | $\frac{2n}{2n+5}$ | $n+1$ | |
| | $\Pi^{\mathsf{f}}_{left,\vartheta}$ | $\langle Pickup(b_1, right, \mathsf{r1}), \ldots, Pickup(b_n, right, \mathsf{r1}), Drop(b_1, right, \mathsf{r2}), \ldots, Drop(b_n, right, \mathsf{r2})\rangle$ | $\frac{2n}{2n+5}$ | $n+1$ | |
| $h^{\mathfrak{I}}$ | $\Pi^{\mathsf{i}}_{b}$ | $\langle Pickup(b, right, \mathsf{r1})^1, Pickup(b, right, \mathsf{r1})^2, Move(\mathsf{r1}, \mathsf{r2}), Drop(b, left, \mathsf{r2})^2\rangle$ | $\frac{3}{n+1} + \frac{1}{n}$ | $n$ | $\frac{4n+1}{n+1}$ |
| $h^{\mathfrak{FI}}$ | $\Pi^{\mathsf{f}}_{robot}$ | $\langle Pickup(b_1, right, \mathsf{r1}), \ldots, Pickup(b_n, right, \mathsf{r1}), Move(\mathsf{r1}, \mathsf{r2}), Drop(b_1, right, \mathsf{r2}), \ldots, Drop(b_n, right, \mathsf{r2})\rangle$ | $\frac{2n}{3n+6} + \frac{1}{n+1}$ | 1 | $\frac{4n}{3} + \frac{4n+6}{3n+6}$ |
| | $\Pi^{\mathsf{f}}_{right,\vartheta}$ | $\langle Pickup(b_1, left, \mathsf{r1}), \ldots, Pickup(b_n, left, \mathsf{r1}), Drop(b_1, left, \mathsf{r2}), \ldots, Drop(b_n, left, \mathsf{r2})\rangle$ | $\frac{2n}{3n+6}$ | $n+1$ | |
| | $\Pi^{\mathsf{f}}_{left,\vartheta}$ | $\langle Pickup(b_1, right, \mathsf{r1}), \ldots, Pickup(b_n, right, \mathsf{r1}), Drop(b_1, right, \mathsf{r2}), \ldots, Drop(b_n, right, \mathsf{r2})\rangle$ | $\frac{2n}{3n+6}$ | $n+1$ | |
| | $\Pi^{\mathsf{i}}_{b}$ | $\langle Pickup(b, right, \mathsf{r1})^1, Pickup(b, right, \mathsf{r1})^2, Move(\mathsf{r1}, \mathsf{r2}), Drop(b, left, \mathsf{r2})^2\rangle$ | $\frac{3}{3n+6} + \frac{1}{n+1}$ | $n$ | |

Assuming $n > 0$ balls should be moved from one room to another, the cost of the optimal plan for the original task is $3n - 1$ when $n$ is even, and $3n$ when $n$ is odd. Therefore, the asymptotic performance ratios for the heuristics $h^{\mathfrak{F}}, h^{\mathfrak{I}}, h^{\mathfrak{FI}}$ on Gripper are 2/3, 0, and 4/9, respectively.

## 9.2 Logistics

Each Logistics task consists of some $k$ cities, $x$ airplanes, $y$ trucks and $n$ packages. Each city $i$ is associated with a set $L_i = \{l_i^1 \ldots, l_i^{\alpha_i}\}$ of locations within that city; the union of the locations of all the cities is denoted by $L = \bigcup_{i=1}^{k} L_i$. In addition, precisely one location in each city is an airport, and the set of airports is $L^A = \{l_1^1 \ldots, l_k^1\} \subseteq L$. Each truck can move only within the city in which it is located, and airplanes can fly between airports. The airplanes are denoted by $U = \{u_1, \ldots, u_x\}$, the trucks by $T = \{t_1, \ldots, t_y\}$, and the packages by $P = \{p_1, \ldots, p_n\}$. Let $T_i = \{t \in T \mid I[t] \in L_i\}$ denote the trucks of city $i$, and $P = P_1 \cup P_2 \cup P_3 \cup P_4 \cup P_5$ denote a partition of the packages as follows:

- each package in $P_1 = \{p \in P \mid I[p], G[p] \in L^A\}$ is both initially at an airport and needs to be moved to another airport,
- each package in $P_2 = \{p \in P \mid I[p] \in L^A \cap L_i, G[p] \in L_j \setminus L^A, i \neq j\}$ is initially at an airport and needs to be moved to a non-airport location in another city,
- each package in $P_3 = \{p \in P \mid I[p] \in L_i, G[p] \in L_i\}$ needs to be moved within one city,
- each package in $P_4 = \{p \in P \mid I[p] \in L_i \setminus L^A, G[p] \in L^A \setminus L_i\}$ needs to be moved from a non-airport location in one city to the airport of some other city, and
- each package in $P_5 = \{p \in P \mid I[p] \in L_i \setminus L^A, G[p] \in L_j \setminus L^A, i \neq j\}$ needs to be moved from a non-airport location in one city to a non-airport location in another city.

A natural Logistics task description in $\textsc{sas}^+$ is as follows.

- Variables $V = U \cup T \cup P$ with domains

$$\forall u \in U : \mathcal{D}(u) = L^A,$$
$$\forall 1 \leq i \leq k, \forall t \in T_i : \mathcal{D}(t) = L_i,$$
$$\forall p \in P : \mathcal{D}(p) = L \cup U \cup T.$$

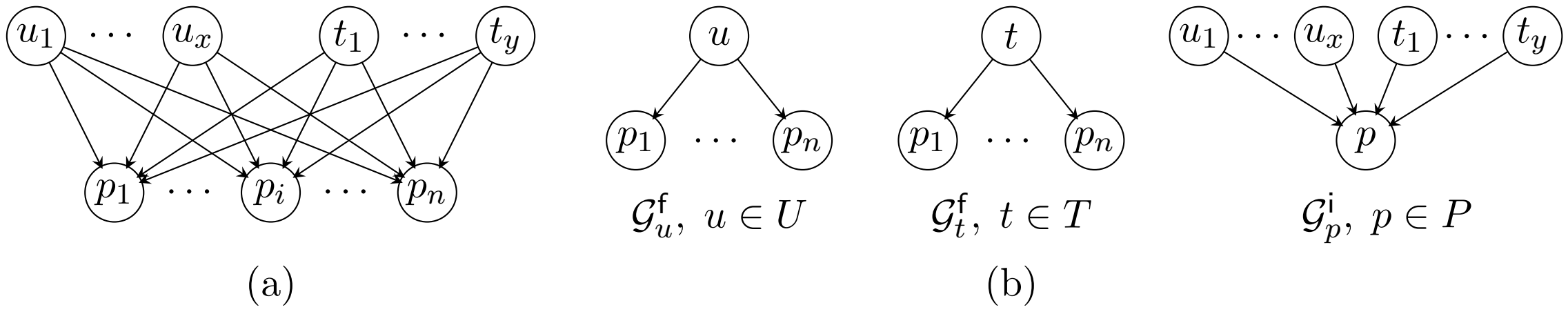

Figure 11: Logistics's (a) causal graph and (b) the corresponding collection of $v$-forks and $v$-iforks

- Initial state $I \in (L^A)^x \times L_1 \times \cdots \times L_k \times (L)^n$.
- Goal $G = \{p_1 : l_1, \ldots, p_n : l_n\} \in (L)^n$.
- Actions

$$A = \bigcup_{i=1}^{k} \bigcup_{l \in L_i} \bigcup_{t \in T_i} \left[ \{Lt(p,t,l), Ut(p,t,l) \mid p \in P\} \cup \{Mt(t,l,l') \mid, l' \in L_i \setminus \{l\}\} \right]$$
$$\cup \bigcup_{l \in L^A} \bigcup_{u \in U} \left[ \{La(p,u,l), Ua(p,u,l) \mid p \in P\} \cup \{Ma(u,l,l') \mid l' \in L^A \setminus \{l\}\} \right],$$

where

  - load package $p$ onto truck $t$ in location $l$: $Lt(p,t,l) = \langle \{p : l, t : l\}, \{p : t\} \rangle$,
  - unload package $p$ from truck $t$ in location $l$: $Ut(p,t,l) = \langle \{p : t, t : l\}, \{p : l\} \rangle$,
  - move truck $t$ from location $l$ to location $l'$: $Mt(t,l,l') = \langle \{t : l\}, \{t : l'\} \rangle$,
  - load package $p$ onto airplane $u$ in $l$: $La(p,u,l) = \langle \{p : l, u : l\}, \{p : u\} \rangle$,
  - unload package $p$ from airplane $u$ into $l$: $Ua(p,u,l) = \langle \{p : u, u : l\}, \{p : l\} \rangle$, and
  - move airplane $u$ from location $l$ to $l'$: $Ma(u,l,l') = \langle \{u : l\}, \{u : l'\} \rangle$.

The (parametrized in $n$, $x$, and $y$) causal graph of Logistics tasks is depicted in Figure 11a.

### 9.2.1 Fork Decomposition

Since the variables $u \in U$ and $t \in T$ have no goal value, the collection of $v$-forks and $v$-iforks is as in Figure 11b. The domains of the inverted-fork sinks are all abstracted as in Eq. 15 (“distance-from-initial-value”), while the domains of the fork roots are abstracted

| Action | $\Pi^{\mathsf{f}}_{u,l}$ | $\Pi^{\mathsf{f}}_{u,l'}$ | $\Pi^{\mathsf{f}}_{u,l''}$ | $\Pi^{\mathsf{f}}_{u',l}$ | $\Pi^{\mathsf{f}}_{t,l}$ | $\Pi^{\mathsf{f}}_{t,l'}$ | $\Pi^{\mathsf{f}}_{t,l''}$ | $\Pi^{\mathsf{f}}_{t',l}$ | $\Pi^{\mathsf{i}}_{p,m}$ | $\mathbf{\Pi}_{\mathcal{F}}$ | $\mathbf{\Pi}_{\mathcal{I}}$ | $\mathbf{\Pi}_{\mathcal{FI}}$ |
|---|---|---|---|---|---|---|---|---|---|---|---|---|
| $Mt(t,l,l')$ | 0 | 0 | 0 | 0 | 1 | 1 | 0 | 0 | 1 | $\frac{1}{2}$ | $\frac{1}{n^{\mathsf{i}}}$ | $\frac{1}{2+n^{\mathsf{i}}}$ |
| $Ma(u,l,l')$ | 1 | 1 | 0 | 0 | 0 | 0 | 0 | 0 | 1 | $\frac{1}{2}$ | $\frac{1}{n^{\mathsf{i}}}$ | $\frac{1}{2+n^{\mathsf{i}}}$ |

(a)

| | | | | | $I[p] \in L^A \cap L_i$ | | | | $I[p] \in L_i \setminus L^A$ | | | | | | | | |
|---|---|---|---|---|---|---|---|---|---|---|---|---|---|---|---|---|---|
| | | | | | $p \in P_1$ | $p \in P_2$ | | $p \in P_3$ | $p \in P_3$ | $p \in P_4$ | | $p \in P_5$ | | | | | |
| Action | | $\Pi^{\mathsf{f}}_{u,l}$ | $\Pi^{\mathsf{f}}_{t,l}$ | $\Pi^{\mathsf{i}}_{p',m}$ | $\Pi^{\mathsf{i}}_{p,1}$ | $\Pi^{\mathsf{i}}_{p,1}$ | $\Pi^{\mathsf{i}}_{p,2}$ | $\Pi^{\mathsf{i}}_{p,1}$ | $\Pi^{\mathsf{i}}_{p,1}$ | $\Pi^{\mathsf{i}}_{p,1}$ | $\Pi^{\mathsf{i}}_{p,2}$ | $\Pi^{\mathsf{i}}_{p,1}$ | $\Pi^{\mathsf{i}}_{p,2}$ | $\Pi^{\mathsf{i}}_{p,3}$ | $\mathbf{\Pi}_{\mathcal{F}}$ | $\mathbf{\Pi}_{\mathcal{I}}$ | $\mathbf{\Pi}_{\mathcal{FI}}$ |
| $Lt(p,t,l)$, $Ut(p,t,l)$ | $l \in L_i$ | 1 | 1 | 0 | 1 | 1 | 0 | 1 | 1 | 1 | 0 | 1 | 0 | 0 | $\frac{1}{n^{\mathsf{f}}}$ | 1 | $\frac{1}{n^{\mathsf{f}}+1}$ |
| | $l \in L_j$ | 1 | 1 | 0 | 0 | 0 | 1 | 0 | 0 | 0 | 0 | 0 | 0 | 1 | $\frac{1}{n^{\mathsf{f}}}$ | 1 | $\frac{1}{n^{\mathsf{f}}+1}$ |
| $La(p,u,l)$, $Ua(p,u,l)$ | | 1 | 1 | 0 | 1 | 1 | 0 | 1 | 0 | 0 | 1 | 0 | 1 | 0 | $\frac{1}{n^{\mathsf{f}}}$ | 1 | $\frac{1}{n^{\mathsf{f}}+1}$ |

(b)

Figure 12: Number of representatives of each original LOGISTICS action in each abstract task, as well as the partition of the action costs between these representatives; tables (a) and (b) capture the move and load/unload actions, respectively

as in Eq. 16 ("leave-one-out"). Thus, we have

$$\mathbf{\Pi}_{\mathcal{F}} = \bigcup_{u \in U} \bigcup_{l \in L^A} \{\Pi^{\mathsf{f}}_{u,l}\} \cup \bigcup_{i=1}^{k} \bigcup_{t \in T_i} \bigcup_{l \in L_i} \{\Pi^{\mathsf{f}}_{t,l}\},$$

$$\mathbf{\Pi}_{\mathcal{I}} = \bigcup_{p \in P} \{\Pi^{\mathsf{i}}_{p,1}\} \cup \bigcup_{p \in P_2 \cup P_4 \cup P_5} \{\Pi^{\mathsf{i}}_{p,2}\} \cup \bigcup_{p \in P_5} \{\Pi^{\mathsf{i}}_{p,3}\},$$

$$\mathbf{\Pi}_{\mathcal{FI}} = \bigcup_{u \in U} \bigcup_{l \in L^A} \{\Pi^{\mathsf{f}}_{u,l}\} \cup \bigcup_{i=1}^{k} \bigcup_{t \in T_i} \bigcup_{l \in L_i} \{\Pi^{\mathsf{f}}_{t,l}\} \cup \bigcup_{p \in P} \{\Pi^{\mathsf{i}}_{p,1}\} \cup \bigcup_{p \in P_2 \cup P_4 \cup P_5} \{\Pi^{\mathsf{i}}_{p,2}\} \cup \bigcup_{p \in P_5} \{\Pi^{\mathsf{i}}_{p,3}\}.$$

The total number of forks is $n^{\mathsf{f}} = |\mathbf{\Pi}_{\mathcal{F}}| = |U| \cdot |L^A| + \sum_{i=1}^{k} |T_i| \cdot |L_i|$, and the total number of inverted forks is $n^{\mathsf{i}} = |\mathbf{\Pi}_{\mathcal{I}}| = |P_1| + 2 \cdot |P_2| + |P_3| + 2 \cdot |P_4| + 3 \cdot |P_5|$. For each action $a \in A$, the number of its representatives in each abstract task, as well as the cost assigned to each such representative, are given in Figure 12. Each row in the tables of Figure 12 corresponds to a certain LOGISTICS action, each column (except for the last three) represents an abstract task, and each entry captures the number of representatives an action has in the corresponding task. The last three columns show the portion of the total cost that is given to an action representative in each task, in each of the three heuristics in question.

### 9.2.2 LOWER BOUND

Note that any optimal plan for a LOGISTICS task contains at least as many load/unload actions as move actions. Thus, the following lemma provides us with the lower bound of 1/2 for all three heuristics in question.

**Lemma 1** *For any* Logistics *task,* $h^{\mathfrak{F}}$*,* $h^{\mathfrak{I}}$*, and* $h^{\mathfrak{FI}}$ *account for the full cost of the load/unload actions required by any optimal plan for that task.*

**Proof:** For any Logistics task, all the optimal plans for that task contain the same amount of load/unload actions for each package $p \in P$ as follows.

**$p \in P_1$:** 2 actions — one load onto an airplane, and one unload from that airplane,

**$p \in P_2$:** 4 actions — one load onto an airplane, one unload from that airplane, one load onto a truck, and one unload from that truck,

**$p \in P_3$:** 2 actions — one load onto a truck, and one unload from that truck,

**$p \in P_4$:** 4 actions — one load onto a truck, one unload from that truck, one load onto an airplane, and one unload from that airplane, and

**$p \in P_5$:** 6 actions — two loads onto some trucks, two unloads from these trucks, one load onto an airplane, and one unload from that airplane.

Consider the fork-decomposition $\mathbf{\Pi}_{\mathfrak{F}}$. Any optimal plan for each of the abstract tasks will contain the number of load/unload actions exactly as above (the effects of these actions remain unchanged in these tasks). The cost of each representative of each load/unload action is $\frac{1}{n^{\mathsf{f}}}$, and there are $n^{\mathsf{f}}$ abstract tasks. Therefore, the heuristic $h^{\mathfrak{F}}$ fully accounts for the cost of the required load/unload actions.

Now consider the fork-decomposition $\mathbf{\Pi}_{\mathfrak{I}}$. With $m$ being the domain-decomposition index of the abstraction, any optimal plan for the abstract task $\Pi^{\mathsf{i}}_{p,m}$ will include one load and one unload actions as follows.

**$p \in P_1$:** one load onto an airplane and one unload from that airplane,

**$p \in P_2, m=1$:** one load onto an airplane and one unload from that airplane,

**$p \in P_2, m=2$:** one load onto a truck and one unload from that truck,

**$p \in P_3$:** one load onto a truck and one unload from that truck,

**$p \in P_4, m=1$:** one load onto a truck and one unload from that truck,

**$p \in P_4, m=2$:** one load onto an airplane, and one unload from that airplane,

**$p \in P_5, m=1$:** one load onto a truck and one unload from that truck,

**$p \in P_5, m=2$:** one load onto an airplane and one unload from that airplane, and

**$p \in P_5, m=3$:** one load onto a truck and one unload from that truck.

The cost of each representative of load/unload actions is 1, and thus the heuristic $h^{\mathfrak{I}}$ fully accounts for the cost of the required load/unload actions.

Finally, consider the fork-decomposition $\mathbf{\Pi}_{\mathfrak{FI}}$. Any optimal plan for each of the fork-structured abstract tasks will contain the same number of load/unload actions as for $\mathbf{\Pi}_{\mathfrak{F}}$. The cost of each representative of load/unload actions is $\frac{1}{n^{\mathsf{f}}+1}$ and there are $n^{\mathsf{f}}$ such abstract tasks. In addition, each of these load/unload actions will also appear in exactly one inverted fork-structured abstract task. Therefore the heuristic $h^{\mathfrak{FI}}$ also fully accounts for the cost of the required load/unload actions. □

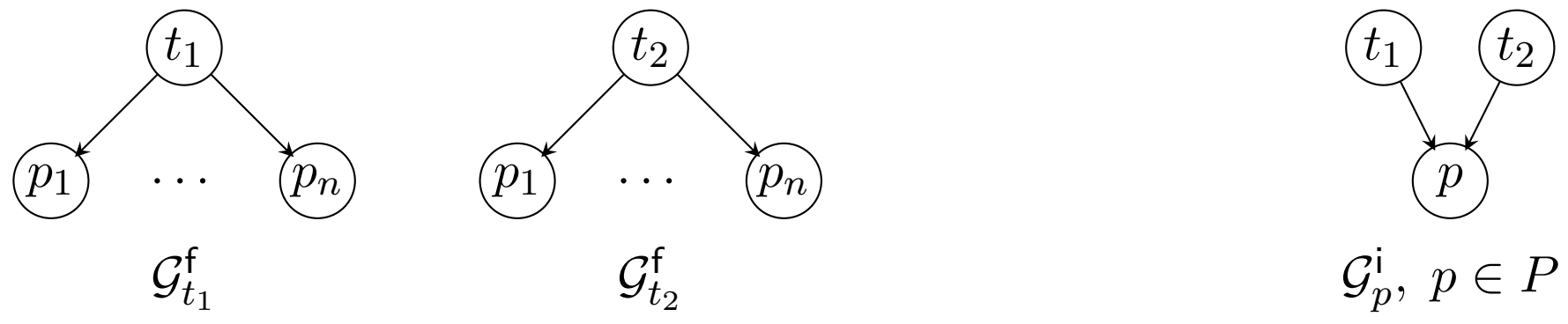

Figure 13: Collection of $v$-forks and $v$-iforks for the Logistics task used for the proof of the upper bound of 1/2

#### 9.2.3 Upper Bound

An instance on which all three heuristics achieve exactly 1/2 consists of two trucks $t_1, t_2$, no airplanes, one city, and $n$ packages such that the initial and goal locations of all the packages are all pairwise different, and both trucks are initially located at yet another location. More formally, if $L = \{l_i\}_{i=0}^{2n}$, and $T = \{t_1, t_2\}$, then the SAS$^+$ encoding for this Logistics task is as follows.

- Variables $V = \{t_1, t_2, p_1, \ldots, p_n\}$ with domains

$$\forall t \in T: \ \mathcal{D}(t) = L,$$
$$\forall p \in P: \ \mathcal{D}(p) = L \cup T.$$

- Initial state $I = \{t_1 : l_0, t_2 : l_0, p_1 : l_1, \ldots, p_n : l_n\}$.
- Goal $G = \{p_1 : l_{n+1}, \ldots, p_n : l_{2n}\}$.
- Actions $A = \{Lt(p, t, l), Ut(p, t, l) \mid l \in L, t \in T, p \in P\} \cup \{Mt(t, l, l') \mid t \in T, \{l, l'\} \subseteq L\}$.

The collection of $v$-forks and $v$-iforks for this task is depicted in Figure 13. The domains of the inverted-fork sinks are all abstracted as in Eq. 15 ("distance-from-initial-value"), while the domains of the fork roots are abstracted as in Eq. 16 ("leave-one-out"), and therefore we have

$$\mathbf{\Pi}_{\mathcal{F}} = \{\Pi^{\mathsf{f}}_{t_1,l}\Pi^{\mathsf{f}}_{t_2,l} \mid l \in L\},$$
$$\mathbf{\Pi}_{\mathcal{I}} = \{\Pi^{\mathsf{i}}_{p,1} \mid p \in P\},$$
$$\mathbf{\Pi}_{\mathcal{FI}} = \{\Pi^{\mathsf{f}}_{t_1,l}\Pi^{\mathsf{f}}_{t_2,l} \mid l \in L\} \cup \{\Pi^{\mathsf{i}}_{p,1} \mid p \in P\}.$$

The total number of forks is thus $n^{\mathsf{f}} = 4n + 2$ and the total number of inverted forks is $n^{\mathsf{i}} = n$. The partition of the action costs for Logistics tasks is described in Figure 12. Here we have $P = P_3$ and thus the action cost partition is as follows.

| Action | $\Pi^{\mathsf{f}}_{t,l}$ | $\Pi^{\mathsf{f}}_{t,l'}$ | $\Pi^{\mathsf{f}}_{t,l''}$ | $\Pi^{\mathsf{f}}_{t',l^*}$ | $\Pi^{\mathsf{i}}_{p,1}$ | $\Pi^{\mathsf{i}}_{p',1}$ | $\mathbf{\Pi}_{\mathcal{F}}$ | $\mathbf{\Pi}_{\mathcal{I}}$ | $\mathbf{\Pi}_{\mathcal{FI}}$ |
|---|---|---|---|---|---|---|---|---|---|
| $Mt(t, l, l')$ | 1 | 1 | 0 | 0 | 1 | 0 | $\frac{1}{2}$ | $\frac{1}{n}$ | $\frac{1}{n+2}$ |
| $Lt(p, t, l)$ | 1 | 1 | 1 | 1 | 1 | 0 | $\frac{1}{4n+2}$ | 1 | $\frac{1}{4n+3}$ |
| $Ut(p, t, l)$ | 1 | 1 | 1 | 1 | 1 | 0 | $\frac{1}{4n+2}$ | 1 | $\frac{1}{4n+3}$ |

Given that, the optimal plans for the abstract task are

| $h$ | task | optimal plan | cost | # | $h(I)$ |
|---|---|---|---|---|---|
| $h^{\mathcal{F}}$ | $\Pi^{\mathsf{f}}_{t_1,l}$ | $\langle Lt(p_1,t_2,l_1),\dots,Lt(p_n,t_2,l_n),Ut(p_1,t_2,l_{n+1}),\dots,Ut(p_n,t_2,l_{2n})\rangle$ | $\frac{2n}{4n+2}$ | $2n+1$ | $2n$ |
| | $\Pi^{\mathsf{f}}_{t_2,l}$ | $\langle Lt(p_1,t_1,l_1),\dots,Lt(p_n,t_1,l_n),Ut(p_1,t_1,l_{n+1}),\dots,Ut(p_n,t_1,l_{2n})\rangle$ | $\frac{2n}{4n+2}$ | $2n+1$ | |
| $h^{\mathcal{I}}$ | $\Pi^{\mathsf{i}}_{p_i,1}$ | $\langle Mt(t_1,l_0,l_i),Lt(p_i,t_1,l_i),Mt(t_1,l_i,l_{n+i}),Ut(p_i,t_1,l_{n+i})\rangle$ | $\frac{2}{n}+2$ | $n$ | $2n+2$ |
| $h^{\mathcal{FI}}$ | $\Pi^{\mathsf{f}}_{t_1,l}$ | $\langle Lt(p_1,t_2,l_1),\dots,Lt(p_n,t_2,l_n),Ut(p_1,t_2,l_{n+1}),\dots,Ut(p_n,t_2,l_{2n})\rangle$ | $\frac{2n}{4n+3}$ | $2n+1$ | $2n+\frac{2n}{n+2}$ |
| | $\Pi^{\mathsf{f}}_{t_2,l}$ | $\langle Lt(p_1,t_1,l_1),\dots,Lt(p_n,t_1,l_n),Ut(p_1,t_1,l_{n+1}),\dots,Ut(p_n,t_1,l_{2n})\rangle$ | $\frac{2n}{4n+3}$ | $2n+1$ | |
| | $\Pi^{\mathsf{i}}_{p_i,1}$ | $\langle Mt(t_1,l_0,l_i),Lt(p_i,t_1,l_i),Mt(t_1,l_i,l_{n+i}),Ut(p_i,t_1,l_{n+i})\rangle$ | $\frac{2}{n+2}+\frac{2}{4n+3}$ | $n$ | |

while an optimal plan for the original task, e.g., $\langle Mt(t_1,l_0,l_1),Lt(p_1,t_1,l_1),Mt(t_1,l_1,l_2),Lt(p_2,t_1,l_2),$ $Mt(t_1,l_2,l_3),\dots,Lt(p_n,t_1,l_n),Mt(t_1,l_n,l_{n+1}),Ut(p_1,t_1,l_{n+1}),Mt(t_1,l_{n+1},l_{n+2}),Ut(p_2,t_1,l_{n+2}),$ $Mt(t_1,l_{n+2},l_{n+3}),\dots,Ut(p_n,t_1,l_{2n})\rangle$, has the cost of $4n$, providing us with the upper bound of $1/2$ for all three heuristics. Putting our lower and upper bounds together, the asymptotic ratio of all three heuristics in question is $1/2$.

### 9.3 Blocksworld

Each Blocksworld task consists of a table $\mathsf{table}$, a crane $c$, and $n+1$ blocks $B = \{b_1,\dots,b_{n+1}\}$. Each block can be either on the table, or on top of some other block, or held by the crane. The crane can pick up a block if it currently holds nothing, and that block has no other block on top of it. The crane can drop the held block on the table or on top of some other block.

Consider now a Blocksworld task as follows. The blocks initially form a tower $b_1,\dots,b_n,b_{n+1}$ with $b_{n+1}$ being on the table, and the goal is to move them to form a tower $b_1,\dots,b_{n-1},b_{n+1},b_n$ with $b_n$ being on the table. That is, the goal is to swap the lowest two blocks of the tower. A natural description of this task in SAS$^+$ is as follows.

- Variables $V = \{b, clear_b \mid b \in B\} \cup \{c\}$ with domains

$$\begin{aligned} \mathcal{D}(c) &= \{\mathsf{empty}\} \cup B, \\ \forall b \in B:\ \mathcal{D}(b) &= \{\mathsf{table},\mathsf{c}\} \cup B \setminus \{b\}, \\ \mathcal{D}(clear_b) &= \{\mathsf{yes},\mathsf{no}\}. \end{aligned}$$

- Initial state

$$\begin{aligned} I = \ & \{c\!:\!\mathsf{empty}, b_{n+1}\!:\!\mathsf{table}, clear_{b_1}\!:\!\mathsf{yes}\} \bigcup \\ & \{b_i\!:\!b_{i+1} \mid 1 \le i \le n\} \bigcup \\ & \{clear_b\!:\!\mathsf{no} \mid b \in B \setminus \{b_1\}\}\,. \end{aligned}$$

- Goal $G = \{b_n\!:\!\mathsf{table}, b_{n+1}\!:\!b_n, b_{n-1}\!:\!b_{n+1}\} \cup \{b_i\!:\!b_{i+1} \mid 1 \le i \le n-2\}$.

- Actions $A = \{P_T(b), D_T(b) \mid b \in B\} \cup \{P(b,b'), D(b,b') \mid \{b,b'\} \subseteq B\}$ where
  - pick block $b$ from the table: $P_T(b) = \langle\{c\!:\!\mathsf{empty}, b\!:\!\mathsf{table}, clear_b\!:\!\mathsf{yes}\}, \{c\mathsf{b}, b\!:\!\mathsf{c}\}\rangle$,
  - pick block $b$ from block $b'$:
    $P(b,b') = \langle\{c\!:\!\mathsf{empty}, b\!:\!\mathsf{b'}, clear_b\!:\!\mathsf{yes}, clear_{b'}\!:\!\mathsf{no}\}, \{c\!:\!\mathsf{b}, b\!:\!\mathsf{c}, clear_{b'}\!:\!\mathsf{yes}\}\rangle$,

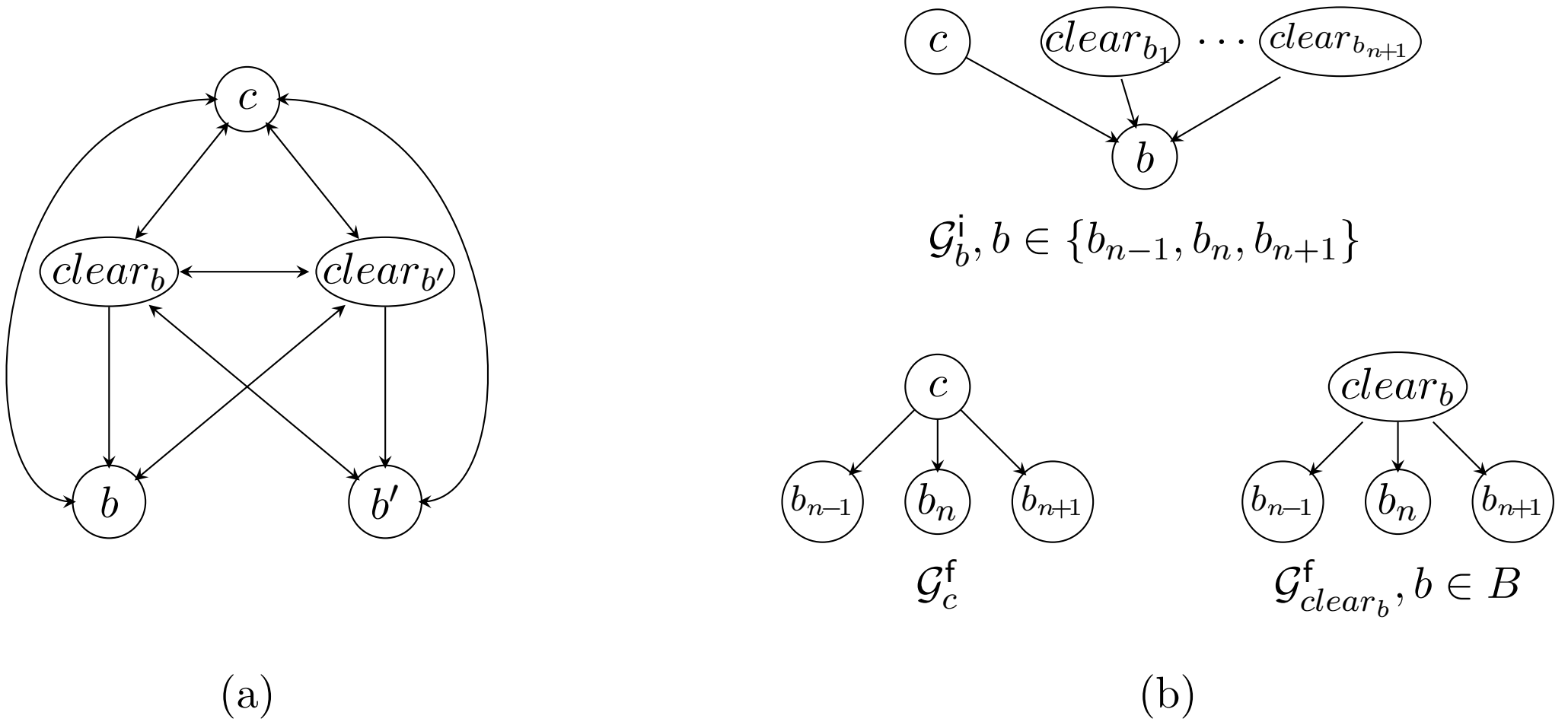

Figure 14: (a) Causal graph and (b) the corresponding collection of $v$-forks and $v$-iforks for the Blocksworld task used in the proof

- drop block $b$ on the table: $D_T(b) = \langle\{c\colon \mathsf{b}, b\colon \mathsf{c}\}, \{c\colon \mathsf{empty}, b\colon \mathsf{table}\}\rangle$, and
- drop block $b$ on block $b'$:
  $D(b, b') = \langle\{c\colon \mathsf{b}, b\colon \mathsf{c}, clear_{b'}\colon \mathsf{yes}\}, \{c\colon \mathsf{empty}, b\colon \mathsf{b}', clear_{b'}\colon \mathsf{no}\}\rangle$.

A schematic version of the causal graph of this task is depicted in Figure 14a. Since only the variables $b_{n-1}, b_n, b_{n+1}$ have goal values that are different from their values in the initial state, the collection of $v$-forks and $v$-iforks is as in Figure 14b. After the ("leave-one-out," Eq. 16) domain abstraction of the variable $c$, $c$-fork $\mathcal{G}^{\mathsf{f}}_c$ breaks down into $n+2$ abstract tasks. The sinks of $v$-iforks $\mathcal{G}^{\mathsf{i}}_{b_{n-1}}$, $\mathcal{G}^{\mathsf{i}}_{b_n}$, and $\mathcal{G}^{\mathsf{i}}_{b_{n+1}}$ also go through the process of domain decomposition ("distance-from-initial-value," Eq. 15). However, due to the structure of the domain transition graphs of the block variables, domain decomposition here results in only a single abstract task for each of the $v$-iforks. Thus we have

$$\begin{aligned}
\mathbf{\Pi}_{\mathcal{F}} &= \{\Pi^{\mathsf{f}}_{c,\mathsf{empty}}\} \cup \{\Pi^{\mathsf{f}}_{c,b} \mid b \in B\} \cup \{\Pi^{\mathsf{f}}_{clear_b} \mid b \in B\},\\
\mathbf{\Pi}_{\mathcal{I}} &= \{\Pi^{\mathsf{i}}_{b_{n-1},1}, \Pi^{\mathsf{i}}_{b_n,1}, \Pi^{\mathsf{i}}_{b_{n+1},1}\},\\
\mathbf{\Pi}_{\mathcal{F}\mathcal{I}} &= \{\Pi^{\mathsf{f}}_{c,\mathsf{empty}}\} \cup \{\Pi^{\mathsf{f}}_{c,b} \mid b \in B\} \cup \{\Pi^{\mathsf{f}}_{clear_b} \mid b \in B\} \cup \Pi^{\mathsf{i}}_{b_{n-1},1}, \Pi^{\mathsf{i}}_{b_n,1}, \Pi^{\mathsf{i}}_{b_{n+1},1}\}.
\end{aligned}$$

It is technically straightforward to verify that, for each abstract task in $\mathbf{\Pi}_{\mathcal{F}}$, $\mathbf{\Pi}_{\mathcal{I}}$, and $\mathbf{\Pi}_{\mathcal{F}\mathcal{I}}$, there exists a plan that (i) involves only the representatives of the actions

$$\{P(b_{n-1}, b_n), D_T(b_{n-1}), P(b_n, b_{n+1}), D_T(b_n), P_T(b_{n+1}), D(b_{n+1}, b_n), P_T(b_{n-1}), D(b_{n-1}, b_{n+1})\}, \tag{21}$$

and (ii) involves each representative of each original action at most once. Even if together these plans account for the total cost of all eight actions in Eq. 21, the total cost of all these plans (and thus the estimates of all the three heuristics) is upper-bounded by 8, while an optimal plan for the original task, e.g., $\langle P(b_1, b_2), D_T(b_1), P(b_2, b_3), D_T(b_2), \ldots, P(b_n, b_{n+1}), D_T(b_n),$ $P_T(b_{n+1}), D(b_{n+1}, b_n), P_T(b_{n-1}), D(b_{n-1}, b_{n+1}), P_T(b_{n-2}), D(b_{n-2}, b_{n-1}), \ldots, P_T(b_1), D(b_1, b_2)\rangle$, has a cost of $4n$. Hence, the asymptotic performance ratio of all three heuristics on the Blocksworld domain is 0.

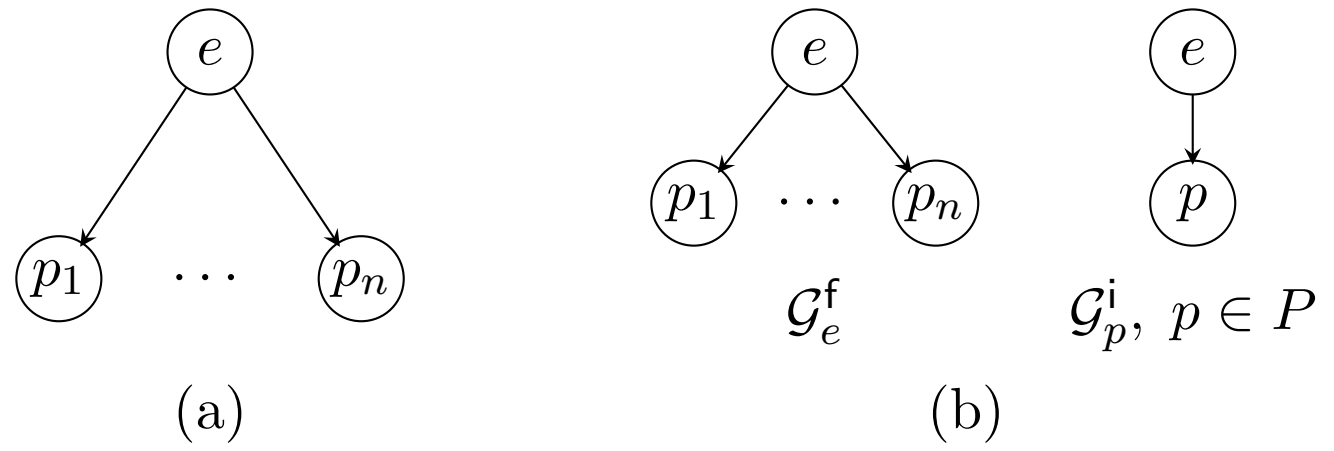

Figure 15: Miconic's (a) causal graph and (b) the corresponding collection of $v$-forks and $v$-iforks

### 9.4 Miconic

Each Miconic task consists of one elevator $e$, a set of floors $F$, and the passengers $P$. The elevator can move between $|F|$ floors and on each floor it can load and/or unload passengers. A natural SAS$^+$ description of a Miconic task is as follows.

- Variables $V = \{e\} \cup P$ with domains

$$\begin{aligned} &\mathcal{D}(e) = F, \\ \forall p \in P: \ &\mathcal{D}(p) = F \cup \{\mathsf{e}\}. \end{aligned}$$

- Initial state $I = \{e\!:\!f_e\} \cup \{p\!:\!f_p \mid p \in P\} \in (F)^{|P|+1}$.
- Goal $G = \{p\!:\!f'_p \mid p \in P\} \in (F)^{|P|}$.
- Actions $A = \{In(p,f), Out(p,f) \mid f \in F, p \in P\} \cup \{Move(f,f') \mid \{f,f'\} \subseteq F\}$, where
  - load passenger $p$ into $e$ on floor $f$: $In(p,f) = \langle\{e\!:\!f, p\!:\!f\}, \{p\!:\!\mathsf{e}\}\rangle$,
  - unload passenger $p$ from $e$ to floor $f$: $Out(p,f) = \langle\{e\!:\!f, p\!:\!\mathsf{e}\}, \{p\!:\!f\}\rangle$, and
  - move elevator from floor $f$ to floor $f'$: $Move(f,f') = \langle\{e\!:\!f\}, \{e\!:\!f'\}\rangle$.

The (parametrized in $n$) causal graph of Miconic tasks is depicted in Figure 15a, and Figure 15b depicts the corresponding collection of $v$-forks and $v$-iforks. The domains of the inverted-fork sinks are all abstracted as in Eq. 15 ("distance-from-initial-value"), and the domains of the fork roots are abstracted as in Eq. 16 ("leave-one-out"). Thus, we have

$$\begin{aligned} \mathbf{\Pi}_{\mathcal{F}} &= \{\Pi^{\mathsf{f}}_{e,f} \mid f \in F\}, \\ \mathbf{\Pi}_{\mathcal{I}} &= \{\Pi^{\mathsf{i}}_{p,1} \mid p \in P\}, \\ \mathbf{\Pi}_{\mathcal{FI}} &= \{\Pi^{\mathsf{f}}_{e,f} \mid f \in F\} \cup \{\Pi^{\mathsf{i}}_{p,1} \mid p \in P\}. \end{aligned}$$

The total number of the fork-structured abstract tasks is thus $n^{\mathsf{f}} = |\mathbf{\Pi}_{\mathcal{F}}| = |F|$ and the total number of the inverted fork structured abstract tasks is $n^{\mathsf{i}} = |\mathbf{\Pi}_{\mathcal{I}}| = |P|$. For each action $a \in A$, the number of its representatives in each abstract task, as well as the cost assigned to each such representative, are given in Table 7.

| Action | $\Pi^{\mathsf{f}}_{e,f}$ | $\Pi^{\mathsf{f}}_{e,f'}$ | $\Pi^{\mathsf{f}}_{e,f''}$ | $\Pi^{\mathsf{i}}_{p,1}$ | $\Pi^{\mathsf{i}}_{p',1}$ | $\mathbf{\Pi}_{\mathfrak{F}}$ | $\mathbf{\Pi}_{\mathfrak{I}}$ | $\mathbf{\Pi}_{\mathfrak{F}\mathfrak{I}}$ |
|---|---|---|---|---|---|---|---|---|
| $Move(f, f')$ | 1 | 1 | 0 | 1 | 1 | $\frac{1}{2}$ | $\frac{1}{n^{\mathsf{i}}}$ | $\frac{1}{2+n^{\mathsf{i}}}$ |
| $In(p, f)$ | 1 | 1 | 1 | 1 | 0 | $\frac{1}{n^{\mathsf{f}}}$ | 1 | $\frac{1}{n^{\mathsf{f}}+1}$ |
| $In(p', f)$ | 1 | 1 | 1 | 0 | 1 | $\frac{1}{n^{\mathsf{f}}}$ | 1 | $\frac{1}{n^{\mathsf{f}}+1}$ |
| $Out(p, f)$ | 1 | 1 | 1 | 1 | 0 | $\frac{1}{n^{\mathsf{f}}}$ | 1 | $\frac{1}{n^{\mathsf{f}}+1}$ |
| $Out(p', f)$ | 1 | 1 | 1 | 0 | 1 | $\frac{1}{n^{\mathsf{f}}}$ | 1 | $\frac{1}{n^{\mathsf{f}}+1}$ |

Table 7: Number of representatives for each original Miconic action in each abstract task, as well as the partition of the action costs among these representatives

### 9.4.1 Lower Bounds

First, as Miconic is a special case of the Logistics domain, Lemma 1 applies here analogously, with each package in $P_3$ corresponding to a passenger. Thus, for each $p \in P$, all three heuristics account for the full cost of the load/unload actions required by any optimal plan for that task.

Let us now focus on the abstract tasks $\mathbf{\Pi}_{\mathfrak{F}} = \{\Pi^{\mathsf{f}}_{e,f} \mid f \in F\}$. Recall that the task $\Pi^{\mathsf{f}}_{e,f}$ is induced by an $e$-fork and, in terms of domain decomposition, distinguishes between being at floor $f$ and being somewhere else. Without loss of generality, the set of floors $F$ can be restricted to the initial and the goal values of the variables, and this because no optimal plan will move the elevator to or from a floor $f$ that is neither an initial nor a goal location of a passenger or the elevator. Let $F_I = \{I[p] \mid p \in P\}$ and $F_G = \{G[p] \mid p \in P\}$. The costs of the optimal plans for each abstract task $\Pi^{\mathsf{f}}_{e,f}$ are as follows.

$f \in F_I \cap F_G$ : Let $p, p' \in P$ be a pair of passengers with initial and goal locations in $f$, respectively; that is, $I[p] = G[p'] = f$. If $f = I[e]$, then any plan for $\Pi^{\mathsf{f}}_{e,f}$ has to move the elevator from $f$ in order to load passenger $p'$, and then move the elevator back to $f$ in order to unload passenger $p'$. Therefore the cost of any plan for $\Pi^{\mathsf{f}}_{e,f}$ is at least $\frac{2|P|}{|F|} + 1$, where (see the last three columns of Table 7) the first component of the summation comes from summing the costs of the representatives of the load/unload actions for all the passengers, and the second component is the sum of the costs of representatives of the two respective move actions. Similarly, if $f \neq I[e]$, then any plan for $\Pi^{\mathsf{f}}_{e,f}$ has to move the elevator to $f$ in order to load passenger $p$, and then move the elevator from $f$ in order to unload $p$. Therefore, here as well, the cost of any plan for $\Pi^{\mathsf{f}}_{e,f}$ is at least $\frac{2|P|}{|F|} + 1$.

$f \in F_I \setminus F_G$ : Let $p \in P$ be a passenger initially at $f$, that is, $I[p] = f$. If $f = I[e]$, then any plan for $\Pi^{\mathsf{f}}_{e,f}$ has to move the elevator from $f$ in order to unload $p$, and thus the cost of any plan for $\Pi^{\mathsf{f}}_{e,f}$ is at least $\frac{2|P|}{|F|} + \frac{1}{2}$. Otherwise, if $f \neq I[e]$, then any plan for $\Pi^{\mathsf{f}}_{e,f}$ has to move the elevator to $f$ in order to load $p$, and then move the elevator from $f$ in order to unload $p$. Hence, in this case, the cost of any plan for $\Pi^{\mathsf{f}}_{e,f}$ is at least $\frac{2|P|}{|F|} + 1$.

$f \in F_G \setminus F_I$ : Let $p \in P$ be a passenger who must arrive at floor $f$, that is, $G[p] = f$. If $f = I[e]$, then any plan for $\Pi^{\mathsf{f}}_{e,f}$ has to move the elevator from $f$ in order to load $p$, and then move the elevator back to $f$ in order to unload $p$. Hence, here as well, the cost of any plan for $\Pi^{\mathsf{f}}_{e,f}$ is at least $\frac{2|P|}{|F|} + 1$. Otherwise, if $f \neq I[e]$, then any plan for $\Pi^{\mathsf{f}}_{e,f}$ has to move the elevator to $f$ in order to unload $p$, and thus the cost of any plan for $\Pi^{\mathsf{f}}_{e,f}$ is at least $\frac{2|P|}{|F|} + \frac{1}{2}$.

$f \notin F_G \cup F_I$ : If $f = I[e]$, then any plan for $\Pi^{\mathsf{f}}_{e,f}$ has to include a move from $f$ in order to load/unload the passengers, and thus the cost of any plan for $\Pi^{\mathsf{f}}_{e,f}$ is at least $\frac{2|P|}{|F|} + \frac{1}{2}$. Otherwise, if $f \neq I[e]$, the elevator is initially "in the set of all other locations," and thus the cost of any plan for $\Pi^{\mathsf{f}}_{e,f}$ is at least $\frac{2|P|}{|F|}$.

Putting this case-by-case analysis together, we have

$$h^{\mathcal{F}}(I) \geq \begin{cases} 2|P| + |F_I \cap F_G| + |F_I \setminus F_G| + \frac{|F_G \setminus F_I|}{2}, & I[e] \in F_I \cap F_G \\ 2|P| + |F_I \cap F_G| + |F_I \setminus F_G| - 1 + \frac{1}{2} + \frac{|F_G \setminus F_I|}{2}, & I[e] \in F_I \setminus F_G \\ 2|P| + |F_I \cap F_G| + |F_I \setminus F_G| + 1 + \frac{|F_G \setminus F_I| - 1}{2}, & I[e] \in F_G \setminus F_I \\ 2|P| + |F_I \cap F_G| + |F_I \setminus F_G| + \frac{|F_G \setminus F_I| - 1}{2} + \frac{1}{2}, & I[e] \notin F_G \cup F_I \end{cases}.$$

Note that the value in the second case is the lowest. This gives us a lower bound on the $h^{\mathcal{F}}$ estimate as in Eq. 22.

$$h^{\mathcal{F}}(I) \geq 2|P| + |F_I \setminus F_G| + \frac{|F_G \setminus F_I|}{2} + |F_I \cap F_G| - \frac{1}{2}. \tag{22}$$

Now, let us provide an upper bound on the length (= cost) of the optimal plan for a Miconic task. First, let $P' \subseteq P$ denote the set of passengers with both initial and goal locations in $F_I \cap F_G$. Let $m(P', F_I \cap F_G)$ denote the length of the optimal traversal of the floors $F_I \cap F_G$ such that, for each passenger $p \in P'$, a visit of $I[p]$ comes before some visit of $G[p]$. Given that, on a case-by-case basis, a (not necessarily optimal) plan for the Miconic task at hand is as follows.

$I[e] \in F_I \cap F_G$ : Collect all the passengers at $I[e]$ if any, then traverse all the floors in $F_I \setminus F_G$ and collect passengers from these floors, then move the elevator to the first floor $f$ on the optimal path $\pi$ traversing the floors $F_I \cap F_G$, drop off the passengers whose destination is $f$, collect the new passengers if any, keep moving along $\pi$ while collecting and dropping off passengers at their initial and target floors, and then traverse $F_G \setminus F_I$, dropping off the remaining passengers at their destinations. The cost of such a plan (and thus of the optimal plan) is upper-bounded as in Eq. 23 below.

$$h^*(I) \leq 2|P| + |F_I \setminus F_G| + m(P', F_I \cap F_G) + |F_G \setminus F_I|. \tag{23}$$

$I[e] \in F_I \setminus F_G$ : Collect all the passengers at $I[e]$ if any, then traverse all the floors in $F_I \setminus F_G$ and collect passengers from these floors while making sure that this traversal ends up at the first floor $f$ of the optimal path $\pi$ traversing the floors $F_I \cap F_G$, then follow $\pi$ while collecting and dropping passengers off at their initial and target floors, and then traverse $F_G \setminus F_I$, dropping the remaining passengers off at their destinations. As in the first case, the cost of such a plan is upper-bounded as in Eq. 23.

$I[e] \notin F_I$ : Traverse the floors $F_I \setminus F_G$ and collect all the passengers from these floors, then move along the optimal path $\pi$ traversing the floors $F_I \cap F_G$ while collecting and dropping off passengers at their initial and target floors, and then traverse the floors $F_G \setminus F_I$, dropping the remaining passengers off at their destinations. Here as well, the cost of such a plan is upper-bounded by the expression in Eq. 23.

**Lemma 2** *For any* Miconic *task with passengers* $P$*, we have* $\frac{h^{\mathfrak{F}}(I)}{h^*(I)} \geq \frac{5|P|-1}{6|P|}$.

**Proof:** Recall that $P' \subseteq P$ is the set of all passengers with both initial and goal locations in $F_I \cap F_G$. First we give two upper bounds on the length of the optimal traversal of the floors $F_I \cap F_G$ such that, for each passenger $p \in P'$, a visit of $I[p]$ comes before some visit of $G[p]$. From Theorem 5.3.3 of Helmert (2008) we have

$$m(P', F_I \cap F_G) = |F_I \cap F_G| + m^*(\mathcal{G}'), \tag{24}$$

where $m^*(\mathcal{G}')$ is the size of the minimum feedback vertex set of the directed graph $\mathcal{G}' = (\mathcal{V}', \mathcal{E}')$, with $\mathcal{V}' = F_I \cap F_G$ and $\mathcal{E}'$ containing an arc from $f$ to $f'$ if and only if a passenger $p \in P'$ is initially at floor $f$ and should arrive at floor $f'$.

Note that $m^*(\mathcal{G}')$ is trivially bounded by the number of graph nodes $\mathcal{V}'$. In addition, observe that, for any order of the nodes $\mathcal{V}'$, the arcs $\mathcal{E}'$ can be partitioned into "forward" and "backward" arcs, and one of these subsets must contain no more than $\frac{|\mathcal{E}'|}{2}$ arcs. Removing from $\mathcal{G}'$ all the nodes that are origins of the arcs in that smaller subset of $\mathcal{E}'$ results in a directed acyclic graph. Hence, the set of removed nodes is a (not necessarily minimum) feedback vertex set of $\mathcal{G}'$, and the size of this set is no larger than $\frac{|\mathcal{E}'|}{2}$. Putting these two bounds on $m^*(\mathcal{G}')$ together with Eq. 24 we obtain

$$m(P', F_I \cap F_G) \leq \min\left\{2|F_I \cap F_G|,\ \ |F_I \cap F_G| + \frac{|P'|}{2}\right\}. \tag{25}$$

From the disjointness of $F_G \setminus F_I$ and $F_I \cap F_G$, and the fact that the goal of all the passengers in $P'$ is in $F_I$, we have $|F_G \setminus F_I| \leq |P| - |P'|$. From Eqs. 22 and 23 we have

$$\frac{h^{\mathfrak{F}}}{h^*} \geq \frac{2|P| + |F_I \setminus F_G| + \frac{|F_G \setminus F_I|}{2} + |F_I \cap F_G| - \frac{1}{2}}{2|P| + |F_I \setminus F_G| + |F_G \setminus F_I| + m(P', F_I \cap F_G)}. \tag{26}$$

As we are interested in a lower bound on the ratio $\frac{h^{\mathfrak{F}}}{h^*}$, the right-hand side of the inequality should be minimized, and thus we can safely set $|F_I \setminus F_G| = 0$ and $|F_G \setminus F_I| = |P| - |P'|$, obtaining

$$\frac{h^{\mathfrak{F}}}{h^*} \geq \frac{2|P| + \frac{|P|-|P'|}{2} + |F_I \cap F_G| - \frac{1}{2}}{2|P| + |P| - |P'| + m(P', F_I \cap F_G)} = \frac{5|P| - |P'| + 2|F_I \cap F_G| - 1}{6|P| - 2|P'| + 2m(P', F_I \cap F_G)}. \tag{27}$$

Let us examine the right-most expression in Eq. 27 with respect to the two upper bounds on $m(P', F_I \cap F_G)$ as in Eq. 25.

- If the minimum is obtained on $2|F_I \cap F_G|$, then $m(P', F_I \cap F_G) \leq 2|F_I \cap F_G| \leq |F_I \cap F_G| + \frac{|P'|}{2}$, where the last inequality can be reformulated as

$$2|F_I \cap F_G| - |P'| \leq 0.$$

This allows us to provide a lower bound on the right-most expression in Eq. 27, and thus on $\frac{h^{\mathcal{F}}}{h^*}$ as

$$\frac{h^{\mathcal{F}}}{h^*} \geq \frac{5|P| - |P'| + 2|F_I \cap F_G| - 1}{6|P| - 2|P'| + 2m(P', F_I \cap F_G)} \geq \frac{5|P| + (2|F_I \cap F_G| - |P'|) - 1}{6|P| + 2(2|F_I \cap F_G| - |P'|)} \geq \frac{5|P| - 1}{6|P|}. \tag{28}$$

- If the minimum is obtained on $|F_I \cap F_G| + \frac{|P'|}{2}$, then $m(P', F_I \cap F_G) \leq |F_I \cap F_G| + \frac{|P'|}{2} < 2|F_I \cap F_G|$, where the last inequality can be reformulated as

$$2|F_I \cap F_G| - |P'| > 0.$$

This again allows us to provide a lower bound on $\frac{h^{\mathcal{F}}}{h^*}$ via Eq. 27 as

$$\frac{h^{\mathcal{F}}}{h^*} \geq \frac{5|P| - |P'| + 2|F_I \cap F_G| - 1}{6|P| - 2|P'| + 2m(P', F_I \cap F_G)} \geq \frac{5|P| + (2|F_I \cap F_G| - |P'|) - 1}{6|P| + (2|F_I \cap F_G| - |P'|)} \geq \frac{5|P| - 1}{6|P|}. \tag{29}$$

Note that both lower bounds on $\frac{h^{\mathcal{F}}}{h^*}$ in Eq. 28 and Eq. 29 are as required by the claim of the lemma. □

### 9.4.2 Upper Bounds

A Miconic task on which the heuristic $h^{\mathcal{F}}$ achieves the performance ratio of exactly 5/6 consists of an elevator $e$, floors $F = \{f_i\}_{i=0}^{n}$, passengers $P = \{p_i\}_{i=1}^{n}$, all the passengers and the elevator being initially at $f_0$, and the target floors of the passengers all being pairwise disjoint. The SAS$^+$ encoding for the Miconic task is as follows.

- Variables $V = \{e\} \cup P$ with the domains $\mathcal{D}(e) = F$ and $\forall p \in P : \mathcal{D}(p) = F \cup \{\mathsf{e}\}$.
- Initial state $I = \{e : f_0, p_1 : f_0, \ldots, p_n : f_0\}$.
- Goal $G = \{p_1 : f_1, \ldots, p_n : f_n\}$.
- Actions $A = \{In(p, f), Out(p, f) \mid f \in F, p \in P\} \cup \{Move(f, f') \mid \{f, f'\} \subseteq F\}$.

The causal graph of this task and the corresponding collection of $v$-forks (consisting of only one $e$-fork) are depicted in Figure 15. The domain of $e$ is abstracted as in Eq. 16 ("leave-one-out"), providing us with

$$\mathbf{\Pi}_{\mathcal{F}} = \{\Pi^{\mathsf{f}}_{e,f_0}, \Pi^{\mathsf{f}}_{e,f_1}, \ldots, \Pi^{\mathsf{f}}_{e,f_n}\}.$$

The costs of the action representatives in these abstract tasks are given in Table 7 with $n^{\mathsf{f}} = n + 1$. The optimal plans for the abstract tasks in $\mathbf{\Pi}_{\mathcal{F}}$ are

| task | optimal plan | cost | # | $h^{\mathcal{F}}(I)$ |
|---|---|---|---|---|
| $\Pi^{\mathsf{f}}_{e,f_0}$ | $\langle In(p_1, f_0), \ldots, In(p_n, f_0), Move(f_0, f_1), Out(p_1, f_1), \ldots, Out(p_n, f_n)\rangle$ | $\frac{1}{2} + \frac{2n}{n+1}$ | $n+1$ | $\frac{5n+1}{2}$ |
| $\Pi^{\mathsf{f}}_{e,f_1}$ | $\langle In(p_1, f_0), \ldots, In(p_n, f_0), Out(p_2, f_2), \ldots, Out(p_n, f_n), Move(f_0, f_1), Out(p_1, f_1)\rangle$ | $\frac{1}{2} + \frac{2n}{n+1}$ | | |
| $\Pi^{\mathsf{f}}_{e,f_n}$ | $\langle In(p_1, f_0), \ldots, In(p_n, f_0), Out(p_1, f_1), \ldots, Out(p_{n-1}, f_{n-1}), Move(f_0, f_n), Out(p_n, f_n)\rangle$ | $\frac{1}{2} + \frac{2n}{n+1}$ | | |

while an optimal plan for the original task, $\langle In(p_1, f_0), \ldots, In(p_n, f_0), Move(f_0, f_1), Out(p_1, f_1), Move(f_1, f_2), Out(p_2, f_2), Move(f_2, f_3), \ldots, Out(p_n, f_n)\rangle$, has a cost of $3n$, providing us with the upper bound of 5/6 for the $h^{\mathfrak{F}}$ heuristic in Miconic. Putting this upper bound together with the previously obtained lower bound of 5/6, we conclude that the asymptotic performance ratio of $h^{\mathfrak{F}}$ in Miconic is 5/6.

A Miconic task on which the heuristics $h^{\mathfrak{I}}$ and $h^{\mathfrak{FI}}$ achieve exactly 1/2 consists of an elevator $e$, floors $F = \{f_i\}_{i=0}^{2n}$, passengers $P = \{p_i\}_{i=1}^{n}$, and the initial and target floors for all the passengers and the elevator being pairwise disjoint. The task description in SAS$^+$ is as follows.

- Variables $V = \{e\} \cup P$ with the domains $\mathcal{D}(e) = F$ and $\forall p \in P : \mathcal{D}(p) = F \cup \{\mathsf{e}\}$.
- Initial state $I = \{e : f_0, p_1 : f_1, \ldots, p_n : f_n\}$.
- Goal $G = \{p_1 : f_{n+1}, \ldots, p_n : f_{2n}\}$.
- Actions $A = \{In(p, f), Out(p, f) \mid f \in F, p \in P\} \cup \{Move(f, f') \mid \{f, f'\} \subseteq F\}$.

The causal graph of this task and the corresponding collection of $v$-forks and $v$-iforks are depicted in Figure 15. The domains of the inverted-fork sinks are all abstracted as in Eq. 15 ("distance-from-initial-value"), and the domains of the fork roots are all abstracted as in Eq. 16 ("leave-one-out"). This provides us with

$$\mathbf{\Pi}_{\mathfrak{I}} = \{\Pi^{\mathsf{i}}_{p_1,1}, \ldots, \Pi^{\mathsf{i}}_{p_n,1}\},$$
$$\mathbf{\Pi}_{\mathfrak{FI}} = \{\Pi^{\mathsf{f}}_{e,f_0}, \Pi^{\mathsf{f}}_{e,f_1}, \ldots, \Pi^{\mathsf{f}}_{e,f_n}, \Pi^{\mathsf{f}}_{e,f_{n+1}}, \ldots, \Pi^{\mathsf{f}}_{e,f_{2n}}, \Pi^{\mathsf{i}}_{p_1,1}, \ldots, \Pi^{\mathsf{i}}_{p_n,1}\}.$$

The costs of the action representatives in these abstract tasks are given in Table 7 with $n^{\mathsf{f}} = 2n + 1$ and $n^{\mathsf{i}} = n$. The optimal plans for the abstract tasks in $\mathbf{\Pi}_{\mathfrak{I}}$ and $\mathbf{\Pi}_{\mathfrak{FI}}$ are

| $h$ | task | optimal plan | cost | # | $h(I)$ |
|---|---|---|---|---|---|
| $h^{\mathfrak{I}}$ | $\Pi^{\mathsf{i}}_{p_i,1}$ | $\langle Move(f_0, f_i), In(p_i, f_i), Move(f_i, f_{n+i}), Out(p_i, f_{n+i})\rangle$ | $\frac{2}{n} + 2$ | $n$ | $2n + 2$ |
| $h^{\mathfrak{FI}}$ | $\Pi^{\mathsf{f}}_{e,f_0}$ | $\langle Move(f_0, f_1), In(p_1, f_1), \ldots, In(p_n, f_n), Out(p_1, f_{n+1}), \ldots, Out(p_n, f_{2n})\rangle$ | $\frac{1}{n+2} + \frac{2n}{2n+2}$ | 1 | $2n + \frac{5n+1}{n+2}$ |
| | $\Pi^{\mathsf{f}}_{e,f_1}$ | $\langle Move(f_0, f_1), In(p_1, f_1), Move(f_1, f_2), In(p_2, f_2), \ldots, In(p_n, f_n), Out(p_1, f_{n+1}), \ldots, Out(p_n, f_{2n})\rangle$ | $\frac{2}{n+2} + \frac{2n}{2n+2}$ | $n$ | |
| | $\Pi^{\mathsf{f}}_{e,f_n}$ | $\langle Move(f_0, f_n), In(p_n, f_n), Move(f_n, f_1), In(p_1, f_1), \ldots, In(p_{n-1}, f_{n-1}), Out(p_1, f_{n+1}), \ldots, Out(p_n, f_{2n})\rangle$ | $\frac{2}{n+2} + \frac{2n}{2n+2}$ | | |
| | $\Pi^{\mathsf{f}}_{e,f_{n+1}}$ | $\langle In(p_1, f_1), \ldots, In(p_n, f_n), Out(p_2, f_{n+2}), \ldots, Out(p_n, f_{2n}), Move(f_0, f_{n+1}), Out(p_1, f_{n+1})\rangle$ | $\frac{1}{n+2} + \frac{2n}{2n+2}$ | $n$ | |
| | $\Pi^{\mathsf{f}}_{e,f_{2n}}$ | $\langle In(p_1, f_1), \ldots, In(p_n, f_n), Out(p_1, f_{n+1}), \ldots, Out(p_{n-1}, f_{2n-1}), Move(f_0, f_{2n}), Out(p_n, f_{2n})\rangle$ | $\frac{1}{n+2} + \frac{2n}{2n+2}$ | | |
| | $\Pi^{\mathsf{i}}_{p_i,1}$ | $\langle Move(f_0, f_i), In(p_i, f_i), Move(f_i, f_{n+i}), Out(p_i, f_{n+i})\rangle$ | $\frac{2}{n+2} + \frac{2}{2n+2}$ | $n$ | |

while an optimal plan for the original task, $\langle Move(f_0, f_1), In(p_1, f_1), Move(f_1, f_2), In(p_2, f_2), Move(f_2, f_3), \ldots, In(p_n, f_n), Move(f_n, f_{n+1}), Out(p_1, f_{n+1}), Move(f_{n+1}, f_{n+2}), Out(p_2, f_{n+2}), Move(f_{n+2}, f_{n+3}), \ldots, Out(p_n, f_{2n})\rangle$, has the cost of $4n$, providing us with the upper bound of 1/2 for the $h^{\mathfrak{I}}$ and $h^{\mathfrak{FI}}$ heuristics in Miconic. Putting this upper bound together with the previously obtained lower bound of 1/2, we conclude that the asymptotic performance ratio of $h^{\mathfrak{I}}$ and $h^{\mathfrak{FI}}$ in Miconic is 1/2.

### 9.5 Satellite

The Satellite domain is quite complex. A Satellite tasks consists of some satellites $\mathcal{S}$, each $s \in \mathcal{S}$ with a finite set of instruments $\mathcal{I}_s$ onboard, $\mathcal{I} = \bigcup_{s\in\mathcal{S}} \mathcal{I}_s$. There is a set of image modes $\mathcal{M}$, and for each mode $m \in \mathcal{M}$, there is a set $\mathcal{I}_m \subseteq \mathcal{I}$ of instruments supporting mode $m$. Likewise, there is a set of directions $\mathcal{L}$, image objectives $\mathcal{O} \subseteq \mathcal{L}\times\mathcal{M}$, and functions $cal : \mathcal{I} \mapsto \mathcal{L}$, $p_0 : \mathcal{S} \mapsto \mathcal{L}$, and $p_* : \mathcal{S}_0 \mapsto \mathcal{L}$ with $\mathcal{S}_0 \subseteq \mathcal{S}$, where $cal$ is the calibration target direction function, $p_0$ is the initial direction function, and $p_*$ is the goal pointing direction function.

Let us denote by $\mathcal{O}_i = \{o = (d, m) \in \mathcal{O} \mid i \in \mathcal{I}_m\}$ the subset of all images that can be taken by instrument $i$, by $\mathcal{O}^s = \bigcup_{i\in\mathcal{I}_s} \mathcal{O}_i$ the subset of all images that can be taken by instruments on satellite $s$, and by $\mathcal{S}_m = \{s \mid \mathcal{I}_s \cap \mathcal{I}_m \neq \emptyset\}$ the subset of all satellites that can take images in mode $m$. The problem description in SAS$^+$ is as follows.

- Variables $V = \mathcal{S} \cup \{On_i, C_i \mid i \in \mathcal{I}\} \cup \mathcal{O}$ with domains

$$\begin{aligned} \forall s \in \mathcal{S} :\ & \mathcal{D}(s) = \mathcal{L}, \\ \forall i \in \mathcal{I} :\ & \mathcal{D}(On_i) = \mathcal{D}(C_i) = \{0, 1\}, \\ \forall o \in \mathcal{O} :\ & \mathcal{D}(o) = \{0, 1\}. \end{aligned}$$

- Initial state $I = \{s : p_0(s) \mid s \in \mathcal{S}\} \cup \{On_i : 0, C_i : 0 \mid i \in \mathcal{I}\} \cup \{o : 0 \mid o \in \mathcal{O}\}$.
- Goal $G = \{s : p_*(s) \mid s \in \mathcal{S}_0\} \cup \{o : 1 \mid o \in \mathcal{O}\}$.
- Actions

$$\begin{aligned} A = \bigcup_{s\in\mathcal{S}} &\big(\{Turn(s, d, d') \mid \{d, d'\} \subseteq \mathcal{L}\} \cup \{SwOn(i, s), Cal(i, s), SwOff(i) \mid i \in \mathcal{I}_s\}\big) \cup \\ &\{TakeIm(o, d, s, i) \mid o = (d, m) \in \mathcal{O}, s \in \mathcal{S}_m, i \in \mathcal{I}_m \cap \mathcal{I}_s\}, \end{aligned}$$

  where

  - turn satellite: $Turn(s, d, d') = \langle\{s : d\}, \{s : d'\}\rangle$,
  - power on instrument: $SwOn(i, s) = \langle\{On_{i'} : 0 \mid i' \in \mathcal{I}_s\}, \{On_i : 1\}\rangle$,
  - power off instrument: $SwOff(i) = \langle\{On_i : 1\}, \{On_i : 0, C_i : 0\}\rangle$,
  - calibrate instrument: $Cal(i, s) = \langle\{C_i : 0, On_i : 1, s : cal(i)\}, \{C_i : 1\}\rangle$, and
  - take an image: $TakeIm(o, d, s, i) = \langle\{o : 0, C_i : 1, s : d\}, \{o : 1\}\rangle$.

#### 9.5.1 Fork Decomposition

The causal graph of an example Satellite task and a representative subset of the collection of $v$-forks and $v$-iforks are depicted in Figure 16. Since the variables $\{On_i, C_i \mid i \in \mathcal{I}\} \cup \mathcal{S} \setminus \mathcal{S}_0$ have no goal value, the collection of $v$-forks and $v$-iforks will be as follows in the general case.

- For each satellite $s \in \mathcal{S}$, an $s$-fork with the leaves $\mathcal{O}^s$.

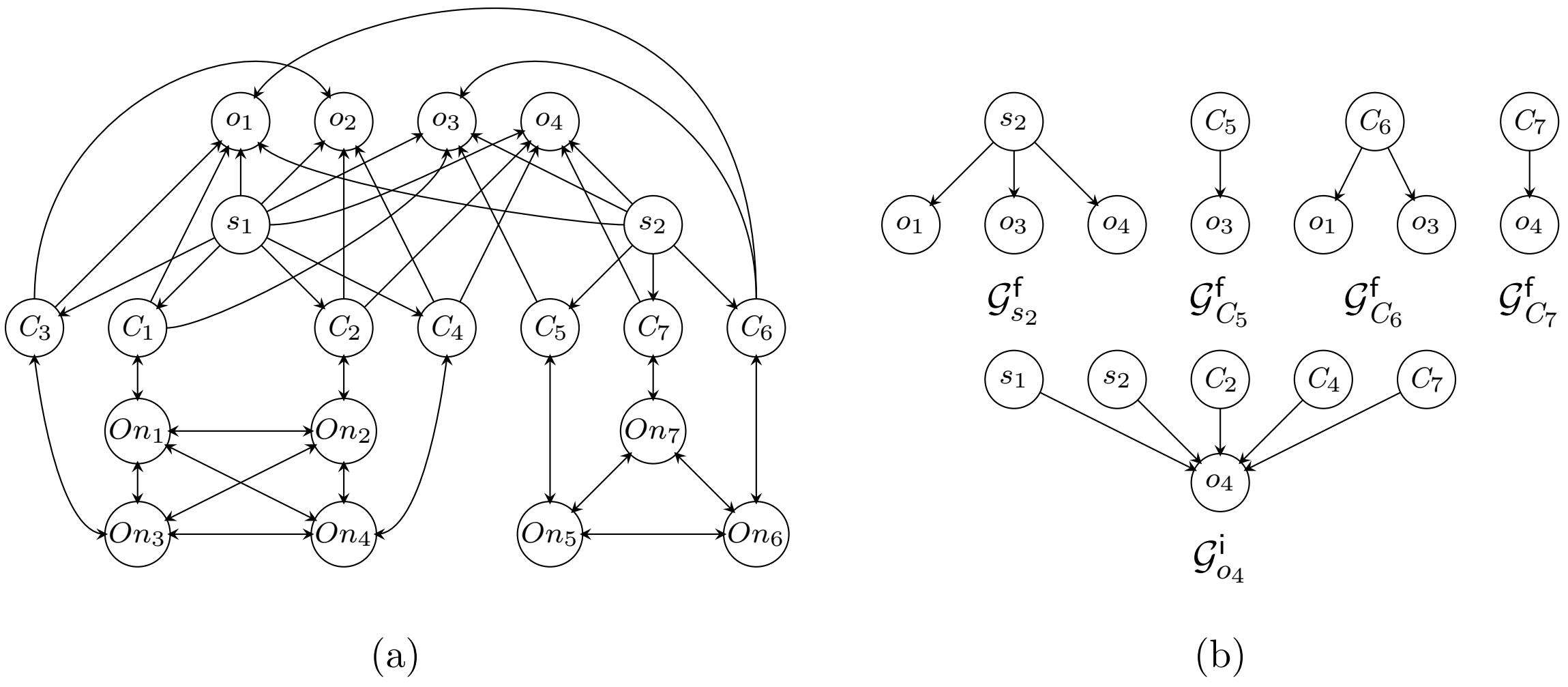

Figure 16: Satellite example task (a) causal graph and (b) a representative subset of the collection of $v$-forks and $v$-iforks

- For each instrument $i \in \mathcal{I}$, a $C_i$-fork with the leaves $\mathcal{O}_i$.
- For each image objective $o = (d, m) \in \mathcal{O}$, a $o$-ifork with the parents $\{C_i \mid i \in \mathcal{I}_m\} \cup \mathcal{S}_m$.

The root domains of all forks rooted at instruments $i \in \mathcal{I}$ and of all the inverted-fork sinks are binary in the first place, and the root domains of the forks rooted at satellites $s \in \mathcal{S}$ are abstracted as in Eq. 16 (“leave-one-out”). This provides us with

$$\begin{aligned} \mathbf{\Pi}_{\mathfrak{F}} &= \{\Pi^{\mathsf{f}}_{s,d} \mid s \in \mathcal{S}, d \in \mathcal{L}\} \cup \{\Pi^{\mathsf{f}}_{C_i} \mid i \in \mathcal{I}\}, \\ \mathbf{\Pi}_{\mathfrak{I}} &= \{\Pi^{\mathsf{i}}_{o} \mid o \in \mathcal{O}\}, \\ \mathbf{\Pi}_{\mathfrak{FI}} &= \{\Pi^{\mathsf{f}}_{s,d} \mid s \in \mathcal{S}, d \in \mathcal{L}\} \cup \{\Pi^{\mathsf{f}}_{C_i} \mid i \in \mathcal{I}\} \cup \{\Pi^{\mathsf{i}}_{o} \mid o \in \mathcal{O}\}. \end{aligned}$$

The total number of forks is thus $n^{\mathsf{f}} = |\mathcal{S}| \cdot |\mathcal{L}| + |\mathcal{I}|$ and the total number of inverted forks is $n^{\mathsf{i}} = |\mathcal{O}|$. For each action $a \in A$, the number of its representatives in each abstract task, as well as the cost assigned to each such representative, are given in Figure 17.

#### 9.5.2 Lower Bounds

First, note that any optimal plan for a Satellite task contains at most 6 actions per image objective $o \in \mathcal{O}$ and one action per satellite $s \in \mathcal{S}_0$ such that $I[s] \neq G[s]$. Now we show that each of the three heuristics fully account for the cost of at least one action per image objective $o \in \mathcal{O}$ and one action per such a satellite. This will provide us with the lower bound of 1/6 on the asymptotic performance ratios of our three heuristics.

**Lemma 3** *For any* Satellite *task,* $h^{\mathfrak{F}}$*,* $h^{\mathfrak{I}}$*, and* $h^{\mathfrak{FI}}$ *fully account for the cost of at least one* **Take Image** *action* $TakeIm(o, d, s, i)$ *for each image objective* $o \in \mathcal{O}$*.*

**Proof:** For an image objective $o = (d, m) \in \mathcal{O}$, some actions $TakeIm(o, d, s, i) = \langle \{o : 0, C_i : 1, s : d\}, \{o : 1\} \rangle$ will appear in optimal plans for $|\mathcal{S}_m| \cdot |\mathcal{L}|$ fork abstract tasks rooted

| Action | $\Pi^{\mathsf{f}}_{s,d}$ | $\Pi^{\mathsf{f}}_{s,d'}$ | $\Pi^{\mathsf{f}}_{s,d''}$ | $\Pi^{\mathsf{f}}_{s',d^*}$ | $\Pi^{\mathsf{f}}_{C_i}$ | $\Pi^{\mathsf{f}}_{C_{i'}}$ | $o \in \mathcal{O}_i$: $\Pi^{\mathsf{i}}_o$ | $o \in \mathcal{O}^s \setminus \mathcal{O}_i$: $\Pi^{\mathsf{i}}_o$ | $o \notin \mathcal{O}^s$: $\Pi^{\mathsf{i}}_o$ | $\mathbf{\Pi}_{\mathcal{F}}$ | $\mathbf{\Pi}_{\mathcal{I}}$ | $\mathbf{\Pi}_{\mathcal{FI}}$ |
|---|---|---|---|---|---|---|---|---|---|---|---|---|
| $Turn(s,d,d')$ | 1 | 1 | 0 | 0 | 0 | 0 | 1 | 1 | 0 | $\frac{1}{2}$ | $\frac{1}{\lvert\mathcal{O}^s\rvert}$ | $\frac{1}{\lvert\mathcal{O}^s\rvert+2}$ |
| $SwOn(i,s)$ | 0 | 0 | 0 | 0 | 0 | 0 | 0 | 0 | 0 | 0 | 0 | 0 |
| $Cal(i,s)$ | 0 | 0 | 0 | 0 | 1 | 0 | 1 | 0 | 0 | 1 | $\frac{1}{\lvert\mathcal{O}_i\rvert}$ | $\frac{1}{\lvert\mathcal{O}_i\rvert+1}$ |
| $SwOff(i)$ | 0 | 0 | 0 | 0 | 1 | 0 | 1 | 0 | 0 | 1 | $\frac{1}{\lvert\mathcal{O}_i\rvert}$ | $\frac{1}{\lvert\mathcal{O}_i\rvert+1}$ |

(a)

| Action | $s' \in \mathcal{S}_m$: $\Pi^{\mathsf{f}}_{s',d'}$ | $s' \notin \mathcal{S}_m$: $\Pi^{\mathsf{f}}_{s',d'}$ | $i' \in \mathcal{I}_m$: $\Pi^{\mathsf{f}}_{C_{i'}}$ | $i' \notin \mathcal{I}_m$: $\Pi^{\mathsf{f}}_{C_{i'}}$ | $\Pi^{\mathsf{i}}_o$ | $\Pi^{\mathsf{i}}_{o'}$ | $\mathbf{\Pi}_{\mathcal{F}}$ | $\mathbf{\Pi}_{\mathcal{I}}$ | $\mathbf{\Pi}_{\mathcal{FI}}$ |
|---|---|---|---|---|---|---|---|---|---|
| $TakeIm(o,d,s,i)$, $o=(d,m)$ | 1 | 0 | 1 | 0 | 1 | 0 | $\frac{1}{\lvert\mathcal{S}_m\rvert\cdot\lvert\mathcal{L}\rvert+\lvert\mathcal{I}_m\rvert}$ | 1 | $\frac{1}{\lvert\mathcal{S}_m\rvert\cdot\lvert\mathcal{L}\rvert+\lvert\mathcal{I}_m\rvert+1}$ |

(b)

Figure 17: Number of representatives for each original Satellite action in each abstract task, as well as the partition of the action costs between these representatives; table (a) shows **Turn**, **Switch On**, **Switch Off**, and **Calibrate** actions, and table (b) shows **Take Image** actions

in satellites, $|\mathcal{I}_m|$ fork abstract tasks rooted in instrument calibration status variables $C_i$, and one inverted-fork abstract task with sink $o$. Together with the costs of the action representatives in the abstract problems (see Figure 17), we have

$h^{\mathcal{F}}$ : cost of each representative is $\frac{1}{|\mathcal{S}_m|\cdot|\mathcal{L}|+|\mathcal{I}_m|}$ and there are $|\mathcal{S}_m|\cdot|\mathcal{L}|+|\mathcal{I}_m|$ fork abstract tasks,

$h^{\mathcal{I}}$ : cost of each representative is 1 and there is one inverted fork abstract task, and

$h^{\mathcal{FI}}$ : cost of each representative is $\frac{1}{|\mathcal{S}_m|\cdot|\mathcal{L}|+|\mathcal{I}_m|+1}$ and there are $|\mathcal{S}_m|\cdot|\mathcal{L}|+|\mathcal{I}_m|+1$ abstract tasks.

Therefore, for each $o \in \mathcal{O}$, the cost of one $TakeIm(o,d,s,i)$ action will be fully accounted for by each of the three heuristics. □

**Lemma 4** *For any* Satellite *task, $h^{\mathcal{F}}$, $h^{\mathcal{I}}$, and $h^{\mathcal{FI}}$ fully account for the cost of at least one* **Turn** *action $Turn(s,d,d')$ for each $s \in \mathcal{S}_0$ such that $I[s] \neq G[s]$.*

**Proof:** If $s \in \mathcal{S}_0$ is a satellite with $I[s] \neq G[s]$, then an action $Turn(s,I[s],d')$ will appear in any optimal plan for $\Pi^{\mathsf{f}}_{s,I[s]}$, an action $Turn(s,d,G[s])$ will appear in any optimal plan for $\Pi^{\mathsf{f}}_{s,G[s]}$, and for each $o \in \mathcal{O}^s$, an action $Turn(s,d,G[s])$ will appear in any optimal plan for $\Pi^{\mathsf{i}}_o$. Together with the costs of the action representatives in the abstract problems (see Figure 17) we have

$h^{\mathcal{F}}$ : cost of each representative is $\frac{1}{2}$ and there are 2 fork abstract tasks,

$h^{\mathcal{I}}$ : cost of each representative is $\frac{1}{|\mathcal{O}^s|}$ and there are $|\mathcal{O}^s|$ inverted fork abstract tasks, and

$h^{\mathcal{F}\mathcal{I}}$ : cost of each representative is $\frac{1}{|\mathcal{O}^s|+2}$ and there are $|\mathcal{O}^s| + 2$ abstract tasks.

Therefore, for each $s \in \mathcal{S}_0$ such that $I[s] \neq G[s]$, the cost of one $Turn(s, d, d')$ action will be fully accounted for by each of the three heuristics. □

Together, Lemmas 3 and 4 imply that, for $h \in \{h^{\mathcal{F}}, h^{\mathcal{I}}, h^{\mathcal{F}\mathcal{I}}\}$, on Satellite we have $\frac{h}{h^*} \geq 1/6$.

### 9.5.3 Upper Bound

A Satellite task on which all three heuristics achieve the ratio of exactly 1/6 consists of two identical satellites $\mathcal{S} = \{s, s'\}$ with $l$ instruments each, $\mathcal{I} = \mathcal{I}_s \cup \mathcal{I}_{s'} = \{1, \dots, l\} \cup \{l + 1, \dots, 2l\}$, such that instruments $\{i, l+i\}$ have two modes each: $m_0$ and $m_i$. There is a set of $n+1$ directions $\mathcal{L} = \{d_I, d_1, \dots, d_n\}$ and a set of $n$ image objectives $\mathcal{O} = \{o_1, \dots, o_n\}, o_i = (d_I, m_i)$ for $1 \leq i \leq l$ and $o_i = (d_i, m_0)$ for $l < i \leq n$. The calibration direction of instruments $\{i, l+i\}$ is $d_i$. The SAS$^+$ encoding for this planning task is as follows.

- Variables $V = \mathcal{S} \cup \mathcal{O} \cup \{On_i, C_i \mid i \in \mathcal{I}\}$.
- Initial state $I = \{s\!:\!d_I \mid s \in \mathcal{S}\} \cup \{On_i\!:\!0, C_i\!:\!0 \mid i \in \mathcal{I}\} \cup \{o\!:\!0 \mid o \in \mathcal{O}\}$.
- Goal $G = \{o\!:\!1 \mid o \in \mathcal{O}\}$.
- Actions

$$A = \bigcup_{s \in \mathcal{S}} \left(\{Turn(s, d, d') \mid \{d, d'\} \subseteq \mathcal{L}\} \cup \{SwOn(i, s), Cal(i, s), SwOff(i) \mid i \in \mathcal{I}_s\}\right) \cup \bigcup_{s \in \mathcal{S}} \left( \{TakeIm((d_I, m_i), d_I, s, i) \mid i \in \mathcal{I}_s\} \cup \bigcup_{j=l+1}^{n} \{TakeIm((d_j, m_0), d_j, s, i) \mid i \in \mathcal{I}_s\} \right).$$

The causal graph of this task is depicted in Figure 18a. The state variables $\{On_i, C_i \mid i \in \mathcal{I}\} \cup \mathcal{S}$ have no goal value, and thus the collection of $v$-forks and $v$-iforks for this task is as in Figure 18b. The domains of the inverted-fork sinks are binary, and the domains of the fork roots are abstracted as in Eq. 16 ("leave-one-out"). This provides us with

$$\begin{aligned} \mathbf{\Pi}_{\mathcal{F}} &= \{\Pi^{\mathsf{f}}_{s,d}, \Pi^{\mathsf{f}}_{s',d} \mid d \in \mathcal{L}\} \cup \{\Pi^{\mathsf{f}}_{C_i} \mid i \in \mathcal{I}\}, \\ \mathbf{\Pi}_{\mathcal{I}} &= \{\Pi^{\mathsf{i}}_{o} \mid o \in \mathcal{O}\}, \\ \mathbf{\Pi}_{\mathcal{F}\mathcal{I}} &= \{\Pi^{\mathsf{f}}_{s,d}, \Pi^{\mathsf{f}}_{s',d} \mid d \in \mathcal{L}\} \cup \{\Pi^{\mathsf{f}}_{C_i} \mid i \in \mathcal{I}\} \cup \{\Pi^{\mathsf{i}}_{o} \mid o \in \mathcal{O}\}. \end{aligned}$$

The total number of forks in this task is $n^{\mathsf{f}} = 2n + 2l + 2$ and the total number of inverted forks is $n^{\mathsf{i}} = n$. The costs of the action representatives in each abstract task are given in Figure 17, where $|\mathcal{O}^s| = |\mathcal{O}^{s'}| = |\mathcal{O}| = n$, $|\mathcal{O}_i| = n - l + 1$, $|\mathcal{S}_m| = 2$, $|\mathcal{I}_{m_0}| = 2l$, $|\mathcal{I}_{m_i}| = 2$, and $|\mathcal{L}| = n + 1$.

The optimal plans per abstract task are depicted in Table 8, while an optimal plan for the original problem, $\langle SwOn(1, s), Turn(s, d_I, d_1), Cal(1, s), Turn(s, d_1, d_I), TakeIm(o_1, d_I, s, 1),$

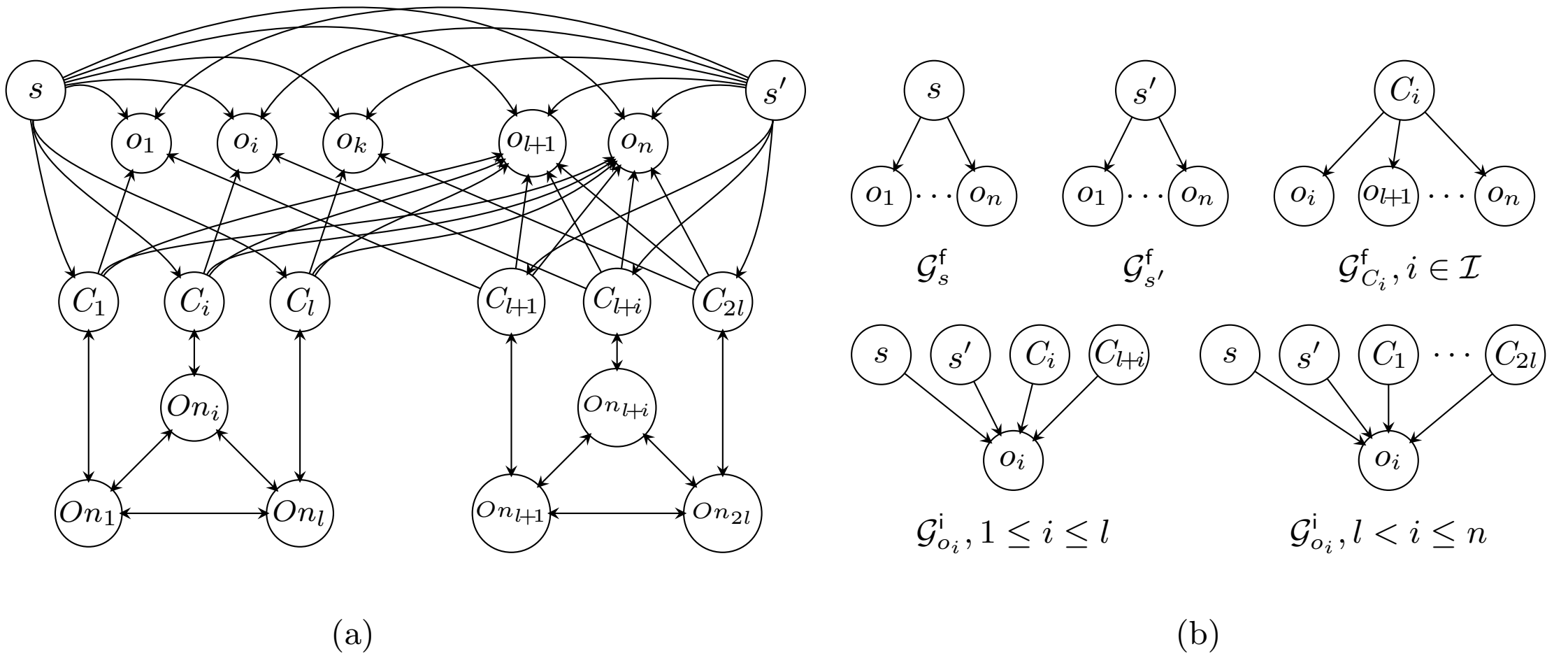

Figure 18: (a) Causal graph and (b) the corresponding collection of $v$-forks and $v$-iforks for the Satellite task used in the proof of the upper bound of 1/6

| $h$ | task | optimal plan | cost | # | $h(I)$ |
|---|---|---|---|---|---|
| $h^{\mathcal{F}}$ | $\Pi^{\mathsf{f}}_{s,d}$ | $\langle TakeIm(o_1, d_I, s', l{+}1), \ldots, TakeIm(o_l, d_I, s', 2l), TakeIm(o_{l+1}, d_{l+1}, s', 2l), \ldots, TakeIm(o_n, d_n, s', 2l)\rangle$ | $\frac{l}{2n+4} + \frac{n-l}{2n+2l+2}$ | $n+1$ | $n$ |
| | $\Pi^{\mathsf{f}}_{s',d}$ | $\langle TakeIm(o_1, d_I, s, 1), \ldots, TakeIm(o_l, d_I, s, l), TakeIm(o_{l+1}, d_{l+1}, s, l), \ldots, TakeIm(o_n, d_n, s, l)\rangle$ | $\frac{l}{2n+4} + \frac{n-l}{2n+2l+2}$ | $n+1$ | |
| | $\Pi^{\mathsf{f}}_{C_i}, i \in \mathcal{I}_s$ | $\langle TakeIm(o_i, d_I, s', l+i), TakeIm(o_{l+1}, d_{l+1}, s', 2l), \ldots, TakeIm(o_n, d_n, s', 2l)\rangle$ | $\frac{1}{2n+4} + \frac{n-l}{2n+2l+2}$ | $l$ | |
| | $\Pi^{\mathsf{f}}_{C_i}, i \in \mathcal{I}_{s'}$ | $\langle TakeIm(o_i, d_I, s, i), TakeIm(o_{l+1}, d_{l+1}, s, l), \ldots, TakeIm(o_n, d_n, s, l)\rangle$ | $\frac{1}{2n+4} + \frac{n-l}{2n+2l+2}$ | $l$ | |
| $h^{\mathcal{I}}$ | $\Pi^{\mathsf{i}}_{o_j}, 1 \le j \le l$ | $\langle Turn(s, d_I, d_j), Cal(j, s), Turn(s, d_j, d_I), TakeIm(o_j, d_I, s, j)\rangle$ | $\frac{2}{n} + \frac{1}{n-l+1} + 1$ | $l$ | $n + 2 + \frac{n}{n-l+1}$ |
| | $\Pi^{\mathsf{i}}_{o_j}, l < j \le n$ | $\langle Turn(s, d_I, d_1), Cal(1, s), Turn(s, d_1, d_j), TakeIm(o_j, d_I, s, 1)\rangle$ | $\frac{2}{n} + \frac{1}{n-l+1} + 1$ | $n-l$ | |
| $h^{\mathcal{FI}}$ | $\Pi^{\mathsf{f}}_{s,d}$ | $\langle TakeIm(o_1, d_I, s', l+1), \ldots, TakeIm(o_l, d_I, s', 2l), TakeIm(o_{l+1}, d_{l+1}, s', 2l), \ldots, TakeIm(o_n, d_n, s', 2l)\rangle$ | $\frac{l}{2n+5} + \frac{n-l}{2n+2l+3}$ | $n+1$ | $n + \frac{2n}{n+2} + \frac{n}{n-l+2}$ |
| | $\Pi^{\mathsf{f}}_{s',d}$ | $\langle TakeIm(o_1, d_I, s, 1), \ldots, TakeIm(o_l, d_I, s, l), TakeIm(o_{l+1}, d_{l+1}, s, l), \ldots, TakeIm(o_n, d_n, s, l)\rangle$ | $\frac{l}{2n+5} + \frac{n-l}{2n+2l+3}$ | $n+1$ | |
| | $\Pi^{\mathsf{f}}_{C_i}, i \in \mathcal{I}_s$ | $\langle TakeIm(o_i, d_I, s', l+i), TakeIm(o_{l+1}, d_{l+1}, s', 2l), \ldots, TakeIm(o_n, d_n, s', 2l)\rangle$ | $\frac{1}{2n+5} + \frac{n-l}{2n+2l+3}$ | $l$ | |
| | $\Pi^{\mathsf{f}}_{C_i}, i \in \mathcal{I}_{s'}$ | $\langle TakeIm(o_i, d_I, s, i), TakeIm(o_{l+1}, d_{l+1}, s, l), \ldots, TakeIm(o_n, d_n, s, l)\rangle$ | $\frac{1}{2n+5} + \frac{n-l}{2n+2l+3}$ | $l$ | |
| | $\Pi^{\mathsf{i}}_{o_j}, 1 \le j \le l$ | $\langle Turn(s, d_I, d_j), Cal(j, s), Turn(s, d_j, d_I), TakeIm(o_j, d_I, s, j)\rangle$ | $\frac{2}{n+2} + \frac{1}{n-l+2} + \frac{1}{2n+5}$ | $l$ | |
| | $\Pi^{\mathsf{i}}_{o_j}, l < j \le n$ | $\langle Turn(s, d_I, d_1), Cal(1, s), Turn(s, d_1, d_j), TakeIm(o_j, d_I, s, 1)\rangle$ | $\frac{2}{n+2} + \frac{1}{n-l+2} + \frac{1}{2n+2l+3}$ | $n-l$ | |

Table 8: Optimal plans for the abstract tasks and the overall heuristic estimates for the Satellite task used in the proof of the upper bound of 1/6

$SwOff(1), \ldots SwOn(l-1, s), Turn(s, d_I, d_{l-1}), Cal(l-1, s), Turn(s, d_{l-1}, d_I), TakeIm(o_{l-1}, d_I, s, l-1),$ $SwOff(l-1), SwOn(l, s), Turn(s, d_I, d_l), Cal(l, s), Turn(s, d_l, d_I), TakeIm(o_l, d_I, s, l), Turn(s, d_I, d_{l+1}),$ $TakeIm(o_{l+1}, d_{l+1}, s, l), \ldots, Turn(s, d_{n-1}, d_n), TakeIm(o_n, d_n, s, l)\rangle$, has the cost of $4l + 2n - 1$. For

$l = n - \sqrt{n}$, this provides us with the asymptotic performance ratio of 1/6 for all three heuristics.

## 10. Summary

We considered heuristic search for cost-optimal planning and introduced a domain-independent framework for devising admissible heuristics using additive implicit abstractions. Each such implicit abstraction corresponds to abstracting the planning task at hand by an instance of a tractable fragment of optimal planning. The key motivation for our investigation was to escape the restriction of explicit abstractions, such as pattern-database and merge-and-shrink abstractions, to abstract spaces of a fixed size. We presented a concrete scheme for additive implicit abstractions by decomposing the planning task along its causal graph and suggested a concrete realization of this idea, called fork-decomposition, that is based on two novel fragments of tractable cost-optimal planning. We then studied the induced admissible heuristics both formally and empirically, and showed that they favorably compete in informativeness with the state-of-the-art admissible heuristics both in theory and in practice. Our empirical evaluation stressed the tradeoff between the accuracy of the heuristics and runtime complexity of computing them. To alleviate the problem of expensive per-search-node runtime complexity of fork-decomposition heuristics, we showed that an equivalent of the explicit abstractions' notion of "database" exists also for the fork-decomposition abstractions, and this despite their exponential-size abstract spaces. Our subsequent empirical evaluation of heuristic search with such databases for the fork-decomposition heuristics showed that it favorably competes with the state of the art of cost-optimal planning.

The basic principles of the implicit abstraction framework motivate further research in numerous directions, most importantly in (i) discovering new islands of tractability of optimal planning, and (ii) abstracting the general planning tasks into such islands. Likewise, there is promise in combining implicit abstractions with other techniques for deriving admissible heuristic estimates. A first step towards combining implicit abstractions with polynomial-time discoverable landmarks of the planning tasks has recently been taken by Domshlak, Katz, and Lefler (2010). We believe that various combinations of such techniques might well improve the informativeness of the heuristics, and this without substantially increasing their runtime complexity.

## Acknowledgments

The work of both authors was partly supported by Israel Science Foundation grants 670/07 and 1101/07.

## Appendix A. Detailed Results of Empirical Evaluation

| | | $h^{\mathfrak{F}}$ | | $h^{\mathfrak{I}}$ | | $h^{\mathfrak{FI}}$ | | *MS*-$10^4$ | | *MS*-$10^5$ | | $\text{HSP}^*_{\text{F}}$ | | blind | | $h_{\max}$ | |
|---|---|---|---|---|---|---|---|---|---|---|---|---|---|---|---|---|---|
| task | $h^*$ | nodes | time | nodes | time | nodes | time | nodes | time | nodes | time | nodes | time | nodes | time | nodes | time |
| airport-ipc4 | | | | | | | | | | | | | | | | | |
| 01 | 8 | 10 | 0.01 | 9 | 0.00 | 9 | 0.00 | 9 | 0.00 | 9 | 0.00 | 9 | 0.72 | 11 | 0.00 | 9 | 0.00 |
| 02 | 9 | 12 | 0.03 | 15 | 0.01 | 15 | 0.03 | 10 | 0.00 | 10 | 0.00 | 10 | 1.23 | 13 | 0.00 | 10 | 0.00 |
| 03 | 17 | 86 | 0.25 | 133 | 0.07 | 93 | 0.31 | 18 | 0.04 | 18 | 0.03 | 29 | 5.10 | 164 | 0.00 | 57 | 0.00 |
| 04 | 20 | 22 | 0.02 | 21 | 0.02 | 21 | 0.02 | 21 | 0.02 | 21 | 0.01 | 21 | 1.32 | 23 | 0.00 | 21 | 0.00 |
| 05 | 21 | 23 | 1.29 | 30 | 0.06 | 27 | 1.43 | 22 | 0.01 | 22 | 0.01 | 22 | 46.54 | 27 | 0.00 | 22 | 0.00 |
| 06 | 41 | 513 | 36.72 | 639 | 1.54 | 567 | 45.25 | 42 | 0.16 | 42 | 0.17 | 42 | 123.13 | 738 | 0.01 | 418 | 0.02 |
| 07 | 41 | 514 | 37.00 | 632 | 1.53 | 550 | 44.15 | 42 | 0.17 | 42 | 0.17 | 42 | 117.56 | 742 | 0.01 | 405 | 0.02 |
| 08 | 62 | | | 21544 | 166.51 | | | 24372 | 25.42 | 9623 | 1549.13 | 203 | 602.09 | 27032 | 0.28 | 9687 | 0.90 |
| 09 | 71 | | | | | | | 152408 | 64.92 | 89525 | 466.14 | 12956 | 993.07 | 175717 | 2.47 | 56484 | 7.62 |
| 10 | 18 | 19 | 0.02 | 19 | 0.02 | 19 | 0.03 | 19 | 0.02 | 19 | 0.01 | 19 | 2.45 | 21 | 0.00 | 19 | 0.00 |
| 11 | 21 | 23 | 1.90 | 30 | 0.08 | 27 | 2.13 | 22 | 0.02 | 22 | 0.01 | 22 | 65.36 | 27 | 0.00 | 22 | 0.01 |
| 12 | 39 | 475 | 54.18 | 728 | 2.76 | 568 | 71.23 | 40 | 0.21 | 40 | 0.21 | 40 | 169.02 | 873 | 0.01 | 392 | 0.03 |
| 13 | 37 | 434 | 47.48 | 663 | 2.60 | 479 | 59.82 | 38 | 0.20 | 38 | 0.21 | 38 | 134.87 | 822 | 0.01 | 342 | 0.03 |
| 14 | 60 | | | 25110 | 334.72 | | | 30637 | 51.23 | 8968 | 238.16 | 62 | 714.76 | 35384 | 0.39 | 9196 | 1.11 |
| 15 | 58 | | | 23317 | 307.60 | | | 28798 | 46.20 | 8931 | 267.81 | 59 | 647.05 | 33798 | 0.38 | 8200 | 1.01 |
| 16 | 79 | | | | | | | 1031524 | 200.95 | 305340 | 1077.90 | | | 1247467 | 19.72 | 221993 | 49.03 |
| 17 | 88 | | | | | | | | | | | | | | | 1043661 | 310.89 |
| 19 | 90 | | | | | | | | | | | | | | | 831632 | 253.21 |
| 21 | 101 | | | | | | | 7326 | 372.92 | 102 | 10.28 | | | 18809 | 0.42 | 3184 | 1.12 |
| 22 | 148 | | | | | | | 1119943 | 762.02 | | | | | | | 159967 | 105.29 |
| 36 | 109 | | | | | | | 34365 | 853.70 | | | | | 63061 | 1.44 | | |
| blocks-ipc2 | | | | | | | | | | | | | | | | | |
| 04-0 | 6 | 15 | 0.01 | 46 | 0.01 | 17 | 0.01 | 7 | 0.03 | 7 | 0.03 | 7 | 0.36 | 93 | 0.00 | 25 | 0.00 |
| 04-1 | 10 | 14 | 0.01 | 31 | 0.00 | 15 | 0.00 | 11 | 0.04 | 11 | 0.03 | 11 | 0.39 | 66 | 0.00 | 23 | 0.00 |
| 04-2 | 6 | 7 | 0.01 | 26 | 0.00 | 10 | 0.00 | 7 | 0.04 | 7 | 0.03 | 7 | 0.38 | 63 | 0.00 | 18 | 0.00 |
| 05-0 | 12 | 32 | 0.03 | 302 | 0.06 | 113 | 0.08 | 13 | 0.30 | 13 | 0.96 | 13 | 1.32 | 467 | 0.00 | 145 | 0.00 |
| 05-1 | 10 | 37 | 0.03 | 280 | 0.06 | 98 | 0.07 | 11 | 0.29 | 11 | 0.96 | 11 | 1.36 | 567 | 0.00 | 135 | 0.00 |
| 05-2 | 16 | 152 | 0.09 | 596 | 0.10 | 348 | 0.18 | 17 | 0.29 | 17 | 0.95 | 17 | 1.49 | 792 | 0.00 | 297 | 0.00 |
| 06-0 | 12 | 33 | 0.04 | 766 | 0.27 | 207 | 0.25 | 13 | 0.95 | 13 | 8.56 | 13 | 4.10 | 1826 | 0.00 | 276 | 0.00 |
| 06-1 | 10 | 41 | 0.07 | 2395 | 0.74 | 578 | 0.78 | 11 | 0.90 | 11 | 8.34 | 11 | 4.17 | 4887 | 0.01 | 755 | 0.01 |
| 06-2 | 20 | 855 | 0.80 | 5444 | 1.23 | 3352 | 2.88 | 733 | 0.87 | 85 | 8.84 | 31 | 4.29 | 6385 | 0.02 | 2556 | 0.03 |
| 07-0 | 20 | 278 | 0.56 | 20183 | 8.26 | 4022 | 8.18 | 577 | 1.93 | 144 | 23.32 | 22 | 11.47 | 37157 | 0.14 | 5943 | 0.11 |
| 07-1 | 22 | 6910 | 11.22 | 59207 | 17.37 | 38539 | 49.71 | 10071 | 1.70 | 1835 | 21.05 | 174 | 11.25 | 63376 | 0.21 | 33194 | 0.46 |
| 07-2 | 20 | 1458 | 2.85 | 46009 | 15.05 | 18854 | 29.61 | 1855 | 1.59 | 782 | 20.37 | 90 | 10.99 | 55218 | 0.19 | 18293 | 0.29 |
| 08-0 | 18 | 1533 | 4.79 | 344157 | 179.42 | 69830 | 208.07 | 5557 | 3.67 | 678 | 36.80 | 25 | 26.00 | 519107 | 2.28 | 94671 | 2.07 |
| 08-1 | 20 | 10040 | 27.97 | 517514 | 236.64 | 191352 | 475.33 | 45711 | 3.88 | 11827 | 33.49 | 151 | 26.57 | 636498 | 2.60 | 199901 | 3.85 |
| 08-2 | 16 | 479 | 1.79 | 237140 | 136.18 | 32567 | 110.76 | 277 | 3.63 | 54 | 32.53 | 17 | 25.85 | 433144 | 1.93 | 52717 | 1.30 |
| 09-0 | 30 | | | | | | | 1233374 | 16.00 | 971409 | 77.74 | 464 | 56.76 | 7984649 | 36.76 | 3840589 | 85.00 |
| 09-1 | 28 | 3435 | 18.17 | | | | | 95068 | 7.35 | 58873 | 63.15 | 82 | 56.98 | 5914572 | 29.73 | 1200345 | 32.06 |
| 09-2 | 26 | 6379 | 35.22 | | | | | 161719 | 13.54 | 20050 | 82.45 | 81 | 57.02 | 5963160 | 30.02 | 1211463 | 32.15 |
| 10-0 | 34 | | | | | | | | | | | 1800 | 114.26 | | | | |
| 10-1 | 32 | | | | | | | | | 12063665 | 228.76 | 1835 | 115.19 | | | | |
| 10-2 | 34 | | | | | | | | | | | 3685 | 116.75 | | | | |
| 11-0 | 32 | | | | | | | | | 7046739 | 141.44 | 2678 | 213.32 | | | | |
| 11-1 | 30 | | | | | | | | | | | 1510 | 203.79 | | | | |
| 11-2 | 34 | | | | | | | | | | | 3984 | 213.97 | | | | |
| 12-0 | 34 | | | | | | | | | | | 1184 | 370.06 | | | | |
| 12-1 | 34 | | | | | | | | | | | 614 | 382.34 | | | | |
| 13-0 | 42 | | | | | | | | | | | 83996 | 860.45 | | | | |
| 13-1 | 44 | | | | | | | | | | | 163438 | 1104.27 | | | | |
| 14-0 | 38 | | | | | | | | | | | 2779 | 1063.02 | | | | |
| 14-1 | 36 | | | | | | | | | | | 7154 | 1087.40 | | | | |
| depots-ipc3 | | | | | | | | | | | | | | | | | |
| 01 | 10 | 114 | 0.24 | 279 | 0.11 | 161 | 0.32 | 11 | 0.00 | 11 | 0.00 | 45 | 0.77 | 329 | 0.00 | 136 | 0.00 |
| 02 | 15 | 1134 | 10.82 | 9344 | 12.40 | 2638 | 22.68 | 738 | 3.24 | 16 | 1.14 | 898 | 11.56 | 15404 | 0.11 | 3771 | 0.17 |
| 03 | 27 | | | | | | | 348288 | 20.69 | 239313 | 222.35 | 103089 | 247.13 | 2930398 | 27.20 | 1204646 | 97.62 |
| 04 | 30 | | | | | | | 1284048 | 52.05 | 1273762 | 529.34 | | | | | | |
| 07 | 21 | | | | | | | 211820 | 37.54 | | | 41328 | 324.19 | 6501100 | 71.58 | 1331701 | 166.76 |
| 10 | 24 | | | | | | | 3241083 | 157.52 | | | | | | | | |
| 13 | 25 | | | | | | | 1427824 | 116.06 | | | | | | | | |
| grid-ipc1 | | | | | | | | | | | | | | | | | |
| 01 | 14 | 571 | 60.28 | 1117 | 9.49 | 472 | 55.87 | 660 | 8.63 | 467 | 121.10 | | | 6446 | 0.08 | 190 | 0.10 |
| 02 | 26 | | | | | | | 3392724 | 50.35 | 3244132 | 241.94 | | | | | 664016 | 231.26 |

Table 9: Runtimes of cost-optimal heuristic-search planners on the Airport, Blocksworld, Depots, and Grid domains. The description of the planners is given in Section 6; here the fork-decomposition heuristics are computed fully online. Column *task* denotes problem instance, column $h^*$ denotes optimal solution length. Other columns capture the run *time* and number of expanded *nodes*.

| | | $h^{\mathfrak{F}}$ | | $h^{\mathfrak{I}}$ | | $h^{\mathfrak{FI}}$ | | $MS$-$10^4$ | | $MS$-$10^5$ | | $\mathrm{HSP}^*_{\mathrm{F}}$ | | blind | | $h_{\max}$ | |
|---|---|---|---|---|---|---|---|---|---|---|---|---|---|---|---|---|---|
| task | $h^*$ | nodes | time | nodes | time | nodes | time | nodes | time | nodes | time | nodes | time | nodes | time | nodes | time |
| driverlog-ipc3 | | | | | | | | | | | | | | | | | |
| 01 | 7 | 49 | 0.05 | 37 | 0.01 | 37 | 0.04 | 8 | 0.04 | 8 | 0.03 | 44 | 0.47 | 182 | 0.00 | 20 | 0.00 |
| 02 | 19 | 15713 | 18.27 | 18452 | 10.29 | 15794 | 23.80 | 20 | 0.13 | 20 | 0.26 | 15998 | 4.55 | 68927 | 0.36 | 54283 | 0.52 |
| 03 | 12 | 164 | 0.25 | 190 | 0.13 | 163 | 0.31 | 13 | 0.16 | 13 | 0.25 | 863 | 1.25 | 16031 | 0.09 | 2498 | 0.03 |
| 04 | 16 | 6161 | 19.15 | 10778 | 17.14 | 7665 | 29.88 | 17 | 0.49 | 17 | 2.41 | 22933 | 12.20 | 999991 | 8.12 | 393673 | 6.56 |
| 05 | 18 | 13640 | 45.02 | 11400 | 18.91 | 10984 | 46.16 | 2614 | 0.60 | 19 | 4.58 | 24877 | 18.77 | 6290803 | 61.57 | 1724611 | 34.73 |
| 06 | 11 | 608 | 5.21 | 795 | 3.60 | 492 | 6.05 | 291 | 1.35 | 12 | 9.72 | 3804 | 10.08 | 681757 | 7.64 | 54451 | 1.71 |
| 07 | 13 | 864 | 9.56 | 1730 | 7.71 | 1006 | 13.80 | 14 | 1.42 | 14 | 15.35 | 25801 | 41.34 | 6349767 | 81.53 | 493480 | 17.31 |
| 08 | 22 | | | | | | | 287823 | 7.34 | 2952 | 20.31 | | | | | | |
| 09 | 22 | | | 198651 | 849.04 | | | 15504 | 1.70 | 23 | 10.43 | | | | | | |
| 10 | 17 | 4304 | 199.81 | 16099 | 85.74 | 4037 | 200.52 | 18 | 1.64 | 18 | 18.54 | 18234 | 68.22 | | | | |
| 11 | 19 | 43395 | 1421.90 | 41445 | 186.53 | 39069 | 1395.51 | 34137 | 1.99 | 10790 | 17.01 | 559623 | 1193.00 | | | 6141130 | 330.22 |
| 13 | 26 | | | | | | | 1298884 | 19.52 | 870875 | 35.33 | | | | | | |
| freecell-ipc3 | | | | | | | | | | | | | | | | | |
| 01 | 8 | 234 | 1.54 | 974 | 4.88 | 274 | 3.25 | 87 | 3.12 | 9 | 38.74 | 9 | 13.01 | 3437 | 0.03 | 1043 | 0.15 |
| 02 | 14 | 30960 | 107.07 | 75150 | 230.54 | 37131 | 224.62 | 31487 | 40.40 | | | 466 | 70.29 | 130883 | 1.46 | 41864 | 10.77 |
| 03 | 18 | 197647 | 877.16 | | | | | 95805 | 140.96 | | | 1589 | 169.39 | 944843 | 11.45 | 210503 | 75.62 |
| 04 | 26 | | | | | | | 943074 | 86.78 | | | 15848 | 341.02 | 3021326 | 38.80 | 600525 | 247.70 |
| 05 | 30 | | | | | | | 5950977 | 243.74 | | | 40642 | 916.44 | | | 1408035 | 1062.25 |
| gripper-ipc1 | | | | | | | | | | | | | | | | | |
| 01 | 11 | 214 | 0.04 | 240 | 0.02 | 214 | 0.05 | 12 | 0.00 | 12 | 0.00 | 33 | 0.11 | 236 | 0.00 | 208 | 0.00 |
| 02 | 17 | 1768 | 0.54 | 1832 | 0.36 | 1803 | 0.75 | 18 | 0.11 | 18 | 0.08 | 680 | 0.37 | 1826 | 0.01 | 1760 | 0.01 |
| 03 | 23 | 11626 | 5.38 | 11736 | 4.05 | 11689 | 8.11 | 11514 | 0.47 | 2094 | 1.75 | 7370 | 1.52 | 11736 | 0.04 | 11616 | 0.08 |
| 04 | 29 | 68380 | 43.58 | 68558 | 35.24 | 68479 | 70.72 | 68380 | 1.24 | 68190 | 8.05 | 55568 | 10.29 | 68558 | 0.27 | 68368 | 0.56 |
| 05 | 35 | 376510 | 328.10 | 376784 | 296.59 | 376653 | 560.93 | 376510 | 3.52 | 376510 | 19.46 | 344386 | 79.96 | 376772 | 1.59 | 376496 | 3.51 |
| 06 | 41 | | | | | | | 1982032 | 13.42 | 1982032 | 42.16 | 1911592 | 577.49 | 1982394 | 9.59 | 1982016 | 21.57 |
| 07 | 47 | | | | | | | 10091986 | 61.66 | 10091986 | 106.84 | | | 10092464 | 51.10 | 10091968 | 119.64 |
| logistics-ipc1 | | | | | | | | | | | | | | | | | |
| 01 | 26 | | | | | | | 1918881 | 41.03 | 949586 | 34.82 | 211955 | 1700.26 | | | | |
| 05 | 22 | 3293 | 945.35 | | | | | 768161 | 18.69 | 609393 | 35.27 | | | | | | |
| 31 | 13 | 436 | 9.67 | 1981 | 2.53 | 1284 | 21.84 | 494 | 0.42 | 14 | 2.11 | 481 | 6.58 | 155645 | 1.66 | 32282 | 0.57 |
| 32 | 20 | 392 | 2.57 | 2704 | 2.24 | 962 | 5.53 | 21 | 0.16 | 21 | 0.72 | 9598 | 7.08 | 245325 | 2.07 | 81156 | 1.00 |
| 33 | 27 | | | | | | | | | 529338 | 32.55 | | | | | | |
| logistics-ipc2 | | | | | | | | | | | | | | | | | |
| 04-0 | 20 | 21 | 0.02 | 193 | 0.06 | 65 | 0.06 | 21 | 0.03 | 21 | 0.05 | 21 | 0.34 | 11246 | 0.05 | 4884 | 0.03 |
| 04-1 | 19 | 20 | 0.03 | 570 | 0.13 | 293 | 0.16 | 20 | 0.03 | 20 | 0.04 | 20 | 0.37 | 9249 | 0.04 | 4185 | 0.03 |
| 04-2 | 15 | 16 | 0.02 | 117 | 0.03 | 79 | 0.05 | 16 | 0.04 | 16 | 0.05 | 16 | 0.36 | 4955 | 0.02 | 1205 | 0.01 |
| 05-0 | 27 | 28 | 0.05 | 2550 | 0.98 | 1171 | 1.09 | 28 | 0.10 | 28 | 0.38 | 28 | 0.58 | 109525 | 0.64 | 74694 | 0.59 |
| 05-1 | 17 | 18 | 0.03 | 675 | 0.19 | 427 | 0.31 | 18 | 0.10 | 18 | 0.38 | 18 | 0.72 | 22307 | 0.13 | 6199 | 0.05 |
| 05-2 | 8 | 9 | 0.02 | 24 | 0.01 | 13 | 0.02 | 9 | 0.09 | 9 | 0.38 | 9 | 0.78 | 1031 | 0.00 | 280 | 0.00 |
| 06-0 | 25 | 26 | 0.06 | 4249 | 1.85 | 2461 | 2.54 | 26 | 0.18 | 26 | 1.23 | 26 | 1.03 | 490207 | 3.40 | 202229 | 1.92 |
| 06-1 | 14 | 15 | 0.03 | 181 | 0.09 | 99 | 0.13 | 15 | 0.18 | 15 | 1.26 | 15 | 1.16 | 24881 | 0.16 | 3604 | 0.03 |
| 06-2 | 25 | 26 | 0.05 | 2752 | 1.22 | 1394 | 1.51 | 26 | 0.19 | 26 | 1.26 | 26 | 1.03 | 476661 | 3.32 | 200012 | 1.98 |
| 06-9 | 24 | 25 | 0.04 | 2395 | 0.94 | 1428 | 1.34 | 25 | 0.18 | 25 | 1.22 | 25 | 1.02 | 422557 | 2.95 | 133521 | 1.29 |
| 07-0 | 36 | 37 | 0.42 | 251287 | 203.64 | 98053 | 386.80 | 525 | 0.65 | 37 | 4.87 | 24317 | 35.46 | | | | |
| 07-1 | 44 | 1689 | 10.08 | | | | | 666324 | 8.83 | 49 | 4.94 | 362179 | 453.06 | | | | |
| 08-0 | 31 | 32 | 0.42 | 82476 | 78.73 | 35805 | 161.33 | 1042 | 0.96 | 32 | 6.90 | 14890 | 33.50 | | | | |
| 08-1 | 44 | 45 | 0.66 | 1183608 | 1306.92 | | | 16708 | 1.15 | 45 | 7.21 | 114155 | 198.84 | | | | |
| 09-0 | 36 | 37 | 0.54 | 351538 | 407.06 | 167038 | 883.68 | 20950 | 1.56 | 37 | 9.46 | 32017 | 83.16 | | | | |
| 09-1 | 30 | 31 | 0.50 | 59336 | 80.88 | 25359 | 168.73 | 31 | 1.27 | 31 | 9.43 | 6720 | 26.48 | | | | |
| 10-0 | 45 | 46 | 2.26 | | | | | | | 668834 | 29.73 | | | | | | |
| 10-1 | 42 | 43 | 2.10 | | | | | | | 1457130 | 43.00 | | | | | | |
| 11-0 | 48 | 697 | 26.78 | | | | | | | 701106 | 37.42 | | | | | | |
| 11-1 | 60 | 21959 | 696.23 | | | | | | | | | | | | | | |
| 12-0 | 42 | 43 | 2.78 | | | | | | | 775996 | 43.56 | | | | | | |
| 12-1 | 68 | | | | | | | | | 2222340 | 87.47 | | | | | | |
| mprime-ipc1 | | | | | | | | | | | | | | | | | |
| 01 | 5 | 196 | 0.19 | 10 | 0.03 | 24 | 0.07 | 6 | 2.00 | 6 | 20.45 | 108 | 49.59 | 3636 | 0.07 | 68 | 0.04 |
| 02 | 7 | 11604 | 422.83 | 44045 | 1620.68 | 2565 | 242.83 | 3317 | 88.58 | | | | | | | 12606 | 36.65 |
| 03 | 4 | 427 | 35.09 | 7 | 0.50 | 11 | 3.15 | 36 | 33.64 | 5 | 463.85 | | | 9868 | 0.67 | 5 | 0.07 |
| 04 | 8 | 3836 | 6.62 | 1775 | 1.17 | 1093 | 3.44 | 9 | 6.09 | 9 | 82.71 | 19076 | 781.74 | 599590 | 23.58 | 200 | 0.24 |
| 05 | 11 | | | | | | | 1705009 | 127.53 | | | | | | | 1488157 | 1638.78 |
| 07 | 5 | 3314 | 14.91 | 47 | 0.15 | 346 | 3.07 | 1667 | 46.72 | | | | | 18744 | 0.56 | 11 | 0.04 |
| 08 | 6 | | | | | | | 1469752 | 403.45 | | | | | | | 7650 | 84.33 |
| 09 | 8 | 19838 | 454.91 | 100188 | 1798.69 | 5227 | 284.13 | 21993 | 36.25 | | | | | 2197646 | 71.69 | 19023 | 30.26 |
| 11 | 7 | 9 | 0.16 | 219 | 0.54 | 8 | 0.16 | 8 | 4.69 | 8 | 62.68 | 22 | 394.26 | 73260 | 2.21 | 915 | 0.54 |
| 12 | 6 | 16320 | 192.10 | 8118 | 46.69 | 5243 | 95.01 | 34763 | 11.45 | 42055 | 143.27 | 25665 | 724.12 | 108652 | 3.50 | 1520 | 1.78 |
| 15 | 6 | | | | | | | | | | | | | | | 1039 | 178.55 |
| 16 | 6 | 252 | 171.97 | | | 448 | 447.49 | 473 | 81.42 | | | | | 425144 | 32.17 | 7962 | 35.65 |
| 17 | 4 | | | 453 | 671.03 | | | | | | | | | 172736 | 42.48 | 5 | 1.06 |
| 19 | 6 | | | | | | | 123039 | 313.25 | | | | | | | 36013 | 533.75 |
| 21 | 6 | | | | | | | | | | | | | 1503293 | 103.23 | 15250 | 101.75 |
| 25 | 4 | 75 | 0.10 | 30 | 0.04 | 29 | 0.08 | 5 | 0.48 | 5 | 2.75 | 85 | 8.71 | 383 | 0.00 | 6 | 0.00 |
| 26 | 6 | | | | | | | 172432 | 46.33 | 189154 | 454.69 | | | 819590 | 61.01 | 440 | 2.69 |
| 27 | 5 | 54 | 2.28 | 1772 | 33.82 | 9 | 1.31 | 6 | 11.59 | 6 | 154.43 | | | 84079 | 3.50 | 831 | 2.08 |
| 28 | 7 | 8 | 0.03 | 403 | 0.23 | 37 | 0.08 | 8 | 1.88 | 8 | 22.55 | 128 | 146.80 | 17333 | 0.25 | 211 | 0.06 |
| 29 | 4 | 182 | 4.53 | 56 | 1.11 | 32 | 1.79 | 5 | 14.92 | 5 | 201.40 | | | 3187 | 0.17 | 7 | 0.10 |
| 31 | 4 | 248 | 52.86 | 46 | 7.83 | 19 | 11.79 | 419 | 99.87 | | | | | 3584 | 0.19 | 11 | 0.17 |
| 32 | 7 | 31759 | 133.33 | 12436 | 34.94 | 11839 | 95.52 | 19429 | 21.61 | 7269 | 292.37 | 11073 | 1701.00 | 115479 | 2.75 | 3096 | 1.74 |
| 34 | 4 | 234 | 11.65 | 46 | 2.13 | 23 | 3.08 | 450 | 151.69 | | | | | 3618 | 0.19 | 11 | 0.18 |
| 35 | 5 | 392 | 3.09 | 290 | 2.54 | 84 | 1.89 | 359 | 3.63 | 6 | 43.43 | 706 | 96.55 | 2476 | 0.05 | 44 | 0.03 |

Table 10: Similar to Table 9 for the Driverlog, Freecell, Gripper, Logistics-ipc1, Logistics-ipc2, and Mprime domains.

| | | $h^{\mathcal{F}}$ | | $h^{\mathcal{I}}$ | | $h^{\mathcal{FI}}$ | | $MS$-$10^4$ | | $MS$-$10^5$ | | $\mathrm{HSP}^*_{\mathrm{F}}$ | | blind | | $h_{\max}$ | |
|---|---|---|---|---|---|---|---|---|---|---|---|---|---|---|---|---|---|
| task | $h^*$ | nodes | time | nodes | time | nodes | time | nodes | time | nodes | time | nodes | time | nodes | time | nodes | time |
| miconic-strips-ipc2 | | | | | | | | | | | | | | | | | |
| 01-0 | 4 | 5 | 0.00 | 5 | 0.00 | 5 | 0.00 | 5 | 0.00 | 5 | 0.00 | 5 | 0.01 | 5 | 0.00 | 5 | 0.00 |
| 01-1 | 3 | 5 | 0.00 | 5 | 0.00 | 5 | 0.00 | 4 | 0.00 | 4 | 0.00 | 4 | 0.00 | 5 | 0.00 | 4 | 0.00 |
| 01-2 | 4 | 5 | 0.00 | 5 | 0.00 | 5 | 0.00 | 5 | 0.00 | 5 | 0.00 | 5 | 0.01 | 5 | 0.00 | 5 | 0.00 |
| 01-3 | 4 | 5 | 0.00 | 5 | 0.00 | 5 | 0.00 | 5 | 0.00 | 5 | 0.00 | 5 | 0.00 | 5 | 0.00 | 5 | 0.00 |
| 01-4 | 4 | 5 | 0.00 | 5 | 0.00 | 5 | 0.00 | 5 | 0.00 | 5 | 0.00 | 5 | 0.01 | 5 | 0.00 | 5 | 0.00 |
| 02-0 | 7 | 19 | 0.00 | 22 | 0.00 | 19 | 0.00 | 8 | 0.00 | 8 | 0.00 | 26 | 0.01 | 30 | 0.00 | 20 | 0.00 |
| 02-1 | 7 | 21 | 0.00 | 23 | 0.00 | 21 | 0.00 | 8 | 0.00 | 8 | 0.00 | 26 | 0.01 | 30 | 0.00 | 22 | 0.00 |
| 02-2 | 7 | 21 | 0.00 | 23 | 0.00 | 21 | 0.00 | 8 | 0.00 | 8 | 0.00 | 27 | 0.00 | 30 | 0.00 | 22 | 0.00 |
| 02-3 | 7 | 24 | 0.01 | 24 | 0.00 | 24 | 0.00 | 8 | 0.00 | 8 | 0.00 | 20 | 0.01 | 26 | 0.00 | 17 | 0.00 |
| 02-4 | 7 | 19 | 0.00 | 22 | 0.00 | 19 | 0.00 | 8 | 0.00 | 8 | 0.00 | 23 | 0.01 | 31 | 0.00 | 20 | 0.00 |
| 03-0 | 10 | 86 | 0.01 | 129 | 0.01 | 98 | 0.01 | 11 | 0.00 | 11 | 0.00 | 100 | 0.03 | 193 | 0.00 | 105 | 0.00 |
| 03-1 | 11 | 120 | 0.01 | 168 | 0.01 | 147 | 0.01 | 12 | 0.00 | 12 | 0.00 | 140 | 0.02 | 218 | 0.00 | 150 | 0.00 |
| 03-2 | 10 | 137 | 0.01 | 143 | 0.01 | 137 | 0.01 | 11 | 0.00 | 11 | 0.00 | 122 | 0.02 | 164 | 0.00 | 92 | 0.00 |
| 03-3 | 10 | 96 | 0.01 | 153 | 0.01 | 117 | 0.01 | 11 | 0.00 | 11 | 0.00 | 131 | 0.02 | 197 | 0.00 | 130 | 0.00 |
| 03-4 | 10 | 103 | 0.01 | 149 | 0.01 | 115 | 0.01 | 11 | 0.00 | 11 | 0.00 | 114 | 0.02 | 190 | 0.00 | 114 | 0.00 |
| 04-0 | 14 | 524 | 0.06 | 843 | 0.08 | 686 | 0.12 | 15 | 0.01 | 15 | 0.01 | 669 | 0.10 | 1182 | 0.00 | 866 | 0.00 |
| 04-1 | 13 | 505 | 0.06 | 817 | 0.08 | 663 | 0.12 | 14 | 0.01 | 14 | 0.01 | 634 | 0.11 | 1176 | 0.00 | 860 | 0.00 |
| 04-2 | 15 | 685 | 0.08 | 942 | 0.09 | 802 | 0.13 | 16 | 0.01 | 16 | 0.01 | 822 | 0.12 | 1277 | 0.00 | 969 | 0.00 |
| 04-3 | 15 | 681 | 0.07 | 942 | 0.09 | 798 | 0.13 | 16 | 0.01 | 16 | 0.01 | 820 | 0.12 | 1319 | 0.00 | 970 | 0.00 |
| 04-4 | 15 | 685 | 0.07 | 942 | 0.09 | 802 | 0.13 | 16 | 0.01 | 16 | 0.01 | 821 | 0.12 | 1334 | 0.00 | 969 | 0.00 |
| 05-0 | 17 | 2468 | 0.37 | 4009 | 0.66 | 3307 | 0.93 | 18 | 0.06 | 18 | 0.05 | 2829 | 0.44 | 6350 | 0.03 | 4387 | 0.03 |
| 05-1 | 17 | 2807 | 0.42 | 4345 | 0.71 | 3677 | 1.01 | 18 | 0.06 | 18 | 0.05 | 3260 | 0.49 | 6602 | 0.03 | 4664 | 0.03 |
| 05-2 | 15 | 1596 | 0.29 | 2981 | 0.55 | 2275 | 0.73 | 16 | 0.06 | 16 | 0.05 | 1594 | 0.32 | 5565 | 0.03 | 3524 | 0.03 |
| 05-3 | 17 | 2256 | 0.36 | 3799 | 0.62 | 3104 | 0.87 | 18 | 0.06 | 18 | 0.05 | 2568 | 0.42 | 5944 | 0.03 | 4140 | 0.03 |
| 05-4 | 18 | 3210 | 0.46 | 4732 | 0.78 | 4267 | 1.11 | 19 | 0.06 | 19 | 0.05 | 3953 | 0.55 | 6949 | 0.04 | 5268 | 0.04 |
| 06-0 | 19 | 9379 | 1.98 | 17665 | 4.74 | 13531 | 5.90 | 20 | 0.18 | 20 | 0.32 | 9312 | 1.76 | 30786 | 0.20 | 21194 | 0.20 |
| 06-1 | 19 | 9106 | 1.93 | 18134 | 4.75 | 14052 | 5.94 | 20 | 0.18 | 20 | 0.32 | 10252 | 1.96 | 30093 | 0.20 | 21255 | 0.20 |
| 06-2 | 20 | 10900 | 2.19 | 19084 | 4.90 | 15111 | 6.28 | 21 | 0.18 | 21 | 0.32 | 11247 | 2.11 | 32390 | 0.21 | 21694 | 0.21 |
| 06-3 | 20 | 12127 | 2.43 | 21708 | 5.69 | 17807 | 7.19 | 21 | 0.17 | 21 | 0.32 | 14216 | 2.56 | 32574 | 0.21 | 24552 | 0.23 |
| 06-4 | 21 | 13784 | 2.62 | 23255 | 5.93 | 19536 | 7.66 | 22 | 0.17 | 22 | 0.32 | 16880 | 3.04 | 33793 | 0.22 | 26167 | 0.24 |
| 07-0 | 23 | 53662 | 13.29 | 96092 | 37.56 | 79449 | 46.76 | 24 | 0.32 | 24 | 1.75 | 56686 | 14.31 | 155466 | 1.22 | 116685 | 1.32 |
| 07-1 | 24 | 56328 | 13.86 | 99109 | 38.56 | 83677 | 47.49 | 7001 | 0.38 | 25 | 1.75 | 63035 | 16.33 | 164470 | 1.29 | 118494 | 1.33 |
| 07-2 | 22 | 48141 | 12.52 | 96139 | 38.02 | 78471 | 46.17 | 1646 | 0.33 | 23 | 1.71 | 55751 | 13.98 | 161342 | 1.27 | 119688 | 1.36 |
| 07-3 | 22 | 46867 | 12.11 | 93117 | 36.63 | 75424 | 44.43 | 1861 | 0.33 | 23 | 1.74 | 53121 | 13.27 | 155176 | 1.23 | 114649 | 1.30 |
| 07-4 | 25 | 84250 | 18.24 | 126595 | 46.11 | 111984 | 61.34 | 23159 | 0.52 | 26 | 1.71 | 96327 | 24.76 | 168219 | 1.33 | 140128 | 1.58 |
| 08-0 | 27 | 272580 | 81.51 | 485051 | 267.27 | 408114 | 317.78 | 41629 | 0.91 | 28 | 4.18 | 290649 | 104.18 | 755255 | 7.16 | 594032 | 7.95 |
| 08-1 | 27 | 284415 | 86.93 | 527216 | 288.07 | 446837 | 347.43 | 42679 | 0.90 | 28 | 4.25 | 339177 | 123.10 | 794365 | 7.56 | 636587 | 8.66 |
| 08-2 | 26 | 207931 | 66.37 | 414294 | 235.89 | 330993 | 271.03 | 37744 | 0.86 | 27 | 4.25 | 204614 | 73.39 | 731622 | 6.92 | 534711 | 7.37 |
| 08-3 | 28 | 369479 | 104.29 | 598031 | 320.33 | 527216 | 392.87 | 140453 | 1.94 | 29 | 4.21 | 435617 | 160.49 | 833421 | 7.97 | 690267 | 9.29 |
| 08-4 | 27 | 297516 | 87.65 | 507910 | 278.64 | 431432 | 333.91 | 62933 | 1.16 | 28 | 4.12 | 315339 | 111.84 | 771608 | 7.33 | 613253 | 8.43 |
| 09-0 | 31 | 1461729 | 497.72 | | | | | 684737 | 9.07 | 126918 | 8.89 | 1555286 | 794.93 | 3685552 | 41.04 | 3006991 | 49.12 |
| 09-1 | 30 | 1207894 | 438.69 | 2335166 | 1787.13 | | | 406041 | 5.61 | 100937 | 8.73 | 1344815 | 683.05 | 3649801 | 40.32 | 2893803 | 47.54 |
| 09-2 | 30 | 1294691 | 460.11 | 2340411 | 1791.16 | | | 442547 | 6.06 | 82946 | 8.63 | 1357681 | 692.11 | 3576134 | 39.61 | 2895182 | 47.26 |
| 09-3 | 32 | 1840936 | 589.09 | | | | | 765455 | 10.00 | 277302 | 11.14 | 2083168 | 1051.95 | 3796035 | 42.13 | 3304570 | 53.29 |
| 09-4 | 28 | 1252484 | 467.94 | | | | | 317692 | 4.65 | 29 | 7.03 | 1231554 | 605.01 | 3589382 | 39.29 | 2956995 | 48.84 |
| 10-0 | 33 | | | | | | | 2436164 | 35.24 | 863244 | 23.76 | | | 15804498 | 200.90 | 13267920 | 250.58 |
| 10-1 | 32 | | | | | | | 2340169 | 34.09 | 335745 | 15.68 | | | 16472633 | 208.39 | 13720664 | 256.89 |
| 10-2 | 32 | | | | | | | 1735477 | 25.29 | 486286 | 17.72 | | | 15867374 | 201.01 | 12497087 | 236.89 |
| 10-3 | 34 | | | | | | | 3952148 | 55.86 | 940556 | 24.24 | | | 16309701 | 208.42 | 13801989 | 262.53 |
| 10-4 | 33 | | | | | | | 2715866 | 39.44 | 625559 | 19.91 | | | 16472551 | 209.13 | 13925654 | 262.57 |
| 11-0 | 37 | | | | | | | 11473359 | 183.60 | 4724980 | 93.56 | | | | | | |
| 11-1 | 34 | | | | | | | 7535468 | 124.80 | 1934943 | 47.91 | | | | | | |
| 11-2 | 38 | | | | | | | 14645785 | 233.68 | 6330198 | 120.71 | | | | | | |
| 11-3 | 38 | | | | | | | | | 5809711 | 110.10 | | | | | | |
| 11-4 | 35 | | | | | | | 5853546 | 95.56 | 1082086 | 32.22 | | | | | | |
| mystery-ipc1 | | | | | | | | | | | | | | | | | |
| 01 | 5 | 7 | 0.01 | 6 | 0.00 | 6 | 0.01 | 6 | 0.20 | 6 | 1.79 | 10 | 5.38 | 30 | 0.00 | 8 | 0.00 |
| 02 | 7 | 2404 | 64.94 | 8012 | 234.10 | 722 | 47.50 | 1672 | 82.70 | | | | | 770852 | 21.85 | 2368 | 4.47 |
| 03 | 4 | 73 | 1.92 | 7 | 0.12 | 11 | 0.59 | 5 | 16.46 | 5 | 193.75 | 65 | 811.87 | 507 | 0.02 | 5 | 0.03 |
| 04 | $\infty$ | 0 | 0.01 | | | 0 | 0.00 | | | | | | | | | | |
| 07 | $\infty$ | 0 | 0.00 | 0 | 0.00 | 0 | 0.00 | 0 | 0.00 | 0 | 0.00 | 0 | 0.00 | 0 | 0.00 | 0 | 0.00 |
| 09 | 8 | 3049 | 47.68 | 10764 | 137.61 | 1215 | 40.75 | 3165 | 29.34 | | | 3868 | 670.08 | 138289 | 2.18 | 1458 | 1.44 |
| 11 | 7 | 9 | 0.02 | 33 | 0.03 | 8 | 0.02 | 8 | 1.51 | 8 | 16.59 | 34 | 41.20 | 426 | 0.00 | 19 | 0.00 |
| 12 | $\infty$ | | | 2093419 | 938.05 | | | 2102777 | 14.61 | 2102729 | 27.84 | | | 2102777 | 15.09 | 1177842 | 21.87 |
| 15 | 6 | | | | | | | | | | | | | 279973 | 13.21 | 135 | 2.62 |
| 16 | $\infty$ | 0 | 0.14 | | | 0 | 0.19 | | | | | | | | | | |
| 17 | 4 | 354 | 200.98 | 85 | 26.31 | 83 | 90.17 | 198 | 445.85 | | | | | 5400 | 0.41 | 5 | 0.35 |
| 18 | $\infty$ | 0 | 0.00 | 0 | 0.00 | 0 | 0.00 | 0 | 0.00 | 0 | 0.00 | 0 | 0.00 | 0 | 0.00 | 0 | 0.00 |
| 19 | 6 | | | 4968 | 183.24 | | | 12478 | 96.38 | | | | | 133871 | 3.65 | 1516 | 5.44 |
| 20 | 7 | | | | | | | 285069 | 59.22 | 547246 | 578.39 | | | 686125 | 23.28 | 718 | 3.76 |
| 24 | $\infty$ | 0 | 0.13 | | | 0 | 0.30 | | | | | | | | | | |
| 25 | 4 | 9 | 0.02 | 10 | 0.01 | 9 | 0.02 | 5 | 0.10 | 5 | 0.10 | 14 | 1.22 | 31 | 0.00 | 6 | 0.00 |
| 26 | 6 | 1807 | 50.40 | 1835 | 25.34 | 1344 | 60.20 | 2526 | 5.94 | 346 | 70.78 | 3107 | 291.36 | 8455 | 0.10 | 37 | 0.05 |
| 27 | 5 | 14 | 0.27 | 159 | 1.61 | 6 | 0.22 | 6 | 4.80 | 6 | 80.48 | 7 | 243.78 | 2174 | 0.03 | 73 | 0.04 |
| 28 | 7 | 8 | 0.01 | 47 | 0.02 | 15 | 0.02 | 8 | 0.63 | 8 | 6.77 | 31 | 16.67 | 843 | 0.00 | 32 | 0.00 |
| 29 | 4 | 31 | 0.26 | 14 | 0.10 | 10 | 0.17 | 5 | 8.94 | 5 | 107.10 | 27 | 536.30 | 153 | 0.01 | 7 | 0.02 |
| 30 | 9 | | | | | | | 42112 | 28.07 | 44893 | 357.07 | | | 1977063 | 38.26 | 26686 | 28.27 |

Table 11: Similar to Table 9 for the Miconic and Mystery domains.

| | | $h^{\mathfrak{F}}$ | | $h^{\mathfrak{I}}$ | | $h^{\mathfrak{FI}}$ | | $MS$-$10^4$ | | $MS$-$10^5$ | | $\text{HSP}^*_{\text{F}}$ | | blind | | $h_{\max}$ | |
|---|---|---|---|---|---|---|---|---|---|---|---|---|---|---|---|---|---|
| task | $h^*$ | nodes | time | nodes | time | nodes | time | nodes | time | nodes | time | nodes | time | nodes | time | nodes | time |
| openstacks-ipc5 | | | | | | | | | | | | | | | | | |
| 01 | 23 | 2264 | 0.49 | 3895 | 1.19 | 3070 | 1.36 | 24 | 0.05 | 24 | 0.06 | 2000 | 1.02 | 4822 | 0.01 | 4016 | 0.03 |
| 02 | 23 | 2617 | 0.56 | 4485 | 1.32 | 3561 | 1.57 | 24 | 0.06 | 24 | 0.06 | 2378 | 1.07 | 5501 | 0.02 | 4594 | 0.04 |
| 03 | 23 | 2264 | 0.49 | 3895 | 1.15 | 3070 | 1.36 | 24 | 0.06 | 24 | 0.06 | 2000 | 1.02 | 4822 | 0.01 | 4016 | 0.03 |
| 04 | 23 | 2264 | 0.49 | 3895 | 1.15 | 3070 | 1.36 | 24 | 0.06 | 24 | 0.06 | 2000 | 1.02 | 4822 | 0.02 | 4016 | 0.03 |
| 05 | 23 | 2264 | 0.48 | 3895 | 1.15 | 3070 | 1.35 | 24 | 0.06 | 24 | 0.05 | 2000 | 1.02 | 4822 | 0.01 | 4016 | 0.03 |
| 06 | 45 | 366768 | 255.00 | 779710 | 1599.86 | 587482 | 1498.20 | 621008 | 4.85 | 279614 | 7.86 | 379735 | 217.37 | 882874 | 4.91 | 822514 | 18.71 |
| 07 | 46 | 410728 | 277.99 | 760668 | 1546.44 | 606782 | 1515.46 | 594758 | 4.69 | 264535 | 7.34 | 405564 | 226.32 | 836647 | 4.62 | 787163 | 17.81 |
| pathways-ipc5 | | | | | | | | | | | | | | | | | |
| 01 | 6 | 1624 | 0.03 | 1299 | 0.02 | 1299 | 0.03 | 7 | 1.14 | 7 | 0.79 | 1405 | 0.28 | 1624 | 0.00 | 36 | 0.00 |
| 02 | 12 | 2755 | 0.08 | 2307 | 0.06 | 2437 | 0.09 | 1946 | 2.56 | 13 | 42.11 | 990 | 0.29 | 2984 | 0.02 | 348 | 0.01 |
| 03 | 18 | 44928 | 2.59 | 20416 | 1.06 | 29106 | 2.14 | 21671 | 6.43 | 14901 | 129.23 | 14772 | 6.99 | 87189 | 1.06 | 4346 | 0.16 |
| 04 | 17 | 126950 | 11.45 | 33788 | 2.97 | 58738 | 7.07 | | | 98484 | 288.39 | 34206 | 27.00 | 456143 | 8.22 | 104068 | 2.61 |
| pipesworld-notankage-ipc4 | | | | | | | | | | | | | | | | | |
| 01 | 5 | 121 | 0.15 | 109 | 0.05 | 121 | 0.18 | 6 | 0.04 | 6 | 0.04 | 6 | 2.79 | 121 | 0.00 | 13 | 0.00 |
| 02 | 12 | 1413 | 2.05 | 1542 | 0.86 | 1413 | 2.42 | 169 | 0.30 | 13 | 0.17 | 435 | 3.07 | 1808 | 0.01 | 792 | 0.02 |
| 03 | 8 | 1742 | 5.26 | 3001 | 3.31 | 1742 | 6.43 | 9 | 1.15 | 9 | 0.69 | 128 | 3.84 | 3293 | 0.02 | 262 | 0.02 |
| 04 | 11 | 7007 | 24.71 | 8911 | 12.43 | 7007 | 30.79 | 651 | 1.95 | 12 | 7.05 | 812 | 8.84 | 16088 | 0.11 | 2925 | 0.13 |
| 05 | 8 | 4093 | 27.45 | 6805 | 19.74 | 4093 | 35.40 | 77 | 5.63 | 9 | 21.15 | 155 | 16.53 | 11128 | 0.12 | 1121 | 0.15 |
| 06 | 10 | 12401 | 105.37 | 27377 | 103.75 | 12401 | 140.53 | 1299 | 5.26 | 61 | 39.31 | 1151 | 23.41 | 49905 | 0.48 | 7102 | 0.72 |
| 07 | 8 | 4370 | 71.75 | 9168 | 68.10 | 4370 | 105.53 | 233 | 19.78 | 9 | 59.70 | 185 | 29.88 | 46502 | 0.57 | 2631 | 0.48 |
| 08 | 10 | 18851 | 406.67 | 56189 | 483.28 | 20584 | 600.94 | 561 | 12.42 | 497 | 94.69 | 1673 | 48.84 | 273585 | 3.39 | 22874 | 3.58 |
| 09 | 13 | | | | | | | 104875 | 25.48 | | | 10478 | 74.26 | 5513309 | 80.62 | 321861 | 68.99 |
| 10 | 18 | | | | | | | 2982520 | 66.89 | | | 689832 | 1439.64 | | | 11121245 | 1579.77 |
| 11 | 20 | | | 472950 | 1577.22 | | | 90598 | 9.20 | 52159 | 43.24 | 108503 | 625.52 | 710123 | 3.86 | 107061 | 14.51 |
| 12 | 24 | | | | | | | 594661 | 12.41 | 416184 | 109.43 | 433296 | 1117.57 | 2467804 | 13.83 | 464982 | 56.82 |
| 13 | 16 | | | 117475 | 899.72 | | | 12835 | 34.28 | | | 24224 | 1019.65 | 481045 | 3.14 | 33417 | 6.38 |
| 14 | 30 | | | | | | | 13255718 | 119.54 | | | | | | | | |
| 15 | 26 | | | | | | | 648132 | 65.43 | | | | | 4921698 | 34.90 | 555619 | 105.49 |
| 17 | 22 | | | | | | | 3200672 | 90.07 | | | | | | | | |
| 19 | 24 | | | | | | | 8767431 | 150.88 | | | | | | | | |
| 21 | 14 | 23833 | 1663.46 | 49035 | 495.53 | | | 3992 | 18.13 | 948 | 159.63 | | | 157782 | 1.31 | 8966 | 2.42 |
| 23 | 18 | | | | | | | 296506 | 49.11 | 104750 | 256.13 | | | | | 481859 | 229.00 |
| 24 | 24 | | | | | | | 7315150 | 142.82 | | | | | | | | |
| 41 | 12 | | | | | | | | | | | | | | | 114257 | 250.18 |
| pipesworld-tankage-ipc4 | | | | | | | | | | | | | | | | | |
| 01 | 5 | 77 | 0.13 | 126 | 0.07 | 105 | 0.20 | 6 | 3.54 | 6 | 0.13 | 6 | 3.88 | 128 | 0.00 | 13 | 0.01 |
| 02 | 12 | 960 | 1.20 | 1005 | 0.60 | 960 | 1.55 | 110 | 3.04 | 13 | 0.20 | 179 | 6.04 | 1012 | 0.01 | 659 | 0.02 |
| 03 | 8 | 20803 | 155.53 | 52139 | 158.91 | 20803 | 207.57 | 244 | 22.64 | 9 | 36.89 | 818 | 24.47 | 52983 | 0.77 | 1802 | 1.33 |
| 04 | 11 | 110284 | 1004.10 | 157722 | 668.67 | 110284 | 1408.50 | 3892 | 16.68 | 12 | 155.03 | 8116 | 64.68 | 221429 | 3.06 | 41540 | 14.49 |
| 05 | 8 | 6531 | 73.63 | 13148 | 79.04 | 6531 | 112.61 | 376 | 15.46 | 9 | 120.06 | 313 | 59.99 | 12764 | 0.21 | 2834 | 1.61 |
| 06 | 10 | 20171 | 329.40 | 43583 | 310.24 | 20171 | 460.45 | 1794 | 328.18 | 11 | 201.44 | 3102 | 97.31 | 58487 | 0.87 | 15746 | 6.61 |
| 07 | 8 | | | | | | | | | | | 2695 | 339.76 | 5404036 | 198.08 | 104531 | 420.47 |
| 08 | 11 | | | | | | | 96043 | 191.77 | | | | | | | | |
| 11 | 22 | | | | | | | 660104 | 28.60 | 660102 | 162.93 | | | 4116344 | 30.67 | 752867 | 334.42 |
| 13 | 16 | | | | | | | 188517 | 122.11 | | | | | | | | |
| 15 | 30 | | | | | | | 2546587 | 141.12 | | | | | | | | |
| 17 | 44 | | | | | | | 12850247 | 352.46 | | | | | | | | |
| 21 | 14 | | | | | | | 13241 | 69.80 | | | | | 4423951 | 65.44 | 126845 | 222.23 |
| 31 | 39 | | | | | | | 1357801 | 124.64 | | | | | 1726598 | 13.56 | 919764 | 381.66 |
| tpp-ipc5 | | | | | | | | | | | | | | | | | |
| 01 | 5 | 6 | 0.00 | 6 | 0.00 | 6 | 0.00 | 6 | 0.00 | 6 | 0.00 | 6 | 0.01 | 7 | 0.00 | 6 | 0.00 |
| 02 | 8 | 9 | 0.00 | 11 | 0.00 | 9 | 0.00 | 9 | 0.00 | 9 | 0.00 | 9 | 0.01 | 26 | 0.00 | 16 | 0.00 |
| 03 | 11 | 12 | 0.00 | 27 | 0.00 | 16 | 0.00 | 12 | 0.00 | 12 | 0.00 | 12 | 0.03 | 116 | 0.00 | 83 | 0.00 |
| 04 | 14 | 15 | 0.01 | 78 | 0.01 | 47 | 0.01 | 15 | 0.01 | 15 | 0.00 | 15 | 0.07 | 494 | 0.00 | 430 | 0.00 |
| 05 | 19 | 623 | 0.52 | 5110 | 1.36 | 1455 | 1.21 | 20 | 0.36 | 20 | 0.77 | 624 | 0.48 | 24698 | 0.12 | 17398 | 0.15 |
| 06 | 25 | | | | | | | 947059 | 14.22 | 74798 | 23.97 | | | | | 9267024 | 216.69 |
| trucks-ipc5 | | | | | | | | | | | | | | | | | |
| 01 | 13 | 1691 | 0.41 | 1027 | 0.22 | 1039 | 0.40 | 14 | 0.03 | 14 | 0.02 | 285 | 0.56 | 5774 | 0.02 | 402 | 0.01 |
| 02 | 17 | 9624 | 2.68 | 2898 | 0.57 | 2957 | 1.35 | 4192 | 0.22 | 18 | 0.17 | 1413 | 1.04 | 28348 | 0.14 | 939 | 0.03 |
| 03 | 20 | 80693 | 71.37 | 20752 | 19.93 | 22236 | 31.25 | 199405 | 2.89 | 173790 | 6.88 | 4049 | 4.43 | 379582 | 2.97 | 9465 | 0.40 |
| 04 | 23 | 1753866 | 1237.60 | 1205793 | 850.34 | 1315672 | 1394.88 | 2591561 | 29.17 | 2568634 | 56.96 | 8817 | 7.75 | 2990366 | 26.65 | 209140 | 9.43 |
| 05 | 25 | | | | | | | 23444940 | 392.99 | | | 14744 | 23.12 | | | 1248571 | 90.78 |
| 06 | 30 | | | | | | | | | | | 308920 | 343.47 | | | | |
| 07 | 23 | 2134728 | 1313.60 | 719751 | 408.75 | 755608 | 820.55 | 7575415 | 88.91 | 8080496 | 117.13 | 43270 | 27.62 | 12410588 | 117.92 | 223011 | 19.34 |
| 08 | 25 | | | | | | | | | | | 49663 | 47.61 | | | 3106944 | 403.36 |
| 09 | 28 | | | | | | | | | | | 233577 | 248.21 | | | | |

Table 12: Similar to Table 9 for the Openstacks, Pathways, Pipesworld-NoTankage, Pipesworld-Tankage, TPP, and Trucks domains.

| | | $h^{\mathcal{F}}$ | | $h^{\mathcal{I}}$ | | $h^{\mathcal{FI}}$ | | $MS$-$10^4$ | | $MS$-$10^5$ | | $\text{HSP}^*_\text{F}$ | | blind | | $h_{\max}$ | |
|---|---|---|---|---|---|---|---|---|---|---|---|---|---|---|---|---|---|
| task | $h^*$ | nodes | time | nodes | time | nodes | time | nodes | time | nodes | time | nodes | time | nodes | time | nodes | time |
| psr-small-ipc4 | | | | | | | | | | | | | | | | | |
| 01 | 8 | 10 | 0.00 | 10 | 0.00 | 10 | 0.00 | 9 | 0.00 | 9 | 0.00 | 9 | 0.01 | 11 | 0.00 | 9 | 0.00 |
| 02 | 11 | 52 | 0.01 | 55 | 0.00 | 52 | 0.01 | 12 | 0.00 | 12 | 0.00 | 20 | 0.08 | 71 | 0.00 | 47 | 0.00 |
| 03 | 11 | 31 | 0.01 | 31 | 0.00 | 31 | 0.00 | 12 | 0.00 | 12 | 0.00 | 20 | 0.04 | 33 | 0.00 | 28 | 0.00 |
| 04 | 10 | 66 | 0.04 | 91 | 0.03 | 73 | 0.06 | 11 | 0.00 | 11 | 0.00 | 12 | 0.34 | 332 | 0.00 | 102 | 0.00 |
| 05 | 11 | 75 | 0.01 | 79 | 0.01 | 75 | 0.02 | 12 | 0.00 | 12 | 0.00 | 23 | 0.11 | 154 | 0.00 | 69 | 0.00 |
| 06 | 8 | 10 | 0.00 | 10 | 0.00 | 10 | 0.00 | 9 | 0.00 | 9 | 0.00 | 9 | 0.01 | 11 | 0.00 | 9 | 0.00 |
| 07 | 11 | 61 | 0.01 | 61 | 0.00 | 61 | 0.01 | 12 | 0.00 | 12 | 0.00 | 26 | 0.09 | 122 | 0.00 | 62 | 0.00 |
| 08 | 8 | 24 | 0.01 | 29 | 0.00 | 25 | 0.01 | 9 | 0.00 | 9 | 0.00 | 9 | 0.12 | 128 | 0.00 | 52 | 0.00 |
| 09 | 8 | 18 | 0.01 | 19 | 0.00 | 18 | 0.00 | 9 | 0.00 | 9 | 0.00 | 9 | 0.06 | 49 | 0.00 | 20 | 0.00 |
| 10 | 7 | 131 | 0.20 | 183 | 0.18 | 155 | 0.32 | 8 | 0.04 | 8 | 0.04 | 18 | 1.04 | 1358 | 0.00 | 376 | 0.01 |
| 11 | 19 | 149 | 0.03 | 149 | 0.02 | 149 | 0.04 | 20 | 0.00 | 20 | 0.00 | 96 | 0.19 | 153 | 0.00 | 142 | 0.00 |
| 12 | 16 | 120 | 0.03 | 123 | 0.02 | 120 | 0.04 | 17 | 0.00 | 17 | 0.00 | 40 | 0.17 | 153 | 0.00 | 113 | 0.00 |
| 13 | 15 | 90 | 0.02 | 90 | 0.01 | 90 | 0.02 | 16 | 0.00 | 16 | 0.00 | 59 | 0.16 | 95 | 0.00 | 86 | 0.00 |
| 14 | 9 | 19 | 0.00 | 19 | 0.00 | 19 | 0.00 | 10 | 0.00 | 10 | 0.00 | 13 | 0.06 | 27 | 0.00 | 18 | 0.00 |
| 15 | 10 | 1200 | 6.55 | 708 | 6.25 | 769 | 9.91 | 11 | 0.46 | 11 | 2.58 | 356 | 18.99 | 3562 | 0.02 | 324 | 0.02 |
| 16 | 25 | 2328 | 0.65 | 2158 | 0.34 | 2176 | 0.85 | 975 | 0.11 | 26 | 0.12 | 2287 | 1.34 | 2742 | 0.01 | 1876 | 0.01 |
| 17 | 9 | 15 | 0.00 | 15 | 0.00 | 15 | 0.00 | 10 | 0.00 | 10 | 0.00 | 13 | 0.03 | 16 | 0.00 | 14 | 0.00 |
| 18 | 12 | 85 | 0.03 | 90 | 0.01 | 85 | 0.03 | 13 | 0.00 | 13 | 0.00 | 29 | 0.21 | 158 | 0.00 | 91 | 0.00 |
| 19 | 25 | 8025 | 4.31 | 7856 | 2.19 | 7876 | 5.80 | 2910 | 0.27 | 26 | 0.77 | 6338 | 4.46 | 9009 | 0.04 | 6925 | 0.08 |
| 20 | 17 | 80 | 0.02 | 80 | 0.01 | 80 | 0.02 | 18 | 0.00 | 18 | 0.00 | 52 | 0.18 | 84 | 0.00 | 75 | 0.00 |
| 21 | 10 | 28 | 0.01 | 28 | 0.00 | 28 | 0.01 | 11 | 0.00 | 11 | 0.00 | 21 | 0.12 | 42 | 0.00 | 31 | 0.00 |
| 22 | 33 | 163299 | 405.65 | 176058 | 245.42 | 168685 | 617.45 | 34 | 0.28 | 34 | 0.87 | 22315 | 8.16 | 189516 | 0.67 | 177138 | 1.43 |
| 23 | 12 | 77 | 0.04 | 93 | 0.03 | 77 | 0.06 | 13 | 0.00 | 13 | 0.01 | 30 | 0.43 | 200 | 0.00 | 116 | 0.00 |
| 24 | 10 | 28 | 0.01 | 28 | 0.00 | 28 | 0.01 | 11 | 0.00 | 11 | 0.00 | 21 | 0.12 | 42 | 0.00 | 31 | 0.00 |
| 25 | 9 | 485 | 84.24 | 463 | 145.38 | 482 | 213.42 | 10 | 5.42 | 10 | 37.93 | 28 | 780.38 | 8913 | 0.12 | 854 | 0.18 |
| 26 | 17 | 144 | 0.05 | 150 | 0.03 | 146 | 0.06 | 18 | 0.00 | 18 | 0.00 | 52 | 0.28 | 182 | 0.00 | 142 | 0.00 |
| 27 | 21 | 616 | 0.33 | 675 | 0.21 | 650 | 0.49 | 22 | 0.01 | 22 | 0.01 | 179 | 0.85 | 773 | 0.00 | 616 | 0.00 |
| 28 | 14 | 79 | 0.02 | 79 | 0.01 | 79 | 0.02 | 15 | 0.00 | 15 | 0.00 | 49 | 0.29 | 95 | 0.00 | 79 | 0.00 |
| 29 | 21 | 142772 | 436.34 | 187319 | 307.77 | 159325 | 709.89 | 22 | 0.39 | 22 | 1.43 | 3337 | 7.12 | 244499 | 1.27 | 192459 | 2.32 |
| 30 | 22 | 1791 | 1.25 | 1982 | 0.80 | 1883 | 1.90 | 23 | 0.01 | 23 | 0.02 | 393 | 1.35 | 2295 | 0.01 | 1834 | 0.01 |
| 31 | 19 | 11278 | 25.93 | 6810 | 38.66 | 8297 | 53.43 | 2647 | 0.89 | 723 | 6.55 | 7530 | 32.97 | 53911 | 0.25 | 16766 | 0.36 |
| 32 | 24 | 431 | 0.17 | 431 | 0.10 | 431 | 0.25 | 25 | 0.00 | 25 | 0.00 | 352 | 0.74 | 435 | 0.00 | 424 | 0.00 |
| 33 | 21 | 1480 | 0.84 | 1436 | 0.30 | 1391 | 1.00 | 446 | 0.26 | 22 | 0.63 | 947 | 2.29 | 2291 | 0.01 | 1073 | 0.01 |
| 34 | 21 | 223 | 0.07 | 223 | 0.04 | 223 | 0.09 | 22 | 0.00 | 22 | 0.00 | 158 | 0.50 | 227 | 0.00 | 216 | 0.00 |
| 35 | 22 | 65965 | 160.36 | 63186 | 39.55 | 68281 | 199.30 | 24021 | 0.83 | 11113 | 6.36 | 7448 | 8.27 | 165170 | 0.63 | 61548 | 1.06 |
| 36 | 22 | 571766 | 392.49 | 371834 | 786.06 | 458402 | 1094.61 | 48350 | 2.98 | 2783 | 14.07 | 188564 | 111.99 | 1669788 | 9.44 | 717884 | 18.27 |
| 37 | 23 | 1307 | 1.29 | 1417 | 0.95 | 1363 | 2.10 | 24 | 0.02 | 24 | 0.01 | 277 | 2.10 | 1532 | 0.00 | 1342 | 0.01 |
| 38 | 13 | 301 | 0.20 | 372 | 0.15 | 326 | 0.32 | 14 | 0.01 | 14 | 0.01 | 33 | 0.74 | 562 | 0.00 | 357 | 0.00 |
| 39 | 23 | 2486 | 2.49 | 2942 | 1.64 | 2682 | 3.91 | 24 | 0.08 | 24 | 0.07 | 146 | 1.78 | 4103 | 0.01 | 2597 | 0.02 |
| 40 | 20 | | | 182608 | 1384.90 | | | 38837 | 1.88 | 7767 | 12.86 | 23371 | 87.91 | 1036992 | 6.74 | 229210 | 9.51 |
| 41 | 10 | 31 | 0.01 | 34 | 0.00 | 31 | 0.01 | 11 | 0.00 | 11 | 0.00 | 21 | 0.16 | 54 | 0.00 | 35 | 0.00 |
| 42 | 30 | 1855 | 0.50 | 1747 | 0.17 | 1739 | 0.59 | 1117 | 0.18 | 31 | 0.18 | 1773 | 1.29 | 1908 | 0.01 | 1636 | 0.01 |
| 43 | 20 | 328 | 0.09 | 328 | 0.05 | 328 | 0.12 | 21 | 0.00 | 21 | 0.00 | 256 | 0.50 | 333 | 0.00 | 315 | 0.00 |
| 44 | 19 | 2990 | 3.25 | 3430 | 2.30 | 3121 | 5.24 | 20 | 0.05 | 20 | 0.05 | 407 | 2.18 | 4142 | 0.01 | 3235 | 0.02 |
| 45 | 20 | 347 | 0.16 | 376 | 0.11 | 359 | 0.25 | 21 | 0.01 | 21 | 0.00 | 121 | 0.74 | 434 | 0.00 | 358 | 0.00 |
| 46 | 34 | 60888 | 51.77 | 61842 | 21.14 | 61563 | 68.33 | 36941 | 0.67 | 32582 | 4.05 | 19865 | 6.91 | 80785 | 0.25 | 65984 | 0.63 |
| 47 | 27 | 4104 | 5.27 | 4522 | 3.93 | 4284 | 8.70 | 28 | 0.04 | 28 | 0.04 | 515 | 2.32 | 5075 | 0.01 | 4406 | 0.02 |
| 48 | 37 | | | | | | | 129627 | 2.37 | 2500 | 11.08 | 200559 | 101.21 | | | 19020089 | 286.02 |
| 49 | 47 | | | | | | | 2048368 | 15.84 | 594399 | 23.32 | 2772875 | 1408.64 | | | | |
| 50 | 23 | 637 | 0.39 | 659 | 0.26 | 645 | 0.60 | 24 | 0.02 | 24 | 0.02 | 390 | 1.40 | 690 | 0.00 | 642 | 0.00 |
| rovers-ipc5 | | | | | | | | | | | | | | | | | |
| 01 | 10 | 147 | 0.01 | 147 | 0.01 | 147 | 0.02 | 11 | 0.03 | 11 | 0.03 | 48 | 0.07 | 1104 | 0.00 | 283 | 0.00 |
| 02 | 8 | 44 | 0.01 | 44 | 0.01 | 44 | 0.01 | 9 | 0.00 | 9 | 0.00 | 16 | 0.03 | 254 | 0.00 | 129 | 0.00 |
| 03 | 11 | 672 | 0.11 | 419 | 0.05 | 448 | 0.10 | 12 | 0.11 | 12 | 0.12 | 804 | 0.16 | 3543 | 0.02 | 757 | 0.00 |
| 04 | 8 | 47 | 0.02 | 20 | 0.00 | 24 | 0.01 | 9 | 0.04 | 9 | 0.04 | 58 | 0.08 | 897 | 0.00 | 223 | 0.00 |
| 05 | 22 | 808084 | 237.13 | 410712 | 123.64 | 522937 | 231.28 | 617267 | 11.48 | 375808 | 18.46 | 298400 | 101.65 | 8559690 | 126.19 | 4318309 | 81.53 |
| 07 | 18 | | | 741649 | 517.18 | 1682245 | 1780.27 | 3280884 | 51.02 | 2212903 | 59.20 | 1459792 | 866.93 | | | 9618062 | 199.91 |
| 12 | 19 | | | | | | | | | 5187273 | 166.77 | | | | | | |
| satellite-ipc4 | | | | | | | | | | | | | | | | | |
| 01 | 9 | 24 | 0.00 | 32 | 0.00 | 29 | 0.00 | 10 | 0.00 | 10 | 0.00 | 46 | 0.06 | 89 | 0.00 | 59 | 0.00 |
| 02 | 13 | 86 | 0.02 | 337 | 0.10 | 241 | 0.13 | 14 | 0.01 | 14 | 0.01 | 646 | 0.21 | 1728 | 0.01 | 940 | 0.00 |
| 03 | 11 | 2249 | 1.24 | 656 | 0.53 | 728 | 0.82 | 12 | 0.56 | 12 | 0.64 | 1945 | 0.93 | 15185 | 0.17 | 6822 | 0.11 |
| 04 | 17 | 9817 | 10.65 | 14860 | 24.90 | 11250 | 26.18 | 4152 | 0.99 | 18 | 4.43 | 15890 | 9.50 | 345663 | 4.70 | 180815 | 3.37 |
| 05 | 15 | 279569 | 1251.83 | 46453 | 515.80 | 61692 | 877.26 | 81972 | 7.26 | 148667 | 69.28 | 267513 | 565.18 | | | | |
| 06 | 20 | 1496577 | 968.24 | 1572327 | 1721.87 | | | 2769229 | 74.73 | 307962 | 32.52 | | | | | 10751017 | 371.43 |
| zenotravel-ipc3 | | | | | | | | | | | | | | | | | |
| 01 | 1 | 2 | 0.01 | 2 | 0.00 | 2 | 0.01 | 2 | 0.00 | 2 | 0.00 | 2 | 0.45 | 2 | 0.00 | 2 | 0.00 |
| 02 | 6 | 17 | 0.02 | 18 | 0.02 | 17 | 0.02 | 7 | 0.00 | 7 | 0.00 | 9 | 0.46 | 58 | 0.00 | 22 | 0.00 |
| 03 | 6 | 28 | 0.08 | 18 | 0.12 | 12 | 0.11 | 7 | 0.21 | 7 | 0.90 | 40 | 3.42 | 5160 | 0.04 | 492 | 0.02 |
| 04 | 8 | 99 | 0.15 | 88 | 0.26 | 81 | 0.30 | 9 | 0.20 | 9 | 0.89 | 215 | 3.44 | 5256 | 0.03 | 665 | 0.01 |
| 05 | 11 | 177 | 0.32 | 220 | 0.22 | 136 | 0.36 | 12 | 0.25 | 12 | 1.90 | 422 | 7.70 | 82289 | 0.63 | 12466 | 0.33 |
| 06 | 11 | 2287 | 5.51 | 1144 | 2.00 | 504 | 2.40 | 12 | 0.38 | 12 | 3.54 | 1957 | 11.81 | 596531 | 5.90 | 85931 | 2.47 |
| 07 | 15 | 5088 | 9.63 | 4234 | 5.56 | 4199 | 10.58 | 16 | 0.38 | 16 | 3.48 | 34890 | 30.36 | 405626 | 3.56 | 115348 | 2.60 |
| 08 | 11 | 3268 | 43.96 | 1026 | 8.92 | 1655 | 30.06 | 14354 | 2.00 | 12 | 14.48 | 83533 | 292.05 | | | 687846 | 50.76 |
| 09 | 21 | | | | | | | 2517035 | 51.18 | 611457 | 30.47 | | | | | | |
| 10 | 22 | | | | | | | 1322871 | 34.84 | 137872 | 25.44 | | | | | | |
| 11 | 14 | | | 76904 | 1090.67 | | | 310030 | 11.28 | 110726 | 26.65 | | | | | | |

Table 13: Similar to Table 9 for the PSR, Rovers, Satellite, and Zenotravel domains.

| | | $h^{\mathcal{F}}$ | | $h^{\mathcal{I}}$ | | $h^{\mathcal{FI}}$ | | $MS$-$10^4$ | | $MS$-$10^5$ | | $\text{HSP}^*_{\text{F}}$ | | blind | | $h_{\max}$ | |
|---|---|---|---|---|---|---|---|---|---|---|---|---|---|---|---|---|---|
| task | $h^*$ | nodes | time | nodes | time | nodes | time | nodes | time | nodes | time | nodes | time | nodes | time | nodes | time |
| schedule-strips | | | | | | | | | | | | | | | | | |
| 02-0 | 3 | 5 | 0.15 | 5 | 0.14 | 5 | 0.22 | 4 | 511.10 | 4 | 1743.32 | 5 | 577.39 | 76 | 0.02 | 5 | 0.09 |
| 02-1 | 2 | 3 | 0.16 | 4 | 0.11 | 3 | 0.18 | 3 | 104.98 | | | 3 | 754.26 | 6 | 0.02 | 3 | 0.07 |
| 02-2 | 2 | 3 | 0.32 | 3 | 0.17 | 3 | 0.40 | 3 | 231.99 | | | 3 | 495.56 | 5 | 0.02 | 3 | 0.07 |
| 02-3 | 3 | 26 | 0.50 | 37 | 0.76 | 26 | 0.61 | 4 | 56.51 | | | 4 | 658.90 | 529 | 0.03 | 95 | 0.45 |
| 02-4 | 3 | 68 | 1.34 | 188 | 2.24 | 220 | 7.20 | | | | | 4 | 484.62 | 543 | 0.03 | 108 | 0.44 |
| 02-5 | 2 | 3 | 0.33 | 3 | 0.14 | 3 | 0.38 | 3 | 363.11 | | | 3 | 667.32 | 3 | 0.03 | 3 | 0.07 |
| 02-6 | 2 | 3 | 0.14 | 5 | 0.12 | 3 | 0.17 | 3 | 121.84 | | | 3 | 697.42 | 6 | 0.02 | 3 | 0.06 |
| 02-7 | 2 | 3 | 0.30 | 3 | 0.13 | 3 | 0.34 | 3 | 323.77 | | | 3 | 604.06 | 13 | 0.02 | 3 | 0.07 |
| 02-8 | 2 | 3 | 0.32 | 3 | 0.14 | 3 | 0.38 | 3 | 316.53 | | | 3 | 668.79 | 8 | 0.02 | 3 | 0.07 |
| 02-9 | 3 | 5 | 0.15 | 5 | 0.14 | 5 | 0.22 | 4 | 251.46 | | | 5 | 577.16 | 76 | 0.03 | 5 | 0.09 |
| 03-0 | 4 | 40 | 2.72 | 407 | 12.16 | 140 | 14.55 | | | | | | | 11915 | 0.60 | 1127 | 8.98 |
| 03-1 | 2 | 3 | 0.51 | 3 | 0.35 | 3 | 0.72 | | | | | | | 31 | 0.04 | 25 | 0.37 |
| 03-2 | 4 | 27 | 1.16 | 50 | 1.83 | 33 | 2.33 | 5 | 191.03 | | | | | 3617 | 0.23 | 1228 | 9.56 |
| 03-3 | 4 | 15 | 0.79 | 91 | 2.39 | 15 | 0.96 | 5 | 259.13 | | | | | 3379 | 0.23 | 170 | 1.85 |
| 03-4 | 3 | 4 | 1.11 | 16 | 2.08 | 4 | 1.52 | | | | | 4 | 1223.90 | 301 | 0.06 | 22 | 0.27 |
| 03-5 | 4 | 73 | 6.13 | 471 | 16.71 | 74 | 8.32 | 5 | 682.30 | | | | | 12217 | 0.64 | 1175 | 12.43 |
| 03-6 | 4 | 72 | 1.27 | 75 | 1.80 | 69 | 1.33 | 5 | 121.58 | | | | | 2663 | 0.19 | 1542 | 11.73 |
| 03-7 | 4 | 28 | 1.05 | 50 | 1.83 | 28 | 1.43 | 5 | 195.72 | | | | | 12859 | 0.68 | 1323 | 13.47 |
| 03-8 | 4 | 273 | 11.53 | 266 | 11.46 | 273 | 17.48 | | | | | | | 12616 | 0.65 | 1590 | 11.13 |
| 03-9 | 4 | 8 | 0.96 | 31 | 1.77 | 14 | 2.13 | 5 | 235.48 | | | | | 4339 | 0.27 | 913 | 7.69 |
| 04-0 | 5 | 373 | 13.91 | 1498 | 74.46 | 167 | 24.60 | | | | | | | 312193 | 26.88 | 22993 | 273.38 |
| 04-1 | 6 | 17559 | 1373.80 | 10707 | 626.54 | | | 7 | 1115.76 | | | | | 552069 | 49.79 | | |
| 04-2 | 5 | 209 | 9.88 | 406 | 20.85 | 66 | 5.30 | | | | | | | 47696 | 4.97 | 9703 | 131.69 |
| 04-3 | 5 | 142 | 10.47 | 674 | 33.29 | 251 | 29.28 | 6 | 267.29 | | | | | 89272 | 8.74 | 12941 | 163.84 |
| 04-4 | 5 | 921 | 64.48 | 450 | 46.95 | 574 | 116.65 | | | | | | | 62013 | 6.03 | 13614 | 168.07 |
| 04-5 | 6 | 483 | 47.25 | 4544 | 268.77 | 850 | 187.46 | 7 | 837.68 | | | | | | | 107978 | 1399.99 |
| 04-6 | 6 | 779 | 27.09 | 11610 | 361.74 | 1834 | 102.68 | 7 | 459.19 | | | | | | | 107115 | 1001.40 |
| 04-7 | 5 | 99 | 18.48 | 424 | 38.04 | 163 | 40.04 | 6 | 936.68 | | | | | 61327 | 5.97 | 8683 | 103.50 |
| 04-8 | 5 | 102 | 16.01 | 573 | 31.87 | 111 | 23.35 | 6 | 711.65 | | | | | 340467 | 29.56 | 15122 | 181.98 |
| 04-9 | 4 | 1043 | 80.06 | 996 | 76.64 | 1050 | 143.48 | 5 | 316.22 | | | | | 41673 | 4.27 | 5480 | 83.69 |
| 05-0 | 5 | 163 | 41.61 | 483 | 63.23 | 167 | 62.53 | | | | | | | 143350 | 22.71 | 43336 | 751.35 |
| 05-1 | 6 | 2701 | 213.92 | | | 1257 | 286.28 | | | | | | | | | | |
| 05-3 | 7 | | | | | 13622 | 1693.68 | | | | | | | | | | |
| 05-4 | 6 | 989 | 100.02 | 3433 | 229.05 | 582 | 100.05 | | | | | | | | | | |
| 05-5 | 6 | 198 | 21.67 | 9550 | 767.94 | 347 | 68.64 | | | | | | | | | 120602 | 989.42 |
| 05-6 | 7 | 6033 | 743.61 | | | 10325 | 1508.56 | | | | | | | | | | |
| 05-7 | 6 | 944 | 131.19 | 17562 | 1446.20 | 2107 | 379.70 | | | | | | | | | | |
| 05-8 | 7 | 1190 | 172.59 | | | 2709 | 730.54 | | | | | | | | | | |
| 05-9 | 6 | 1537 | 140.49 | 15829 | 1248.19 | 2717 | 547.56 | | | | | | | | | | |
| 06-2 | 6 | 888 | 243.14 | | | 1709 | 730.36 | | | | | | | | | | |
| 06-4 | 8 | 11535 | 1776.87 | | | | | | | | | | | | | | |
| 07-0 | 7 | 2489 | 786.76 | | | | | | | | | | | | | | |
| 07-9 | 8 | 6829 | 1559.86 | | | | | | | | | | | | | | |

Table 14: Similar to Table 9 for the (non-IPC) Schedule-STRIPS domain.

| | | $h^{\mathcal{F}}$ | | $h^{\mathcal{I}}$ | | $h^{\mathcal{FI}}$ | | *MS*-$10^4$ | | *MS*-$10^5$ | | HSP$^*_{\mathrm{F}}$ | | blind | | $h_{\max}$ | |
|---|---|---|---|---|---|---|---|---|---|---|---|---|---|---|---|---|---|
| task | $h^*$ | nodes | time | nodes | time | nodes | time | nodes | time | nodes | time | nodes | time | nodes | time | nodes | time |
| airport-ipc4 | | | | | | | | | | | | | | | | | |
| 01 | 8 | 10 | 0.01 | 9 | 0.00 | 9 | 0.00 | 9 | 0.00 | 9 | 0.00 | 9 | 0.72 | 11 | 0.00 | 9 | 0.00 |
| 02 | 9 | 12 | 0.01 | 15 | 0.00 | 15 | 0.01 | 10 | 0.00 | 10 | 0.00 | 10 | 1.23 | 13 | 0.00 | 10 | 0.00 |
| 03 | 17 | 86 | 0.02 | 133 | 0.01 | 93 | 0.02 | 18 | 0.04 | 18 | 0.03 | 29 | 5.10 | 164 | 0.00 | 57 | 0.00 |
| 04 | 20 | 22 | 0.01 | 21 | 0.00 | 21 | 0.01 | 21 | 0.02 | 21 | 0.01 | 21 | 1.32 | 23 | 0.00 | 21 | 0.00 |
| 05 | 21 | 23 | 0.08 | 30 | 0.02 | 27 | 0.09 | 22 | 0.01 | 22 | 0.01 | 22 | 46.54 | 27 | 0.00 | 22 | 0.00 |
| 06 | 41 | 513 | 0.16 | 639 | 0.06 | 567 | 0.19 | 42 | 0.16 | 42 | 0.17 | 42 | 123.13 | 738 | 0.01 | 418 | 0.02 |
| 07 | 41 | 514 | 0.15 | 632 | 0.05 | 550 | 0.19 | 42 | 0.17 | 42 | 0.17 | 42 | 117.56 | 742 | 0.01 | 405 | 0.02 |
| 08 | 62 | 12733 | 1.89 | 21544 | 1.36 | 14398 | 4.02 | 24372 | 25.42 | 9623 | 1549.13 | 203 | 602.09 | 27032 | 0.28 | 9687 | 0.90 |
| 09 | 71 | 88670 | 16.58 | 136717 | 9.60 | 90412 | 38.78 | 152408 | 64.92 | 89525 | 466.14 | 12956 | 993.07 | 175717 | 2.47 | 56484 | 7.62 |
| 10 | 18 | 19 | 0.01 | 19 | 0.01 | 19 | 0.01 | 19 | 0.02 | 19 | 0.01 | 19 | 2.45 | 21 | 0.00 | 19 | 0.00 |
| 11 | 21 | 23 | 0.10 | 30 | 0.03 | 27 | 0.12 | 22 | 0.02 | 22 | 0.01 | 22 | 65.36 | 27 | 0.00 | 22 | 0.01 |
| 12 | 39 | 475 | 0.20 | 728 | 0.07 | 568 | 0.25 | 40 | 0.21 | 40 | 0.21 | 40 | 169.02 | 873 | 0.01 | 392 | 0.03 |
| 13 | 37 | 434 | 0.20 | 663 | 0.07 | 479 | 0.24 | 38 | 0.20 | 38 | 0.21 | 38 | 134.87 | 822 | 0.01 | 342 | 0.03 |
| 14 | 60 | 12040 | 2.90 | 25110 | 1.86 | 15948 | 4.64 | 30637 | 51.23 | 8968 | 238.16 | 62 | 714.76 | 35384 | 0.39 | 9196 | 1.11 |
| 15 | 58 | 11477 | 2.74 | 23317 | 1.71 | 14557 | 4.25 | 28798 | 46.20 | 8931 | 267.81 | 59 | 647.05 | 33798 | 0.38 | 8200 | 1.01 |
| 16 | 79 | 267277 | 77.39 | 824491 | 97.12 | 353592 | 114.58 | 1031524 | 200.95 | 305340 | 1077.90 | | | 1247467 | 19.72 | 221993 | 49.03 |
| 17 | 88 | 2460667 | 708.82 | | | 2678689 | 1235.79 | | | | | | | | | 1043661 | 310.89 |
| 19 | 90 | 1354353 | 592.53 | 3400142 | 492.06 | 1462739 | 660.17 | | | | | | | | | 831632 | 253.21 |
| 21 | 101 | 5156 | 48.29 | 11259 | 3.72 | 4773 | 51.13 | 7326 | 372.92 | 102 | 10.28 | | | 18809 | 0.42 | 3184 | 1.12 |
| 22 | 148 | 606648 | 1110.09 | 1063668 | 318.90 | 477836 | 1082.91 | 1119943 | 762.02 | | | | | | | 159967 | 105.29 |
| 36 | 109 | 9504 | 129.73 | 34986 | 14.41 | 9436 | 140.75 | 34365 | 853.70 | | | | | 63061 | 1.44 | | |
| 37 | 142 | 37873 | 820.33 | | | | | | | | | | | | | | |
| blocks-ipc2 | | | | | | | | | | | | | | | | | |
| 04-0 | 6 | 15 | 0.00 | 46 | 0.00 | 17 | 0.00 | 7 | 0.03 | 7 | 0.03 | 7 | 0.36 | 93 | 0.00 | 25 | 0.00 |
| 04-1 | 10 | 14 | 0.00 | 31 | 0.00 | 15 | 0.00 | 11 | 0.04 | 11 | 0.03 | 11 | 0.39 | 66 | 0.00 | 23 | 0.00 |
| 04-2 | 6 | 7 | 0.00 | 26 | 0.00 | 10 | 0.00 | 7 | 0.04 | 7 | 0.03 | 7 | 0.38 | 63 | 0.00 | 18 | 0.00 |
| 05-0 | 12 | 32 | 0.00 | 302 | 0.01 | 113 | 0.00 | 13 | 0.30 | 13 | 0.96 | 13 | 1.32 | 467 | 0.00 | 145 | 0.00 |
| 05-1 | 10 | 37 | 0.00 | 280 | 0.00 | 98 | 0.00 | 11 | 0.29 | 11 | 0.96 | 11 | 1.36 | 567 | 0.00 | 135 | 0.00 |
| 05-2 | 16 | 152 | 0.00 | 596 | 0.00 | 348 | 0.01 | 17 | 0.29 | 17 | 0.95 | 17 | 1.49 | 792 | 0.00 | 297 | 0.00 |
| 06-0 | 12 | 33 | 0.00 | 766 | 0.01 | 207 | 0.01 | 13 | 0.95 | 13 | 8.56 | 13 | 4.10 | 1826 | 0.00 | 276 | 0.00 |
| 06-1 | 10 | 41 | 0.00 | 2395 | 0.03 | 578 | 0.02 | 11 | 0.90 | 11 | 8.34 | 11 | 4.17 | 4887 | 0.01 | 755 | 0.01 |
| 06-2 | 20 | 855 | 0.01 | 5444 | 0.05 | 3352 | 0.06 | 733 | 0.87 | 85 | 8.84 | 31 | 4.29 | 6385 | 0.02 | 2556 | 0.03 |
| 07-0 | 20 | 278 | 0.01 | 20183 | 0.28 | 4022 | 0.12 | 577 | 1.93 | 144 | 23.32 | 22 | 11.47 | 37157 | 0.14 | 5943 | 0.11 |
| 07-1 | 22 | 6910 | 0.10 | 59207 | 0.60 | 38539 | 0.67 | 10071 | 1.70 | 1835 | 21.05 | 174 | 11.25 | 63376 | 0.21 | 33194 | 0.46 |
| 07-2 | 20 | 1458 | 0.02 | 46009 | 0.52 | 18854 | 0.39 | 1855 | 1.59 | 782 | 20.37 | 90 | 10.99 | 55218 | 0.19 | 18293 | 0.29 |
| 08-0 | 18 | 1533 | 0.03 | 344157 | 5.46 | 69830 | 2.09 | 5557 | 3.67 | 678 | 36.80 | 25 | 26.00 | 519107 | 2.28 | 94671 | 2.07 |
| 08-1 | 20 | 10040 | 0.17 | 517514 | 7.22 | 191352 | 4.91 | 45711 | 3.88 | 11827 | 33.49 | 151 | 26.57 | 636498 | 2.60 | 199901 | 3.85 |
| 08-2 | 16 | 479 | 0.02 | 237140 | 4.08 | 32567 | 1.09 | 277 | 3.63 | 54 | 32.53 | 17 | 25.85 | 433144 | 1.93 | 52717 | 1.30 |
| 09-0 | 30 | 134185 | 3.10 | 7405904 | 117.14 | 4346535 | 118.23 | 1233374 | 16.00 | 971409 | 77.74 | 464 | 56.76 | 7984649 | 36.76 | 3840589 | 85.00 |
| 09-1 | 28 | 3435 | 0.09 | 4145371 | 77.54 | 917197 | 33.32 | 95068 | 7.35 | 58873 | 63.15 | 82 | 56.98 | 5914572 | 29.73 | 1200345 | 32.06 |
| 09-2 | 26 | 6379 | 0.17 | 4145278 | 78.21 | 923365 | 33.79 | 161719 | 13.54 | 20050 | 82.45 | 81 | 57.02 | 5963160 | 30.02 | 1211463 | 32.15 |
| 10-0 | 34 | 1524599 | 36.52 | | | | | | | | | 1800 | 114.26 | | | | |
| 10-1 | 32 | 610206 | 15.79 | | | | | | | 12063665 | 228.76 | 1835 | 115.19 | | | | |
| 10-2 | 34 | 1516087 | 37.71 | | | | | | | | | 3685 | 116.75 | | | | |
| 11-0 | 32 | | | | | | | | | 7046739 | 141.44 | 2678 | 213.32 | | | | |
| 11-1 | 30 | | | | | | | | | | | 1510 | 203.79 | | | | |
| 11-2 | 34 | | | | | | | | | | | 3984 | 213.97 | | | | |
| 12-0 | 34 | | | | | | | | | | | 1184 | 370.06 | | | | |
| 12-1 | 34 | | | | | | | | | | | 614 | 382.34 | | | | |
| 13-0 | 42 | | | | | | | | | | | 83996 | 860.45 | | | | |
| 13-1 | 44 | | | | | | | | | | | 163438 | 1104.27 | | | | |
| 14-0 | 38 | | | | | | | | | | | 2779 | 1063.02 | | | | |
| 14-1 | 36 | | | | | | | | | | | 7154 | 1087.40 | | | | |
| depots-ipc3 | | | | | | | | | | | | | | | | | |
| 01 | 10 | 114 | 0.01 | 279 | 0.01 | 161 | 0.02 | 11 | 0.00 | 11 | 0.00 | 45 | 0.77 | 329 | 0.00 | 136 | 0.00 |
| 02 | 15 | 1134 | 0.08 | 9344 | 0.31 | 2638 | 0.22 | 738 | 3.24 | 16 | 1.14 | 898 | 11.56 | 15404 | 0.11 | 3771 | 0.17 |
| 03 | 27 | 134428 | 8.59 | 2520703 | 159.84 | 581726 | 66.43 | 348288 | 20.69 | 239313 | 222.35 | 103089 | 247.13 | 2930398 | 27.20 | 1204646 | 97.62 |
| 04 | 30 | 1254545 | 101.18 | | | 5835295 | 923.87 | 1284048 | 52.05 | 1273762 | 529.34 | | | | | | |
| 07 | 21 | 109765 | 9.17 | 4271196 | 336.59 | 487961 | 76.02 | 211820 | 37.54 | | | 41328 | 324.19 | 6501100 | 71.58 | 1331701 | 166.76 |
| 10 | 24 | 2964635 | 283.55 | | | 6081478 | 1187.66 | 3241083 | 157.52 | | | | | | | | |
| 13 | 25 | 1003709 | 152.30 | | | 8161872 | 1559.21 | 1427824 | 116.06 | | | | | | | | |
| driverlog-ipc3 | | | | | | | | | | | | | | | | | |
| 01 | 7 | 49 | 0.00 | 37 | 0.00 | 37 | 0.00 | 8 | 0.04 | 8 | 0.03 | 44 | 0.47 | 182 | 0.00 | 20 | 0.00 |
| 02 | 19 | 15713 | 0.42 | 18452 | 0.27 | 15794 | 0.55 | 20 | 0.13 | 20 | 0.26 | 15998 | 4.55 | 68927 | 0.36 | 54283 | 0.52 |
| 03 | 12 | 164 | 0.00 | 190 | 0.00 | 163 | 0.01 | 13 | 0.16 | 13 | 0.25 | 863 | 1.25 | 16031 | 0.09 | 2498 | 0.03 |
| 04 | 16 | 6161 | 0.42 | 10778 | 0.30 | 7665 | 0.62 | 17 | 0.49 | 17 | 2.41 | 22933 | 12.20 | 999991 | 8.12 | 393673 | 6.56 |
| 05 | 18 | 13640 | 1.01 | 11400 | 0.36 | 10984 | 1.07 | 2614 | 0.60 | 19 | 4.58 | 24877 | 18.77 | 6290803 | 61.57 | 1724611 | 34.73 |
| 06 | 11 | 608 | 0.09 | 795 | 0.06 | 492 | 0.11 | 291 | 1.35 | 12 | 9.72 | 3804 | 10.08 | 681757 | 7.64 | 54451 | 1.71 |
| 07 | 13 | 864 | 0.14 | 1730 | 0.11 | 1006 | 0.21 | 14 | 1.42 | 14 | 15.35 | 25801 | 41.34 | 6349767 | 81.53 | 493480 | 17.31 |
| 08 | 22 | 669994 | 75.74 | 1181268 | 61.32 | 694996 | 104.59 | 287823 | 7.34 | 2952 | 20.31 | | | | | | |
| 09 | 22 | 150255 | 14.72 | 198651 | 11.44 | 164109 | 23.06 | 15504 | 1.70 | 23 | 10.43 | | | | | | |
| 10 | 17 | 4304 | 0.44 | 16099 | 1.21 | 4037 | 0.69 | 18 | 1.64 | 18 | 18.54 | 18234 | 68.22 | | | | |
| 11 | 19 | 43395 | 4.99 | 41445 | 2.22 | 39069 | 5.90 | 34137 | 1.99 | 10790 | 17.01 | 559623 | 1193.00 | | | 6141130 | 330.22 |
| 13 | 26 | 1303099 | 325.71 | 1014865 | 144.64 | 1098694 | 422.20 | 1298884 | 19.52 | 870875 | 35.33 | | | | | | |

Table 15: Runtimes of cost-optimal heuristic-search planners on the Airport, Blocksworld, Depots, and Driverlog domains. The description of the planners is given in Section 6; here the fork-decomposition heuristics are via structural-pattern databases. Column *task* denotes problem instance, column $h^*$ denotes optimal solution length. Other columns capture the run *time* and number of expanded *nodes*.

| | | $h^{\mathfrak{F}}$ | | $h^{\mathfrak{I}}$ | | $h^{\mathfrak{FI}}$ | | $MS$-$10^4$ | | $MS$-$10^5$ | | $\text{HSP}^*_{\text{F}}$ | | blind | | $h_{\max}$ | |
|---|---|---|---|---|---|---|---|---|---|---|---|---|---|---|---|---|---|
| task | $h^*$ | nodes | time | nodes | time | nodes | time | nodes | time | nodes | time | nodes | time | nodes | time | nodes | time |
| freecell-ipc3 | | | | | | | | | | | | | | | | | |
| 01 | 8 | 234 | 0.10 | 974 | 0.15 | 274 | 0.17 | 87 | 3.12 | 9 | 38.74 | 9 | 13.01 | 3437 | 0.03 | 1043 | 0.15 |
| 02 | 14 | 30960 | 1.95 | 75150 | 5.53 | 37131 | 4.79 | 31487 | 40.40 | | | 466 | 70.29 | 130883 | 1.46 | 41864 | 10.77 |
| 03 | 18 | 197647 | 14.41 | 533995 | 78.27 | 240161 | 51.24 | 95805 | 140.96 | | | 1589 | 169.39 | 944843 | 11.45 | 210503 | 75.62 |
| 04 | 26 | 997836 | 60.67 | 1921470 | 232.95 | 1218329 | 213.02 | 943074 | 86.78 | | | 15848 | 341.02 | 3021326 | 38.80 | 600525 | 247.70 |
| 05 | 30 | 6510089 | 448.22 | | | | | 5950977 | 243.74 | | | 40642 | 916.44 | | | 1408035 | 1062.25 |
| grid-ipc1 | | | | | | | | | | | | | | | | | |
| 01 | 14 | 571 | 0.60 | 1117 | 0.34 | 472 | 0.78 | 660 | 8.63 | 467 | 121.10 | | | 6446 | 0.08 | 190 | 0.10 |
| 02 | 26 | 3330274 | 1078.55 | | | | | 3392724 | 50.35 | 3244132 | 241.94 | | | | | 664016 | 231.26 |
| gripper-ipc1 | | | | | | | | | | | | | | | | | |
| 01 | 11 | 214 | 0.00 | 240 | 0.00 | 214 | 0.00 | 12 | 0.00 | 12 | 0.00 | 33 | 0.11 | 236 | 0.00 | 208 | 0.00 |
| 02 | 17 | 1768 | 0.02 | 1832 | 0.01 | 1803 | 0.03 | 18 | 0.11 | 18 | 0.08 | 680 | 0.37 | 1826 | 0.01 | 1760 | 0.01 |
| 03 | 23 | 11626 | 0.19 | 11736 | 0.08 | 11689 | 0.22 | 11514 | 0.47 | 2094 | 1.75 | 7370 | 1.52 | 11736 | 0.04 | 11616 | 0.08 |
| 04 | 29 | 68380 | 1.46 | 68558 | 0.51 | 68479 | 1.63 | 68380 | 1.24 | 68190 | 8.05 | 55568 | 10.29 | 68558 | 0.27 | 68368 | 0.56 |
| 05 | 35 | 376510 | 10.07 | 376784 | 3.20 | 376653 | 11.11 | 376510 | 3.52 | 376510 | 19.46 | 344386 | 79.96 | 376772 | 1.59 | 376496 | 3.51 |
| 06 | 41 | 1982032 | 70.91 | 1982408 | 19.08 | 1982227 | 77.81 | 1982032 | 13.42 | 1982032 | 42.16 | 1911592 | 577.49 | 1982394 | 9.59 | 1982016 | 21.57 |
| 07 | 47 | 10091986 | 438.41 | 10092464 | 105.67 | 10092241 | 478.67 | 10091986 | 61.66 | 10091986 | 106.84 | | | 10092464 | 51.10 | 10091968 | 119.64 |
| logistics-ipc1 | | | | | | | | | | | | | | | | | |
| 01 | 26 | 77763 | 7.14 | 1469610 | 95.49 | 830292 | 98.59 | 1918881 | 41.03 | 949586 | 34.82 | 211955 | 1700.26 | | | | |
| 05 | 22 | 3293 | 0.46 | 850312 | 42.43 | 173477 | 18.19 | 768161 | 18.69 | 609393 | 35.27 | | | | | | |
| 31 | 13 | 436 | 0.03 | 1981 | 0.07 | 1284 | 0.09 | 494 | 0.42 | 14 | 2.11 | 481 | 6.58 | 155645 | 1.66 | 32282 | 0.57 |
| 32 | 20 | 392 | 0.01 | 2704 | 0.07 | 962 | 0.05 | 21 | 0.16 | 21 | 0.72 | 9598 | 7.08 | 245325 | 2.07 | 81156 | 1.00 |
| 33 | 27 | 312180 | 27.19 | | | 3617185 | 427.52 | | | 529338 | 32.55 | | | | | | |
| 35 | 30 | 477883 | 183.08 | | | | | | | | | | | | | | |
| logistics-ipc2 | | | | | | | | | | | | | | | | | |
| 04-0 | 20 | 21 | 0.00 | 193 | 0.00 | 65 | 0.00 | 21 | 0.03 | 21 | 0.05 | 21 | 0.34 | 11246 | 0.05 | 4884 | 0.03 |
| 04-1 | 19 | 20 | 0.00 | 570 | 0.01 | 293 | 0.00 | 20 | 0.03 | 20 | 0.04 | 20 | 0.37 | 9249 | 0.04 | 4185 | 0.03 |
| 04-2 | 15 | 16 | 0.00 | 117 | 0.00 | 79 | 0.00 | 16 | 0.04 | 16 | 0.05 | 16 | 0.36 | 4955 | 0.02 | 1205 | 0.01 |
| 05-0 | 27 | 28 | 0.00 | 2550 | 0.05 | 1171 | 0.03 | 28 | 0.10 | 28 | 0.38 | 28 | 0.58 | 109525 | 0.64 | 74694 | 0.59 |
| 05-1 | 17 | 18 | 0.00 | 675 | 0.01 | 427 | 0.01 | 18 | 0.10 | 18 | 0.38 | 18 | 0.72 | 22307 | 0.13 | 6199 | 0.05 |
| 05-2 | 8 | 9 | 0.00 | 24 | 0.00 | 13 | 0.00 | 9 | 0.09 | 9 | 0.38 | 9 | 0.78 | 1031 | 0.00 | 280 | 0.00 |
| 06-0 | 25 | 26 | 0.00 | 4249 | 0.09 | 2461 | 0.07 | 26 | 0.18 | 26 | 1.23 | 26 | 1.03 | 490207 | 3.40 | 202229 | 1.92 |
| 06-1 | 14 | 15 | 0.00 | 181 | 0.00 | 99 | 0.00 | 15 | 0.18 | 15 | 1.26 | 15 | 1.16 | 24881 | 0.16 | 3604 | 0.03 |
| 06-2 | 25 | 26 | 0.00 | 2752 | 0.06 | 1394 | 0.04 | 26 | 0.19 | 26 | 1.26 | 26 | 1.03 | 476661 | 3.32 | 200012 | 1.98 |
| 06-9 | 24 | 25 | 0.00 | 2395 | 0.04 | 1428 | 0.04 | 25 | 0.18 | 25 | 1.22 | 25 | 1.02 | 422557 | 2.95 | 133521 | 1.29 |
| 07-0 | 36 | 37 | 0.00 | 251287 | 7.52 | 98053 | 4.59 | 525 | 0.65 | 37 | 4.87 | 24317 | 35.46 | | | | |
| 07-1 | 44 | 1689 | 0.07 | 3532213 | 99.33 | 1705009 | 72.35 | 666324 | 8.83 | 49 | 4.94 | 362179 | 453.06 | | | | |
| 08-0 | 31 | 32 | 0.00 | 82476 | 2.69 | 35805 | 1.78 | 1042 | 0.96 | 32 | 6.90 | 14890 | 33.50 | | | | |
| 08-1 | 44 | 45 | 0.01 | 1183608 | 45.72 | 462244 | 25.36 | 16708 | 1.15 | 45 | 7.21 | 114155 | 198.84 | | | | |
| 09-0 | 36 | 37 | 0.00 | 351538 | 13.75 | 167038 | 9.76 | 20950 | 1.56 | 37 | 9.46 | 32017 | 83.16 | | | | |
| 09-1 | 30 | 31 | 0.00 | 59336 | 2.48 | 25359 | 1.73 | 31 | 1.27 | 31 | 9.43 | 6720 | 26.48 | | | | |
| 10-0 | 45 | 46 | 0.01 | | | | | | | 668834 | 29.73 | | | | | | |
| 10-1 | 42 | 43 | 0.01 | | | | | | | 1457130 | 43.00 | | | | | | |
| 11-0 | 48 | 697 | 0.09 | | | | | | | 701106 | 37.42 | | | | | | |
| 11-1 | 60 | 21959 | 2.22 | | | | | | | | | | | | | | |
| 12-0 | 42 | 43 | 0.02 | | | | | | | 775996 | 43.56 | | | | | | |
| 12-1 | 68 | 106534 | 11.64 | | | | | | | 2222340 | 87.47 | | | | | | |
| mprime-ipc1 | | | | | | | | | | | | | | | | | |
| 01 | 5 | 196 | 0.02 | 10 | 0.01 | 24 | 0.01 | 6 | 2.00 | 6 | 20.45 | 108 | 49.59 | 3636 | 0.07 | 68 | 0.04 |
| 02 | 7 | 11604 | 2.72 | 44045 | 80.68 | 2565 | 4.20 | 3317 | 88.58 | | | | | | | 12606 | 36.65 |
| 03 | 4 | 427 | 0.27 | 7 | 0.08 | 11 | 0.16 | 36 | 33.64 | 5 | 463.85 | | | 9868 | 0.67 | 5 | 0.07 |
| 04 | 8 | 3836 | 0.22 | 1775 | 0.10 | 1093 | 0.09 | 9 | 6.09 | 9 | 82.71 | 19076 | 781.74 | 599590 | 23.58 | 200 | 0.24 |
| 05 | 11 | 1745027 | 195.08 | | | 604756 | 592.60 | 1705009 | 127.53 | | | | | | | 1488157 | 1638.78 |
| 07 | 5 | 3314 | 0.25 | 47 | 0.03 | 346 | 0.08 | 1667 | 46.72 | | | | | 18744 | 0.56 | 11 | 0.04 |
| 08 | 6 | 485381 | 491.53 | 1376780 | 1426.21 | | | 1469752 | 403.45 | | | | | | | 7650 | 84.33 |
| 09 | 8 | 19838 | 2.92 | 100188 | 74.85 | 5227 | 6.31 | 21993 | 36.25 | | | | | 2197646 | 71.69 | 19023 | 30.26 |
| 11 | 7 | 9 | 0.02 | 219 | 0.03 | 8 | 0.03 | 8 | 4.69 | 8 | 62.68 | 22 | 394.26 | 73260 | 2.21 | 915 | 0.54 |
| 12 | 6 | 16320 | 1.89 | 8118 | 0.73 | 5243 | 1.13 | 34763 | 11.45 | 42055 | 143.27 | 25665 | 724.12 | 108652 | 3.50 | 1520 | 1.78 |
| 15 | 6 | | | | | | | | | | | | | | | 1039 | 178.55 |
| 16 | 6 | 252 | 0.76 | 51590 | 135.00 | 448 | 2.76 | 473 | 81.42 | | | | | 425144 | 32.17 | 7962 | 35.65 |
| 17 | 4 | 2746 | 10.47 | 453 | 18.78 | 451 | 21.40 | | | | | | | 172736 | 42.48 | 5 | 1.06 |
| 19 | 6 | 727401 | 521.78 | 95361 | 485.79 | | | 123039 | 313.25 | | | | | | | 36013 | 533.75 |
| 21 | 6 | 174221 | 55.09 | 34022 | 47.43 | 169400 | 392.30 | | | | | | | 1503293 | 103.23 | 15250 | 101.75 |
| 25 | 4 | 75 | 0.01 | 30 | 0.01 | 29 | 0.01 | 5 | 0.48 | 5 | 2.75 | 85 | 8.71 | 383 | 0.00 | 6 | 0.00 |
| 26 | 6 | 77622 | 24.69 | 147854 | 48.25 | 68239 | 106.35 | 172432 | 46.33 | 189154 | 454.69 | | | 819590 | 61.01 | 440 | 2.69 |
| 27 | 5 | 54 | 0.16 | 1772 | 1.50 | 9 | 0.18 | 6 | 11.59 | 6 | 154.43 | | | 84079 | 3.50 | 831 | 2.08 |
| 28 | 7 | 8 | 0.01 | 403 | 0.02 | 37 | 0.02 | 8 | 1.88 | 8 | 22.55 | 128 | 146.80 | 17333 | 0.25 | 211 | 0.06 |
| 29 | 4 | 182 | 0.12 | 56 | 0.08 | 32 | 0.11 | 5 | 14.92 | 5 | 201.40 | | | 3187 | 0.17 | 7 | 0.10 |
| 31 | 4 | 248 | 0.51 | 46 | 0.68 | 19 | 1.00 | 419 | 99.87 | | | | | 3584 | 0.19 | 11 | 0.17 |
| 32 | 7 | 31759 | 1.73 | 12436 | 1.46 | 11839 | 1.93 | 19429 | 21.61 | 7269 | 292.37 | 11073 | 1701.00 | 115479 | 2.75 | 3096 | 1.74 |
| 34 | 4 | 234 | 0.26 | 46 | 0.16 | 23 | 0.28 | 450 | 151.69 | | | | | 3618 | 0.19 | 11 | 0.18 |
| 35 | 5 | 392 | 0.07 | 290 | 0.06 | 84 | 0.08 | 359 | 3.63 | 6 | 43.43 | 706 | 96.55 | 2476 | 0.05 | 44 | 0.03 |

Table 16: Similar to Table 15 for the FREECELL, GRID, GRIPPER, LOGISTICS-IPC1, LOGISTICS-IPC2, and MPRIME domains.

| | | $h^{\mathcal{F}}$ | | $h^{\mathcal{I}}$ | | $h^{\mathcal{FI}}$ | | $MS$-$10^4$ | | $MS$-$10^5$ | | $\text{HSP}^*_{\text{F}}$ | | blind | | $h_{\max}$ | |
|---|---|---|---|---|---|---|---|---|---|---|---|---|---|---|---|---|---|
| task | $h^*$ | nodes | time | nodes | time | nodes | time | nodes | time | nodes | time | nodes | time | nodes | time | nodes | time |
| miconic-strips-ipc2 | | | | | | | | | | | | | | | | | |
| 01-0 | 4 | 5 | 0.00 | 5 | 0.00 | 5 | 0.00 | 5 | 0.00 | 5 | 0.00 | 5 | 0.01 | 5 | 0.00 | 5 | 0.00 |
| 01-1 | 3 | 5 | 0.00 | 5 | 0.00 | 5 | 0.00 | 4 | 0.00 | 4 | 0.00 | 4 | 0.00 | 5 | 0.00 | 4 | 0.00 |
| 01-2 | 4 | 5 | 0.00 | 5 | 0.00 | 5 | 0.00 | 5 | 0.00 | 5 | 0.00 | 5 | 0.01 | 5 | 0.00 | 5 | 0.00 |
| 01-3 | 4 | 5 | 0.00 | 5 | 0.00 | 5 | 0.00 | 5 | 0.00 | 5 | 0.00 | 5 | 0.00 | 5 | 0.00 | 5 | 0.00 |
| 01-4 | 4 | 5 | 0.00 | 5 | 0.00 | 5 | 0.00 | 5 | 0.00 | 5 | 0.00 | 5 | 0.01 | 5 | 0.00 | 5 | 0.00 |
| 02-0 | 7 | 19 | 0.00 | 22 | 0.00 | 19 | 0.00 | 8 | 0.00 | 8 | 0.00 | 26 | 0.01 | 30 | 0.00 | 20 | 0.00 |
| 02-1 | 7 | 21 | 0.00 | 23 | 0.00 | 21 | 0.00 | 8 | 0.00 | 8 | 0.00 | 26 | 0.01 | 30 | 0.00 | 22 | 0.00 |
| 02-2 | 7 | 21 | 0.00 | 23 | 0.00 | 21 | 0.00 | 8 | 0.00 | 8 | 0.00 | 27 | 0.00 | 30 | 0.00 | 22 | 0.00 |
| 02-3 | 7 | 24 | 0.00 | 24 | 0.00 | 24 | 0.00 | 8 | 0.00 | 8 | 0.00 | 20 | 0.01 | 26 | 0.00 | 17 | 0.00 |
| 02-4 | 7 | 19 | 0.00 | 22 | 0.00 | 19 | 0.00 | 8 | 0.00 | 8 | 0.00 | 23 | 0.01 | 31 | 0.00 | 20 | 0.00 |
| 03-0 | 10 | 86 | 0.00 | 129 | 0.00 | 98 | 0.00 | 11 | 0.00 | 11 | 0.00 | 100 | 0.03 | 193 | 0.00 | 105 | 0.00 |
| 03-1 | 11 | 120 | 0.00 | 168 | 0.00 | 147 | 0.00 | 12 | 0.00 | 12 | 0.00 | 140 | 0.02 | 218 | 0.00 | 150 | 0.00 |
| 03-2 | 10 | 137 | 0.00 | 143 | 0.00 | 137 | 0.00 | 11 | 0.00 | 11 | 0.00 | 122 | 0.02 | 164 | 0.00 | 92 | 0.00 |
| 03-3 | 10 | 96 | 0.00 | 153 | 0.00 | 117 | 0.00 | 11 | 0.00 | 11 | 0.00 | 131 | 0.02 | 197 | 0.00 | 130 | 0.00 |
| 03-4 | 10 | 103 | 0.00 | 149 | 0.00 | 115 | 0.00 | 11 | 0.00 | 11 | 0.00 | 114 | 0.02 | 190 | 0.00 | 114 | 0.00 |
| 04-0 | 14 | 524 | 0.00 | 843 | 0.00 | 686 | 0.01 | 15 | 0.01 | 15 | 0.01 | 669 | 0.10 | 1182 | 0.00 | 866 | 0.00 |
| 04-1 | 13 | 505 | 0.00 | 817 | 0.00 | 663 | 0.01 | 14 | 0.01 | 14 | 0.01 | 634 | 0.11 | 1176 | 0.00 | 860 | 0.00 |
| 04-2 | 15 | 685 | 0.00 | 942 | 0.00 | 802 | 0.01 | 16 | 0.01 | 16 | 0.01 | 822 | 0.12 | 1277 | 0.00 | 969 | 0.00 |
| 04-3 | 15 | 681 | 0.00 | 942 | 0.00 | 798 | 0.01 | 16 | 0.01 | 16 | 0.01 | 820 | 0.12 | 1319 | 0.00 | 970 | 0.00 |
| 04-4 | 15 | 685 | 0.00 | 942 | 0.00 | 802 | 0.01 | 16 | 0.01 | 16 | 0.01 | 821 | 0.12 | 1334 | 0.00 | 969 | 0.00 |
| 05-0 | 17 | 2468 | 0.03 | 4009 | 0.03 | 3307 | 0.05 | 18 | 0.06 | 18 | 0.05 | 2829 | 0.44 | 6350 | 0.03 | 4387 | 0.03 |
| 05-1 | 17 | 2807 | 0.04 | 4345 | 0.03 | 3677 | 0.06 | 18 | 0.06 | 18 | 0.05 | 3260 | 0.49 | 6602 | 0.03 | 4664 | 0.03 |
| 05-2 | 15 | 1596 | 0.02 | 2981 | 0.02 | 2275 | 0.04 | 16 | 0.06 | 16 | 0.05 | 1594 | 0.32 | 5565 | 0.03 | 3524 | 0.03 |
| 05-3 | 17 | 2256 | 0.03 | 3799 | 0.03 | 3104 | 0.05 | 18 | 0.06 | 18 | 0.05 | 2568 | 0.42 | 5944 | 0.03 | 4140 | 0.03 |
| 05-4 | 18 | 3210 | 0.04 | 4732 | 0.03 | 4267 | 0.06 | 19 | 0.06 | 19 | 0.05 | 3953 | 0.55 | 6949 | 0.04 | 5268 | 0.04 |
| 06-0 | 19 | 9379 | 0.18 | 17665 | 0.15 | 13531 | 0.26 | 20 | 0.18 | 20 | 0.32 | 9312 | 1.76 | 30786 | 0.20 | 21194 | 0.20 |
| 06-1 | 19 | 9106 | 0.17 | 18134 | 0.15 | 14052 | 0.27 | 20 | 0.18 | 20 | 0.32 | 10252 | 1.96 | 30093 | 0.20 | 21255 | 0.20 |
| 06-2 | 20 | 10900 | 0.20 | 19084 | 0.16 | 15111 | 0.28 | 21 | 0.18 | 21 | 0.32 | 11247 | 2.11 | 32390 | 0.21 | 21694 | 0.21 |
| 06-3 | 20 | 12127 | 0.23 | 21708 | 0.18 | 17807 | 0.33 | 21 | 0.17 | 21 | 0.32 | 14216 | 2.56 | 32574 | 0.21 | 24552 | 0.23 |
| 06-4 | 21 | 13784 | 0.24 | 23255 | 0.19 | 19536 | 0.35 | 22 | 0.17 | 22 | 0.32 | 16880 | 3.04 | 33793 | 0.22 | 26167 | 0.24 |
| 07-0 | 23 | 53662 | 1.19 | 96092 | 0.97 | 79449 | 1.76 | 24 | 0.32 | 24 | 1.75 | 56686 | 14.31 | 155466 | 1.22 | 116685 | 1.32 |
| 07-1 | 24 | 56328 | 1.24 | 99109 | 0.96 | 83677 | 1.83 | 7001 | 0.38 | 25 | 1.75 | 63035 | 16.33 | 164470 | 1.29 | 118494 | 1.33 |
| 07-2 | 22 | 48141 | 1.10 | 96139 | 0.94 | 78471 | 1.77 | 1646 | 0.33 | 23 | 1.71 | 55751 | 13.98 | 161342 | 1.27 | 119688 | 1.36 |
| 07-3 | 22 | 46867 | 1.08 | 93117 | 0.92 | 75424 | 1.69 | 1861 | 0.33 | 23 | 1.74 | 53121 | 13.27 | 155176 | 1.23 | 114649 | 1.30 |
| 07-4 | 25 | 84250 | 1.70 | 126595 | 1.22 | 111984 | 2.36 | 23159 | 0.52 | 26 | 1.71 | 96327 | 24.76 | 168219 | 1.33 | 140128 | 1.58 |
| 08-0 | 27 | 272580 | 7.05 | 485051 | 5.51 | 408114 | 10.53 | 41629 | 0.91 | 28 | 4.18 | 290649 | 104.18 | 755255 | 7.16 | 594032 | 7.95 |
| 08-1 | 27 | 284415 | 7.56 | 527216 | 6.01 | 446837 | 11.58 | 42679 | 0.90 | 28 | 4.25 | 339177 | 123.10 | 794365 | 7.56 | 636587 | 8.66 |
| 08-2 | 26 | 207931 | 5.60 | 414294 | 4.79 | 330993 | 8.90 | 37744 | 0.86 | 27 | 4.25 | 204614 | 73.39 | 731622 | 6.92 | 534711 | 7.37 |
| 08-3 | 28 | 369479 | 9.25 | 598031 | 6.74 | 527216 | 13.30 | 140453 | 1.94 | 29 | 4.21 | 435617 | 160.49 | 833421 | 7.97 | 690267 | 9.29 |
| 08-4 | 27 | 297516 | 7.74 | 507910 | 5.79 | 431432 | 11.04 | 62933 | 1.16 | 28 | 4.12 | 315339 | 111.84 | 771608 | 7.33 | 613253 | 8.43 |
| 09-0 | 31 | 1461729 | 43.82 | 2491975 | 32.67 | 2138656 | 63.58 | 684737 | 9.07 | 126918 | 8.89 | 1555286 | 794.93 | 3685552 | 41.04 | 3006991 | 49.12 |
| 09-1 | 30 | 1207894 | 37.47 | 2335166 | 30.76 | 1952916 | 59.39 | 406041 | 5.61 | 100937 | 8.73 | 1344815 | 683.05 | 3649801 | 40.32 | 2893803 | 47.54 |
| 09-2 | 30 | 1294691 | 40.03 | 2340411 | 30.97 | 1972234 | 59.25 | 442547 | 6.06 | 82946 | 8.63 | 1357681 | 692.11 | 3576134 | 39.61 | 2895182 | 47.26 |
| 09-3 | 32 | 1840936 | 52.68 | 2889342 | 38.12 | 2571844 | 74.47 | 765455 | 10.00 | 277302 | 11.14 | 2083168 | 1051.95 | 3796035 | 42.13 | 3304570 | 53.29 |
| 09-4 | 28 | 1252484 | 40.34 | 2352633 | 31.35 | 1944297 | 59.37 | 317692 | 4.65 | 29 | 7.03 | 1231554 | 605.01 | 3589382 | 39.29 | 2956995 | 48.84 |
| 10-0 | 33 | 5716041 | 202.37 | 10316603 | 153.80 | 8774563 | 300.08 | 2436164 | 35.24 | 863244 | 23.76 | | | 15804498 | 200.90 | 13267920 | 250.58 |
| 10-1 | 32 | 5601282 | 201.43 | 10789013 | 162.69 | 9144153 | 315.23 | 2340169 | 34.09 | 335745 | 15.68 | | | 16472633 | 208.39 | 13720664 | 256.89 |
| 10-2 | 32 | 4153191 | 155.86 | 9148616 | 138.69 | 7466572 | 265.86 | 1735477 | 25.29 | 486286 | 17.72 | | | 15867374 | 201.01 | 12497087 | 236.89 |
| 10-3 | 34 | 6108094 | 214.68 | 10960203 | 167.10 | 9400386 | 320.13 | 3952148 | 55.86 | 940556 | 24.24 | | | 16309701 | 208.42 | 13801989 | 262.53 |
| 10-4 | 33 | 5920127 | 211.40 | 11075136 | 170.82 | 9448049 | 322.74 | 2715866 | 39.44 | 625559 | 19.91 | | | 16472551 | 209.13 | 13925654 | 262.57 |
| 11-0 | 37 | | | | | | | 11473359 | 183.60 | 4724980 | 93.56 | | | | | | |
| 11-1 | 34 | 15349953 | 668.77 | | | | | 7535468 | 124.80 | 1934943 | 47.91 | | | | | | |
| 11-2 | 38 | | | | | | | 14645785 | 233.68 | 6330198 | 120.71 | | | | | | |
| 11-3 | 38 | | | | | | | | | 5809711 | 110.10 | | | | | | |
| 11-4 | 35 | | | | | | | 5853546 | 95.56 | 1082086 | 32.22 | | | | | | |
| mystery-ipc1 | | | | | | | | | | | | | | | | | |
| 01 | 5 | 7 | 0.00 | 6 | 0.00 | 6 | 0.00 | 6 | 0.20 | 6 | 1.79 | 10 | 5.38 | 30 | 0.00 | 8 | 0.00 |
| 02 | 7 | 2404 | 0.50 | 8012 | 11.19 | 722 | 1.01 | 1672 | 82.70 | | | | | 770852 | 21.85 | 2368 | 4.47 |
| 03 | 4 | 73 | 0.08 | 7 | 0.04 | 11 | 0.10 | 5 | 16.46 | 5 | 193.75 | 65 | 811.87 | 507 | 0.02 | 5 | 0.03 |
| 04 | ∞ | 0 | 0.00 | | | 0 | 0.00 | | | | | | | | | | |
| 07 | ∞ | 0 | 0.00 | 0 | 0.00 | 0 | 0.00 | 0 | 0.00 | 0 | 0.00 | 0 | 0.00 | 0 | 0.00 | 0 | 0.00 |
| 09 | 8 | 3049 | 0.37 | 10764 | 5.66 | 1215 | 1.01 | 3165 | 29.34 | | | 3868 | 670.08 | 138289 | 2.18 | 1458 | 1.44 |
| 11 | 7 | 9 | 0.01 | 33 | 0.01 | 8 | 0.01 | 8 | 1.51 | 8 | 16.59 | 34 | 41.20 | 426 | 0.00 | 19 | 0.00 |
| 12 | ∞ | 2102777 | 33.84 | 2093419 | 55.58 | 2093419 | 76.80 | 2102777 | 14.61 | 2102729 | 27.84 | | | 2102777 | 15.09 | 1177842 | 21.87 |
| 15 | 6 | 28271 | 20.21 | 21572 | 41.22 | 5079 | 44.42 | | | | | | | 279973 | 13.21 | 135 | 2.62 |
| 16 | ∞ | 0 | 0.15 | | | 0 | 0.27 | | | | | | | | | | |
| 17 | 4 | 354 | 1.32 | 85 | 2.74 | 83 | 3.59 | 198 | 445.85 | | | | | 5400 | 0.41 | 5 | 0.35 |
| 18 | ∞ | 0 | 0.00 | 0 | 0.00 | 0 | 0.00 | 0 | 0.00 | 0 | 0.00 | 0 | 0.00 | 0 | 0.00 | 0 | 0.00 |
| 19 | 6 | 21717 | 4.87 | 4968 | 5.26 | 16276 | 29.28 | 12478 | 96.38 | | | | | 133871 | 3.65 | 1516 | 5.44 |
| 20 | 7 | 89887 | 46.32 | 84572 | 153.53 | 53114 | 173.34 | 285069 | 59.22 | 547246 | 578.39 | | | 686125 | 23.28 | 718 | 3.76 |
| 24 | ∞ | 0 | 0.13 | | | 0 | 0.30 | | | | | | | | | | |
| 25 | 4 | 9 | 0.00 | 10 | 0.00 | 9 | 0.01 | 5 | 0.10 | 5 | 0.10 | 14 | 1.22 | 31 | 0.00 | 6 | 0.00 |
| 26 | 6 | 1807 | 0.27 | 1835 | 0.30 | 1344 | 0.69 | 2526 | 5.94 | 346 | 70.78 | 3107 | 291.36 | 8455 | 0.10 | 37 | 0.05 |
| 27 | 5 | 14 | 0.05 | 159 | 0.09 | 6 | 0.07 | 6 | 4.80 | 6 | 80.48 | 7 | 243.78 | 2174 | 0.03 | 73 | 0.04 |
| 28 | 7 | 8 | 0.00 | 47 | 0.00 | 15 | 0.00 | 8 | 0.63 | 8 | 6.77 | 31 | 16.67 | 843 | 0.00 | 32 | 0.00 |
| 29 | 4 | 31 | 0.04 | 14 | 0.03 | 10 | 0.06 | 5 | 8.94 | 5 | 107.10 | 27 | 536.30 | 153 | 0.01 | 7 | 0.02 |
| 30 | 9 | 23175 | 5.16 | 76480 | 169.86 | 7232 | 13.30 | 42112 | 28.07 | 44893 | 357.07 | | | 1977063 | 38.26 | 26686 | 28.27 |

Table 17: Similar to Table 15 for the Miconic and Mystery domains.

| | | $h^{\mathfrak{F}}$ | | $h^{\mathfrak{I}}$ | | $h^{\mathfrak{FI}}$ | | $MS$-$10^4$ | | $MS$-$10^5$ | | $\text{HSP}^*_{\text{F}}$ | | blind | | $h_{\max}$ | |
|---|---|---|---|---|---|---|---|---|---|---|---|---|---|---|---|---|---|
| task | $h^*$ | nodes | time | nodes | time | nodes | time | nodes | time | nodes | time | nodes | time | nodes | time | nodes | time |
| openstacks-ipc5 | | | | | | | | | | | | | | | | | |
| 01 | 23 | 2264 | 0.02 | 3895 | 0.03 | 3070 | 0.05 | 24 | 0.05 | 24 | 0.06 | 2000 | 1.02 | 4822 | 0.01 | 4016 | 0.03 |
| 02 | 23 | 2617 | 0.03 | 4485 | 0.04 | 3561 | 0.05 | 24 | 0.06 | 24 | 0.06 | 2378 | 1.07 | 5501 | 0.02 | 4594 | 0.04 |
| 03 | 23 | 2264 | 0.02 | 3895 | 0.03 | 3070 | 0.05 | 24 | 0.06 | 24 | 0.06 | 2000 | 1.02 | 4822 | 0.01 | 4016 | 0.03 |
| 04 | 23 | 2264 | 0.02 | 3895 | 0.03 | 3070 | 0.05 | 24 | 0.06 | 24 | 0.06 | 2000 | 1.02 | 4822 | 0.02 | 4016 | 0.03 |
| 05 | 23 | 2264 | 0.02 | 3895 | 0.03 | 3070 | 0.05 | 24 | 0.06 | 24 | 0.05 | 2000 | 1.02 | 4822 | 0.01 | 4016 | 0.03 |
| 06 | 45 | 366768 | 7.52 | 779710 | 18.93 | 587482 | 22.20 | 621008 | 4.85 | 279614 | 7.86 | 379735 | 217.37 | 882874 | 4.91 | 822514 | 18.71 |
| 07 | 46 | 410728 | 8.23 | 760668 | 18.33 | 606782 | 22.53 | 594758 | 4.69 | 264535 | 7.34 | 405564 | 226.32 | 836647 | 4.62 | 787163 | 17.81 |
| pathways-ipc5 | | | | | | | | | | | | | | | | | |
| 01 | 6 | 1624 | 0.00 | 1299 | 0.00 | 1299 | 0.00 | 7 | 1.14 | 7 | 0.79 | 1405 | 0.28 | 1624 | 0.00 | 36 | 0.00 |
| 02 | 12 | 2755 | 0.02 | 2307 | 0.01 | 2437 | 0.02 | 1946 | 2.56 | 13 | 42.11 | 990 | 0.29 | 2984 | 0.02 | 348 | 0.01 |
| 03 | 18 | 44928 | 0.62 | 20416 | 0.25 | 29106 | 0.43 | 21671 | 6.43 | 14901 | 129.23 | 14772 | 6.99 | 87189 | 1.06 | 4346 | 0.16 |
| 04 | 17 | 126950 | 2.66 | 33788 | 0.59 | 58738 | 1.31 | | | 98484 | 288.39 | 34206 | 27.00 | 456143 | 8.22 | 104068 | 2.61 |
| pipesworld-notankage-ipc4 | | | | | | | | | | | | | | | | | |
| 01 | 5 | 121 | 0.02 | 109 | 0.01 | 121 | 0.02 | 6 | 0.04 | 6 | 0.04 | 6 | 2.79 | 121 | 0.00 | 13 | 0.00 |
| 02 | 12 | 1413 | 0.06 | 1542 | 0.02 | 1413 | 0.08 | 169 | 0.30 | 13 | 0.17 | 435 | 3.07 | 1808 | 0.01 | 792 | 0.02 |
| 03 | 8 | 1742 | 0.14 | 3001 | 0.07 | 1742 | 0.18 | 9 | 1.15 | 9 | 0.69 | 128 | 3.84 | 3293 | 0.02 | 262 | 0.02 |
| 04 | 11 | 7007 | 0.45 | 8911 | 0.22 | 7007 | 0.59 | 651 | 1.95 | 12 | 7.05 | 812 | 8.84 | 16088 | 0.11 | 2925 | 0.13 |
| 05 | 8 | 4093 | 0.49 | 6805 | 0.26 | 4093 | 0.65 | 77 | 5.63 | 9 | 21.15 | 155 | 16.53 | 11128 | 0.12 | 1121 | 0.15 |
| 06 | 10 | 12401 | 1.44 | 27377 | 1.34 | 12401 | 2.03 | 1299 | 5.26 | 61 | 39.31 | 1151 | 23.41 | 49905 | 0.48 | 7102 | 0.72 |
| 07 | 8 | 4370 | 0.97 | 9168 | 0.77 | 4370 | 1.34 | 233 | 19.78 | 9 | 59.70 | 185 | 29.88 | 46502 | 0.57 | 2631 | 0.48 |
| 08 | 10 | 18851 | 3.84 | 56189 | 6.21 | 20584 | 6.42 | 561 | 12.42 | 497 | 94.69 | 1673 | 48.84 | 273585 | 3.39 | 22874 | 3.58 |
| 09 | 13 | 1092472 | 160.71 | 2419903 | 151.99 | 1092472 | 219.75 | 104875 | 25.48 | | | 10478 | 74.26 | 5513309 | 80.62 | 321861 | 68.99 |
| 10 | 18 | | | | | | | 2982520 | 66.89 | | | 689832 | 1439.64 | | | 11121245 | 1579.77 |
| 11 | 20 | 313952 | 27.68 | 472950 | 29.55 | 313952 | 43.90 | 90598 | 9.20 | 52159 | 43.24 | 108503 | 625.52 | 710123 | 3.86 | 107061 | 14.51 |
| 12 | 24 | 684234 | 75.72 | 1319980 | 133.58 | 686186 | 145.41 | 594661 | 12.41 | 416184 | 109.43 | 433296 | 1117.57 | 2467804 | 13.83 | 464982 | 56.82 |
| 13 | 16 | 39998 | 6.02 | 117475 | 18.08 | 40226 | 12.69 | 12835 | 34.28 | | | 24224 | 1019.65 | 481045 | 3.14 | 33417 | 6.38 |
| 14 | 30 | | | | | | | 13255718 | 119.54 | | | | | | | | |
| 15 | 26 | 1594863 | 254.43 | 2588849 | 192.90 | 1594863 | 353.40 | 648132 | 65.43 | | | | | 4921698 | 34.90 | 555619 | 105.49 |
| 17 | 22 | 5437393 | 1588.68 | | | | | 3200672 | 90.07 | | | | | | | | |
| 19 | 24 | | | | | | | 8767431 | 150.88 | | | | | | | | |
| 21 | 14 | 23833 | 4.02 | 49035 | 7.76 | 23833 | 7.87 | 3992 | 18.13 | 948 | 159.63 | | | 157782 | 1.31 | 8966 | 2.42 |
| 23 | 18 | 2285790 | 568.93 | 7047138 | 871.03 | 2282678 | 843.28 | 296506 | 49.11 | 104750 | 256.13 | | | | | 481859 | 229.00 |
| 24 | 24 | | | | | | | 7315150 | 142.82 | | | | | | | | |
| 41 | 12 | 502308 | 370.68 | | | 502308 | 1092.50 | | | | | | | | | 114257 | 250.18 |
| pipesworld-tankage-ipc4 | | | | | | | | | | | | | | | | | |
| 01 | 5 | 77 | 0.02 | 126 | 0.01 | 105 | 0.02 | 6 | 3.54 | 6 | 0.13 | 6 | 3.88 | 128 | 0.00 | 13 | 0.01 |
| 02 | 12 | 960 | 0.05 | 1005 | 0.02 | 960 | 0.06 | 110 | 3.04 | 13 | 0.20 | 179 | 6.04 | 1012 | 0.01 | 659 | 0.02 |
| 03 | 8 | 20803 | 1.89 | 52139 | 2.46 | 20803 | 2.82 | 244 | 22.64 | 9 | 36.89 | 818 | 24.47 | 52983 | 0.77 | 1802 | 1.33 |
| 04 | 11 | 110284 | 8.06 | 157722 | 9.60 | 110284 | 14.05 | 3892 | 16.68 | 12 | 155.03 | 8116 | 64.68 | 221429 | 3.06 | 41540 | 14.49 |
| 05 | 8 | 6531 | 0.86 | 13148 | 1.03 | 6531 | 1.32 | 376 | 15.46 | 9 | 120.06 | 313 | 59.99 | 12764 | 0.21 | 2834 | 1.61 |
| 06 | 10 | 20171 | 2.41 | 43583 | 4.32 | 20171 | 4.41 | 1794 | 328.18 | 11 | 201.44 | 3102 | 97.31 | 58487 | 0.87 | 15746 | 6.61 |
| 07 | 8 | 202706 | 73.83 | 2643752 | 1379.11 | 202706 | 208.81 | | | | | 2695 | 339.76 | 5404036 | 198.08 | 104531 | 420.47 |
| 08 | 11 | | | | | | | 96043 | 191.77 | | | | | | | | |
| 11 | 22 | 2345399 | 296.87 | 2629204 | 662.94 | 2365735 | 838.85 | 660104 | 28.60 | 660102 | 162.93 | | | 4116344 | 30.67 | 752867 | 334.42 |
| 13 | 16 | | | | | | | 188517 | 122.11 | | | | | | | | |
| 15 | 30 | 9652091 | 1721.67 | | | | | 2546587 | 141.12 | | | | | | | | |
| 17 | 44 | | | | | | | 12850247 | 352.46 | | | | | | | | |
| 21 | 14 | 839847 | 250.39 | | | | | 13241 | 69.80 | | | | | 4423951 | 65.44 | 126845 | 222.23 |
| 31 | 39 | 1501847 | 240.38 | 1568963 | 661.88 | 1504072 | 850.16 | 1357801 | 124.64 | | | | | 1726598 | 13.56 | 919764 | 381.66 |
| rovers-ipc5 | | | | | | | | | | | | | | | | | |
| 01 | 10 | 147 | 0.00 | 147 | 0.00 | 147 | 0.00 | 11 | 0.03 | 11 | 0.03 | 48 | 0.07 | 1104 | 0.00 | 283 | 0.00 |
| 02 | 8 | 44 | 0.00 | 44 | 0.00 | 44 | 0.00 | 9 | 0.00 | 9 | 0.00 | 16 | 0.03 | 254 | 0.00 | 129 | 0.00 |
| 03 | 11 | 672 | 0.01 | 419 | 0.00 | 448 | 0.01 | 12 | 0.11 | 12 | 0.12 | 804 | 0.16 | 3543 | 0.02 | 757 | 0.00 |
| 04 | 8 | 47 | 0.00 | 20 | 0.00 | 24 | 0.00 | 9 | 0.04 | 9 | 0.04 | 58 | 0.08 | 897 | 0.00 | 223 | 0.00 |
| 05 | 22 | 808084 | 22.61 | 410712 | 9.23 | 522937 | 18.29 | 617267 | 11.48 | 375808 | 18.46 | 298400 | 101.65 | 8559690 | 126.19 | 4318309 | 81.53 |
| 07 | 18 | 4546797 | 191.34 | 741649 | 21.01 | 1682245 | 102.77 | 3280884 | 51.02 | 2212903 | 59.20 | 1459792 | 866.93 | | | 9618062 | 199.91 |
| 12 | 19 | | | 1529551 | 76.46 | | | | | 5187273 | 166.77 | | | | | | |
| satellite-ipc4 | | | | | | | | | | | | | | | | | |
| 01 | 9 | 24 | 0.00 | 32 | 0.00 | 29 | 0.00 | 10 | 0.00 | 10 | 0.00 | 46 | 0.06 | 89 | 0.00 | 59 | 0.00 |
| 02 | 13 | 86 | 0.00 | 337 | 0.00 | 241 | 0.01 | 14 | 0.01 | 14 | 0.01 | 646 | 0.21 | 1728 | 0.01 | 940 | 0.00 |
| 03 | 11 | 2249 | 0.08 | 656 | 0.01 | 728 | 0.04 | 12 | 0.56 | 12 | 0.64 | 1945 | 0.93 | 15185 | 0.17 | 6822 | 0.11 |
| 04 | 17 | 9817 | 0.57 | 14860 | 0.38 | 11250 | 0.76 | 4152 | 0.99 | 18 | 4.43 | 15890 | 9.50 | 345663 | 4.70 | 180815 | 3.37 |
| 05 | 15 | 279569 | 49.47 | 46453 | 4.92 | 61692 | 18.85 | 81972 | 7.26 | 148667 | 69.28 | 267513 | 565.18 | | | | |
| 06 | 20 | 1496577 | 92.22 | 1572327 | 51.68 | 1518261 | 105.65 | 2769229 | 74.73 | 307962 | 32.52 | | | | | 10751017 | 371.43 |

Table 18: Similar to Table 15 for the Openstacks, Pathways, Pipesworld-NoTankage, Pipesworld-Tankage, Rovers, and Satellite domains.

| | | $h^{\mathfrak{F}}$ | | $h^{\mathfrak{I}}$ | | $h^{\mathfrak{FI}}$ | | *MS*-$10^4$ | | *MS*-$10^5$ | | HSP$^*_{\mathrm{F}}$ | | blind | | $h_{\max}$ | |
|---|---|---|---|---|---|---|---|---|---|---|---|---|---|---|---|---|---|
| task | $h^*$ | nodes | time | nodes | time | nodes | time | nodes | time | nodes | time | nodes | time | nodes | time | nodes | time |
| psr-small-ipc4 | | | | | | | | | | | | | | | | | |
| 01 | 8 | 10 | 0.00 | 10 | 0.00 | 10 | 0.00 | 9 | 0.00 | 9 | 0.00 | 9 | 0.01 | 11 | 0.00 | 9 | 0.00 |
| 02 | 11 | 52 | 0.00 | 55 | 0.00 | 52 | 0.00 | 12 | 0.00 | 12 | 0.00 | 20 | 0.08 | 71 | 0.00 | 47 | 0.00 |
| 03 | 11 | 31 | 0.00 | 31 | 0.00 | 31 | 0.00 | 12 | 0.00 | 12 | 0.00 | 20 | 0.04 | 33 | 0.00 | 28 | 0.00 |
| 04 | 10 | 66 | 0.00 | 91 | 0.00 | 73 | 0.00 | 11 | 0.00 | 11 | 0.00 | 12 | 0.34 | 332 | 0.00 | 102 | 0.00 |
| 05 | 11 | 75 | 0.00 | 79 | 0.00 | 75 | 0.00 | 12 | 0.00 | 12 | 0.00 | 23 | 0.11 | 154 | 0.00 | 69 | 0.00 |
| 06 | 8 | 10 | 0.00 | 10 | 0.00 | 10 | 0.00 | 9 | 0.00 | 9 | 0.00 | 9 | 0.01 | 11 | 0.00 | 9 | 0.00 |
| 07 | 11 | 61 | 0.00 | 61 | 0.00 | 61 | 0.00 | 12 | 0.00 | 12 | 0.00 | 26 | 0.09 | 122 | 0.00 | 62 | 0.00 |
| 08 | 8 | 24 | 0.00 | 29 | 0.00 | 25 | 0.00 | 9 | 0.00 | 9 | 0.00 | 9 | 0.12 | 128 | 0.00 | 52 | 0.00 |
| 09 | 8 | 18 | 0.00 | 19 | 0.00 | 18 | 0.00 | 9 | 0.00 | 9 | 0.00 | 9 | 0.06 | 49 | 0.00 | 20 | 0.00 |
| 10 | 7 | 131 | 0.01 | 183 | 0.00 | 155 | 0.01 | 8 | 0.04 | 8 | 0.04 | 18 | 1.04 | 1358 | 0.00 | 376 | 0.01 |
| 11 | 19 | 149 | 0.00 | 149 | 0.00 | 149 | 0.00 | 20 | 0.00 | 20 | 0.00 | 96 | 0.19 | 153 | 0.00 | 142 | 0.00 |
| 12 | 16 | 120 | 0.00 | 123 | 0.00 | 120 | 0.00 | 17 | 0.00 | 17 | 0.00 | 40 | 0.17 | 153 | 0.00 | 113 | 0.00 |
| 13 | 15 | 90 | 0.00 | 90 | 0.00 | 90 | 0.00 | 16 | 0.00 | 16 | 0.00 | 59 | 0.16 | 95 | 0.00 | 86 | 0.00 |
| 14 | 9 | 19 | 0.00 | 19 | 0.00 | 19 | 0.00 | 10 | 0.00 | 10 | 0.00 | 13 | 0.06 | 27 | 0.00 | 18 | 0.00 |
| 15 | 10 | 1200 | 0.08 | 708 | 0.03 | 769 | 0.09 | 11 | 0.46 | 11 | 2.58 | 356 | 18.99 | 3562 | 0.02 | 324 | 0.02 |
| 16 | 25 | 2328 | 0.02 | 2158 | 0.01 | 2176 | 0.03 | 975 | 0.11 | 26 | 0.12 | 2287 | 1.34 | 2742 | 0.01 | 1876 | 0.01 |
| 17 | 9 | 15 | 0.00 | 15 | 0.00 | 15 | 0.00 | 10 | 0.00 | 10 | 0.00 | 13 | 0.03 | 16 | 0.00 | 14 | 0.00 |
| 18 | 12 | 85 | 0.00 | 90 | 0.00 | 85 | 0.00 | 13 | 0.00 | 13 | 0.00 | 29 | 0.21 | 158 | 0.00 | 91 | 0.00 |
| 19 | 25 | 8025 | 0.11 | 7856 | 0.05 | 7876 | 0.12 | 2910 | 0.27 | 26 | 0.77 | 6338 | 4.46 | 9009 | 0.04 | 6925 | 0.08 |
| 20 | 17 | 80 | 0.00 | 80 | 0.00 | 80 | 0.00 | 18 | 0.00 | 18 | 0.00 | 52 | 0.18 | 84 | 0.00 | 75 | 0.00 |
| 21 | 10 | 28 | 0.00 | 28 | 0.00 | 28 | 0.00 | 11 | 0.00 | 11 | 0.00 | 21 | 0.12 | 42 | 0.00 | 31 | 0.00 |
| 22 | 33 | 163299 | 4.17 | 176058 | 1.56 | 168685 | 5.01 | 34 | 0.28 | 34 | 0.87 | 22315 | 8.16 | 189516 | 0.67 | 177138 | 1.43 |
| 23 | 12 | 77 | 0.00 | 93 | 0.00 | 77 | 0.00 | 13 | 0.00 | 13 | 0.01 | 30 | 0.43 | 200 | 0.00 | 116 | 0.00 |
| 24 | 10 | 28 | 0.00 | 28 | 0.00 | 28 | 0.00 | 11 | 0.00 | 11 | 0.00 | 21 | 0.12 | 42 | 0.00 | 31 | 0.00 |
| 25 | 9 | 485 | 3.06 | 463 | 0.58 | 482 | 3.28 | 10 | 5.42 | 10 | 37.93 | 28 | 780.38 | 8913 | 0.12 | 854 | 0.18 |
| 26 | 17 | 144 | 0.00 | 150 | 0.00 | 146 | 0.00 | 18 | 0.00 | 18 | 0.00 | 52 | 0.28 | 182 | 0.00 | 142 | 0.00 |
| 27 | 21 | 616 | 0.01 | 675 | 0.00 | 650 | 0.01 | 22 | 0.01 | 22 | 0.01 | 179 | 0.85 | 773 | 0.00 | 616 | 0.00 |
| 28 | 14 | 79 | 0.00 | 79 | 0.00 | 79 | 0.00 | 15 | 0.00 | 15 | 0.00 | 49 | 0.29 | 95 | 0.00 | 79 | 0.00 |
| 29 | 21 | 142772 | 4.55 | 187319 | 2.12 | 159325 | 5.80 | 22 | 0.39 | 22 | 1.43 | 3337 | 7.12 | 244499 | 1.27 | 192459 | 2.32 |
| 30 | 22 | 1791 | 0.03 | 1982 | 0.01 | 1883 | 0.04 | 23 | 0.01 | 23 | 0.02 | 393 | 1.35 | 2295 | 0.01 | 1834 | 0.01 |
| 31 | 19 | 11278 | 0.25 | 6810 | 0.08 | 8297 | 0.24 | 2647 | 0.89 | 723 | 6.55 | 7530 | 32.97 | 53911 | 0.25 | 16766 | 0.36 |
| 32 | 24 | 431 | 0.01 | 431 | 0.00 | 431 | 0.01 | 25 | 0.00 | 25 | 0.00 | 352 | 0.74 | 435 | 0.00 | 424 | 0.00 |
| 33 | 21 | 1480 | 0.02 | 1436 | 0.01 | 1391 | 0.03 | 446 | 0.26 | 22 | 0.63 | 947 | 2.29 | 2291 | 0.01 | 1073 | 0.01 |
| 34 | 21 | 223 | 0.00 | 223 | 0.00 | 223 | 0.00 | 22 | 0.00 | 22 | 0.00 | 158 | 0.50 | 227 | 0.00 | 216 | 0.00 |
| 35 | 22 | 65965 | 1.43 | 63186 | 0.46 | 68281 | 1.70 | 24021 | 0.83 | 11113 | 6.36 | 7448 | 8.27 | 165170 | 0.63 | 61548 | 1.06 |
| 36 | 22 | 571766 | 12.62 | 371834 | 3.41 | 458402 | 11.77 | 48350 | 2.98 | 2783 | 14.07 | 188564 | 111.99 | 1669788 | 9.44 | 717884 | 18.27 |
| 37 | 23 | 1307 | 0.03 | 1417 | 0.01 | 1363 | 0.03 | 24 | 0.02 | 24 | 0.01 | 277 | 2.10 | 1532 | 0.00 | 1342 | 0.01 |
| 38 | 13 | 301 | 0.01 | 372 | 0.00 | 326 | 0.01 | 14 | 0.01 | 14 | 0.01 | 33 | 0.74 | 562 | 0.00 | 357 | 0.00 |
| 39 | 23 | 2486 | 0.05 | 2942 | 0.02 | 2682 | 0.07 | 24 | 0.08 | 24 | 0.07 | 146 | 1.78 | 4103 | 0.01 | 2597 | 0.02 |
| 40 | 20 | 259683 | 8.59 | 182608 | 2.70 | 270195 | 11.73 | 38837 | 1.88 | 7767 | 12.86 | 23371 | 87.91 | 1036992 | 6.74 | 229210 | 9.51 |
| 41 | 10 | 31 | 0.00 | 34 | 0.00 | 31 | 0.00 | 11 | 0.00 | 11 | 0.00 | 21 | 0.16 | 54 | 0.00 | 35 | 0.00 |
| 42 | 30 | 1855 | 0.02 | 1747 | 0.01 | 1739 | 0.02 | 1117 | 0.18 | 31 | 0.18 | 1773 | 1.29 | 1908 | 0.01 | 1636 | 0.01 |
| 43 | 20 | 328 | 0.00 | 328 | 0.00 | 328 | 0.00 | 21 | 0.00 | 21 | 0.00 | 256 | 0.50 | 333 | 0.00 | 315 | 0.00 |
| 44 | 19 | 2990 | 0.07 | 3430 | 0.03 | 3121 | 0.08 | 20 | 0.05 | 20 | 0.05 | 407 | 2.18 | 4142 | 0.01 | 3235 | 0.02 |
| 45 | 20 | 347 | 0.00 | 376 | 0.00 | 359 | 0.01 | 21 | 0.01 | 21 | 0.00 | 121 | 0.74 | 434 | 0.00 | 358 | 0.00 |
| 46 | 34 | 60888 | 0.86 | 61842 | 0.31 | 61563 | 0.99 | 36941 | 0.67 | 32582 | 4.05 | 19865 | 6.91 | 80785 | 0.25 | 65984 | 0.63 |
| 47 | 27 | 4104 | 0.09 | 4522 | 0.03 | 4284 | 0.11 | 28 | 0.04 | 28 | 0.04 | 515 | 2.32 | 5075 | 0.01 | 4406 | 0.02 |
| 48 | 37 | 12080249 | 604.43 | 17435137 | 247.20 | 13514084 | 784.80 | 129627 | 2.37 | 2500 | 11.08 | 200559 | 101.21 | | | 19020089 | 286.02 |
| 49 | 47 | | | | | | | 2048368 | 15.84 | 594399 | 23.32 | 2772875 | 1408.64 | | | | |
| 50 | 23 | 637 | 0.01 | 659 | 0.01 | 645 | 0.02 | 24 | 0.02 | 24 | 0.02 | 390 | 1.40 | 690 | 0.00 | 642 | 0.00 |
| tpp-ipc5 | | | | | | | | | | | | | | | | | |
| 01 | 5 | 6 | 0.00 | 6 | 0.00 | 6 | 0.00 | 6 | 0.00 | 6 | 0.00 | 6 | 0.01 | 7 | 0.00 | 6 | 0.00 |
| 02 | 8 | 9 | 0.00 | 11 | 0.00 | 9 | 0.00 | 9 | 0.00 | 9 | 0.00 | 9 | 0.01 | 26 | 0.00 | 16 | 0.00 |
| 03 | 11 | 12 | 0.00 | 27 | 0.00 | 16 | 0.00 | 12 | 0.00 | 12 | 0.00 | 12 | 0.03 | 116 | 0.00 | 83 | 0.00 |
| 04 | 14 | 15 | 0.00 | 78 | 0.00 | 47 | 0.00 | 15 | 0.01 | 15 | 0.00 | 15 | 0.07 | 494 | 0.00 | 430 | 0.00 |
| 05 | 19 | 623 | 0.02 | 5110 | 0.08 | 1455 | 0.05 | 20 | 0.36 | 20 | 0.77 | 624 | 0.48 | 24698 | 0.12 | 17398 | 0.15 |
| 06 | 25 | 5843306 | 179.03 | 6916518 | 95.86 | 6153923 | 222.35 | 947059 | 14.22 | 74798 | 23.97 | | | | | 9267024 | 216.69 |
| trucks-ipc5 | | | | | | | | | | | | | | | | | |
| 01 | 13 | 1691 | 0.03 | 1027 | 0.01 | 1039 | 0.03 | 14 | 0.03 | 14 | 0.02 | 285 | 0.56 | 5774 | 0.02 | 402 | 0.01 |
| 02 | 17 | 9624 | 0.23 | 2898 | 0.04 | 2957 | 0.11 | 4192 | 0.22 | 18 | 0.17 | 1413 | 1.04 | 28348 | 0.14 | 939 | 0.03 |
| 03 | 20 | 80693 | 2.99 | 20752 | 0.44 | 22236 | 1.14 | 199405 | 2.89 | 173790 | 6.88 | 4049 | 4.43 | 379582 | 2.97 | 9465 | 0.40 |
| 04 | 23 | 1753866 | 48.55 | 1205793 | 23.48 | 1315672 | 50.35 | 2591561 | 29.17 | 2568634 | 56.96 | 8817 | 7.75 | 2990366 | 26.65 | 209140 | 9.43 |
| 05 | 25 | 12472562 | 515.50 | 8007189 | 242.98 | 9483222 | 512.55 | 23444940 | 392.99 | | | 14744 | 23.12 | | | 1248571 | 90.78 |
| 06 | 30 | | | | | | | | | | | 308920 | 343.47 | | | | |
| 07 | 23 | 2134728 | 96.15 | 719751 | 16.91 | 755608 | 50.72 | 7575415 | 88.91 | 8080496 | 117.13 | 43270 | 27.62 | 12410588 | 117.92 | 223011 | 19.34 |
| 08 | 25 | | | 5199440 | 221.76 | 6630689 | 687.95 | | | | | 49663 | 47.61 | | | 3106944 | 403.36 |
| 09 | 28 | | | | | | | | | | | 233577 | 248.21 | | | | |
| zenotravel-ipc3 | | | | | | | | | | | | | | | | | |
| 01 | 1 | 2 | 0.00 | 2 | 0.00 | 2 | 0.00 | 2 | 0.00 | 2 | 0.00 | 2 | 0.45 | 2 | 0.00 | 2 | 0.00 |
| 02 | 6 | 17 | 0.00 | 18 | 0.00 | 17 | 0.00 | 7 | 0.00 | 7 | 0.00 | 9 | 0.46 | 58 | 0.00 | 22 | 0.00 |
| 03 | 6 | 28 | 0.01 | 18 | 0.01 | 12 | 0.01 | 7 | 0.21 | 7 | 0.90 | 40 | 3.42 | 5160 | 0.04 | 492 | 0.02 |
| 04 | 8 | 99 | 0.01 | 88 | 0.01 | 81 | 0.01 | 9 | 0.20 | 9 | 0.89 | 215 | 3.44 | 5256 | 0.03 | 665 | 0.01 |
| 05 | 11 | 177 | 0.01 | 220 | 0.01 | 136 | 0.02 | 12 | 0.25 | 12 | 1.90 | 422 | 7.70 | 82289 | 0.63 | 12466 | 0.33 |
| 06 | 11 | 2287 | 0.10 | 1144 | 0.05 | 504 | 0.05 | 12 | 0.38 | 12 | 3.54 | 1957 | 11.81 | 596531 | 5.90 | 85931 | 2.47 |
| 07 | 15 | 5088 | 0.16 | 4234 | 0.09 | 4199 | 0.19 | 16 | 0.38 | 16 | 3.48 | 34890 | 30.36 | 405626 | 3.56 | 115348 | 2.60 |
| 08 | 11 | 3268 | 0.35 | 1026 | 0.12 | 1655 | 0.32 | 14354 | 2.00 | 12 | 14.48 | 83533 | 292.05 | | | 687846 | 50.76 |
| 09 | 21 | 2844771 | 177.70 | 2842546 | 176.05 | 2433822 | 262.84 | 2517035 | 51.18 | 611457 | 30.47 | | | | | | |
| 10 | 22 | 2283679 | 295.65 | 1921903 | 196.38 | 1832871 | 383.99 | 1322871 | 34.84 | 137872 | 25.44 | | | | | | |
| 11 | 14 | 139687 | 18.63 | 76904 | 8.20 | 93782 | 19.51 | 310030 | 11.28 | 110726 | 26.65 | | | | | | |

Table 19: Similar to Table 15 for the PSR, TPP, Trucks, and Zenotravel domains.

| | | $h^{\mathfrak{F}}$ | | $h^{\mathfrak{I}}$ | | $h^{\mathfrak{FI}}$ | | $MS$-$10^4$ | | $MS$-$10^5$ | | HSP$^*_{\mathrm{F}}$ | | blind | | $h_{\max}$ | |
|---|---|---|---|---|---|---|---|---|---|---|---|---|---|---|---|---|---|
| task | $h^*$ | nodes | time | nodes | time | nodes | time | nodes | time | nodes | time | nodes | time | nodes | time | nodes | time |
| schedule-strips | | | | | | | | | | | | | | | | | |
| 02-0 | 3 | 5 | 0.07 | 5 | 0.04 | 5 | 0.08 | 4 | 511.10 | 4 | 1743.32 | 5 | 577.39 | 76 | 0.02 | 5 | 0.09 |
| 02-1 | 2 | 3 | 0.08 | 4 | 0.05 | 3 | 0.10 | 3 | 104.98 | | | 3 | 754.26 | 6 | 0.02 | 3 | 0.07 |
| 02-2 | 2 | 3 | 0.17 | 3 | 0.06 | 3 | 0.19 | 3 | 231.99 | | | 3 | 495.56 | 5 | 0.02 | 3 | 0.07 |
| 02-3 | 3 | 26 | 0.17 | 37 | 0.06 | 26 | 0.18 | 4 | 56.51 | | | 4 | 658.90 | 529 | 0.03 | 95 | 0.45 |
| 02-4 | 3 | 68 | 0.17 | 188 | 0.07 | 220 | 0.26 | | | | | 4 | 484.62 | 543 | 0.03 | 108 | 0.44 |
| 02-5 | 2 | 3 | 0.17 | 3 | 0.05 | 3 | 0.19 | 3 | 363.11 | | | 3 | 667.32 | 3 | 0.03 | 3 | 0.07 |
| 02-6 | 2 | 3 | 0.07 | 5 | 0.04 | 3 | 0.09 | 3 | 121.84 | | | 3 | 697.42 | 6 | 0.02 | 3 | 0.06 |
| 02-7 | 2 | 3 | 0.15 | 3 | 0.05 | 3 | 0.17 | 3 | 323.77 | | | 3 | 604.06 | 13 | 0.02 | 3 | 0.07 |
| 02-8 | 2 | 3 | 0.17 | 3 | 0.05 | 3 | 0.19 | 3 | 316.53 | | | 3 | 668.79 | 8 | 0.02 | 3 | 0.07 |
| 02-9 | 3 | 5 | 0.07 | 5 | 0.04 | 5 | 0.08 | 4 | 251.46 | | | 5 | 577.16 | 76 | 0.03 | 5 | 0.09 |
| 03-0 | 4 | 40 | 0.31 | 407 | 0.16 | 140 | 0.45 | | | | | | | 11915 | 0.60 | 1127 | 8.98 |
| 03-1 | 2 | 3 | 0.22 | 3 | 0.08 | 3 | 0.25 | | | | | | | 31 | 0.04 | 25 | 0.37 |
| 03-2 | 4 | 27 | 0.21 | 50 | 0.09 | 33 | 0.25 | 5 | 191.03 | | | | | 3617 | 0.23 | 1228 | 9.56 |
| 03-3 | 4 | 15 | 0.13 | 91 | 0.09 | 15 | 0.15 | 5 | 259.13 | | | | | 3379 | 0.23 | 170 | 1.85 |
| 03-4 | 3 | 4 | 0.39 | 16 | 0.10 | 4 | 0.44 | | | | | 4 | 1223.90 | 301 | 0.06 | 22 | 0.27 |
| 03-5 | 4 | 73 | 0.38 | 471 | 0.14 | 74 | 0.43 | 5 | 682.30 | | | | | 12217 | 0.64 | 1175 | 12.43 |
| 03-6 | 4 | 72 | 0.12 | 75 | 0.08 | 69 | 0.13 | 5 | 121.58 | | | | | 2663 | 0.19 | 1542 | 11.73 |
| 03-7 | 4 | 28 | 0.23 | 50 | 0.09 | 28 | 0.25 | 5 | 195.72 | | | | | 12859 | 0.68 | 1323 | 13.47 |
| 03-8 | 4 | 273 | 0.43 | 266 | 0.14 | 273 | 0.48 | | | | | | | 12616 | 0.65 | 1590 | 11.13 |
| 03-9 | 4 | 8 | 0.23 | 31 | 0.09 | 14 | 0.27 | 5 | 235.48 | | | | | 4339 | 0.27 | 913 | 7.69 |
| 04-0 | 5 | 373 | 0.45 | 1498 | 0.50 | 167 | 0.54 | | | | | | | 312193 | 26.88 | 22993 | 273.38 |
| 04-1 | 6 | 17559 | 15.45 | 10707 | 3.48 | 17686 | 17.58 | 7 | 1115.76 | | | | | 552069 | 49.79 | | |
| 04-2 | 5 | 209 | 0.40 | 406 | 0.19 | 66 | 0.34 | | | | | | | 47696 | 4.97 | 9703 | 131.69 |
| 04-3 | 5 | 142 | 0.40 | 674 | 0.25 | 251 | 0.58 | 6 | 267.29 | | | | | 89272 | 8.74 | 12941 | 163.84 |
| 04-4 | 5 | 921 | 1.14 | 450 | 0.31 | 574 | 1.39 | | | | | | | 62013 | 6.03 | 13614 | 168.07 |
| 04-5 | 6 | 483 | 0.95 | 4544 | 1.11 | 850 | 2.11 | 7 | 837.68 | | | | | | | 107978 | 1399.99 |
| 04-6 | 6 | 779 | 0.56 | 11610 | 2.44 | 1834 | 1.43 | 7 | 459.19 | | | | | | | 107115 | 1001.40 |
| 04-7 | 5 | 99 | 0.58 | 424 | 0.31 | 163 | 0.78 | 6 | 936.68 | | | | | 61327 | 5.97 | 8683 | 103.50 |
| 04-8 | 5 | 102 | 0.52 | 573 | 0.24 | 111 | 0.60 | 6 | 711.65 | | | | | 340467 | 29.56 | 15122 | 181.98 |
| 04-9 | 4 | 1043 | 1.27 | 996 | 0.67 | 1050 | 1.66 | 5 | 316.22 | | | | | 41673 | 4.27 | 5480 | 83.69 |
| 05-0 | 5 | 163 | 0.86 | 483 | 0.51 | 167 | 1.05 | | | | | | | 143350 | 22.71 | 43336 | 751.35 |
| 05-1 | 6 | 2701 | 2.95 | 18878 | 11.36 | 1257 | 3.10 | | | | | | | | | | |
| 05-2 | 7 | 118855 | 86.65 | | | 158640 | 178.66 | | | | | | | | | | |
| 05-3 | 7 | 27159 | 24.88 | 41447 | 13.08 | 13622 | 16.72 | | | | | | | | | | |
| 05-4 | 6 | 989 | 1.63 | 3433 | 1.29 | 582 | 1.36 | | | | | | | | | | |
| 05-5 | 6 | 198 | 0.61 | 9550 | 4.61 | 347 | 1.05 | | | | | | | | | 120602 | 989.42 |
| 05-6 | 7 | 6033 | 11.16 | 49873 | 16.17 | 10325 | 16.63 | | | | | | | | | | |
| 05-7 | 6 | 944 | 1.92 | 17562 | 9.03 | 2107 | 4.10 | | | | | | | | | | |
| 05-8 | 7 | 1190 | 2.43 | 61539 | 20.22 | 2709 | 7.24 | | | | | | | | | | |
| 05-9 | 6 | 1537 | 2.24 | 15829 | 6.85 | 2717 | 5.45 | | | | | | | | | | |
| 06-2 | 6 | 888 | 3.29 | 26986 | 22.47 | 1709 | 6.91 | | | | | | | | | | |
| 06-4 | 8 | 11535 | 20.81 | | | 56273 | 131.69 | | | | | | | | | | |
| 06-6 | 8 | 15589 | 46.68 | | | 41764 | 133.76 | | | | | | | | | | |
| 07-0 | 7 | 2489 | 9.10 | | | 6995 | 25.49 | | | | | | | | | | |
| 07-7 | 8 | 10726 | 41.01 | | | 38251 | 154.49 | | | | | | | | | | |
| 07-9 | 8 | 6829 | 19.20 | | | 30148 | 109.49 | | | | | | | | | | |

Table 20: Similar to Table 15 for the (non-IPC) Schedule-STRIPS domain.

| | | $h^{\mathfrak{F}}$ | | $h^{\mathfrak{I}}$ | | $h^{\mathfrak{FI}}$ | | $\text{HSP}^*_{\text{F}}$ | | blind | |
|---|---|---|---|---|---|---|---|---|---|---|---|
| task | $h^*$ | nodes | time | nodes | time | nodes | time | nodes | time | nodes | time |
| elevators-strips-ipc6 | | | | | | | | | | | |
| 01 | 42 | 7483 | 0.39 | 10507 | 0.84 | 8333 | 1.03 | 12935 | 30.55 | 26670 | 0.40 |
| 02 | 26 | 2898 | 0.45 | 5184 | 1.12 | 4044 | 1.46 | 4810 | 42.63 | 16162 | 0.49 |
| 03 | 55 | 61649 | 4.00 | 219439 | 13.43 | 139760 | 15.62 | 276441 | 469.96 | 650316 | 11.32 |
| 04 | 40 | 60039 | 10.59 | 294029 | 74.29 | 146396 | 62.26 | 278087 | 885.94 | 1025329 | 29.51 |
| 05 | 55 | 909822 | 68.19 | 3269854 | 290.07 | 2113017 | 317.53 | | | 9567169 | 174.38 |
| 06 | 53 | 716238 | 125.83 | 3869775 | 1167.78 | 1965371 | 965.39 | | | | |
| 11 | 56 | 18313 | 1.34 | 50734 | 4.64 | 31545 | 5.00 | 72109 | 190.25 | 145170 | 2.53 |
| 12 | 54 | 21812 | 3.39 | 78362 | 23.84 | 46386 | 21.36 | 74663 | 325.43 | 152021 | 4.47 |
| 13 | 59 | 186526 | 18.43 | 432280 | 66.00 | 297147 | 68.80 | | | 1426461 | 32.06 |
| 14 | 63 | 248709 | 43.80 | 1325517 | 337.57 | 687420 | 290.86 | | | 6238743 | 199.63 |
| 15 | 66 | 201777 | 31.37 | 2823019 | 570.43 | 1255479 | 425.27 | | | | |
| 18 | 61 | 1057327 | 327.82 | | | | | | | | |
| 21 | 48 | 71003 | 5.99 | 79574 | 9.93 | 66582 | 13.62 | 123510 | 443.99 | 194669 | 4.28 |
| 22 | 54 | 890048 | 112.11 | 859710 | 349.90 | 757718 | 395.63 | | | 1633295 | 57.19 |
| 23 | 69 | 4089071 | 335.39 | 10935187 | 1208.05 | 7542146 | 1319.93 | | | | |
| 24 | 56 | 1430559 | 291.88 | | | | | | | | |
| 25 | 63 | 1384406 | 203.57 | | | 4430537 | 1578.04 | | | | |
| 26 | 48 | 699757 | 249.28 | | | | | | | | |
| openstacks-strips-ipc6 | | | | | | | | | | | |
| 01 | 2 | 209 | 0.00 | 209 | 0.00 | 209 | 0.00 | 49 | 0.37 | 193 | 0.00 |
| 02 | 2 | 769 | 0.02 | 769 | 0.00 | 769 | 0.02 | 144 | 0.73 | 769 | 0.00 |
| 03 | 2 | 1729 | 0.04 | 1729 | 0.01 | 1729 | 0.04 | 317 | 1.32 | 1665 | 0.01 |
| 04 | 3 | 8209 | 0.17 | 8209 | 0.07 | 8209 | 0.20 | 2208 | 2.63 | 8113 | 0.06 |
| 05 | 4 | 16705 | 0.41 | 16705 | 0.18 | 16705 | 0.48 | 4220 | 4.87 | 17151 | 0.16 |
| 06 | 2 | 3658 | 0.11 | 3658 | 0.04 | 3658 | 0.13 | 998 | 5.66 | 3288 | 0.02 |
| 07 | 5 | 195109 | 5.85 | 195109 | 2.45 | 195109 | 6.85 | 61253 | 40.74 | 201137 | 2.31 |
| 08 | 5 | 228847 | 7.77 | 228847 | 3.23 | 228847 | 9.06 | 70808 | 57.12 | 234328 | 2.92 |
| 09 | 3 | 116425 | 5.03 | 116425 | 1.61 | 116425 | 5.77 | 4920 | 18.26 | 114281 | 1.32 |
| 10 | 3 | 77681 | 3.57 | 77681 | 1.10 | 77681 | 4.14 | 5261 | 23.40 | 72673 | 0.76 |
| 11 | 4 | 575677 | 28.75 | 575677 | 9.11 | 575677 | 32.97 | 98783 | 105.44 | 563261 | 7.17 |
| 12 | 3 | 354913 | 19.85 | 354913 | 5.63 | 354913 | 22.85 | 10580 | 43.05 | 341169 | 4.11 |
| 13 | 4 | 2596593 | 150.86 | 2596593 | 46.30 | 2596593 | 172.13 | 398023 | 443.57 | 2547985 | 35.07 |
| 14 | 4 | 1260363 | 81.43 | 1260363 | 23.36 | 1260363 | 93.01 | 157304 | 222.14 | 1233115 | 17.19 |
| 15 | 4 | 11995225 | 867.27 | 11995225 | 245.32 | 11995225 | 987.24 | 711526 | 1034.92 | 11926297 | 184.57 |
| 16 | 4 | 5064737 | 379.45 | 5064737 | 104.37 | 5064737 | 432.44 | 411732 | 671.53 | 4928793 | 75.73 |
| 17 | 4 | 8193065 | 673.91 | 8193065 | 179.00 | 8193065 | 765.15 | 421646 | 745.34 | 8065113 | 128.80 |
| 18 | 3 | 1020905 | 88.15 | 1020905 | 22.24 | 1020905 | 99.67 | 34754 | 186.40 | 953049 | 14.32 |
| 19 | 4 | | | | | | | 812451 | 1731.49 | | |
| 21 | 3 | | | | | | | 473553 | 1018.62 | | |
| 22 | 4 | 1805050 | 204.83 | 1805050 | 48.73 | 1805050 | 233.98 | 173929 | 651.93 | 1536764 | 27.62 |
| parcprinter-strips-ipc6 | | | | | | | | | | | |
| 01 | 169009 | 19 | 0.01 | 15 | 0.00 | 15 | 0.00 | 12 | 0.20 | 20 | 0.00 |
| 02 | 438047 | 240 | 0.02 | 183 | 0.01 | 179 | 0.02 | 19 | 1.44 | 1375 | 0.01 |
| 03 | 807114 | 880 | 0.04 | 821 | 0.01 | 668 | 0.03 | 334 | 0.72 | 4903 | 0.03 |
| 04 | 876094 | 142314 | 13.85 | 77520 | 2.28 | 68116 | 7.56 | 993 | 11.21 | 12302518 | 126.46 |
| 05 | 1145132 | 1780073 | 219.49 | 892002 | 31.78 | 822442 | 115.96 | 6922 | 35.00 | | |
| 06 | 1514200 | 4113487 | 613.02 | 3529327 | 148.95 | 3443221 | 557.75 | 19613 | 115.36 | | |
| 11 | 182808 | 25 | 0.01 | 24 | 0.00 | 24 | 0.01 | 10 | 0.55 | 23 | 0.00 |
| 12 | 510256 | 1183 | 0.07 | 1243 | 0.04 | 1135 | 0.08 | 153 | 3.39 | 5138 | 0.05 |
| 13 | 693064 | 74201 | 5.83 | 144084 | 4.27 | 97683 | 9.25 | 8348 | 18.14 | 1130810 | 12.72 |
| 14 | 1020512 | 4491265 | 463.93 | | | | | 422571 | 792.78 | | |
| 21 | 143411 | 13 | 0.00 | 13 | 0.00 | 13 | 0.00 | 9 | 0.09 | 16 | 0.00 |
| 22 | 375821 | 225 | 0.01 | 303 | 0.01 | 282 | 0.02 | 22 | 0.51 | 2485 | 0.02 |
| 23 | 519232 | 4376 | 0.28 | 15825 | 0.47 | 8778 | 0.63 | 260 | 2.03 | 285823 | 3.32 |
| 24 | 751642 | 96748 | 8.49 | 694503 | 24.62 | 316839 | 31.37 | 2281 | 6.38 | | |
| 25 | 1215840 | | | | | | | 68293 | 145.47 | | |
| 26 | 1216460 | | | | | | | 121897 | 404.98 | | |
| scanalyzer-strips-ipc6 | | | | | | | | | | | |
| 01 | 18 | 19788 | 1.68 | 22012 | 5.20 | 19809 | 6.39 | 21259 | 13.69 | 44047 | 0.68 |
| 02 | 22 | 37182 | 1.88 | 37569 | 4.36 | 37524 | 6.07 | 29253 | 13.92 | 45529 | 0.54 |
| 03 | 26 | 43115 | 1.90 | 43298 | 4.02 | 43298 | 5.71 | 37754 | 14.05 | 45882 | 0.49 |
| 04 | 24 | 3947796 | 687.38 | | | | | | | 10175657 | 314.87 |
| 05 | 30 | 9193480 | 870.50 | | | | | | | 10310817 | 242.27 |
| 06 | 36 | 10140909 | 869.52 | | | | | | | 10321465 | 222.66 |
| 22 | 13 | 46 | 0.14 | 51 | 0.08 | 46 | 0.20 | 6 | 0.05 | 54 | 0.01 |
| 23 | 13 | 46 | 0.15 | 51 | 0.08 | 46 | 0.19 | 6 | 0.05 | 54 | 0.01 |
| 24 | 13 | 46 | 0.14 | 51 | 0.08 | 46 | 0.19 | 6 | 0.05 | 54 | 0.01 |
| 25 | 26 | 8974317 | 834.36 | | | | | | | 10170980 | 113.29 |
| 26 | 30 | 9936832 | 720.23 | | | | | | | 10254740 | 95.91 |
| 27 | 34 | 10202674 | 643.41 | | | | | | | 10294023 | 88.03 |

Table 21: Runtimes of cost-optimal heuristic-search planners on the Elevators, Openstacks-strips-08, Parcprinter, and Scanalyzer domains. The description of the planners is given in Section 6; here the fork-decomposition heuristics are via structural-pattern databases. Column *task* denotes problem instance, column $h^*$ denotes optimal solution length. Other columns capture the run *time* and number of expanded *nodes*.

| | | $h^{\mathfrak{F}}$ | | $h^{\mathfrak{I}}$ | | $h^{\mathfrak{FI}}$ | | $\text{HSP}^*_{\text{F}}$ | | blind | |
|---|---|---|---|---|---|---|---|---|---|---|---|
| task | $h^*$ | nodes | time | nodes | time | nodes | time | nodes | time | nodes | time |
| pegsol-strips-ipc6 | | | | | | | | | | | |
| 01 | 2 | 12 | 0.02 | 10 | 0.02 | 10 | 0.03 | 6 | 0.16 | 11 | 0.00 |
| 02 | 5 | 84 | 0.07 | 83 | 0.12 | 83 | 0.18 | 20 | 5.17 | 66 | 0.01 |
| 03 | 4 | 208 | 0.07 | 209 | 0.12 | 209 | 0.19 | 50 | 6.91 | 174 | 0.00 |
| 04 | 4 | 193 | 0.07 | 181 | 0.12 | 181 | 0.19 | 15 | 1.82 | 192 | 0.01 |
| 05 | 4 | 266 | 0.03 | 251 | 0.02 | 251 | 0.04 | 43 | 5.62 | 242 | 0.01 |
| 06 | 4 | 1343 | 0.16 | 901 | 0.14 | 901 | 0.27 | 247 | 25.68 | 1265 | 0.01 |
| 07 | 3 | 217 | 0.08 | 110 | 0.12 | 110 | 0.18 | 26 | 11.67 | 215 | 0.01 |
| 08 | 6 | 31681 | 2.56 | 25253 | 0.71 | 25253 | 2.89 | 7898 | 28.50 | 30776 | 0.15 |
| 09 | 5 | 3743 | 0.36 | 3951 | 0.20 | 3951 | 0.53 | 757 | 23.02 | 3538 | 0.03 |
| 10 | 6 | 29756 | 2.45 | 28241 | 0.77 | 28241 | 3.21 | 7522 | 28.25 | 29658 | 0.14 |
| 11 | 7 | 13832 | 1.08 | 12881 | 0.38 | 12881 | 1.36 | 5979 | 20.60 | 13430 | 0.06 |
| 12 | 8 | 39340 | 2.98 | 37358 | 0.86 | 37358 | 3.63 | 21133 | 32.73 | 38561 | 0.18 |
| 13 | 9 | 33379 | 2.51 | 33374 | 0.76 | 33374 | 3.09 | 25897 | 33.29 | 32370 | 0.15 |
| 14 | 7 | 63096 | 4.82 | 55127 | 1.29 | 55127 | 5.63 | 17144 | 32.20 | 62047 | 0.29 |
| 15 | 8 | 77932 | 5.84 | 73733 | 1.67 | 73733 | 7.09 | 37810 | 38.72 | 76150 | 0.35 |
| 16 | 8 | 10491 | 0.83 | 10598 | 0.33 | 10598 | 1.10 | 7939 | 27.70 | 10090 | 0.05 |
| 17 | 10 | 299676 | 22.38 | 300972 | 6.38 | 300972 | 27.00 | 282810 | 124.39 | 294396 | 1.44 |
| 18 | 7 | 63247 | 4.93 | 50222 | 1.37 | 50222 | 5.55 | 10358 | 29.81 | 62726 | 0.29 |
| 19 | 8 | 279822 | 20.71 | 257988 | 5.62 | 257988 | 24.49 | 90950 | 61.77 | 275969 | 1.29 |
| 20 | 7 | 329570 | 27.36 | 293860 | 8.56 | 293860 | 36.43 | 83693 | 63.99 | 328583 | 1.63 |
| 21 | 8 | 548254 | 41.78 | 494477 | 11.49 | 494477 | 50.51 | 141906 | 87.89 | 545896 | 2.64 |
| 22 | 6 | 69922 | 5.66 | 48190 | 1.43 | 48190 | 5.95 | 13123 | 30.94 | 69465 | 0.33 |
| 23 | 8 | 1262645 | 97.46 | 954593 | 25.16 | 954593 | 108.53 | 181830 | 114.89 | 1258767 | 6.17 |
| 24 | 8 | 1326517 | 106.00 | 1219589 | 31.83 | 1219589 | 136.27 | 271157 | 157.50 | 1324907 | 6.69 |
| 25 | 8 | 830637 | 68.95 | 899323 | 25.33 | 899323 | 107.61 | 201932 | 122.27 | 830182 | 4.33 |
| 26 | 9 | 7196836 | 553.11 | 6943124 | 177.06 | 6943124 | 719.81 | 2031156 | 1024.04 | 7178802 | 37.78 |
| 27 | 7 | 6092258 | 523.12 | 2121936 | 82.61 | 2121936 | 339.83 | 132701 | 118.20 | 6091864 | 34.53 |
| sokoban-strips-ipc6 | | | | | | | | | | | |
| 01 | 11 | 372 | 0.03 | 287 | 0.01 | 269 | 0.02 | 1079 | 6.17 | 1762 | 0.01 |
| 02 | 9 | 551 | 0.02 | 497 | 0.01 | 509 | 0.02 | 700 | 4.57 | 1348 | 0.00 |
| 03 | 10 | 394 | 0.01 | 177 | 0.00 | 173 | 0.01 | 621 | 2.63 | 1165 | 0.00 |
| 04 | 29 | 130524 | 5.57 | 45048 | 0.28 | 44198 | 2.07 | 282895 | 177.07 | 320446 | 1.43 |
| 05 | 8 | 50 | 0.32 | 203202 | 7.28 | 3073 | 0.96 | | | 9607487 | 81.84 |
| 06 | 9 | 526 | 0.04 | 534 | 0.01 | 526 | 0.04 | 6815 | 6.83 | 10526 | 0.04 |
| 07 | 15 | 47522 | 2.81 | 42195 | 0.38 | 28163 | 1.84 | 75669 | 174.92 | 315405 | 1.49 |
| 08 | 31 | 2114443 | 135.12 | 1204212 | 11.27 | 1080337 | 74.55 | | | 13329538 | 77.70 |
| 09 | 19 | 23083 | 1.47 | 26189 | 0.26 | 16013 | 1.12 | 459188 | 400.45 | 818693 | 4.09 |
| 10 | 30 | 69797 | 3.17 | 21291 | 0.18 | 20741 | 1.07 | 620685 | 315.43 | 852150 | 4.07 |
| 11 | 35 | 271598 | 15.63 | 282061 | 2.09 | 271598 | 16.87 | 440869 | 586.91 | 531305 | 2.71 |
| 12 | 32 | 155166 | 10.98 | 60655 | 0.70 | 46865 | 3.69 | | | 4705742 | 25.21 |
| 13 | 20 | 169436 | 8.93 | 294710 | 3.63 | 169436 | 10.26 | 1631677 | 994.61 | 2363177 | 12.60 |
| 14 | 29 | 20737 | 1.05 | 6984 | 0.07 | 6952 | 0.41 | 178574 | 121.96 | 255203 | 1.17 |
| 15 | 76 | 7943562 | 602.99 | 7742698 | 84.75 | 7456505 | 622.89 | | | 21598353 | 120.25 |
| 16 | 50 | 335238 | 20.20 | 242778 | 1.66 | 240912 | 15.14 | 852948 | 859.44 | 935561 | 4.74 |
| 17 | 37 | 80459 | 4.17 | 40425 | 0.29 | 36889 | 2.05 | 239522 | 220.86 | 317984 | 1.43 |
| 18 | 49 | 2109516 | 156.88 | 119938 | 0.97 | 119784 | 9.18 | | | 7219504 | 39.20 |
| 19 | 47 | 5238957 | 354.84 | 3558809 | 33.60 | 3459314 | 251.38 | | | 23255133 | 130.46 |
| 20 | 2 | 648 | 0.14 | 648 | 0.69 | 648 | 0.79 | | | 649 | 0.01 |
| 21 | 10 | 337852 | 74.21 | 450027 | 14.64 | 76647 | 16.85 | | | | |
| 22 | 44 | 5866700 | 473.77 | 4053413 | 45.18 | 3868663 | 335.31 | | | | |
| 23 | 31 | 3565151 | 222.48 | 3613835 | 50.31 | 2563159 | 181.66 | | | | |
| 24 | 50 | 14504610 | 1151.55 | 2244156 | 30.10 | 1759660 | 154.33 | | | | |
| 25 | 39 | | | 23044275 | 275.91 | 17832156 | 1612.04 | | | | |
| 26 | 33 | | | 12138101 | 152.95 | 10473204 | 996.25 | | | | |
| 27 | 23 | | | | | 8738457 | 1131.26 | | | | |
| 30 | 14 | 2074534 | 679.61 | | | | | | | | |
| transport-strips-ipc6 | | | | | | | | | | | |
| 01 | 54 | 60 | 0.01 | 12 | 0.00 | 16 | 0.01 | 60 | 0.48 | 64 | 0.00 |
| 02 | 131 | 1558 | 0.06 | 874 | 0.10 | 998 | 0.15 | 1567 | 6.36 | 2093 | 0.01 |
| 03 | 250 | 380375 | 10.47 | 225310 | 48.69 | 257608 | 59.28 | 380982 | 274.86 | 408643 | 3.69 |
| 04 | 318 | 3526204 | 164.35 | 1462063 | 714.49 | 1660874 | 856.87 | | | 4204372 | 50.69 |
| 11 | 456 | 135 | 0.02 | 111 | 0.01 | 103 | 0.02 | 135 | 0.94 | 164 | 0.00 |
| 12 | 594 | 14873 | 0.37 | 9976 | 1.41 | 11130 | 1.70 | 14874 | 19.85 | 14796 | 0.12 |
| 13 | 550 | 372845 | 15.07 | 224986 | 74.09 | 246069 | 89.04 | 373133 | 454.55 | 408449 | 4.36 |
| 21 | 478 | 62 | 0.01 | 67 | 0.00 | 62 | 0.01 | 62 | 0.50 | 112 | 0.00 |
| 22 | 632 | 7544 | 0.18 | 4455 | 0.37 | 5408 | 0.54 | 7544 | 7.71 | 7610 | 0.06 |
| 23 | 630 | 100269 | 3.65 | 56897 | 13.82 | 70579 | 19.32 | 100347 | 92.93 | 106548 | 1.07 |
| 24 | 614 | 1587821 | 77.96 | 292004 | 120.98 | 382588 | 196.82 | | | 1663856 | 19.29 |
| woodworking-strips-ipc6 | | | | | | | | | | | |
| 01 | 170 | 4313 | 0.23 | 3716 | 0.10 | 4157 | 0.28 | 119 | 0.85 | 9086 | 0.09 |
| 02 | 185 | 5550 | 0.34 | 5054 | 0.14 | 5408 | 0.41 | 409 | 3.03 | 21076 | 0.30 |
| 03 | 275 | | | | | | | 80794 | 136.95 | | |
| 11 | 130 | 860 | 0.10 | 987 | 0.05 | 897 | 0.13 | 50 | 0.93 | 3487 | 0.05 |
| 12 | 225 | 328229 | 41.44 | 328728 | 16.57 | 328930 | 52.03 | 11665 | 18.49 | 1862476 | 37.91 |
| 13 | 215 | 4413726 | 954.34 | 4125788 | 455.35 | 4404104 | 1297.06 | 113386 | 273.76 | | |
| 21 | 95 | 54 | 0.02 | 54 | 0.02 | 53 | 0.03 | 16 | 0.91 | 227 | 0.00 |
| 22 | 185 | 31189 | 4.66 | 67528 | 3.26 | 38912 | 6.83 | 1931 | 5.96 | 177942 | 3.97 |
| 23 | 195 | 44641 | 8.39 | 155426 | 9.71 | 64840 | 14.42 | 4673 | 9.05 | 962698 | 23.76 |

Table 22: Similar to Table 21 for the Pegsol, Sokoban, Transport, and Woodworking domains.

## References

Bäckström, C., & Nebel, B. (1995). Complexity results for SAS$^+$ planning. *Computational Intelligence*, *11*(4), 625–655.

Bonet, B., & Geffner, H. (2001). Planning as heuristic search. *Artificial Intelligence*, *129*(1–2), 5–33.

Bylander, T. (1994). The computational complexity of propositional STRIPS planning. *Artificial Intelligence*, *69*(1-2), 165–204.

Chen, H., & Gimenez, O. (2008). Causal graphs and structurally restricted planning. In *Proceedings of the 18th International Conference on Automated Planning and Scheduling (ICAPS)*, pp. 36–43, Sydney, Australia.

Clarke, E., Grumberg, O., & Peled, D. (1999). *Model Checking*. MIT Press.

Coles, A. I., Fox, M., Long, D., & Smith, A. J. (2008). Additive-disjunctive heuristics for optimal planning. In *Proceedings of the 18th International Conference on Automated Planning and Scheduling (ICAPS)*, pp. 44–51.

Culberson, J., & Schaeffer, J. (1998). Pattern databases. *Computational Intelligence*, *14*(4), 318–334.

Domshlak, C., & Dinitz, Y. (2001). Multi-agent off-line coordination: Structure and complexity. In *Proceedings of Sixth European Conference on Planning (ECP)*, pp. 277–288.

Domshlak, C., Hoffmann, J., & Sabharwal, A. (2009). Friends or foes? On planning as satisfiability and abstract CNF encodings. *Journal of Artificial Intelligence Research*, *36*, 415–469.

Domshlak, C., Katz, M., & Lefler, S. (2010). When abstractions met landmarks. In *Proceedings of the 20th International Conference on Automated Planning and Scheduling (ICAPS)*, pp. 50–56, Toronto, Canada.

Dräger, K., Finkbeiner, B., & Podelski, A. (2006). Directed model checking with distance-preserving abstractions. In Valmari, A. (Ed.), *Proceedings of the 13th International SPIN Workshop on Model Checking Software*, Vol. 3925 of *Lecture Notes in Computer Science*, pp. 19–36, Berlin Heidelberg. Springer-Verlag.

Edelkamp, S. (2001). Planning with pattern databases. In *Proceedings of the European Conference on Planning (ECP)*, pp. 13–34.

Edelkamp, S. (2002). Symbolic pattern databases in heuristic search planning. In *Proceedings of the International Conference on AI Planning and Scheduling (AIPS)*, pp. 274–293.

Edelkamp, S. (2006). Automated creation of pattern database search heuristics. In *Proceedings of the 4th Workshop on Model Checking and Artificial Intelligence (MoChArt)*.

Edelkamp, S., & Kissmann, P. (2009). Optimal symbolic planning with action costs and preferences. In *Proceedings of the 21st International Joint Conference on Artificial Intelligence (IJCAI)*, pp. 1690–1695, Pasadena, CA, US.

Felner, A., Korf, R. E., & Hanan, S. (2004). Additive pattern database heuristics. *Journal of Artificial Intelligence Research*, *22*, 279–318.

Haslum, P. (2008). Additive and reversed relaxed reachability heuristics revisited. In *Proceedings of the 6th International Planning Competition.*

Haslum, P., Bonet, B., & Geffner, H. (2005). New admissible heuristics for domain-independent planning. In *Proceedings of the Twentieth National Conference on Artificial Intelligence (AAAI)*, pp. 1163–1168.

Haslum, P., Botea, A., Helmert, M., Bonet, B., & Koenig, S. (2007). Domain-independent construction of pattern database heuristics for cost-optimal planning. In *Proceedings of the 19th National Conference on Artificial Intelligence (AAAI)*, pp. 1007–1012.

Haslum, P., & Geffner, H. (2000). Admissible heuristics for optimal planning. In *Proceedings of the Fifth International Conference on Artificial Intelligence Planning Systems (ICAPS)*, pp. 140–149.

Helmert, M. (2003). Complexity results for standard benchmark domains in planning. *Artificial Intelligence*, *146*(2), 219–262.

Helmert, M. (2004). A planning heuristic based on causal graph analysis. In *Proceedings of the 14th International Conference on Automated Planning and Scheduling (ICAPS)*, pp. 161–170, Whistler, Canada.

Helmert, M. (2006). The Fast Downward planning system. *Journal of Artificial Intelligence Research*, *26*, 191–246.

Helmert, M., & Domshlak, C. (2009). Landmarks, critical paths and abstractions: What's the difference anyway?. In *Proceedings of the 19th International Conference on Automated Planning and Scheduling (ICAPS)*, pp. 162–169, Thessaloniki, Greece.

Helmert, M., Haslum, P., & Hoffmann, J. (2007). Flexible abstraction heuristics for optimal sequential planning. In *Proceedings of the 17th International Conference on Automated Planning and Scheduling (ICAPS)*, pp. 176–183, Providence, RI, USA.

Helmert, M., & Mattmüller, R. (2008). Accuracy of admissible heuristic functions in selected planning domains. In *Proceedings of the 23rd AAAI Conference on Artificial Intelligence*, pp. 938–943, Chicago, USA.

Helmert, M. (2008). *Understanding Planning Tasks: Domain Complexity and Heuristic Decomposition*, Vol. 4929 of *Lecture Notes in Computer Science.* Springer.

Hernadvölgyi, I., & Holte, R. (1999). PSVN: A vector representation for production systems. Tech. rep. 1999-07, University of Ottawa.

Jonsson, A. (2007). The role of macros in tractable planning over causal graphs. In *Proceedings of the International Joint Conference on Artificial Intelligence (IJCAI-07)*, pp. 1936–1941.

Jonsson, P., & Bäckström, C. (1998). State-variable planning under structural restrictions: Algorithms and complexity. *Artificial Intelligence*, *100*(1–2), 125–176.

Karpas, E., & Domshlak, C. (2009). Cost-optimal planning with landmarks. In *Proceedings of the International Joint Conference on Artificial Intelligence (IJCAI-09)*, pp. 1728–1733, Pasadena, CA, USA.

Katz, M., & Domshlak, C. (2007a). Structural patterns heuristics. In *ICAPS-07 Workshop on Heuristics for Domain-independent Planning: Progress, Ideas, Limitations, Challenges*, Providence, RI, USA.

Katz, M., & Domshlak, C. (2007b). Structural patterns of tractable sequentially-optimal planning. In *Proceedings of the 17th International Conference on Automated Planning and Scheduling (ICAPS)*, pp. 200–207, Providence, RI, USA.

Katz, M., & Domshlak, C. (2008). Structural patterns heuristics via fork decomposition. In *Proceedings of the 18th International Conference on Automated Planning and Scheduling (ICAPS)*, pp. 182–189, Sydney, Australia.

Katz, M., & Domshlak, C. (2009). Structural-pattern databases. In *Proceedings of the 19th International Conference on Automated Planning and Scheduling (ICAPS)*, pp. 186–193, Thessaloniki, Greece.

Katz, M., & Domshlak, C. (2010). Optimal admissible composition of abstraction heuristics. *Artificial Intelligence*, *174*, 767–798.

Pearl, J. (1984). *Heuristics - Intelligent Search Strategies for Computer Problem Solving*. Addison-Wesley.

Prieditis, A. (1993). Machine discovery of effective admissible heuristics. *Machine Learning*, *12*, 117–141.

Richter, S., Helmert, M., & Westphal, M. (2008). Landmarks revisited. In *Proceedings of the Twenty-Third National Conference on Artificial Intelligence (AAAI)*, pp. 975–982, Chicago, IL, USA.

Yang, F., Culberson, J., & Holte, R. (2007). A general additive search abstraction. Tech. rep. TR07-06, University of Alberta.

Yang, F., Culberson, J., Holte, R., Zahavi, U., & Felner, A. (2008). A general theory of additive state space abstractions. *Journal of Artificial Intelligence Research*, *32*, 631–662.